%% file: main.tex
\documentclass{article}

\usepackage{afterpage}
\usepackage{amsfonts}        % blackboard math symbols
\usepackage{amsmath,amssymb} % define this before the line numbering. 
\usepackage{array}
\usepackage{booktabs}        % for professional tables
\usepackage{graphicx}
\usepackage{lineno}
\usepackage{lipsum}
\usepackage{microtype}
\usepackage{multirow}
\usepackage{skak}
\usepackage{subcaption}
\usepackage{xfrac}

\input{quotes-with-tikz.tex}
\usetikzlibrary{shapes.misc}
\tikzset{cross/.style={cross out, draw=black, minimum size=2*(#1-\pgflinewidth), inner sep=0pt, outer sep=0pt},
cross/.default={1pt}}

\usepackage[accepted]{arxiv}
\newcommand{\figref}[1]{Figure \ref{#1}}

\def\drm{\mathrm{d}}
\def\erm{\mathrm{e}}

\def\Ebb{\mathbb{E}}
\def\Ibb{\mathbb{I}}
\def\Rbb{\mathbb{R}}
\def\Acal{\mathcal{A}}

\def\Gcal{\mathcal{G}}
\def\Hcal{\mathcal{H}}
\def\Mcal{\mathcal{M}}
\def\p{\mathbf{p}}
\def\q{\mathbf{q}}
\def\r{\mathbf{r}}
\def\s{\mathbf{s}}

\def\bookmove{\textcolor{gray}{\emph{(book) }}}
\def\defined{\stackrel{\text{\tiny def}}{=}}

\newcommand{\move}[1]{\textcolor{gray}{\textit{#1.~}}}
\newcommand{\blackmove}[1]{\textcolor{gray}{\textit{#1\ldots~}}}
\newcommand{\specialmove}[1]{\underline{#1}}
\newcommand{\mmove}[1]{#1.~}
\newcommand{\mblackmove}[1]{#1\ldots~}
\newcommand{\scalechessboard}[1]{\scalebox{0.9}{#1}}
\def\gamedrawn{\hfill \textbf{1/2--1/2}}
\def\blackwins{\hfill \textbf{0--1}}
\def\whitewins{\hfill \textbf{1--0}}

\newcommand{\ulrich}[1]{}

\newcommand{\matthias}[1]{}


\icmltitlerunning{Assessing Game Balance with AlphaZero}

\begin{document}

\twocolumn[
\icmltitle{Assessing Game Balance with AlphaZero: \\ Exploring Alternative Rule Sets in Chess}

\vskip 0.1in

\vbox{%
\hsize\textwidth
\linewidth\hsize
\centering
\begin{tabular}[t]{c}
\textbf{Nenad~Toma\v{s}ev}\textsuperscript{*} \\ DeepMind
\end{tabular}
\hspace{25pt}
\begin{tabular}[t]{c}
\textbf{Ulrich Paquet}\textsuperscript{*} \\ DeepMind
\end{tabular}
\hspace{25pt}
\begin{tabular}[t]{c}
\textbf{Demis Hassabis} \\ DeepMind
\end{tabular}
\hspace{25pt}
\begin{tabular}[t]{c}
\textbf{Vladimir Kramnik} \\
% 14th World Chess Champion\textsuperscript{\S}
World Chess Champion \\ 2000--2007\textsuperscript{\S}
\end{tabular}
}

\icmlkeywords{AlphaZero}

\vskip 0.4in
]


% TODO -- subsection capitalization = ?

\begin{abstract}
\input{abstract.tex}
\end{abstract}

\input{introduction.tex}
\input{methods.tex}
\input{results-quantitative.tex}
\input{results-qualitative.tex}
\input{conclusions.tex}

\section*{Acknowledgements}
We would like to thank chess grandmasters Peter Heine Nielsen, and Matthew Sadler for their valuable feedback on our preliminary findings and the early version of the manuscript. Oliver Smith and Kareem Ayoub have been of great help in managing the project. We would also like to thank the team of Chess.com for providing us with a platform to announce and discuss No-castling chess and present annotated games. 

\bibliography{az_balance_bibliography}
\bibliographystyle{icml2018}

\clearpage

\appendix
\input{appendix-math.tex}
\clearpage
\input{appendix-SAN.tex}

\end{document}

%% file: quotes-with-tikz.tex
\usepackage{ifxetex,ifluatex}
\usepackage{etoolbox}
\usepackage[svgnames]{xcolor}
\usepackage{tikz}
\usepackage{framed}

\newif\ifxetexorluatex
\ifxetex
  \xetexorluatextrue
\else
  \ifluatex
    \xetexorluatextrue
  \else
    \xetexorluatexfalse
  \fi
\fi
\ifxetexorluatex%
  \usepackage{fontspec}
  \usepackage{libertine} % or use \setmainfont to choose any font on your system
  \newfontfamily\quotefont[Ligatures=TeX]{Linux Libertine O} % selects Libertine as the quote font
\else
  \usepackage[utf8]{inputenc}
  \usepackage[T1]{fontenc}
  \usepackage{libertine} % or any other font package
  \newcommand*\quotefont{\fontfamily{LinuxLibertineT-LF}} % selects Libertine as the quote font
\fi

\newcommand*\quotesize{60} % if quote size changes, need a way to make shifts relative
\newcommand*{\openquote}
   {\tikz[remember picture,overlay,xshift=-4ex,yshift=-2.5ex]
   \node (OQ) {\quotefont\fontsize{\quotesize}{\quotesize}\selectfont``};\kern0pt}

\newcommand*{\closequote}[1]
  {\tikz[remember picture,overlay,xshift=4ex,yshift={#1}]
   \node (CQ) {\quotefont\fontsize{\quotesize}{\quotesize}\selectfont''};}

\colorlet{shadecolor}{White}

\newcommand*\shadedauthorformat{\emph} % define format for the author argument

\newcommand*\authoralign[1]{%
  \if#1l
    \def\authorfill{}\def\quotefill{\hfill}
  \else
    \if#1r
      \def\authorfill{\hfill}\def\quotefill{}
    \else
      \if#1c
        \gdef\authorfill{\hfill}\def\quotefill{\hfill}
      \else\typeout{Invalid option}
      \fi
    \fi
  \fi}
\newenvironment{shadequote}[2][l]%
{\authoralign{#1}
\ifblank{#2}
   {\def\shadequoteauthor{}\def\yshift{-2ex}\def\quotefill{\hfill}}
   {\def\shadequoteauthor{\par\authorfill\shadedauthorformat{#2}}\def\yshift{2ex}}
\begin{snugshade}\begin{quote}\openquote}
{\shadequoteauthor\quotefill\closequote{\yshift}\end{quote}\end{snugshade}}

%% file: abstract.tex
% Version 2
% - active parts of sentences first
% - readability + flow
% - more of paper represented
% - bringing in Turing's wording

It is non-trivial to design engaging and balanced sets of game rules.
Modern chess has evolved over centuries,
but without a similar recourse to history, 
the consequences of rule changes to game dynamics are difficult to predict.
AlphaZero provides an alternative \emph{in silico} means of game balance assessment.
It is a system 
that can learn near-optimal strategies for any rule set 
from scratch, without any human supervision, by continually learning from its own experience.
In this study we use AlphaZero to creatively explore and design new chess variants.
There is growing interest in chess variants like Fischer Random Chess, because of classical chess's voluminous
opening theory, the high percentage of draws in professional play, and the non-negligible number of games that end while both players are still in their home preparation.
We compare nine other variants that involve atomic changes to the rules of chess.
The changes allow for novel strategic and tactical patterns to emerge, while
keeping the games close to the original.
By learning near-optimal strategies for each variant with AlphaZero,
we determine what games between strong human players
might look like if these variants were adopted.
Qualitatively, several variants are very dynamic.
An analytic comparison show that pieces are valued differently between variants,
and that some variants are more decisive than classical chess.
Our findings demonstrate the rich possibilities that lie beyond the rules of modern chess.

% Version 1

% Designing engaging and balanced sets of game rules is non-trivial, due to difficulties in assessing the consequences of individual changes on game dynamics and appeal.
% AlphaZero is a reinforcement learning system 
% that can learn near-optimal strategies for any rule set 
% from scratch without any human supervision,
% and provides an in silico alternative for game balance assessment.
% In this paper we demonstrate the potential to use AlphaZero
% as a tool for creative exploration and design of new chess variants.
% Given the increasing depth of known chess opening theory, the high percentage of draws in professional play, and the non-negligible number of games that end while both players are still in their home preparation,
% there is increasing interest in chess variants like Fischer Random Chess.
% In this study, we use AlphaZero to explore 9 chess variants that involve atomic changes to the rules of chess.
% The changes keep the game close to the original,
% but allow for novel strategic and tactical patterns to emerge.
% By effectively simulating decades of human play in a matter of hours,
% we are able to determine what the games between strong human players would potentially look like, if these variants were to be adopted.
% In this process, we identified several variants of chess that appear to be very dynamic and interesting. Our findings demonstrate the rich possibilities that lie beyond the modern chess rules.

%% file: introduction.tex
\section{Introduction}

Rule design is a critical part of game development, and small alterations to
game rules can have a large effect on a game's
overall playability and the resulting game dynamics.
Fine-tuning and balancing rule sets in games is often a
laborious and time-consuming process.
Automating the balancing process is an open area of research~\cite{balance_rest_play, ai_board_games},
and machine learning and evolutionary methods have recently been used to help game designers balance 
games more efficiently~\cite{rl_balance, coevol_game_balance, halim14evolutionary, Grau_Moya_2018}. Here we examine the potential of AlphaZero~\cite{Silver1140} to be used as an exploration tool for investigating game balance and game dynamics under different rule sets in board games, taking chess as an example use case.

Popular games often evolve over time and modern-day chess is no exception.
The original game of chess is thought to have been conceived in India in the 6th century,
from where it initially spread to Persia, then the Muslim world and later to Europe and globally. In medieval times, European chess was still largely based on Shatranj, an early variant originating from the Sasanian Empire that was based on the Indian Chatura\.nga~\cite{history_of_chess}.
Notably, the queen and the bishop (alfin) moves were much more restricted,
and the pieces were not as powerful as those in modern chess. Castling did not exist, but the king's leap and the queen's leap existed instead as special first king and queen moves. Apart from checkmate, it was also possible to win by baring the opposite king, leaving the piece isolated with the entirety of its army having been captured.
In Shatranj, stalemate was considered a win,
whereas these days it is considered a draw. The evolution of chess variants over the centuries can be viewed through the lens of changes in search space complexity and the expected final outcome uncertainty throughout the game, the latter being emphasized by modern rules and seen as important for the overall entertainment value~\cite{evol_chess_ref}. Modern chess was introduced in the 15th century, and is one of the most popular games to date, captivating the imagination of players around the world.

The interest in further development of chess has not subsided, especially considering a decreasing number of decisive games in professional chess and an increasing reliance on theory and home preparation with chess engines. This trend, coupled with curiosity and desire to tinker with such an inspiring game, has given rise to many variants of chess that have been proposed over the years~\cite{gollon68variations,variants_encyclopedia,
wikivariants}.
These variants involve alterations to the board, the piece placement, or the rules, to offer players
``something subtle, sparkling, or amusing which cannot be done in ordinary chess''
\cite{beasly98variants}.
Probably the most well-known and popular chess variant is the so-called Chess960
or Fischer Random Chess, where pieces on the first rank are placed in one of 960 random permutations, making theoretical preparation infeasible.

Chess and artificial intelligence are inextricably linked.
\citet{turing53digital} asked,
\emph{``Could one make a machine to play chess, and to improve its play, game by game, profiting from its experience?''}
While computer chess has progressed steadily since the 1950s,
the second part of Alan Turing's question was realised in full only recently.
AlphaZero~\cite{Silver1140} demonstrated state-of-the-art results in playing Go, chess, and shogi.
It achieved its skill without any human supervision by continuously improving its play by learning from self-play games.
In doing so, it showed a unique playing style, later analysed in Game Changer~\cite{game_changer}. This in turn gave rise to new projects like Leela Chess Zero~\cite{leela} and improvements in existing chess engines. CrazyAra~\cite{czech2019learning} employs a related approach for playing the Crazyhouse chess variant, although it involved pre-training from existing human games.
A model-based extension of the original
AlphaZero system
was shown to generalise to domains like Atari, while maintaining its performance on chess even without an exact environment simulator~\cite{schrittwieser2019mastering}. AlphaZero has also shown promise beyond game environments, as a recent application of the model to global optimisation of quantum dynamics suggests~\cite{quantum_az}.

AlphaZero lends itself
naturally to the problem of finding appealing and well-balanced rule sets,
as no prior game knowledge is needed when training AlphaZero on any particular game.
Therefore, we can rapidly explore different rule sets and
characterise the arising style of play through
quantitative and qualitative comparisons.
Here we examine several hypothetical alterations to the rules of chess through the lens of AlphaZero, highlighting variants of the game that could be of potential interest for the chess community. One such variant that we have examined with AlphaZero,
No-castling chess, has been publicly championed by Vladimir Kramnik~\cite{chesscomnocastling}, and has already had its moment in professional play
on 19 December 2019,
when Luke McShane and Gawain Jones played the first-ever grandmaster No-castling match during the London Chess Classic.
This was followed up by the very first No-castling chess tournament in Chennai in January 2020, which resulted in 89\% decisive games~\cite{nocastlingtournament}.

%% file: methods.tex
\section{Methods}

In this section we motivate nine alterations to the modern chess rules,
describe the key components of AlphaZero that are used in the analysis in Section \ref{sec:quantitative}, and outline how AlphaZero was trained for Classical chess and each of the nine variants.

\subsection{Rule Alterations}

There are many ways in which the rules of chess could be altered
and in this work we limit ourselves to considering atomic changes that keep the game as close as possible to classical chess.
In some cases, secondary changes needed to be made to the 50-move rule to avoid 
potentially infinite games. The idea was to try to preserve the symmetry
and the aesthetic appeal of the original game, while hoping to uncover dynamic variants with new opening, middlegame or endgame patterns and a novel body of opening theory. With that in mind, we did not consider any alterations involving changes to the board itself, the number of pieces, or their arrangement. Such changes were outside of the scope of this initial exploration.
Rule alterations that we examine are listed in Table~\ref{tab:variants_listing}.
The variants in Table~\ref{tab:variants_listing} are by no means new to this paper, and many are guised under other names:
Self-capture is sometimes referred to as ``Reform Chess''
or ``Free Capture Chess'',
while Pawn-back is called ``Wren's Game'' by
\citet{variants_encyclopedia}.
None have yet come under intense scrutiny,
and the impact of counting stalemate as a win is a lingering open question in the chess community.

\begin{table*}[t]
\centering
\begin{tabular}{llc}
\toprule
Variant  & Primary rule change & \multicolumn{1}{l}{Secondary rule change} \\
\toprule
No-castling                      & \begin{tabular}[c]{@{}l@{}}Castling is disallowed\\ throughout the game\end{tabular}                                                  & -                                                                                                                      \\
\midrule
No-castling (10)                 & \begin{tabular}[c]{@{}l@{}}Castling is disallowed\\ for the first 10 moves (20 plies)
\end{tabular}                                    & -                                                                                                                      \\
\midrule
Pawn one square                  & Pawns can only move by one square                                                                                                     & -                                                                                                                      \\
\midrule
Stalemate=win                  & \begin{tabular}[c]{@{}l@{}}Forcing stalemate is a win\\ rather than a draw\end{tabular}                                               & -                                                                                                                      \\
\midrule
Torpedo                    & \begin{tabular}[c]{@{}l@{}}Pawns can move by 1 or 2 squares\\anywhere on the board. En passant can \\ consequently happen anywhere on the board.\end{tabular}      & - \\
\midrule
Semi-torpedo & \begin{tabular}[c]{@{}l@{}}Pawns can move by two square\\ both from the 2nd and the 3rd rank\end{tabular}                             & -                                                                                                                      \\
\midrule
Pawn-back                        & \begin{tabular}[c]{@{}l@{}}Pawns can move backwards\\ by one square, but only back to the\\ 2nd/7th rank for White/Black\end{tabular} & \multicolumn{1}{l}{\begin{tabular}[c]{@{}l@{}}Pawn moves do not count\\ towards the 50 move rule\end{tabular}}         \\
\midrule
Pawn-sideways                    & \begin{tabular}[c]{@{}l@{}}Pawns can also move laterally\\ by one square. Captures are \\unchanged, diagonally upwards\end{tabular}      & \multicolumn{1}{l}{\begin{tabular}[c]{@{}l@{}}Sideway pawn moves do not\\ count towards the 50 move rule\end{tabular}} \\
\midrule
Self-capture                     & \begin{tabular}[c]{@{}l@{}}It is possible to capture\\ one's own pieces\end{tabular} & -  \\
\bottomrule
\end{tabular}
\caption{A list of considered alterations to the rules of chess.}
\label{tab:variants_listing}
\end{table*}

Each of the hypothetical rule alterations listed in Table~\ref{tab:variants_listing} could potentially
affect the game either in desired or undesired ways. As an example, consider No-castling chess.
One possible outcome of disallowing castling is that it would result in an aggressive playing style and attacking games, given that the kings are more exposed during the game and it takes time to get them to safety.
Yet, the inability to easily safeguard one's own king might make attacking itself a poor choice, due to the counterattacking opportunities that open up for the defending side.
In Classical chess, players usually castle prior to launching an attack. Therefore, such a change could alternatively be seen as leading to unenterprising play and a much more restrained approach to the game.

Historically, the only way to assess such ideas would have been for a large number of human players to play the game over a long period of time, until enough experience and understanding has been accumulated.
Not only is this a long process, but it also requires
the support of a large number of players to begin with.
With AlphaZero, we can automate this process and simulate
the equivalent of decades of human play
within a day, allowing us to test these hypotheses
\emph{in silico} and observe the emerging patterns and theory for each of the considered variations of the game.

\figref{fig:pos-examples} illustrates each of the variants with an example position.

\begin{figure*}[p]
\centering
\begin{subfigure}[t]{\columnwidth}
\centering\captionsetup{width=.95\columnwidth}
\scalechessboard{\newgame
\fenboard{2rqkb1r/p4pp1/2n1b2p/1ppp4/3P2nP/2N1BNP1/PP1QPPB1/R4K1R w k - 0 12}
\showboard}
\caption{An example from No-castling chess: This is a typical position 
where both kings haven't found immediate safety and remain exposed into the middlegame.}
\label{fig:nocas}
\end{subfigure}
~
\begin{subfigure}[t]{\columnwidth}
\centering\captionsetup{width=.95\columnwidth}
\scalechessboard{\newgame
\fenboard{r3k2r/bpp1qppp/p1npbn2/P3p3/1PB1P3/2PP1N2/2N1QPPP/R1B1K2R b KQkq - 0 1}
\showboard}
\caption{An example from No-castling(10) chess: The play tends to be slower and more strategic, to allow for later castling. Here, on the 11th move, Black castles at the very first opportunity and White castles immediately after as well.}
\label{fig:nocas10}
\end{subfigure}
\\
\phantom{0} % adds a small space between the figures
\begin{subfigure}[t]{\columnwidth}
\centering\captionsetup{width=.95\columnwidth}
\begin{tikzpicture}
\node[anchor=south west,inner sep=0] (board) at (0,0) {
\scalechessboard{\newgame
\fenboard{r2qk2r/pbp2pb1/1p1p2pp/n1n1p3/2P1P3/P2PBNP1/1PQ1NPBP/R3K2R w KQkq - 1 14}
\showboard} };
\coordinate (b2) at (1.06,1.64);
\coordinate (b3-up) at (1.06,2.3);
\coordinate (b4) at (1.06,2.7);
%\draw[dashed,arrows=->,line width=2pt,draw=red!50!white](b2)--(b4);
\draw[arrows=->,line width=2pt,draw=red!40!white](b2)--(b4);
\draw (b3-up) node[cross=6pt,line width=2pt,red] {};
\end{tikzpicture}
\caption{An example from Pawn-one-square chess: Black just moved the knight to a5.
In Classical chess this would seem counter-intuitive due to the potential of playing the pawn to b4, forking the knights.
Here, however, the pawn cannot move to that square in a single move, justifying the manoeuvre.}
\label{fig:qual-pawn-one-square}
\end{subfigure}
~
\begin{subfigure}[t]{\columnwidth}
\centering\captionsetup{width=.95\columnwidth}
\begin{tikzpicture}
\node[anchor=south west,inner sep=0] (board) at (0,0) {
\scalechessboard{\newgame
\fenboard{6N1/8/8/5K1k/8/5N2/8/8 b - - 0 1}
\showboard} };
\coordinate (g4) at (4.22,2.62);
\coordinate (g5) at (4.22,3.25);
\coordinate (g6) at (4.22,3.87);
\coordinate (h4) at (4.87,2.62);
\coordinate (h6) at (4.87,3.87);
\draw (g4) node[cross=5pt,line width=2pt,blue] {};
\draw (g5) node[cross=5pt,line width=2pt,blue] {};
\draw (g6) node[cross=5pt,line width=2pt,blue] {};
\draw (h4) node[cross=5pt,line width=2pt,blue] {};
\draw (h6) node[cross=5pt,line width=2pt,blue] {};
\end{tikzpicture}
\caption{An example from Stalemate=win chess: An endgame position that would have been a draw in Classical chess is now a win instead.}
\label{fig:qual-stalemate-win}
\end{subfigure}
\caption{
Examples of new strategic and tactical themes that arise in the explored chess variants. \figref{fig:qual-torpedo} continues on the following page.
}
\end{figure*}
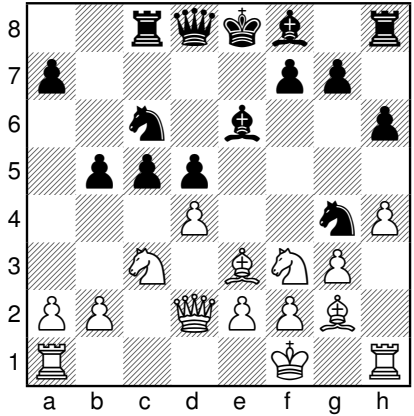
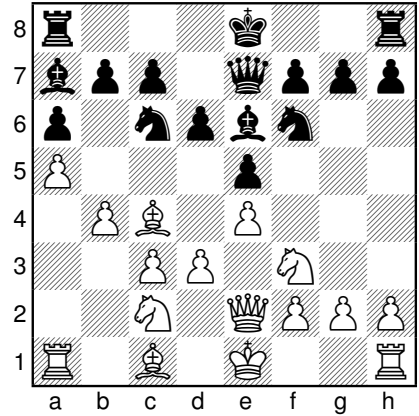
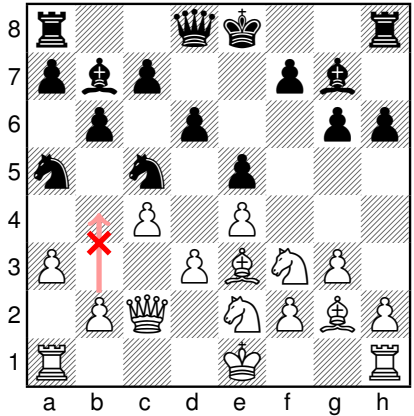
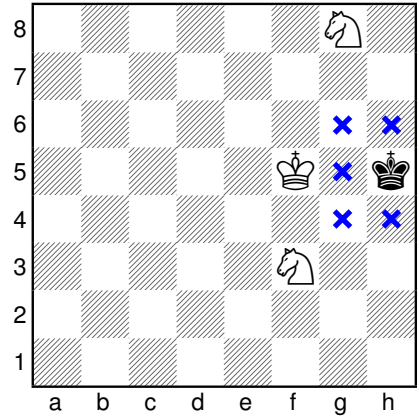

\addtocounter{figure}{-1}

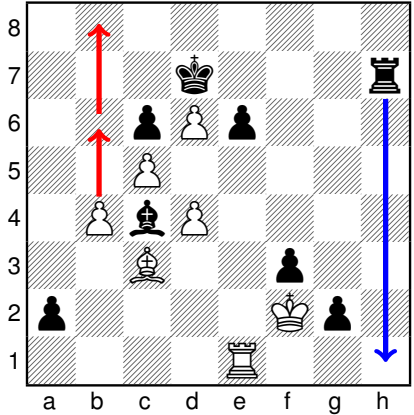
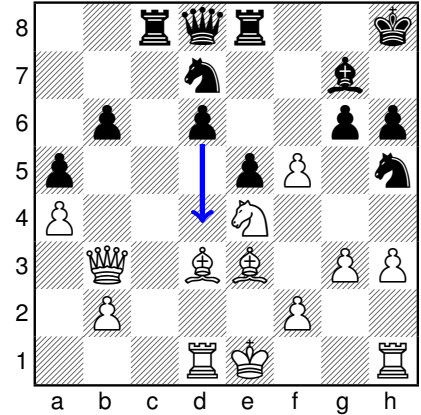
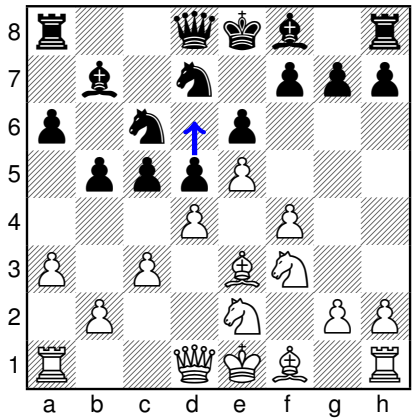
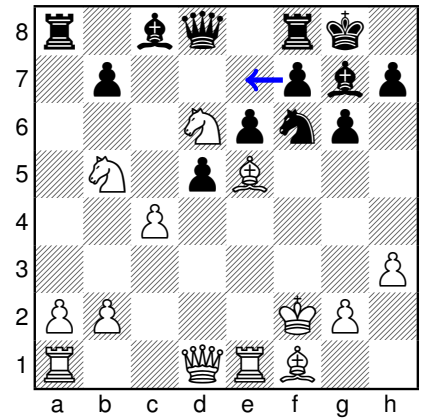
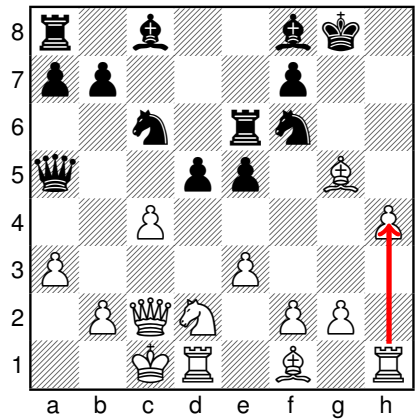
\begin{figure*}[p]
\centering
\begin{subfigure}[t]{\columnwidth}
\addtocounter{subfigure}{4} % works here
\centering\captionsetup{width=.95\columnwidth}
\begin{tikzpicture}
\node[anchor=south west,inner sep=0] (board) at (0,0) {
\scalechessboard{\newgame
\fenboard{8/3k3r/2pPp3/2P5/1PbP4/2B2p2/p4Kp1/4R3 w - - 0 1}
\showboard} };
\coordinate (b4) at (1.06,2.9);
\coordinate (b6-1) at (1.06,3.8);
\coordinate (b6-2) at (1.06,4.0);
\coordinate (b8) at (1.06,5.2);
\coordinate (h1) at (4.87,0.7);
\coordinate (h7) at (4.87,4.2);
\draw[arrows=->,line width=2pt,draw=red](b4)--(b6-1);
\draw[arrows=->,line width=2pt,draw=blue](h7)--(h1);
\draw[arrows=->,line width=2pt,draw=red](b6-2)--(b8);
\end{tikzpicture}
\caption{An example from Torpedo chess: White needs to generate rapid counterplay, and does so with a torpedo move: b4-b6. Black responds with Rh1, to which White promotes to a queen with yet another torpedo move, b6-b8=Q.}
\label{fig:qual-torpedo}
\end{subfigure}
~
\begin{subfigure}[t]{\columnwidth}
\centering\captionsetup{width=.95\columnwidth}
\begin{tikzpicture}
\node[anchor=south west,inner sep=0] (board) at (0,0) {
\scalechessboard{\newgame
\fenboard{2rqr2k/3n2b1/1p1p2pp/p3pP1n/P3N3/1Q1BB1PP/1P3P2/3RK2R w K - 0 1}
\showboard} };
\coordinate (d4) at (2.33,2.55);
\coordinate (d6) at (2.33,3.6);
\draw[arrows=->,line width=2pt,draw=blue](d6)--(d4);
\end{tikzpicture}
\caption{An example from Semi-torpedo chess: The ability to rapidly advance pawns from the 3rd/6th rank enables Black the following energetic option: d6-d4, resulting in a forced tactical sequence.
See Game AZ-19 in Appendix~\ref{sec-semitorpedo} for details.}
\label{fig:qual-semitorpedo}
\end{subfigure}
\\
\phantom{0} % adds a small space between the figures
\begin{subfigure}[t]{\columnwidth}
\centering\captionsetup{width=.95\columnwidth}
\begin{tikzpicture}
\node[anchor=south west,inner sep=0] (board) at (0,0) {
\scalechessboard{\newgame
\fenboard{r2qkb1r/1b1n1ppp/p1n1p3/1pppP3/3P1P2/P1P1BN2/1P2N1PP/R2QKB1R b KQkq - 0 1}
\showboard} };
\coordinate (d5) at (2.33,3.5);
\coordinate (d6) at (2.33,3.95);
\draw[arrows=->,line width=2pt,draw=blue](d5)--(d6);
\end{tikzpicture}
\caption{An example from Pawn-back chess: Here, Black uses this possibility to challenge White's central pawns, while opening up the diagonal for the b7 bishop, by a pawn-back move d5-d6.}
\label{fig:qual-pawn-back}
\end{subfigure}
~
\begin{subfigure}[t]{\columnwidth}
\centering\captionsetup{width=.95\columnwidth}
\begin{tikzpicture}
\node[anchor=south west,inner sep=0] (board) at (0,0) {
\scalechessboard{\newgame
\fenboard{r1bq1rk1/1p3pbp/3Npnp1/1N1pB3/2P5/7P/PP3KP1/R2QRB2 w q - 0 1}
\showboard} };
\coordinate (f7) at (3.4,4.5);
\coordinate (e7) at (2.9,4.5);
\draw[arrows=->,line width=2pt,draw=blue](f7)--(e7);
\end{tikzpicture}
\caption{An example from Pawn-sideways chess: After sacrificing the knight on f2 the previous move, Black utilises a sideways pawn move
f7-e7 for tactical purposes, opening the f-file towards the White king, while attacking the knight on d6.}
\label{fig:qual-pawn-sideways}
\end{subfigure}
\\
\phantom{0} % adds a small space between the figures
\begin{subfigure}[t]{\columnwidth}
\centering\captionsetup{width=.95\columnwidth}
\begin{tikzpicture}
\node[anchor=south west,inner sep=0] (board) at (0,0) {
\scalechessboard{\newgame
\fenboard{r1b2bk1/pp3p2/2n1rn2/q2pp1B1/2P4P/P3P3/1PQN1PP1/2KR1B1R w Kq - 0 1}
\showboard} };
\coordinate (h1) at (4.87,1.0);
\coordinate (h4) at (4.87,2.6);
\draw[arrows=->,line width=2pt,draw=red](h1)--(h4);
\end{tikzpicture}
\caption{An example from Self-capture chess: a self-capture move Rxh4 generates threats against the Black king.}
\label{fig:qual-self-capture}
\end{subfigure}%
\caption{\emph{(Continued from previous page.)}
Examples of new strategic and tactical themes that arise in the explored chess variants.
}
\label{fig:pos-examples}
\end{figure*}

\subsection{Key components of AlphaZero}

AlphaZero is an adaptive learning system that improves through many rounds of self-play \cite{Silver1140}.
It consists of a deep neural network $f_{\theta}$ with weights $\theta$ that compute
\begin{equation} \label{eq:az-neuralnet}
(\p, v) = f_{\theta}(s)    
\end{equation}
for a given position or state $s$.
The network outputs a vector of move probabilities $\p$
with elements $p(s' | s)$ as prior probabilities for considering each move and hence each next state $s'$.\footnote{We've
suppressed notation somewhat; the probabilities are technically over  actions or moves $a$ in state $s$, but as each action $a$ deterministically leads to a separate next position $s'$, we use the concise $p(s' | s)$ in this paper.}
If we denote game outcome numerically by $+1$, for a win,
0 for a draw and $-1$ for a loss,
the network additionally outputs a scalar value $v \in (-1, 1)$ which
estimates the expected outcome of the game from position $s$.

The two predictions in \eqref{eq:az-neuralnet} are used in Monte Carlo tree search (MCTS) to refine the assessment of a board position.
The prior network $\p$ assigns weights to candidate moves at a ``first glance'' of the board,
yielding an order in which moves are searched with MCTS.
The output $v$ can be viewed as a neural network evaluation function for position $s$.
The statistical estimates of the game outcomes after each move are refined through MCTS, which runs repeated simulations of how the game might unfold up to a certain ply depth.
In each MCTS simulation, $f_{\theta}$ is recursively applied to a sequence of positions (or nodes) up to a certain ply depth if they have not been processed in an earlier simulation.
At maximum ply depth, the position is evaluated
with \eqref{eq:az-neuralnet}, and that evaluation is ``backed up'' to the root,
for each node adjusting its ``action selection rule'' to alter
which moves will be selected and expanded in the next MCTS simulation.
After a number of such MCTS simulations, the root move that was visited (or expanded) most is played.

\subsection{Training and evaluation}

We trained AlphaZero from scratch for each of the rule alterations in Table \ref{tab:variants_listing},
with the same set of model hyperparameters. The models were trained for 1 million training steps, with a batch size of 4096 and allowing for an average 0.12 samples per position from self-play games. In order to encourage exploration during training, a small amount of noise was injected in the prior move probabilities~\eqref{eq:az-neuralnet} before search, sampled from a Dirichlet $\mathrm{Dir}(0.3)$ distribution, followed by a renormalization step~\cite{Silver1140}.
Further diversity was promoted by stochastic move selection in the first 30 plies of each of the training self-play games, by selecting the final moves proportionally to the softmax of the MCTS visit counts. The remaining game moves from ply 31 onwards were selected as top moves based on MCTS. Training self-play games were generated using 800 MCTS simulations per move.

The absence of baselines makes it hard to formally assess the strength of each model, which is why it was important to couple the quantitative analysis and metrics observed at training and test time with a qualitative assessment in collaboration with Vladimir Kramnik,
a renowned chess grandmaster and former world chess champion.
As the rule changes that are considered in this study are mostly minor in practical terms,
it is reasonable to assume that the trained models are of similar strength,
although it is equally reasonable to expect that some of them could be further fine-tuned to account for the differences in game length and the average number of legal moves that need to be considered at each position.
Given the nature of the study, the high level of observed play in trained models, and the number of rule alterations considered, we decided not to pursue such a potentially laborious process, as it would not alter any of the high-level conclusions that we present and discuss.

%% file: results-quantitative.tex
% !TeX root = main.tex

\section{Quantitative assessment} \label{sec:quantitative}

There are marked differences between the styles of chess that arises from each of the rule alterations
Aesthetically, each variant has its own appeal, and
we highlight them further in Section \ref{sec:qualitative}.
Here we provide a quantitative comparison between variants, to complement the qualitative observations.
Using a large quantity of self-play games, we infer the expected draw rate and first-move advantage for each variant, expressed as the expected score for White (Section \ref{sec:expected-scores-and-draws}).
We then illustrate how the same opening can lead to vastly different outcomes under different chess variants in Section
\ref{sec:opening-comparison}, and that these opening-specific differences can differ from the aggregate differences across all openings.
An analysis of the utilisation of the newly introduced options made possible by the new rule alterations in Section \ref{sec:special-moves} shows that the non-classical moves are used in a large percentage of games, often multiple times per game, in each of the variants. This suggests that the new options are indeed useful, and contribute to the game. We estimate the diversity of opening play by looking at the opening trees which we construct from AlphaZero's network priors \eqref{eq:az-neuralnet} for the first couple of moves and show that the breadth of opening possibilities in each of these chess variants seems to be inversely related to their relative decisiveness (Section~\ref{sec:diversity}). Sections \ref{sec:opening-tree-kl} and  \ref{sec:additional-moves} highlight the difference in opening play according to the prior distributions of the variants. Rule adjustments, especially those affecting piece mobility, are also expected to affect the relative material value of the pieces. Finally, Section~\ref{sec:piece-value} provides approximations for piece values in each of the variants, computed from a sample of 10,000 fast-play AlphaZero games.

%%%%%%%%%%%%%%%%%%%%%%%%%%%%

\subsection{Self-play games}
\label{sec:self-play-games}

\begin{figure*}[h]
\centering
\begin{subfigure}[t]{\columnwidth}
\centering\captionsetup{width=.95\columnwidth}
\includegraphics[width=\columnwidth]{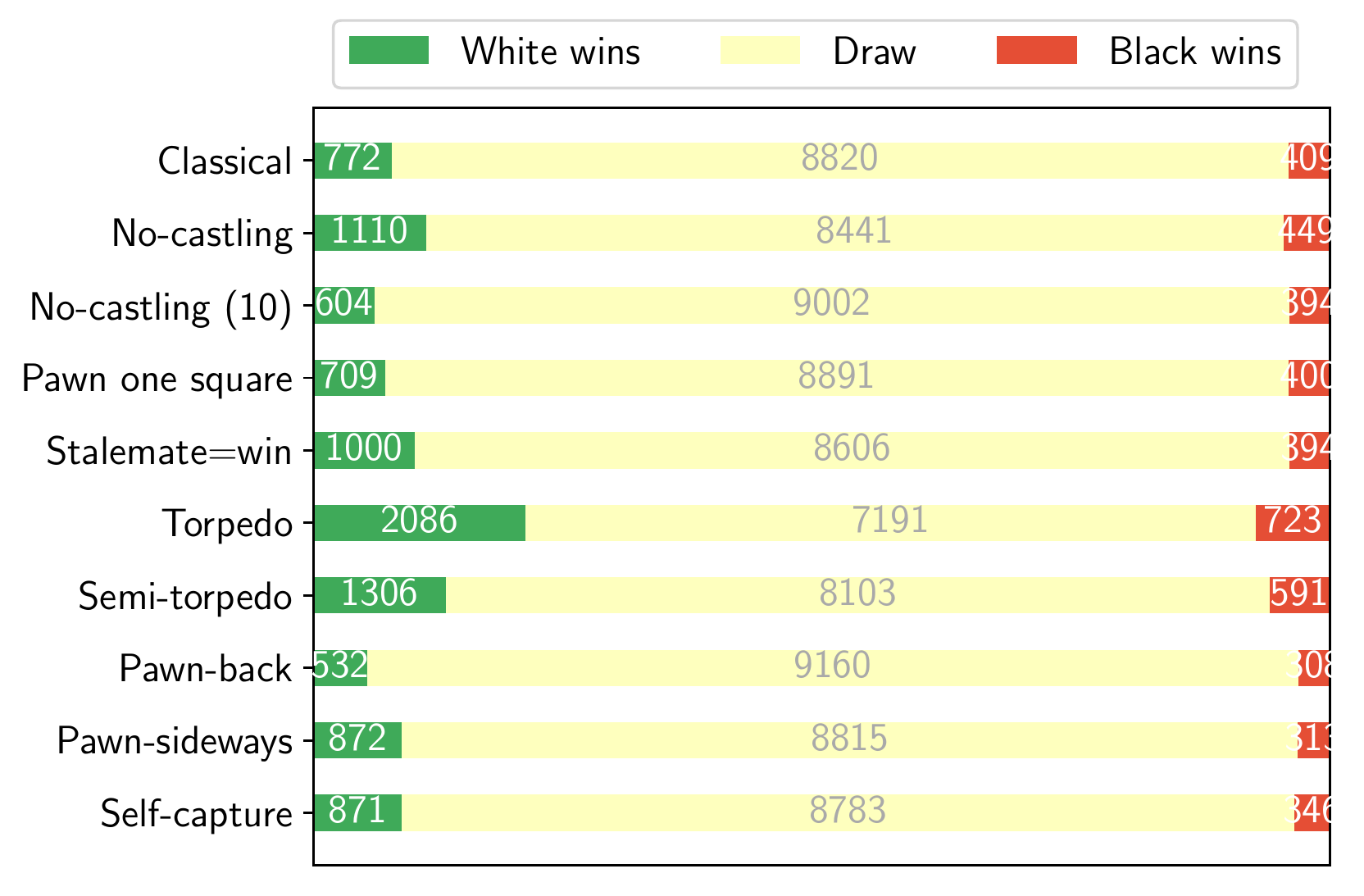}
\caption{The game outcomes of 10,000 AlphaZero games played at 1 second per move for each different chess variant.
}
\label{fig:win-draw-lose-1s}
\end{subfigure}%
~ 
\begin{subfigure}[t]{\columnwidth}
\centering\captionsetup{width=.95\columnwidth}
\includegraphics[width=\columnwidth]{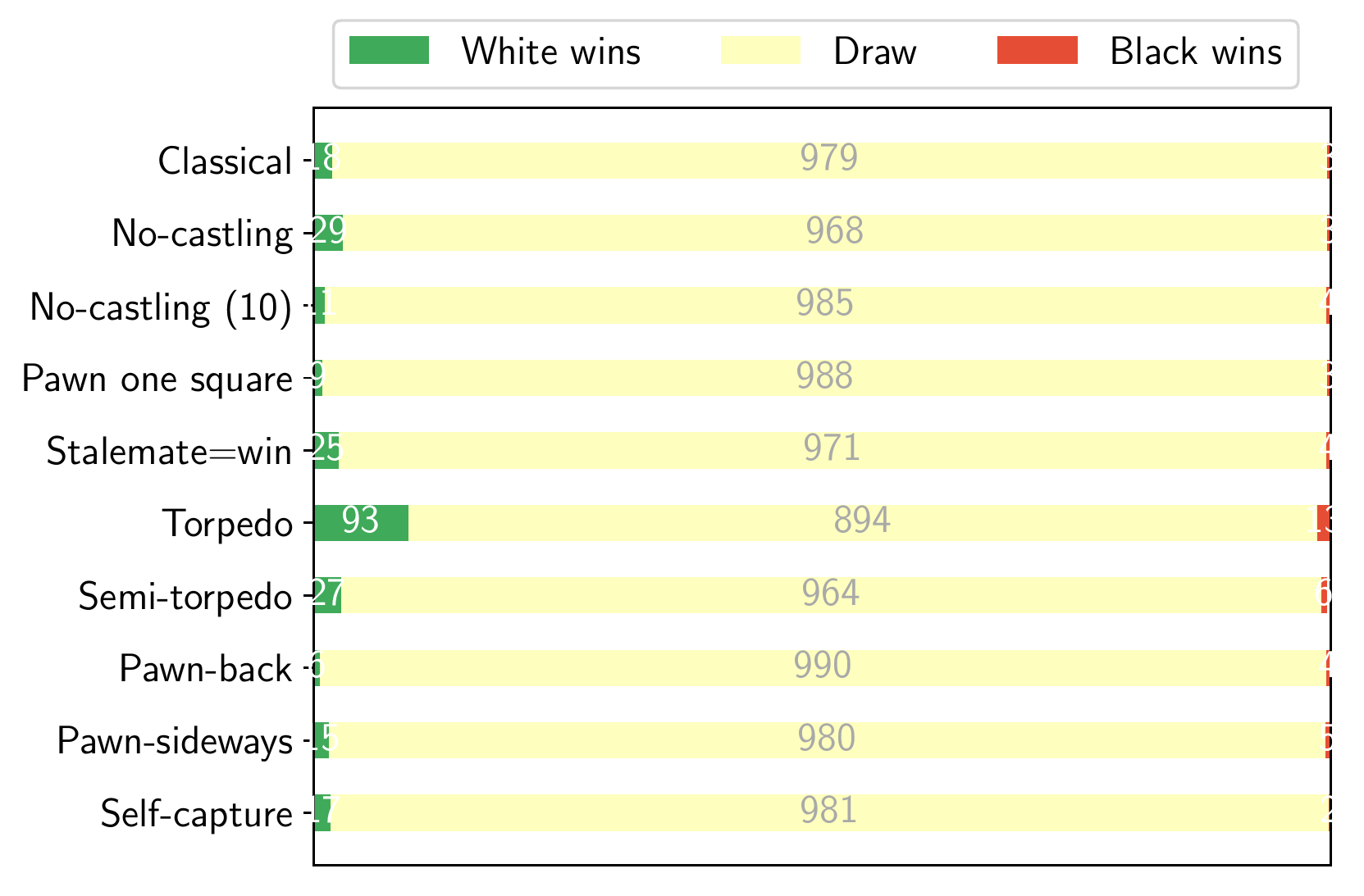}
\caption{The game outcomes of 1,000 AlphaZero games played at 1 minute per move for each different chess variant.}
\label{fig:win-draw-lose-100s}
\end{subfigure}
\caption{AlphaZero self-play game outcomes under different time controls. As moves are determined in a deterministic fashion given the same conditions, diversity was enforced by sampling the first 20 plies in each game proportional to their MCTS visit counts.
Across all variations the percentage of drawn games increases with longer thinking times.
This seems to suggest that the starting position might be theoretically drawn in these chess variants, like in Classical chess, and that some of the variants are simply harder to play, involving more calculation and richer patterns.
}
\label{fig:win-draw-lose}
\end{figure*}

For each chess variant, we generated a diverse set of
$N = 10,\!000$ AlphaZero self-play  games at 1 second
per move, and $N = 1,\!000$ games at 1 minute per move.
The outcomes of the fast self-play games are presented in \figref{fig:win-draw-lose-1s}; the longer games follow in
\figref{fig:win-draw-lose-100s}.
As AlphaZero is approximately deterministic given the same MCTS depth and number of rollouts, we promote diversity in games by sampling the first 20 plies in each game proportional to the softmax of the MCTS visit counts, followed by playing the top moves for the rest of the game.

In addition to that, we generated a set of $N = 1,\!000$ fast-play games from fixed starting positions arising from the Dutch Defence, Chigorin Defence, Alekhine Defence and King's Gambit for each of the variants, as further discussed in Section~\ref{sec:opening-comparison}.

The two sets of diverse self-play games are used
in Section \ref{sec:expected-scores-and-draws} to compare the decisiveness of each variant,
in Section \ref{sec:special-moves} to analyse how many special moves are used, and in
Section \ref{sec:piece-value} to estimate piece values across variants.

A selection of these games is presented in Appendix \ref{sec:chess_appendix}.

%%%%%%%%%%%%%%%%%%%%%%%%%%%%%%%%%%%%%%%%%%%
\subsection{Expected scores and draw rates}
\label{sec:expected-scores-and-draws}

It is widely hypothesised that classical chess is theoretically drawn; that the odds
$\pi =
(\pi_{\mathrm{win}}, \pi_{\mathrm{draw}}, \pi_{\mathrm{lose}})$ of white winning, drawing and losing are $(0, 1, 0)$ at optimal play.
We determine how favourable for white or how
``drawish'' different variants are
by estimating the expected scores and draw rates
at non-optimal play under the \emph{same} conditions.
We keep the conditions that
chess variants are played against themselves with AlphaZero fixed,
like the move selection criteria or Monte Carlo Tree Search (MCTS) evaluation time.

The overall decisiveness in the generated game sets depends on the time controls involved.
We see in
Figures \ref{fig:win-draw-lose-1s} and \ref{fig:win-draw-lose-100s}
that across all variations the percentage of drawn games increases with longer thinking times, and longer thinking times also affect the expected score for White, as shown in Table~\ref{tab:exp_score}.
This suggests that the starting position might be theoretically drawn in these chess variants, like in 
Classical chess, and that some of the variants are simply harder to play, involving more calculation and richer patterns.
We hypothesise that the relative differences in AlphaZero's win rates
might translate to differences in human play, although this hypothesis would need to be practically validated in the future.
Yet, in absence of any existing human games, we can use these results as a preliminary guess of what those results might be,
assuming that what is difficult to calculate for AlphaZero may be difficult for human players as well.

\begin{table}[t]
\centering
\begin{tabular}{llll}
\toprule
Variant & Training %self-play
& 1sec & 1min \\
\toprule
Classical        & $54.1\%$ & $51.8\%$ & $50.8\%$ \\
No castling	     & $55.7\%$ & $53.3\%$ & $51.3\%$ \\
No castling (10) & $52.5\%$ & $51.0\%$	& $50.4\%$ \\
Pawn one square  & $53.5\%$ & $51.6\%$ & $50.3\%$ \\
Stalemate=win    & $54.9\%$ & $53.0\%$ & $51.1\%$ \\
Torpedo          & \textbf{$57.0\%$} & \textbf{$56.8\%$} & \textbf{$54.0\%$} \\
Semi-torpedo     & $54.7\%$ & $53.6\%$ & $50.9\%$ \\
Pawn-back        & $53.0\%$ & $51.1\%$ & $50.1\%$ \\
Pawn-sideways    & $54.8\%$ & $52.8\%$ & $50.5\%$ \\
Self-capture     & $54.2\%$ & $52.6\%$ & $50.8\%$ \\
\bottomrule
\end{tabular}
\caption{Empirical score for White under different game conditions, for each chess variant:
self-play games at the end of model training, 1 second per move games, and 1 minute per move games. Diversity in 1 second per move games and 1 minute per move games was enforced by sampling the first 20 plies in each game proportional to their MCTS visit counts.
}
\label{tab:exp_score}
\end{table}

\subsubsection{Inference for game odds}

To compare variants, we first infer the odds of their outcomes under set playing conditions.
For a given variant, let the game outcomes $\Gcal$ be $n_{\mathrm{win}}$ wins and $n_{\mathrm{lose}}$ losses for white, and
$n_{\mathrm{draw}} = N - n_{\mathrm{win}} - n_{\mathrm{lose}}$ draws.
If we assume a uniform Dirichlet prior on $\pi$ and
multinomial likelihood for winning, drawing or losing,
the posterior distribution is Dirichlet,
\begin{equation} \label{eq:posterior}
p(\pi | \Gcal) = \mathrm{Dir}(n_{\mathrm{win}} + 1, n_{\mathrm{draw}} + 1, n_{\mathrm{lose}} + 1) \ .
\end{equation}
    
\subsubsection{Draw rates}
To compare the decisiveness of chess variants, we infer the probability that variant A has a lower draw rate than variant B, given the games played $\Gcal^{\mathrm{A}}$ and $\Gcal^{\mathrm{B}}$ under the same 
conditions:\footnote{This approach follows \citet[Chapter~37.1]{mackay2003information}.}
\begin{align}
& p(\pi_{\mathrm{draw}}^{\mathrm{A}} < \pi_{\mathrm{draw}}^{\mathrm{B}} ) = \nonumber \\
& \iint 
\Ibb \left[\pi_{\mathrm{draw}}^{\mathrm{A}} < \pi_{\mathrm{draw}}^{\mathrm{B}} \right] \,
p(\pi^{\mathrm{A}} | \Gcal^{\mathrm{A}} ) \,
p(\pi^{\mathrm{B}} | \Gcal^{\mathrm{B}} ) \,
\drm \pi^{\mathrm{A}} \, \drm \pi^{\mathrm{B}} \ .
\label{eq:draw-rates}
\end{align}
The integral is not available in closed form;
we evaluate it with a Monte Carlo estimate
by drawing pairs of samples from
$p(\pi^{\mathrm{A}} | \Gcal^{\mathrm{A}} )$ and
$p(\pi^{\mathrm{B}} | \Gcal^{\mathrm{B}} )$ -- using \eqref{eq:posterior} --
and computing the fraction of times that samples satisfy $\pi_{\mathrm{draw}}^{\mathrm{A}} < \pi_{\mathrm{draw}}^{\mathrm{B}}$.

\begin{figure*}[t]
\centering
\begin{subfigure}[t]{\columnwidth}
\centering\captionsetup{width=.95\columnwidth}
\includegraphics[width=\columnwidth]{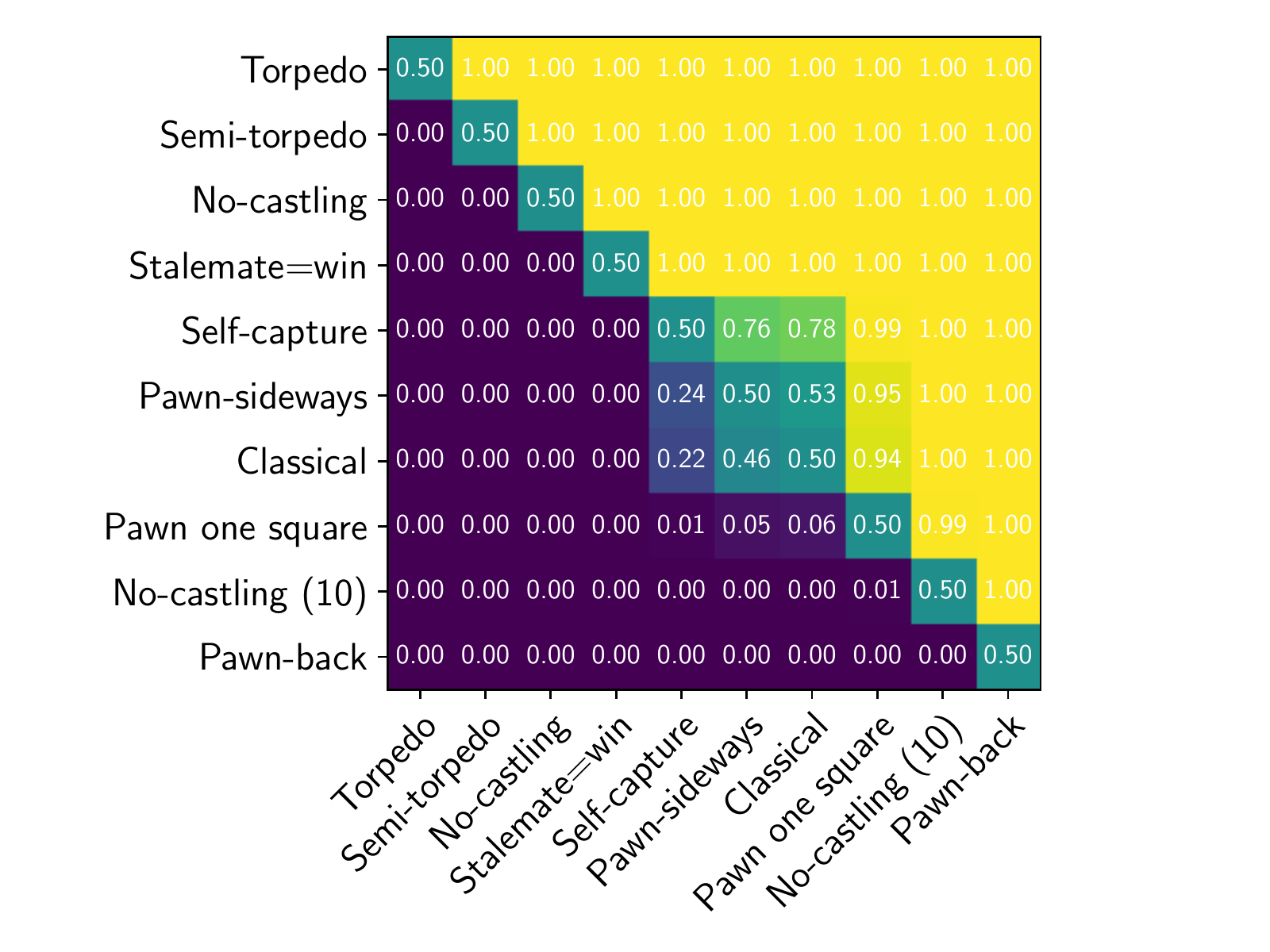}
\caption{A draw rate comparison
$p (\pi_{\mathrm{draw}}^{\mathrm{row}} < \pi_{\mathrm{draw}}^{\mathrm{column}} )$
at approximately 1 seconds per move, on 10,000 AlphaZero games per variation.
}
\label{fig:draw-rate-1s}
\end{subfigure}%
~ 
\begin{subfigure}[t]{\columnwidth}
\centering\captionsetup{width=.95\columnwidth}
\includegraphics[width=\columnwidth]{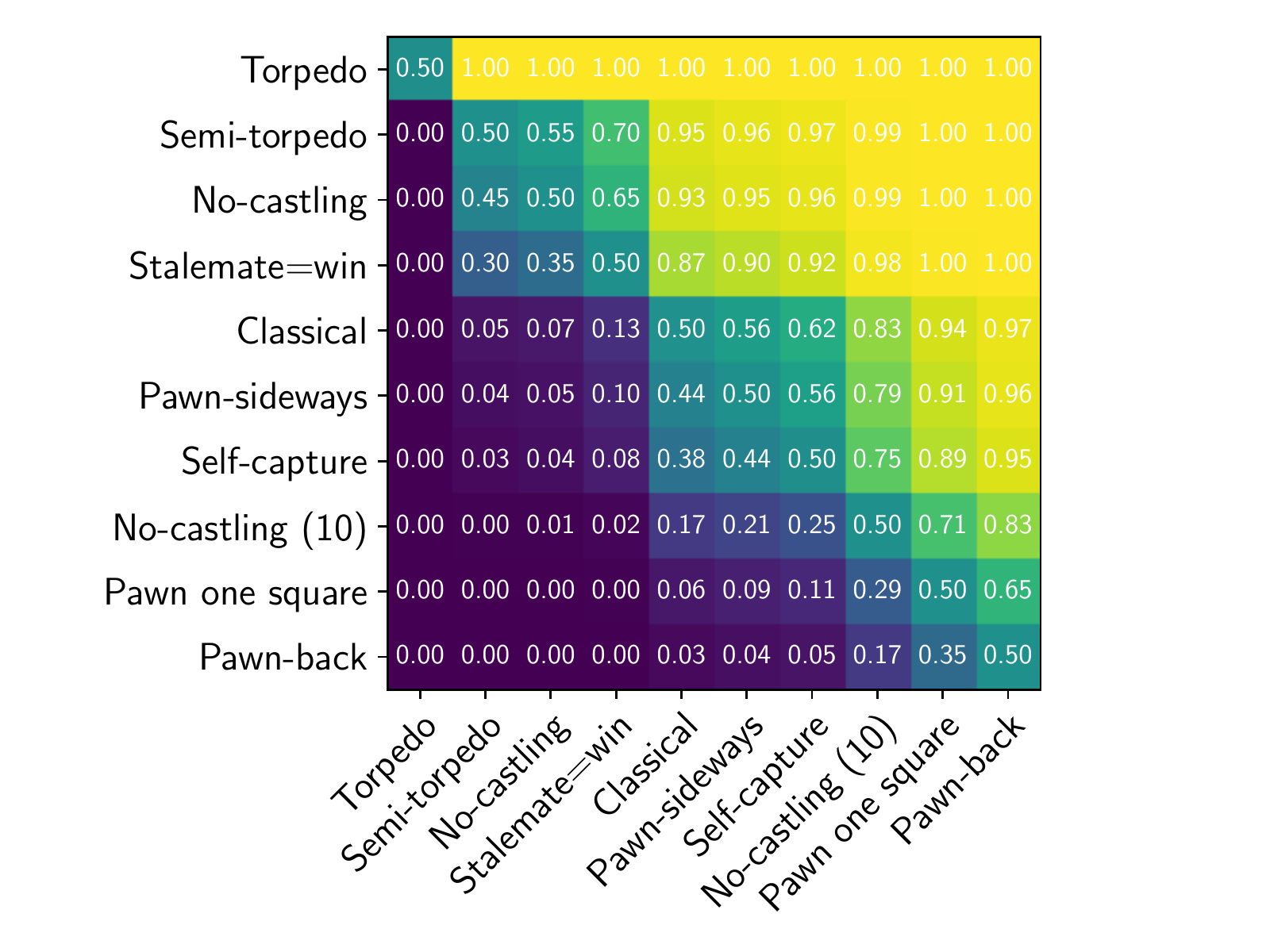}
\caption{A draw rate comparison
$p (\pi_{\mathrm{draw}}^{\mathrm{row}} < \pi_{\mathrm{draw}}^{\mathrm{column}} )$
at approximately 1 minute per move, on 1,000 AlphaZero games per variation.}
\label{fig:draw-rate-100s}
\end{subfigure}
\\
\begin{subfigure}[t]{\columnwidth}
\centering\captionsetup{width=.95\columnwidth}
\includegraphics[width=\columnwidth]{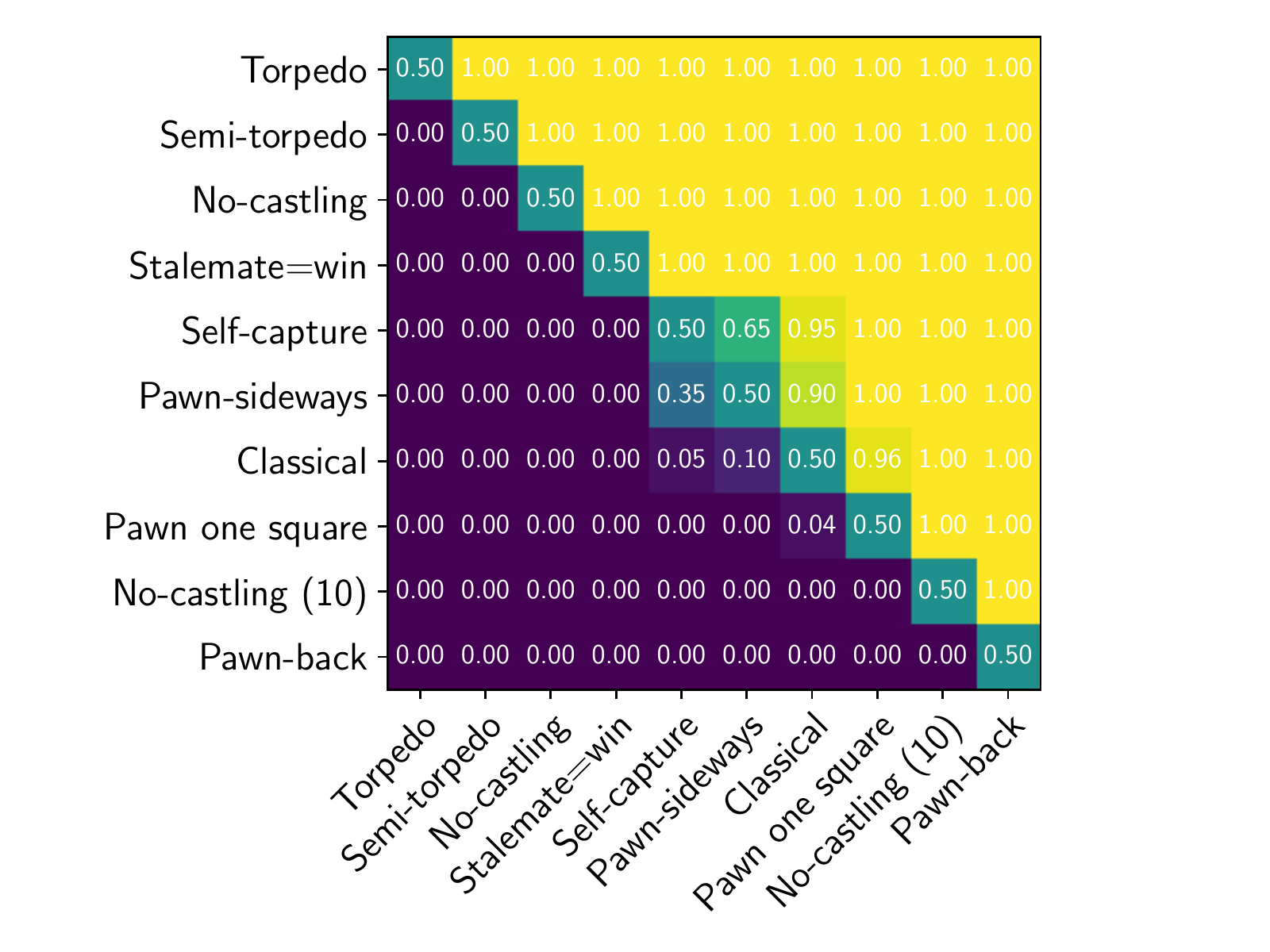}
\caption{A comparison of expected scores
$p(e^{\mathrm{row}} > e^{\mathrm{column}})$
at 1 second per move, on 10,000 games per variation.
}
\label{fig:expected-scores-1s}
\end{subfigure}%
~ 
\begin{subfigure}[t]{\columnwidth}
\centering\captionsetup{width=.95\columnwidth}
\includegraphics[width=\columnwidth]{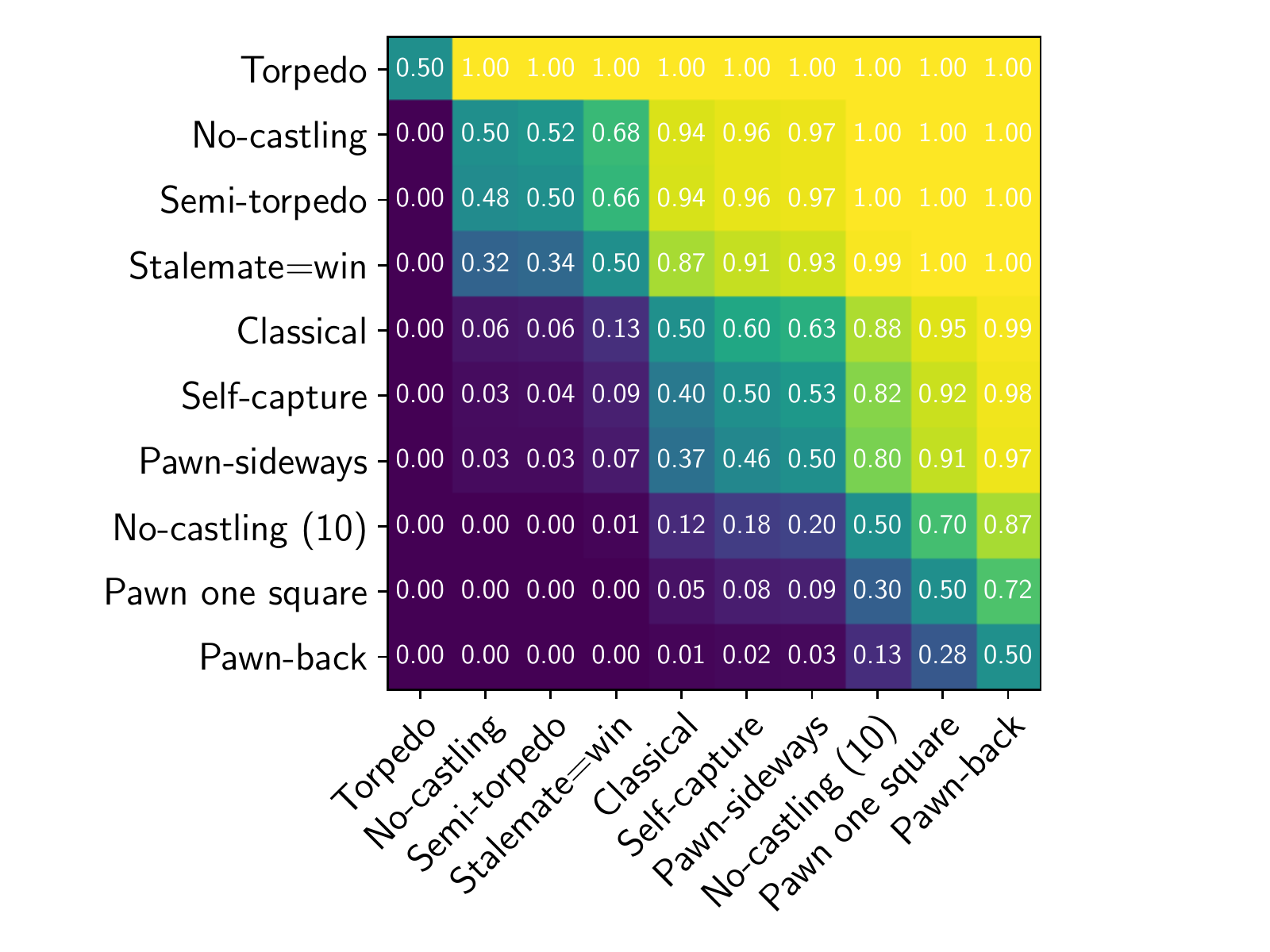}
\caption{A comparison of expected scores
$p(e^{\mathrm{row}} > e^{\mathrm{column}})$
at 1 minute per move, on 1,000 games per variation.}
\label{fig:expected-scores-100s}
\end{subfigure}
\caption{
A comparison of draw rates.
The most decisive chess variants under both time controls are Torpedo, Semi-torpedo, No-castling and Stalemate=win.
These four variants also give White the largest first-move advantage.
}
\label{fig:draw-rate-expected-score}
\end{figure*}

\figref{fig:draw-rate-1s} provides an indication of
the relative \emph{decisiveness} of variants, when played by AlphaZero at approximately 1 second
per move, and
\figref{fig:draw-rate-100s} provides the comparison at 1 minute per move. 
Under both time controls,
the most decisive chess variants we explored are
Torpedo, Semi-torpedo, No-castling and Stalemate=win. Torpedo and Semi-torpedo have increased pawn mobility, allowing for faster, more dynamic play, leading to more decisive outcomes. There are also more moves to consider at each juncture.
No-castling chess makes it harder to evacuate the king to safety, similarly affecting the draw rate. Finally, Stalemate=win removes one important drawing resource for the weaker side, converting a number of important endgame positions from being drawn to being winning for the stronger side. Under the same conditions of play, the slower Pawn one square chess variant
and Pawn-back chess variant are the most drawish.
Pawn-back chess incorporates additional defensive resources, and the ability to go back to protect the weak squares seems to be more important for defending worse positions than it is for attacking -- given that attacking tends to involve moving forward on the board.
\matthias{not sure if this last statement is not a fallacy?
see Kramnik's comment on ``allowing more aggressive play''}

\subsubsection{Expected scores}

The decisiveness of a chess variant under imperfect play does not necessarily have to correspond to the first-move advantage.
In classical chess, White scores higher on average.
Top-level chess players tend to press for an advantage with the White pieces and defend with the Black pieces, looking for opportunities to counter-attack.
The reason is the first-move advantage; it is an initiative that, with good play, persists throughout the opening phase of the game.
This not a universal property that would hold in any game
\matthias{ambiguous, ``under any set of rules''?},
as playing the first move might also disadvantage a player in some types of games. It is therefore important to estimate the effect of the rule changes on the first-move advantage in each chess variant, expressed as the expected score for White.

The expected score for White is defined as:
\begin{equation}
e = \pi_{\mathrm{win}} + \tfrac{1}{2} \pi_{\mathrm{draw}}
\end{equation}
for a particular set of conditions like time controls, the move selection criteria and the AlphaZero model playing the game.
Given the game outcomes $\Gcal^{\mathrm{A}}$ and $\Gcal^{\mathrm{B}}$ of variants A and B, the probability of white having a higher first-move advantage in variant A is
\begin{align}
p (e^{\mathrm{A}} > e^{\mathrm{B}} ) & = % \nonumber \\
% & \qquad
\iint 
\Ibb \left[
\pi_{\mathrm{win}}^{\mathrm{A}} + \tfrac{1}{2} \pi_{\mathrm{draw}}^{\mathrm{A}} > \pi_{\mathrm{win}}^{\mathrm{B}} + \tfrac{1}{2}
\pi_{\mathrm{draw}}^{\mathrm{B}}
\right] \nonumber \\
& \qquad\phantom{\iint}
p(\pi^{\mathrm{A}} | \Gcal^{\mathrm{A}} ) \,
p(\pi^{\mathrm{B}} | \Gcal^{\mathrm{B}} ) \,
\drm \pi^{\mathrm{A}} \, \drm \pi^{\mathrm{B}} \ ,
\end{align}
which we again evaluate with a Monte Carlo estimate.

White's first-move advantage
with approximately 1 second and 1 minute per move in AlphaZero games is
compared in Figures \ref{fig:expected-scores-1s} and \ref{fig:expected-scores-100s} respectively.
The relative ordering of variations follows the ranking in general decisiveness, suggesting that the new chess variants that are more decisive in AlphaZero games are also more advantageous for White,
possibly due to an increase in dynamic attacking options.

%%%%%%%%%%%%%%%%%%%%%%%%%%%%%%%%%%%%%%%%%%%%%%%%%%%
\subsection{Differences in specific openings}
\label{sec:opening-comparison}

\begin{figure*}[t]
\centering
\begin{subfigure}[t]{\columnwidth}
\centering\captionsetup{width=.95\columnwidth}
\includegraphics[width=\columnwidth]{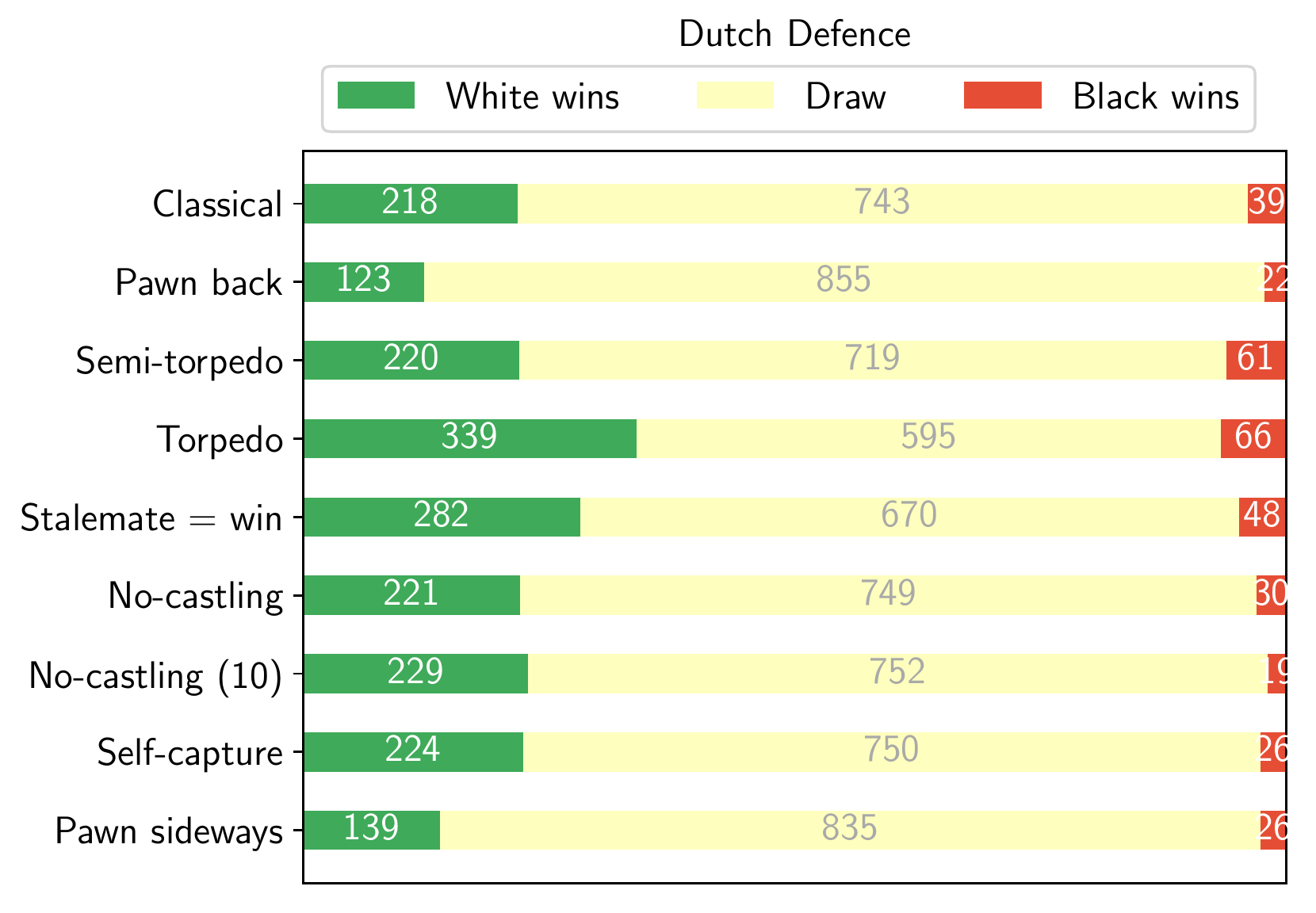}
\caption{Dutch Defence (\mmove{1}d4 f5)}
\label{fig:openings-dutch}
\end{subfigure}%
~ 
\begin{subfigure}[t]{\columnwidth}
\centering\captionsetup{width=.95\columnwidth}
\includegraphics[width=\columnwidth]{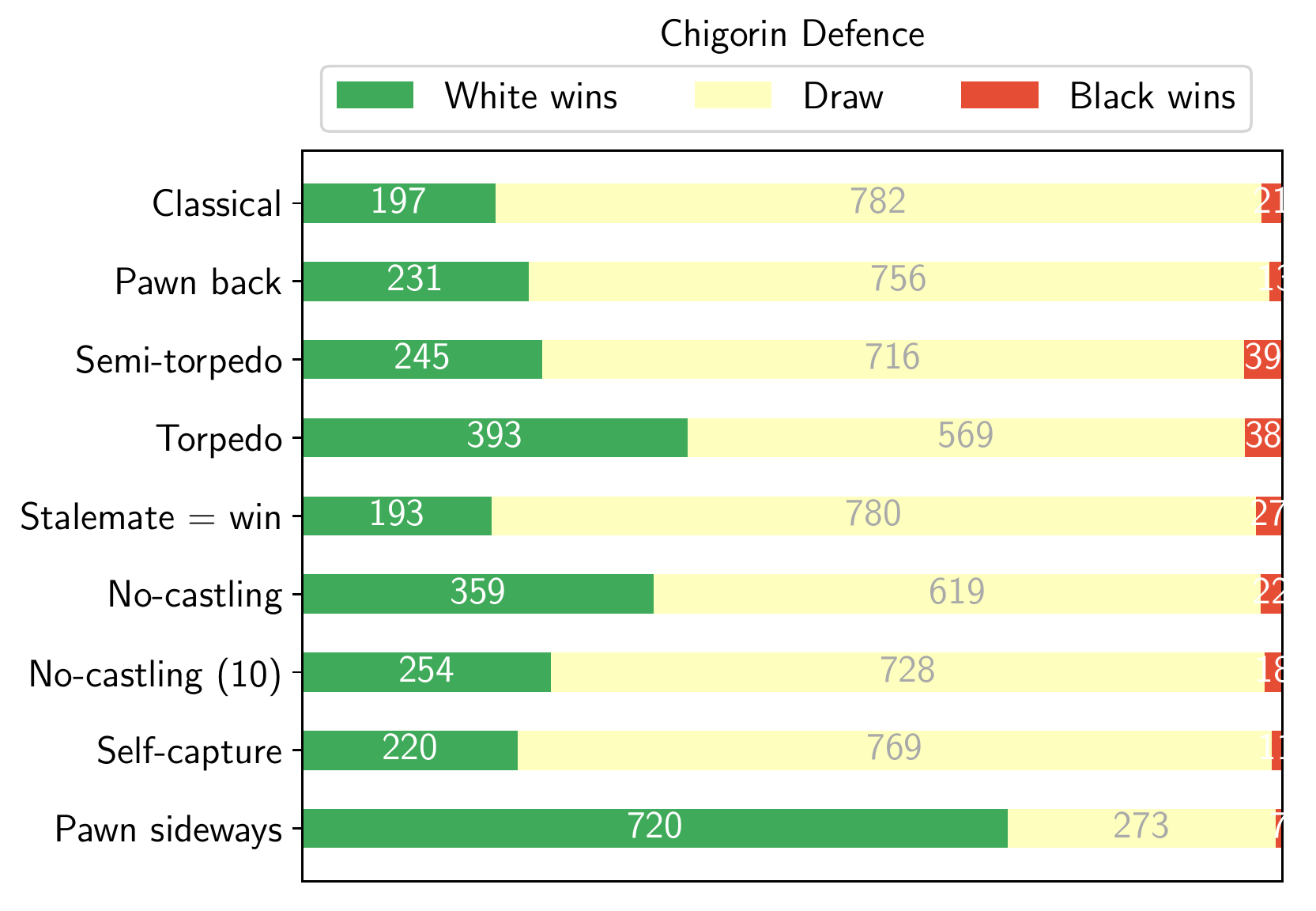}
\caption{Chigorin Defence (\mmove{1}d4 d5 \mmove{2}c4 Nc6)}
\label{fig:openings-chigorin}
\end{subfigure}
\\
\begin{subfigure}[t]{\columnwidth}
\centering\captionsetup{width=.95\columnwidth}
\includegraphics[width=\columnwidth]{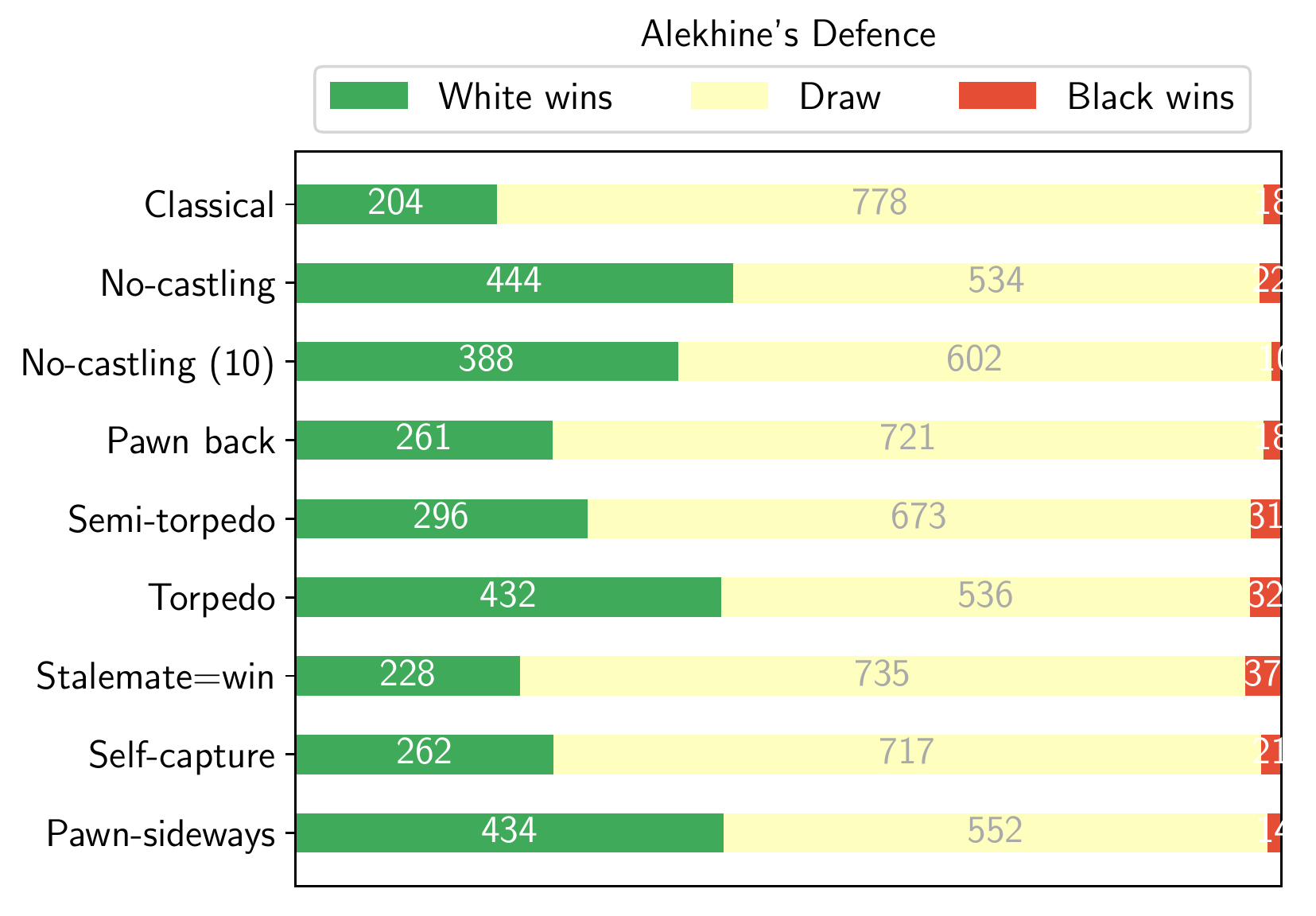}
\caption{Alekhine Defence (\mmove{1}e4 Nf6)
}
\label{fig:openings-alekhine}
\end{subfigure}%
~ 
\begin{subfigure}[t]{\columnwidth}
\centering\captionsetup{width=.95\columnwidth}
\includegraphics[width=\columnwidth]{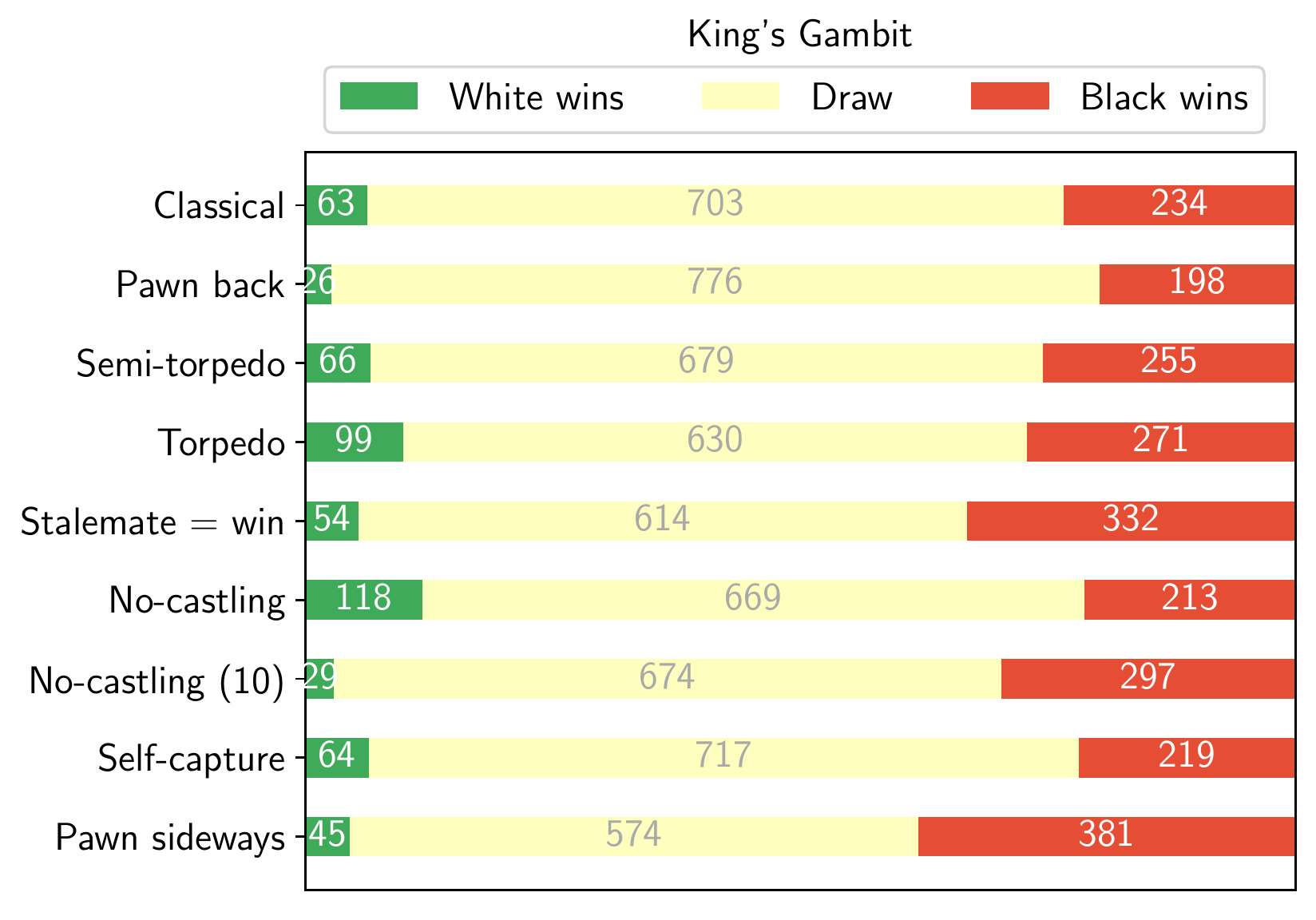}
\caption{King's Gambit (\mmove{1}e4 e5 \mmove{2}f4)}
\label{fig:openings-kings-gambit}
\end{subfigure}
\caption{The same opening position can give vastly different degrees of advantage to either play, depending on the variant under consideration, as shown here by the number of games won, drawn and lost for AlphaZero as White when playing at approximately 1 second per move, for a sample of 1000 games, while always playing the best move without any additional noise being added for play diversity. The stochasticity captured in the results stems from the asynchronous execution of MCTS threads during search. Therefore, these results indicate how favorable the 'main line' continuation is, for each of the following openings: the Dutch Defence, the Chigorin Defence, Alekhine Defence and the King's Gambit. 
}
\label{fig:opening-comparison}
\end{figure*}

To further illustrate how different alterations of the rule set would require players to adjust their opening repertoires, we provide a comparison of how favourable specific opening positions are for the first player, for each of the variants previously introduced in Table \ref{tab:variants_listing}. Figure \ref{fig:opening-comparison} shows the win, draw, and loss percentages for White under 1 second per move, for the Dutch Defence, Chigorin Defence, Alekhine Defence and King's Gambit, on a sample of 1000 self-play games. The only variant we did not include in these comparisons is Pawn one square,
as the lines used in the comparisons involve the double-pawn-moves which are not legal in that variant.

These four opening systems are not considered to be the most principled ways of playing Classical chess.
They are therefore particularly interesting for establishing if a certain rule change pushes the evaluation of each of these openings from ``slightly inferior'' to ``unsound'' or ``unplayable''.

In case of Dutch Defence in \figref{fig:openings-dutch},
we see that it is more favourable for White in Torpedo and Stalemate=win chess than in Classical chess.
This is in line with the overall increase in decisiveness in those variations, but is not more favourable in case of No-castling chess, despite No-castling chess otherwise being more decisive than Classical chess.
We can already see in this one example that the overall differences in decisiveness between variants are not equally distributed across all possible opening lines, and that the evaluation of the difference in the expected score will depend on the style of opening play.

In case of Chigorin Defence in \figref{fig:openings-chigorin},
Pawn-sideways chess seems to be refuting the variation, based on our initial findings. In a smaller sample of games played at 1 minute per move, we have seen a 100\% score being achieved by AlphaZero in this line of Pawn-sideways chess, though these are still preliminary conclusions.
To the human eye the line does not appear to be very forcing;
it is not a short tactical refutation, but results in a fairly long-term strategic advantage, which AlphaZero converts into a win. This line also seems to be harder to defend in No-castling chess and Torpedo, but not in Stalemate=win chess, unlike the Dutch Defence.

The Alekhine Defence in \figref{fig:openings-alekhine}
seems to be less sound in all of the variations considered, compared to Classical chess, with a major increase in decisiveness in Pawn-sideways chess, No-castling chess and Torpedo chess.

Finally, King's Gambit in \figref{fig:openings-kings-gambit}
seems to give a substantial advantage to Black across all chess variants considered, although in No-castling chess and Torpedo chess, White has somewhat better winning chances than in Classical chess. Pawn-sideways chess, again, seems to be the worst of the variants to consider playing this line in. Still, in our preliminary experiments with games at longer thinking times, most games would still ultimately end in a draw. This suggests that it is still likely a playable opening, when played at a very high level with deep calculation.

%%%%%%%%%%%%%%%%%%%%%%%%%%%%%%%%%%%%%%%%
\subsection{Utilisation of special moves}
\label{sec:special-moves}

\matthias{Thought: just having the option of special moves might already change the game without them actually being utilized that much.}

Several of the variants that are explored in this study involve additional move options that are not permitted under the rules of Classical chess, like additional pawn moves and self-captures.
It is not clear from the outset how often these newly introduced moves would be utilised in each of the variants.
Will they make a difference?
We use the set of 10,000 games at 1 second per move from Section \ref{sec:self-play-games}
to quantify how often the additional moves are played.
\matthias{when I read this last sentence, I always wonder ``is it difference for the 1-minute per move case?''}

\subsubsection{Torpedo moves}

In Semi-torpedo chess, $88\%$ of all games have at least one torpedo move, and $1.20\%$ of all moves played in the game are torpedo moves.
In Torpedo chess, these percentages are even higher: $94\%$ of games utilise torpedo moves and these represent $2.40\%$ of all moves played in the game.
Furthermore, $28.7\%$ of games featured pawn promotions with a torpedo move,
highlighting the speed at which a passed pawn can be promoted to a queen.

\subsubsection{Backwards and lateral pawn moves}

In Pawn-back chess, $96.3\%$ of the games involved a backwards pawn move.
In Pawn-sideways chess, $99.6\%$ of games features lateral pawn moves, and a total of $11.4\%$ of all moves in the game were lateral pawn moves, as the reconfiguring of
pawn formations was common in AlphaZero's playing style in this chess variant.

\subsubsection{Self-captures}

In Self-capture chess, $52.5\%$ of games featured self-capture moves, which represented $0.7\%$ of all moves played. The most common self-captures involved sacrificing a pawn ($86.9\%$), although sacrificing a bishop ($5.3\%$) or a knight ($4.5\%$) was not uncommon.
Rook self-capture sacrifices were rare ($2.3\%$) and occasionally AlphaZero would self-capture a
queen ($1\%$), though these were mostly unnecessary captures in winning positions, given that AlphaZero was not incentivised to win in the fastest possible way.

\subsubsection{Winning through stalemate}

In Stalemate=win chess the percentage of all decisive games that were won by stalemate rather than mate in AlphaZero games was $37.2\%$, though this number is inflated due to the fact that AlphaZero would often stylistically stalemate rather than mate the opponent in positions where both are possible.

The percentages listed above suggest that the rule changes featured in these chess variants did indeed leave a trace on how the game is being played,
and that they are useful
\matthias{how is the ``useful'' conclusion drawn?}
additional options that can potentially change the game dynamics.
Yet, it is important to note that the resulting games are still of approximately similar length, as shown in \figref{fig:game-lengths} in Appendix \ref{sec:app-quantitative}, with some changes in the 
empirical duration of decisive games.
This means that playing a game in one of these chess variants is unlikely to prolong or shorten the game by a large amount, meaning that classical time controls should still be appropriate. Note that the numbers in \figref{fig:game-lengths} that correspond to the number of plies in AlphaZero games are an upper bound on game length, since AlphaZero was trained without discounting, and would therefore not play the fastest winning sequence in its decisive games.

%%%%%%%%%%%%%%%%%%%%%%
\subsection{Diversity}
\label{sec:diversity}

For a game to be appealing, it has to be rich enough in options that these options do not get quickly exhausted, as play would then become repetitive.
We use the average information content (entropy)
of the first $T=20$ plies of play from each variant's prior as a surrogate diversity measure.
\matthias{mention again that prior is before MCTS}
\matthias{I would explain this (from next section) here already: ``The prior is a weighted list of \emph{possible moves} for state $s_t$ that are utilised in AlphaZero's MCTS search.
The weights specify how \emph{plausible} each move is before MCTS calculation;''}
The trained AlphaZero policy priors
model the move probabilities of the positions in self-play training data, and reflects the statistics at which opening lines appear there.
An entropy of \emph{zero} corresponds to there being one and only one forcing sequence of moves to be playable for White and Black, all other moves leading to substantially worse positions for each side. A higher entropy implies a wider and more balanced opening tree of variations, leading to a more diverse set of middlegame positions.
The intuition that there would be many more plausible opening lines in slower variants like Pawn one square, holds true experimentally.
In simulation, more decisive variants like Torpedo chess typically have fewer plausibly playable opening lines.

The decomposition of the entropy as a statistical expectation
can help identify whether there exist defensive lines that equalise the game in an almost forcing way.
In Classical chess, one such defensive resource is the Berlin Defence in the Ruy Lopez, taking the sting out of \mmove{1}e4.
We show in Section \ref{sec:berlin} that
AlphaZero, when trained on Classical chess, expresses a strong preference for the Berlin Defence, similarly to the human consensus on the solidity of the Berlin endgame. Without the option to castle, this particular line disappears in No-castling chess.

\subsubsection{Average information content}

The prior network from \eqref{eq:az-neuralnet}
defines the probability of \emph{a priori} considering move $a_t$ in state $s_t$, but as move $a_t$ leads to state $s_{t+1}$ deterministically, we shall abbreviate the prior with $p(s_{t+1} | s_t)$.

\begin{table}[]
\centering
\begin{tabular}{lll}
\toprule
Variant & Entropy & Equivalent 20-ply games \\
\toprule
No-castling & 27.65 & $1.02 \times 10^{12}$ \\   
Torpedo & 27.89 & $1.30 \times 10^{12}$ \\  
Self-capture & 27.94 & $1.36 \times 10^{12}$ \\
No-castling (10) & 27.97 & $1.40 \times 10^{12}$ \\ 
Classical & 28.58 & $2.58 \times 10^{12}$ \\
Stalemate=win & 29.01 & $3.97 \times 10^{12}$ \\
Semi-torpedo & 31.63 & $5.45 \times 10^{13}$ \\
Pawn-back & 32.30 & $1.07 \times 10^{14}$ \\
Pawn-sideways & 34.16 & $6.85 \times 10^{14}$ \\
Pawn one square & 38.95 & $8.24 \times 10^{16}$ \\
\midrule
Uniform random & 64.96 & $1.63 \times 10^{28}$ \\
\bottomrule
\end{tabular}
\caption{The average information content in nats in the first 20 plies of the AlphaZero prior for each chess variant.
The uniform random baseline assumes an equal probability for each move in Classical chess,
and provides rough indication of the ratio between ``plausible'' and ``possible'' games according to the AlphaZero prior.
The uniform random baseline depends on the number of legal moves per position, and is marginally different but of the same magnitude for other variations.}
\label{tab:entropy}
\end{table}

The prior is a weighted list of \emph{possible moves} for state $s_t$ that are utilised in AlphaZero's MCTS search.
The weights specify how \emph{plausible} each move is before MCTS calculation; they specify candidates for consideration.
In information-theoretic terms, the entropy 
\begin{equation} \label{eq:one-move-entropy}
H(s_t) = - \sum_{s_{t+1}} p(s_{t+1} | s_t) \log p(s_{t+1} | s_t) 
\end{equation}
is a function of state $s_t$ and
represents the number of nats (or bits, if $\log_2$ is used) that are needed to encode the weighted moves in position $s_t$.

If there are $M(s_t)$ legal moves in state $s_t$, then the
number of candidate moves $m(s_t)$ -- the number that a top player would realistically consider -- is much smaller than $M(s_t)$.
In \citet{degroot46}'s original framing,
$M(s_t)$ is a player's legal freedom of choice, while
$m(s_t)$ is their objective freedom of choice.
\citet{Iida2003} hypothesise that $m(s_t) \approx \sqrt{M(s_t)}$ on average.
\matthias{this section is a good motivation for why you consider the prior
in your analysis. mention this earlier?}
Because $p(s_{t+1} | s_t)$ is a distribution on all legal moves, we define the number of candidate moves
$m(s_t)$ by
\begin{equation} \label{eq:candidate-moves}
m(s_t) = \exp ( H(s_t) ) \ ; 
\end{equation}
it is 
the number of \emph{uniformly weighted} moves that could
be encoded in the \emph{same} number of nats
as $p(s_{t+1} | s_t)$.\footnote{As an illustrative example, if the number of candidate moves
is $m(s_t) = 3$ for some $p(s_{t+1} | s_t)$ that might put non-zero mass on all of its moves, then $m(s_t)$ is also equal to
the number of candidate moves of a probability vector
$\p = [\frac{1}{3}, \frac{1}{3}, \frac{1}{3}, 0, \ldots, 0]$ that puts equal non-zero mass on only three moves.}

We provide insight into the diversity of the prior opening tree through two quantities, the move sequence entropy $\Hcal(t)$ at depth $t$ from the opening position,
and the average number of candidate moves at ply $t$, $\Mcal(t)$.

\paragraph{Move sequence entropy}
Let $\s = \s_{1:t} = [s_1, s_2, \ldots s_t]$  be the sequence of states after $t$ plies,
starting at $s_0$, the initial position.
The prior probability -- without search -- of move sequence $\s_{1:t}$ is
% \begin{equation}
$p(\s_{1:t} | s_0) = \prod_{\tau=1}^t p(s_{\tau} | s_{\tau - 1})$.
% \end{equation}
The entropy of the move sequence is
\begin{align} 
\Hcal(t) & = - \sum_{\s_{1:t}} p(\s_{1:t}) \log p(\s_{1:t}) \nonumber \\
& = \Ebb_{\s_{1:t} \sim p(\s_{1:t})} \Big[- \log p(\s_{1:t}) \Big] \ , \label{eq:entropy}
\end{align}
where the starting position $s_0$ is dropped from notation for brevity.
An entropy $\Hcal(t) = 0$ implies that, according to the prior,
one and only one reasonable opening line could be considered by White and Black up to depth $t$, with all deviations form that line leading to substantially worse positions for the deviating side.
A higher $\Hcal(t)$ implies that we would \emph{a priori} expect a wider opening tree of variations, and consequently a more diverse set of middlegame positions.

\paragraph{Average number of candidate moves}

The entropy of a chess variant's prior opening tree is an unwieldy number that doesn't immediately inform us how many
move options we have in each chess variant.
A more naturally interpretable number is 
the expected number of (good) candidate moves at each ply as the game unfolds.
The average number of candidate moves at ply $t$ is
\begin{equation} \label{eq:average-candidates}
\Mcal(t) = 
\sum_{\s_{1:t}} p(\s_{1:t}) \, m(s_t) =
\Ebb_{\s_{1:t} \sim p(\s_{1:t})} \Big[ m(s_t) \Big] \ .
\end{equation}
Both the sums in \eqref{eq:entropy} and
\eqref{eq:average-candidates} are
over an exponential number of move sequences.
We compute Monte Carlo estimates of $\Hcal(t)$ and $\Mcal(t)$ by sampling $10^4$ sequences from $p(\s)$ and averaging the negative log probabilities of those sequences to obtain $\Hcal(t)$,
or averaging $m(s_t)$ over all samples at depth $t$ to obtain $\Mcal(t)$.
\matthias{summarize results here!}
We defer a presentation of the breakdown of the average number of candidate moves per variant to
\figref{fig:average-candidates} in Appendix \ref{sec:app-quantitative},
and will encounter $\Mcal(t)$ next in \figref{fig:classical-vs-nocastling} when Classical and No-castling chess are compared side by side.

The entropy of the AlphaZero prior opening tree is given in Table \ref{tab:entropy} for each variation.
Similar to the calculation in \eqref{eq:candidate-moves}
we give an estimate of the equivalent number of 20-ply sequences as $\exp(\Hcal(t))$.
As a baseline comparison, we take a prior distribution
for Classical chess where all legal moves are \emph{equally} playable, and
estimate the entropy of the ``Uniform random'' move selection criteria. It affords us a crude estimate of the number of possible classical openings, as opposed to the number of plausibly playable or candidate openings.
The estimates in Table \ref{tab:entropy} for Classical chess and "Uniform random Classical chess'' corroborate the claim
that the number of playable opening lines -- a player's objective freedom of choice -- is roughly the square root of the number of legal opening lines \cite{Iida2003}.

The two variants that have the largest entropy and hence largest opening tree in Table \ref{tab:entropy},
Pawn-sideways and Pawn one square, also happen to be among the most drawish, according to Figures \ref{fig:draw-rate-1s}
and \ref{fig:draw-rate-100s}.
The two variants that have the smallest opening trees under our analysis, No-castling and Torpedo, are also the most decisive and give White some of the largest advantages, according to Figures \ref{fig:draw-rate-1s} to \ref{fig:expected-scores-100s}.
Importantly, we estimate the size of the opening trees of these more decisive versions to still be of the same order of magnitude as that of Classical chess.

\begin{figure}[t]
\centering
\includegraphics[width=\columnwidth]{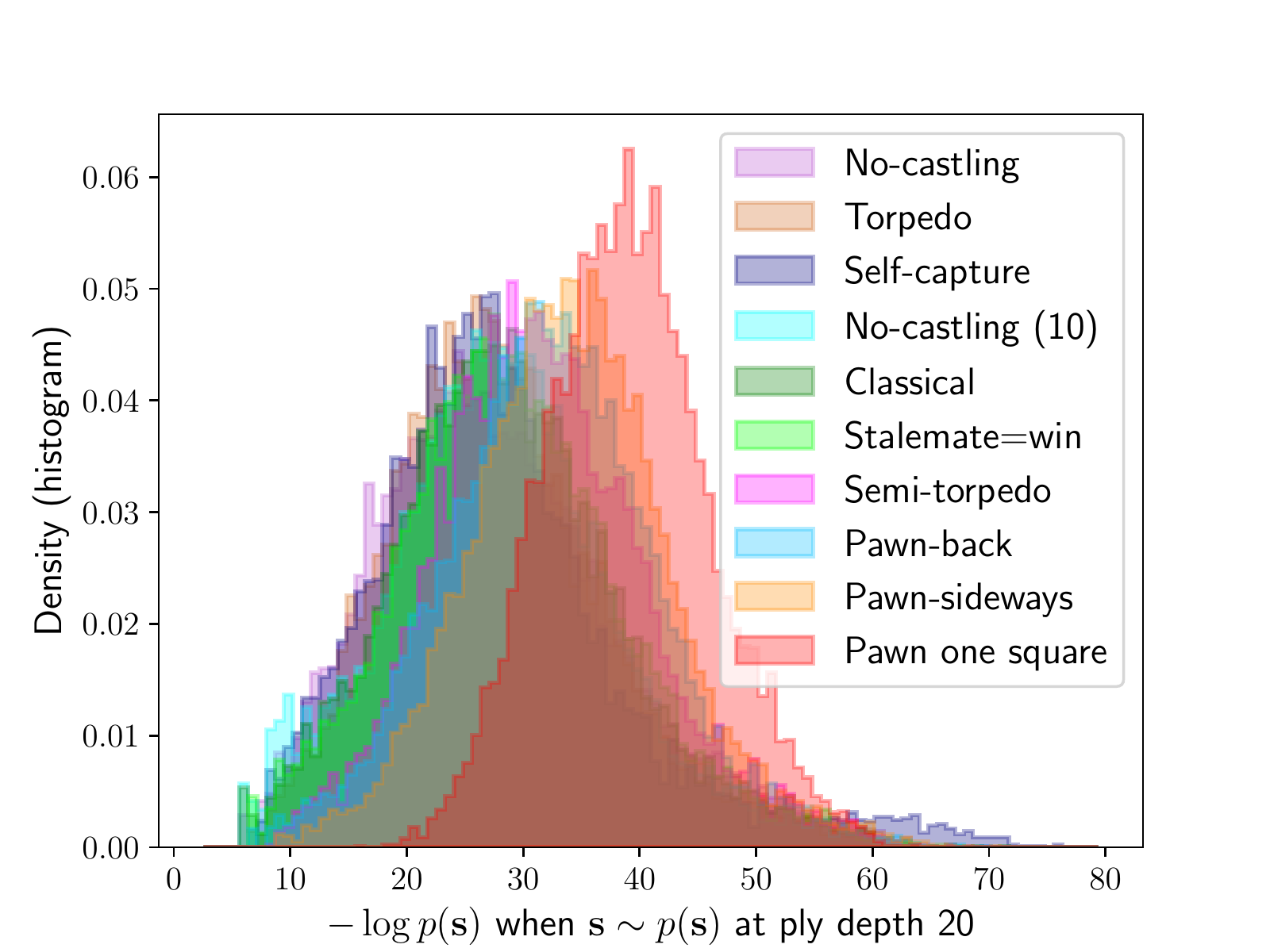}
\caption{
Histograms of $- \log p(\s)$ when $\s \sim p(\s)$ for
each variant. Following \eqref{eq:entropy},
the means of these distributions give the entropies in Table \ref{tab:entropy}.
The individual histograms are separately presented in \figref{fig:entropy-breakdown} in Appendix \ref{sec:app-quantitative}.
}
\label{fig:log-prob-histogram}
\end{figure}

\figref{fig:log-prob-histogram}
(a separate figure for each variant appears in \figref{fig:entropy-breakdown} in Appendix \ref{sec:app-quantitative}) visualises
the density of $- \log p(\s)$ when state sequences $\s$ are drawn from $p(\s)$.
The mean of each density is the entropy of \eqref{eq:entropy}, and an overlap in the histograms of two variants implies that their opening trees
contain a similar number of lines that are considered as candidates with similar odds.
In \figref{fig:log-prob-histogram}, a histogram that is shifted to the left means that fewer move sequences are considered \emph{a priori}, and each has higher probability.
A histogram that is shifted to the right implies that a larger variety of move sequences are \emph{a priori} considered, and each has to be considered with a smaller probability.
``Uniform random'' is shown in \figref{fig:entropy-breakdown-u},
and would appear as a tall narrow spike centred around 64 in this figure.
In the following section, we shall use log probability histograms as a tool to highlight the differences between
Classical and No-castling chess.

%%%%%%%%%%%%%%%%%%%%%%%%%%%%%%%%%%%%%%%%%%%%%%%
\subsubsection{Classical vs. No-castling chess}
\label{sec:berlin}

In Classical chess AlphaZero has a strong preference for playing the Berlin Defence \mblackmove{1}e5 \mmove{2}Nf3 Nc6 \mmove{3}Bb5 Nf6 in response to \mmove{1}e4, and here \mmove{4}O-O is White's main reply, which is not an option in no-castling chess.
Yet, castling is also an integral part of most other lines in the Ruy Lopez, affecting each move when considering relative preferences.
In the absence of castling, AlphaZero does not have as strong
a preference for a particular line for Black after \mmove{1}e4, suggesting either that it is not as easy to fully neutralise White's initiative, or alternatively that there is a larger number of promising defensive options.

\begin{table}[]
\centering
\begin{tabular}{lll}
\toprule
Variant & Entropy & Equiv.~21-ply games \\
\toprule
Classical (e4) & 23.72 & $2.00 \times 10^{10}$ \\
Classical (Nf3) & 29.54 & $6.75 \times 10^{12}$ \\
No-castling (e4) & 27.42 & $8.10 \times 10^{11}$ \\
No-castling (Nf3) & 28.40 & $2.16 \times 10^{12}$ \\
\bottomrule
\end{tabular}
\caption{The average information content in nats
of the AlphaZero prior for Classical and No-castling chess,
estimated on the 20 plies following \mmove{1}e4 and \mmove{1}Nf3.
}
\label{tab:e4-Nf3-entropy}
\end{table}

\begin{figure*}[h!]
\centering
\begin{subfigure}[t]{\columnwidth}
\centering\captionsetup{width=.95\columnwidth}
\includegraphics[width=\columnwidth]{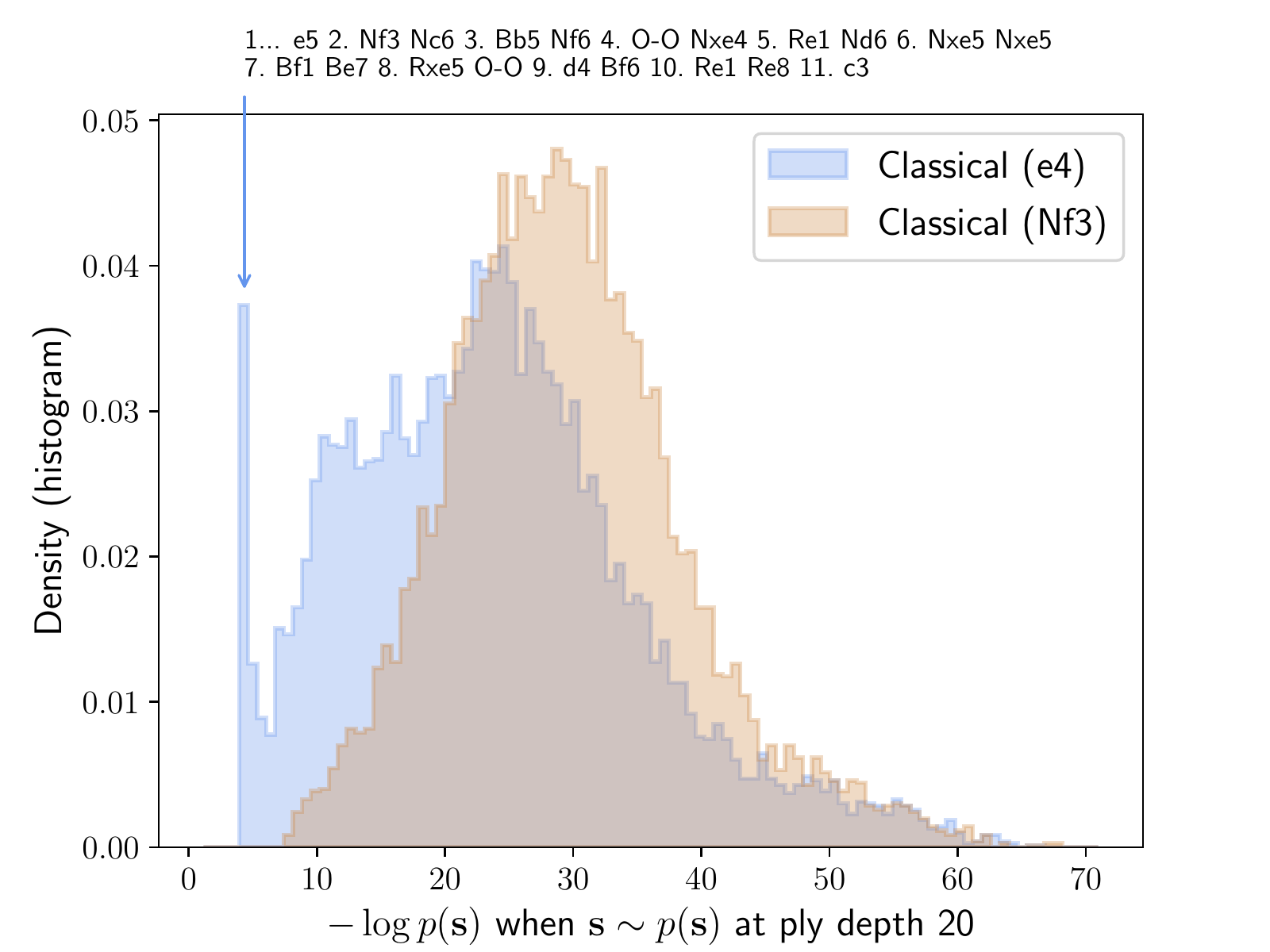}
\caption{The density of
(negative) log likelihoods for opening lines in Classical chess after \mmove{1}e4 and \mmove{1}Nf3
when move sequences are sampled from the AlphaZero prior.
There is a marked difference in overlap between the histograms, suggesting that AlphaZero \emph{a priori} considers ``narrower'' opening lines after \mmove{1}e4 than after \mmove{1}Nf3.
We identify the samples $\s$ at the high likelihood spike with a particular line in the Berlin Defence.
}
\label{fig:log-p-classical-e4-nf3}
\end{subfigure}%
~ 
\begin{subfigure}[t]{\columnwidth}
\centering\captionsetup{width=.95\columnwidth}
\includegraphics[width=\columnwidth]{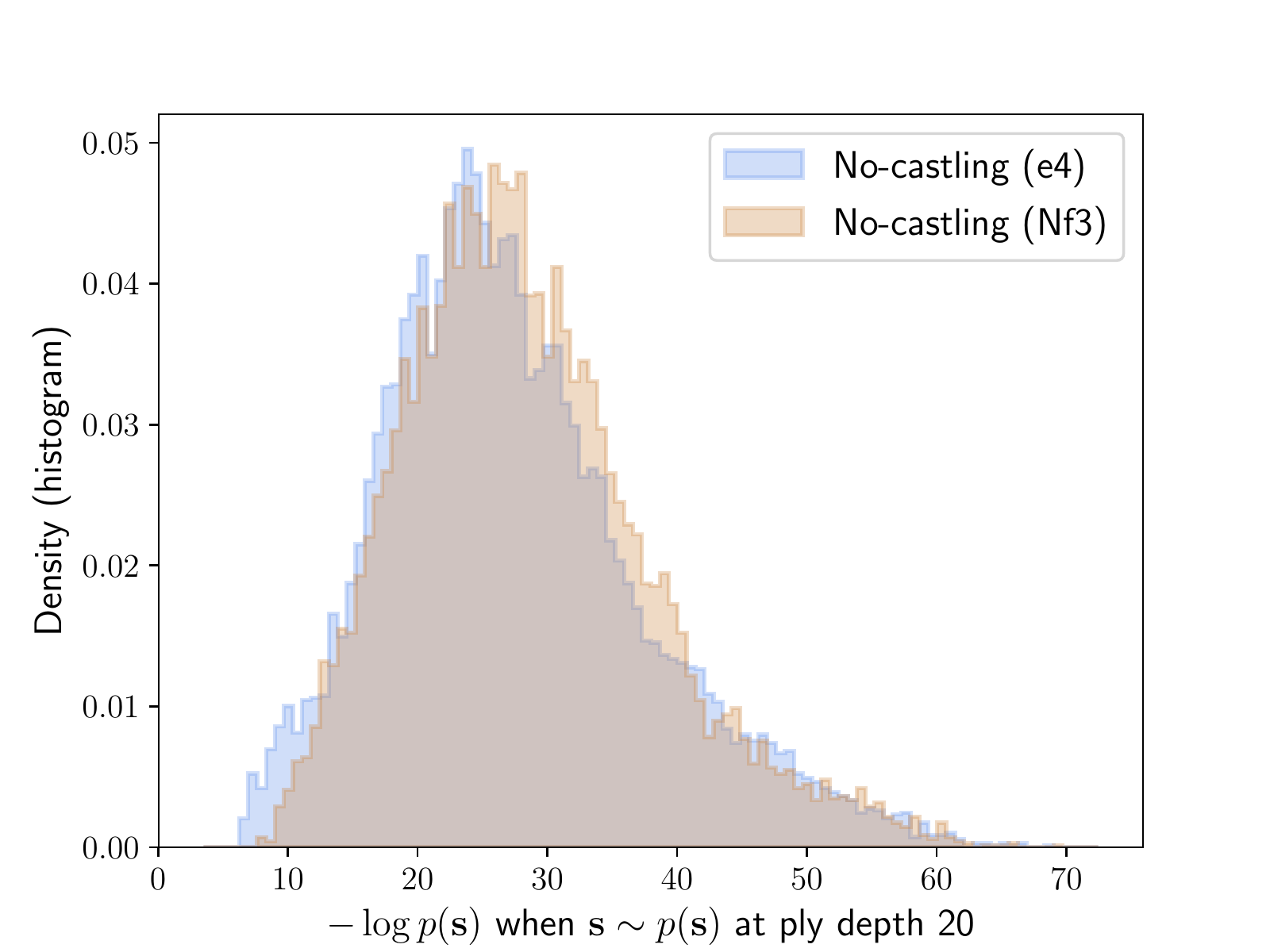}
\caption{The density of
(negative) log likelihoods for opening lines in No-castling chess after \mmove{1}e4 and \mmove{1}Nf3
when move sequences are sampled from the AlphaZero prior. Without the option of castling a king to safety, the prior opening trees after \mmove{1}e4 and \mmove{1}Nf3 have more similar ``distributional footprints'' compared to
Classical chess in \figref{fig:log-p-classical-e4-nf3}.}
\label{fig:log-p-no-castling-e4-nf3}
\end{subfigure}
\\
\begin{subfigure}[t]{\columnwidth}
\centering\captionsetup{width=.95\columnwidth}
\includegraphics[width=\columnwidth]{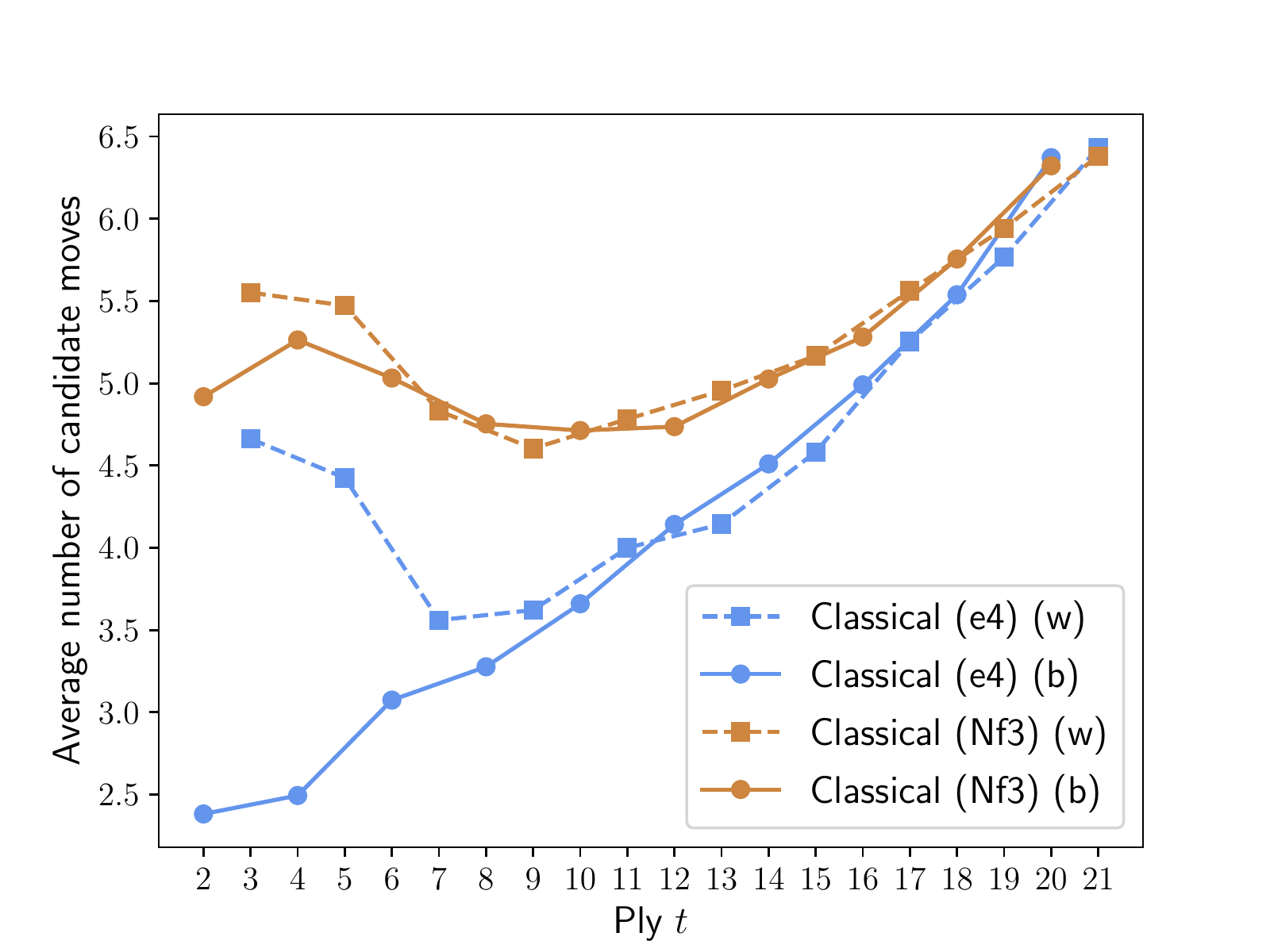}
\caption{The average number of candidate moves $\Mcal(t)$, as computed with
\eqref{eq:average-candidates}, for Classical chess.}
\label{fig:candidates-classical}
\end{subfigure}%
~ 
\begin{subfigure}[t]{\columnwidth}
\centering\captionsetup{width=.95\columnwidth}
\includegraphics[width=\columnwidth]{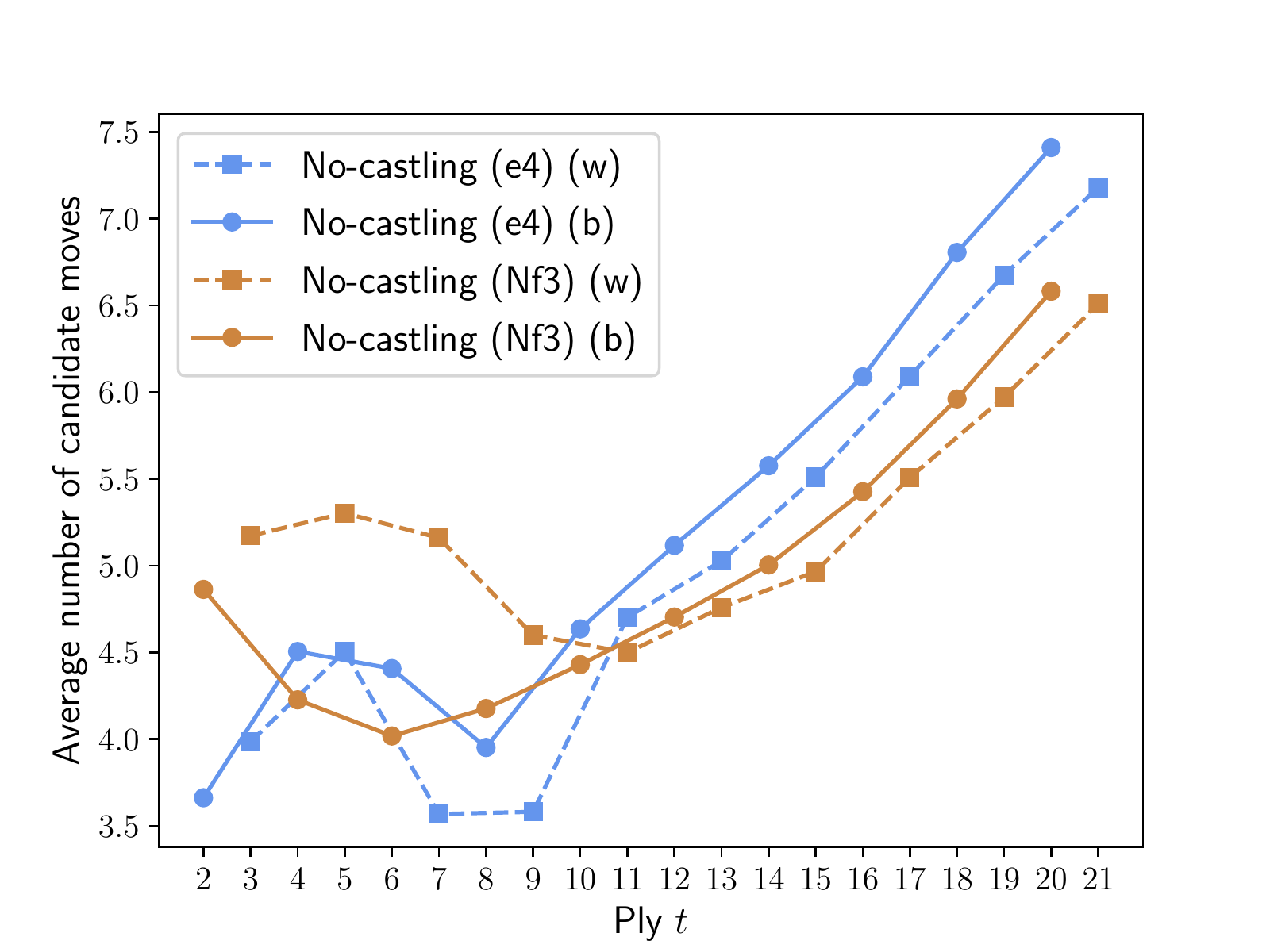}
\caption{The average number of candidate moves $\Mcal(t)$, as computed with
\eqref{eq:average-candidates}, for No-castling chess.}
\label{fig:x}
\end{subfigure}
\caption{The diversity of responses to \mmove{1}e4 and \mmove{1}Nf3 in Classical and No-castling chess, as well as the average number of candidate moves available for White and Black at each ply.
The spike is in the classical chess \mmove{1}e4 response distribution is at
\mblackmove{1}e5 \mmove{2}Nf3 Nc6 \mmove{3}Bb5 Nf6 \mmove{4}O-O Nxe4 \mmove{5}Re1 Nd6
\mmove{6}Nxe5 Nxe5 \mmove{7}Bf1 Be7 \mmove{8}Rxe5 O-O \mmove{9}d4 Bf6
\mmove{10}Re1 Re8 \mmove{11}c3, a known equalising line in the Berlin Defence, leading to drawish positions.}
\label{fig:classical-vs-nocastling}
\end{figure*}

To indicate the difference between Classical and No-castling chess, we compare the prior's opening trees after \mmove{1}e4 and \mmove{1}Nf3 in \figref{fig:classical-vs-nocastling}.
If we examine the density of $- \log p(\s_{2:21} | s_1)$ under $p(\s_{2:21} | s_1)$, where $s_1$ is the board position after either \mmove{1}e4 or \mmove{1}Nf3, we see a marked shift in the characteristics of the AlphaZero
\emph{prior} opening trees
(see Figures \ref{fig:log-p-classical-e4-nf3} and
\ref{fig:log-p-no-castling-e4-nf3}).
Statistically, the AlphaZero prior after \mmove{1}e4 is much more forcing
than after \mmove{1}Nf3 in Classical chess.
This is also evident from the average information content of the 20 plies after \mmove{1}e4 and \mmove{1}Nf3 in Table \ref{tab:e4-Nf3-entropy}.
In No-castling chess, \mmove{1}e4 seems as flexible as \mmove{1}Nf3, with a much wider variety of emerging preferential lines of play in the AlphaZero model. 

\figref{fig:classical-vs-nocastling} additionally shows the average number of candidate moves at each ply.
In Classical chess, White has more options than Black in both lines, the difference slowly diminishing over time as the first-move advantage decreases. 
\mmove{1}Nf3 offers more options, as it is less forcing. In No-castling chess, there seems to be a higher number of effective available moves for both sides after \mmove{1}e4 in the first couple of plies, based on the AlphaZero model. 

The Berlin Defence is a contributing factor to the narrower opening tree footprint we see in \figref{fig:log-p-classical-e4-nf3}.
As defensive tool for Black, Vladimir Kramnik successfully used the Berlin Defence in his World Championship Match with Garry Kasparov in 2000. He describes his choice as follows:

\begin{shadequote}[]{}
\emph{Back in the 90s, the engines of the time seemed to think that White had the advantage in the Berlin endgame, giving evaluations around +1 in White's favour. I thought that things weren't as simple, given that Black's only real problem was the loss of castling rights, and the difficulty of connecting rooks. The first time that I had a deeper look at it was when I was preparing for the match with Kasparov, and I thought that the opening was a good choice against Kasparov's playing style. Pursuing it required a belief in instinct and the human assessment of the position. Nowadays, it is considered to be a very solid opening, and modern engines assess most arising positions as being equal.}
\end{shadequote}

%%%%%%%%%%%%%%%%%%%%%%%%%%%%%%%%%%%%%%%%%%%%%%
\subsection{Differences between opening trees}
\label{sec:opening-tree-kl}

We compare how similar opening trees are by considering
how likely a given sequence of moves is under two variants.
To compare, we define one variant $p$ as the reference variant, and generate a move sequence $\s$ according to its prior.
The Kullback-Leibler divergence is a measure of how likely
such sequences of moves are under the opening book of variant $q$ compared to that of $p$.
Given two distributions $p(\s)$ and $q(\s)$,
the Kullback-Leibler divergence from $q$ to $p$ is 
the relative entropy of variant $p$ with respect to $q$,
\begin{align}
D_{\mathrm{KL}}[p \| q] & = \sum_{\s} p(\s) \log \frac{p(\s)}{q(\s)} \nonumber \\
& = \Ebb_{\s \sim p(\s)} \Big[\log p(\s) - \log q(\s) \Big] \ . \label{eq:kl}
\end{align}
It is the expected number of extra nats (or bits if $\log_2$ is used) that is required to compress move sequences from
variant $p$ using variant $q$'s opening book distribution.
The calculation in \eqref{eq:kl} involves a sum that is exponential in the length of $\s$,
and we estimate it with a Monte Carlo average of $\log p(\s) / q(\s)$ over $10^4$ sampled sequences from $p(\s)$.

A legal move in variant $p$ may be illegal in variant $q$,
in which case there is no way in which sequences in $p$ can be encoded in $q$. The Kullback-Leibler divergence in \eqref{eq:kl} is then infinite. More formally,
this happens when $q(s_{t+1} | s_t)$
puts zero mass on state transitions which are possible in $p$.
We therefore need to ensure that the reference variant $p$
is chosen so that its legal moves are a \emph{subset} of those of $q$.
In Table \ref{tab:kl} we show all divergences with respect to Classical chess, and distinguish between two kinds of variants:
\begin{enumerate}
\item variants that add moves to Classical chess, and whose legal moves are supersets of Classical chess;
\item variants that remove legal moves from Classical chess, and whose moves are subsets of Classical chess.
\end{enumerate}
The legal moves of Stalemate=win correspond to that of Classical chess, and it is included as both a superset and a subset in Table \ref{tab:kl}.
The density of samples from \eqref{eq:kl} is given in \figref{fig:kl} in Appendix \ref{sec:app-quantitative}.
The divergence is largest for variants that introduce the largest number of additional pawn moves or the most restrictions.
Self-capture chess, despite the plethora of additional opportunities for self-capture, is statistically closer to Classical chess because of the low frequency at which the extra moves are played.

\begin{table}[]
\centering
\begin{tabular}{clll}
\toprule
& Variant $p$ & Variant $q$ & $D_{\mathrm{KL}}[p \| q]$
% & Percentage more candidates
\\
\toprule
\parbox[t]{2mm}{\multirow{6}{*}{\rotatebox[origin=c]{90}{Supersets}}}
& Classical & Stalemate=win & 2.59 \\ % & 14\% more \\
& Classical & Self-capture & 5.24 \\ % & 30\% more \\
& Classical & Semi-torpedo & 10.35 \\ % & 68\% more \\
& Classical & Pawn-back & 11.70 \\ % & 79\% more \\
& Classical & Torpedo & 11.89 \\ % & 81\% more \\
& Classical & Pawn-sideways & 24.23 \\ % & 219\% more \\
\toprule
\parbox[t]{2mm}{\multirow{4}{*}{\rotatebox[origin=c]{90}{Subsets}}}
& Stalemate=win & Classical & 2.50 \\ % & 13\% more \\
& No-castling (10) & Classical & 7.17 \\ % & 43\% more \\
& No-castling & Classical & 13.19 \\ % & 93\% more \\
& Pawn one square & Classical & 20.28 \\ % & 176\% more \\
\bottomrule
\end{tabular}
\caption{
Differences in the opening tree of the new chess variants and Classical chess.
These are expressed as Kullback-Leibler (KL) divergences, the direction depending on whether a particular variant is a superset or a subset of Classical chess, based on the rule change. In all cases but Stalemate=win the 
reverse KL divergences are infinite as when there are
legal opening lines $\s$ in variant $p$ that don't exist in $q$, and hence for which $q(\s) = 0$ when $p(\s)$ is not (contributing $- \log 0$ to the divergence).
}
\label{tab:kl}
\end{table}

%%%%%%%%%%%%%%%%%%%%%%%%%%%%%%%%%%%%%%%%%%%%%%%%%%%%%%%%%
\subsection{How much opening theory should be relearned?}
\label{sec:additional-moves}

\begin{figure*}[t]
\centering
\includegraphics[width=1.47\columnwidth]{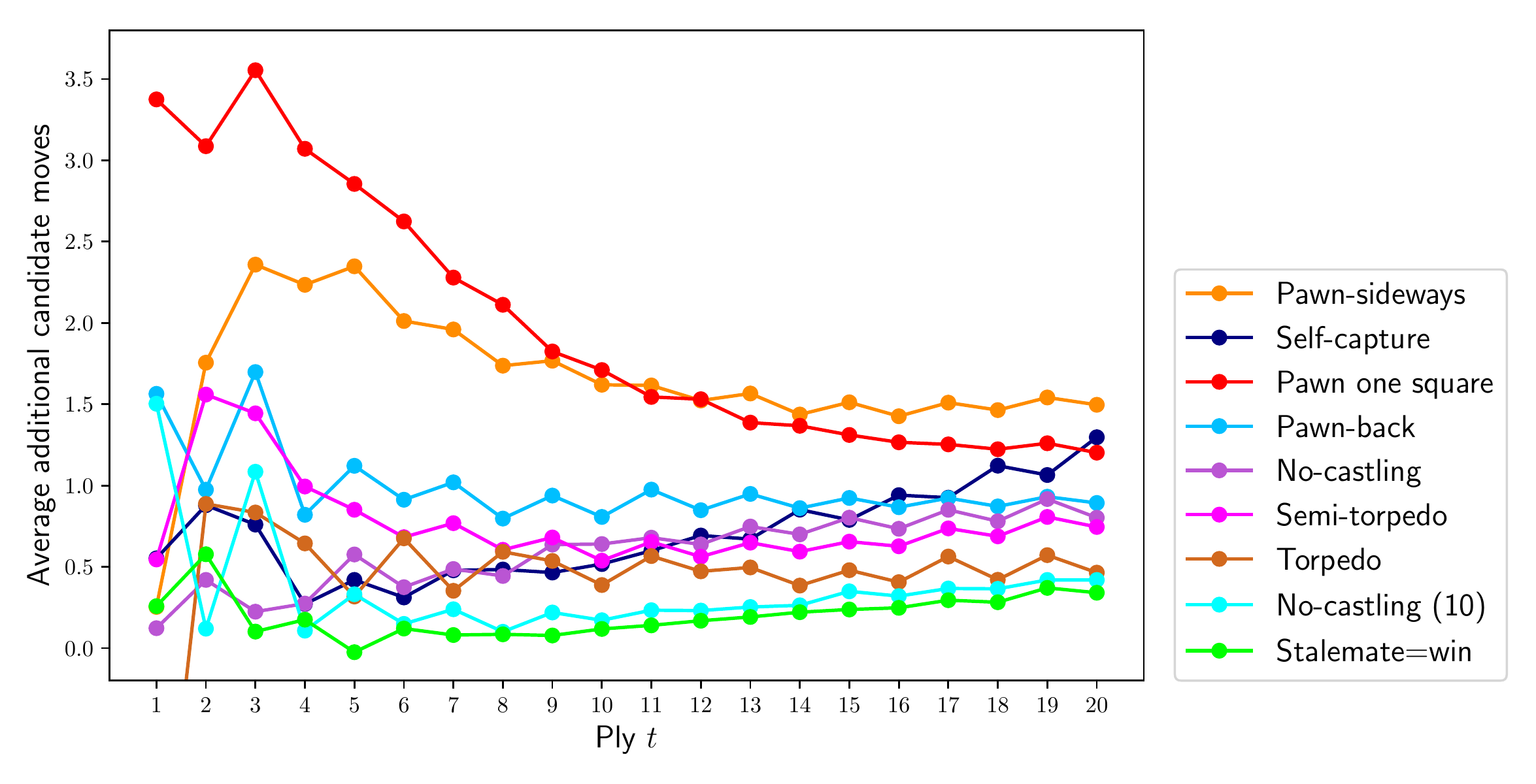}
\caption{
The average number of additional candidate moves
$\Acal_q(t)$ that
a Classical player Q with prior $q(s_{t+1} | s_t)$ should consider in order to match player P's candidate moves from prior $p(\s)$
for each of the evaluated variants; see \eqref{eq:additional-avg}.
\emph{(The order of the variants in the legend matches their ordering at ply $t=20$.)}
}
\label{fig:additional-candidates}
\end{figure*}

Although the relative entropy expresses how many more nats are required to encode prior moves of one variant given another, it does not tell us whether one variant's player is considering the right candidate moves when playing another variant.
How many \emph{more} candidate moves should a player Q,
who was trained on one variant of chess, take into consideration when wanting to play at player P's level in another variation?
Let $q(\s)$ be the candidate prior for the variation that player Q was trained on, and $p(\s)$ the prior for variant P, variant that Q wants to play.
We define the combination of the two priors as the normalized supremum
\begin{equation} \label{eq:combined}
r(s_{t+1} | s_t) = \frac{ \max \big\{ p(s_{t+1} | s_t), q(s_{t+1} | s_t) \big\} }{
\sum_{s_{t+1}'} \max \big\{ p(s_{t+1}' | s_t), q(s_{t+1}' | s_t) \big\} } \ .
\end{equation}
There is a particular reason behind our choice of definition for the combined prior in \eqref{eq:combined}: The number of candidate moves that the combination of players P and Q would consider, is always smaller than the sum of candidate moves that P and Q would consider individually.

Put more formally, define the number of candidate moves for the combined player as the number of uniformly weighed moves that could be encoded in the same number of nats as $r(s_{t+1} | s_t)$,\footnote{The perceptive reader would recognise
equation \eqref{eq:num_candidates_2} as equation
\eqref{eq:candidate-moves}. We restate it here with a subscript to indicate the explicit dependence on the distribution.}
\begin{equation} \label{eq:num_candidates_2}
m_r(s_t) = \exp \left(
- \sum_{s_{t+1}} r(s_{t+1} | s_t) \log r(s_{t+1} | s_t)
\right) \ .
\end{equation}
For any choice of priors $p$ and $q$ the number of candidate moves that are considered by the combined player in state $s_t$ is lower bounded by
\begin{equation} \label{eq:bound}
m_r(s_t) \le m_p(s_t) + m_q(s_t) \ ,
\end{equation}
which we prove in Appendix \ref{sec:bound-proof}.

We now define the difference
\begin{equation}
\mathrm{additional}(s_t) = m_r(s_t) - m_q(s_t)    
\end{equation}
to represent the number of additional candidate moves that player Q should consider, to play at the level of P in position $s_t$.
The additional number of candidates $\mathrm{additional}(s_t)$ is zero when the priors match, $q=p$,
and intuitively Q doesn't need to consider any further candidate moves. The number of additional moves may be negative; intuitively, Q puts enough weight on all candidates that P deems important, and doesn't need to consider any further candidate moves.
The number of additional candidate moves
and is upper bounded by $\mathrm{additional}(s_t) \le m_p(s_t)$ according to \eqref{eq:bound}; at the very worst, Q would additionally have to consider all of P's candidates.

We consider positions up to ply $t$ plies sampled from prior for P, and at ply $t$ evaluate how many additional candidate moves Q should consider on average:
\begin{equation} \label{eq:additional-avg}
\Acal_q(t) = \Ebb_{\s_{1:t} \sim p(\s_{1:t})} \Big[ \mathrm{additional}(s_t) \Big] \ .
\end{equation}
The expectation is estimated with a Monte Carlo average over $10^4$ samples from $p(\s_{1:t})$.

\figref{fig:additional-candidates} shows the average additional number of candidate moves if Q is taken as the Classical chess prior, with P iterating over all other variants.
From the outset, Pawn one square places $60\%$ of its prior mass on \mmove{1}d3, \mmove{1}e3, \mmove{1}c3 and \mmove{1}h3, which together only account for $13\%$ of Classical's prior mass. As pawns are moved from the starting rank and pieces are developed, $\Acal_q(t)$ slowly decreases for Pawn one square.
As the opening progresses, Stalemate=win slowly drifts from zero,
presumably because some board configurations that would lead to drawn endgames under Classical rules might have a different outcome.
Torpedo puts $66\%$ of its prior mass on one move, \mmove{1}d4, whereas the Classical prior is broader (its top move, \mmove{1}d4, occupies $38\%$ of its prior mass). The truncated plot value for Torpedo is $\Acal_q(1) = -1.8$, signifying that the first Classical candidate moves effectively already include those of Torpedo chess.
There is a slow upward drift in the average number of additional candidates that a Classical player has to consider under Self-capture chess as a game progresses. We hypothesise that it can, in part, be ascribed to the number of reasonable self-capturing options increasing toward the middle game.

%%%%%%%%%%%%%%%%%%%%%%%%
\subsection{Material}
\label{sec:piece-value}

Material plays an important role in chess, and is often used to assess whether a particular sequence of piece exchanges and captures is favourable.
Material sacrifices in chess are made either for concrete tactical reasons, e.g.~mating attacks, or to be traded off for long-term positional strengthening of the position.
Understanding the material value of pieces in chess helps players master the game and is one of the very first pieces of chess knowledge taught to beginners.
Changes to the rules of chess affect piece mobility,
and hence also the relative value of pieces.
Without a basic estimate of what the relative piece values
in each variant are, it would be harder for human players to start playing these chess variants.
As a guide, we provide an experimental approximation to piece values
based on outcomes of AlphaZero games under 1 second per move.

% Tie to AZ network --
% It consists of a deep neural network $f_{\theta}$ with weights $\theta$ that compute
% \begin{equation} \label{eq:az-neuralnet}
% (\p, v) = f_{\theta}(s)    
% \end{equation}
% for a given position or state $s$.

We approximate piece values from the weights of a linear model that predicts the game outcome from the difference in numbers of each piece only.
As background, the real AlphaZero evaluation $v$ in $(\p, v) = f_{\theta}(s)$ is the output of a deep neural network with weights $\theta$. The expected game outcome $v$ is the result of a final $\tanh$ activation to ensure an output in $(-1, 1)$.
If  $z \in \{-1, 0, 1 \}$ indicates the playing side's game outcome, AlphaZero's loss function includes the mean squared error $(z - v)^2$ \cite{Silver1140}.
We create a simplified evaluation function $g_w(s)$ that only takes piece counts on the board into consideration.
For a position $s$ we construct a feature vector $d \defined [1, d_{\textrm{\sympawn}},
d_{\textrm{\symknight}}, d_{\textrm{\symbishop}},
d_{\textrm{\symrook}}, d_{\textrm{\symqueen}}]$
that contains
the integer differences between the playing side and their opponent's
number of pawns, knights, bishops, rooks and queens.
We define $g_w$ with weights $w \in \Rbb^6$ as
\begin{equation}
g_w(s) = \tanh(w^T d) \ .
\end{equation}
When trained on the 10,000 AlphaZero self-play board positions from Section 
\ref{sec:self-play-games} for each variant, 
the piece weights $w$ provide an indication of their relative importance.
Let $(s, z) \sim \mathrm{games}$ represent a sample of a position and final
game outcome from a variant's self-play games.
We minimise 
\begin{equation}
\ell(w) = \Ebb_{(s, z) \sim \mathrm{games}} \Big[ \big(z - g_w(s) \big)^2 \Big]
\end{equation}
empirically over $w$, and normalise weights $w$ by $w_{\sympawn}$ to yield the relative
piece values. The recovered piece values for each of the chess variants are given in Table~\ref{tab:piece-value}.

\begin{table}[h]
\centering
\begin{tabular}{llllll}
\toprule
Variant & \sympawn & \symknight & \symbishop & \symrook & \symqueen \\
\toprule
Classical        & 1 & 3.05 & 3.33 & 5.63 & 9.5 \\
No castling	     & 1 & 2.97 & 3.13 & 5.02 & 9.49 \\
No castling (10) & 1 & 3.14 & 3.40 & 5.37 & 9.85 \\
Pawn one square  & 1 & 2.95 & 3.14 & 5.36 & 9.62 \\
Stalemate=win    & 1 & 2.95 & 3.13 & 4.76 & 8.96 \\
Self-capture     & 1 & 3.10 & 3.22 & 5.34 & 9.42 \\
Pawn-back        & 1 & 2.65 & 2.85 & 4.67 & 9.39 \\
Semi-torpedo     & 1 & 2.72 & 2.95 & 4.69 & 8.3 \\
Torpedo          & 1 & 2.25 & 2.46 & 3.58 & 7.12 \\
Pawn-sideways    & 1 & 1.8 & 1.98 & 2.99 & 5.92 \\
\bottomrule
\end{tabular}
\caption{Estimated piece values from AlphaZero self-play games for each variant.
}
\label{tab:piece-value}
\end{table}

In Classical chess, piece values vary based on positional considerations and game stage.
The piece values in Table \ref{tab:piece-value} should not be taken as a gold standard,
as the sample of AlphaZero games that they were estimated on does not fully capture the diversity of human play, and the game lengths do not correspond to that of human games, which tend to be shorter.
For comparison, we have included the piece value estimates that we obtain by applying the same method to Classical chess,
showing that the estimates do not deviate much from the known material values. Over the years, many material systems have been proposed in chess. The most commonly used one~\cite{capablanca2006chess} gives 3--3--5--9 for values of knights, bishops, rooks and queens. Another system~\cite{kaufman} gives 3.25--3.25--5--9.75. Yet, bishops are typically considered to be more valuable than the knights, and there is usually an additive adjustment while in possession of a bishop pair. The rook value varies between 4.5 and 5.5 depending on the system and the queen values span from 8.5 to 10.
The relative piece values estimated on the AlphaZero game sample for Classical chess, 3.05--3.33--5.63--9.5, do not deviate much from the existing systems.
This suggests that the estimates for the new chess variants are likely to be approximately correct as well.

We can see similar piece values estimated for No-castling, No-castling(10), Pawn-one-square chess, Self-capture and Stalemate=win. This is not surprising, given that these variants do not involve a major change in piece mobility. Estimated piece values look quite different in the remaining variations, where pawn mobility has been increased: Pawn-back, Semi-torpedo, Torpedo and Pawn-sideways. In Pawn-sideways chess, minor pieces seem to be worth approximately two pawns, which is in line with our anecdotal observations when analysing AlphaZero games, as such exchanges are frequently made. Like Torpedo chess, pawns become much stronger and more valuable than before. Changes in Pawn-back and Semi-torpedo are not as pronounced.

%% file: results-qualitative.tex
\section{Qualitative assessment} \label{sec:qualitative}

To evaluate the differences in play between the set of chess variations considered in this study, we
couple the quantitative assessment of the variations with expert analysis based on a large set of representative games. While the overall decisiveness and opening diversity add to the appeal of any chess variation, the subjective questions of aesthetic value and the types of positions, moves and patterns that arise are not possible to fully capture quantitatively. For providing a deep qualitative assessment of the appeal of these chess variations, we rely on the experience of chess grandmaster Vladimir Kramnik, an ex-world chess champion and an authority on the game. By characterising typical patterns, we hope to provide players with insights to help them judge for themselves if they would find some of these chess variants interesting enough to try out in practice. What we provide here are preliminary findings.

The detailed qualitative assessment of the chess variants presented in this article, along with typical motifs and illustrative games, is provided in the Appendix (Section~\ref{sec:chess_appendix}).
For this analysis,
we use the 1,000 1-minute per move games of Section \ref{sec:self-play-games}
as well as 200 1-minute per move games from a diverse set of early opening positions that all of the major opening systems.
By looking at the former, we were able to assess AlphaZero's preferred style of play in each chess variant, and by looking at the latter, we could assess how the treatment of different opening lines changes and which of those become more or less promising under each of the rule changes. 
Figure~\ref{fig:pos-examples} shows an illustrative example position for each of the considered chess variants.

What follows is a short summary of the main takeaways from the qualitative analysis for each of the variants, provided by GM Vladimir Kramnik.

\textbf{No-castling chess} is a potentially exciting variant, given that king safety is often compromised for both players, allowing for simultaneous attacking and counter-attacking and the equality, when reached, tends to be dynamic in nature rather than ``dry''. The multitude of approaches to evacuate the king, and their timing, adds complexity to the opening play. \textbf{No-castling (10)}, where castling is not permitted for the first 10 moves (20 plies) is a partial restriction, rather than an absolute one -- which does not change the game to the same extent. Due to castling being such a powerful option, the lines preferred by AlphaZero all tend to involve castling, only delayed -- resulting in a preference for slower, closed positions, and a less attractive style of play. Such partial castling restrictions can be considered if the desire is to sidestep opening theory and preparation, but this may not be of interest for the wider chess audience.

\textbf{Pawn one square chess} variant may appeal to players who enjoy slower, strategic play -- as well as a training tool for understanding pawn structures, due to the transpositional possibilities when setting up the pawns. The reduced pawn mobility makes it harder to launch fast attacks, making the game overall less decisive.

\textbf{Stalemate=win chess} has little effect on the opening and middlegame play, mostly affecting the evaluation of certain endgames. As such, it does not increase decisiveness of the game by much, as it seems to almost always be possible to defend without relying on stalemate as a drawing resource. Therefore, this chess variant is not likely to be useful for sidestepping known theory or for making the game substantially more decisive at the high level. The overall effect of the change seems to be minor.

\textbf{Torpedo and Semi-torpedo chess} both make the game more dynamic and more decisive, and Torpedo chess in particular leads to new motifs and changes in all stages of the game. Creating passed pawns becomes very important, as they are hard to stop. The attacking possibilities make Torpedo chess quite appealing, and it is likely to be of interest for players that enjoy tactical play.

\textbf{Pawn-back chess} makes it possible to regain control of the weakened squares in the position and remove some square weaknesses. It also introduces additional possibilities for opening up diagonals and making squares available for the pieces. Counter-intuitively, even though moving the pieces backwards is usually a defensive manoeuvre, this can make more aggressive options possible, given that pawns can now be pushed further earlier on, as there is always an option of moving them back to cover the weakened squares. AlphaZero has a strong preference for playing the French defence with Black, which is particularly interesting.

\textbf{Pawn-sideways chess} is incredibly complex, resulting in patterns that are at times quite ``alien'' when one is used to classical chess. The pawn structures become very fluid and it is impossible to create permanent pawn weaknesses. Given how important this concept is in classical chess, this chess variant requires us to rethink how we approach any given position, making it very concrete and relying on deep calculation. Restructuring the pawn formation takes time, and players need to use that time for creating other types of advantages. Many of AlphaZero games in this variant have been quite tactical, some involving novel tactics that are not possible under classical rules.

\textbf{Self-capture chess} is quite entertaining, as it introduces additional options for sacrificing material -- and material sacrifices have a certain aesthetic appeal. Self-capture moves can feature in all stages of the game. Not every game involves self-captures, as giving away material is not always required, but they do feature in a substantial percentage of the games, and in some games they occur multiple times. Self-capture moves can be used to open files and squares for the pieces in the attack;
opening up a blockade by sacrificing a pawn in the pawn chain; or in defence, while escaping the mating net.

%% file: conclusions.tex
\section{Conclusions}

We have demonstrated how AlphaZero can be used for prototyping board games and assessing the consequences of rule changes in the game design process, as demonstrated on chess, where we have trained AlphaZero models to evaluate 9 different chess variants, representing atomic changes to the rules of classical chess.
Training an AlphaZero model under these rule changes helped us effectively simulate decades of human play in a matter of hours, and answer the ``what if'' question: what the play would potentially look like under developed theory in each chess variant. We believe that a similar approach could be used for auto-balancing game mechanics in other types of games, including computer games, in cases when a sufficiently performant reinforcement learning system is available.

To assess the consequences of the rule changes, we coupled the quantitative analysis of the trained model and self-play games with a deep qualitative analysis where we identified many new patterns and ideas that are not possible under the rules of classical chess.
We showed that there several chess variants among those considered in this study that are even more decisive than classical chess: Torpedo chess, Semi-torpedo chess, No-castling chess and Stalemate=win chess.

We additionally quantified the arising diversity of opening play and the intersection of opening trees between chess variations, showing how different the opening theory is for each of the rule changes.
There is a negative correlation between the overall opening diversity and decisiveness, as the decisive variants likely require more precise play, with fewer plausible choices per move. For each of the chess variants, we estimated the material value of each of the pieces based on the results of 10,000 AlphaZero games, to provide insight into favourable exchange sequences and make it easier for human players to understand the game.

No-castling chess, being the first variant that we
analysed (chronologically), has already been tried in an experimental blitz grandmaster tournament in Chennai, as well as a couple of longer grandmaster games.
Our assessment suggests that several of the assessed chess variants might be quite appealing to interested players,
and we hope that this study will prove to be a valuable resource for the wider chess community.

%% file: appendix-math.tex
\section{Quantitative Appendix}
\label{sec:app-quantitative}

\subsection{Proof of equation \eqref{eq:bound}}
\label{sec:bound-proof}

Let $\p$ and $\q$ be two vectors with non-negative entries that sum to one. Define $\r$ as a vector with elements
\begin{equation} \label{eq:r-def}
r_i = \frac{\max(p_i, q_i) }{ \sum_{i'} \max(p_{i'}, q_{i'}) } \ .
\end{equation}
We show below that
\begin{equation} \label{eq:bound-appendix}
\erm^{- \sum_i r_i \log r_i} \le \erm^{- \sum_i p_i \log p_i} + \erm^{- \sum_i q_i \log q_i} \ .
\end{equation}
Let $R = \sum_{i} \max(p_{i}, q_{i})$ be the normalizing constant in \eqref{eq:r-def}. It is bounded by $1 \le R \le 2$. We write the entropy as
\begin{align}
& - \sum_i r_i \log r_i \nonumber \\
& \quad = - \frac{1}{R}  \sum_i \max(p_i, q_i) \log \max(p_i, q_i) + \log R \nonumber \\
& \quad = - \frac{1}{R}  \sum_i \max(p_i \log p_i, \, q_i \log q_i) + \log R \nonumber \\
& \quad \le - \sum_i \max(p_i \log p_i, \, q_i \log q_i) + \log R \nonumber \\
& \quad \le - \frac{1}{2} \sum_i p_i \log p_i
- \frac{1}{2} \sum_i q_i \log q_i + \log R
\label{eq:entropy-bound}
\end{align}
where the last inequality in \eqref{eq:entropy-bound}
follows from $\max(a, b) \ge \frac{a + b}{2}$.
Exponentiating \eqref{eq:entropy-bound} and applying
Jensen's inequality yields
\begin{align}
& \erm^{- \sum_i r_i \log r_i} \nonumber \\
& \quad \le R \erm^{\frac{1}{2} (- \sum_i - p_i \log p_i) + \frac{1}{2} (- \sum_i q_i \log q_i)} \nonumber \\
& \quad \le R \left( \frac{1}{2}  \erm^{- \sum_i p_i \log p_i } + \frac{1}{2} \erm^{- \sum_i q_i \log q_i} \right) \nonumber \\
& \quad \le \erm^{- \sum_i p_i \log p_i } + \erm^{- \sum_i q_i \log q_i} \ .
\end{align}
The final line follows from $R / 2 \le 1$ as $1 \le R \le 2$.
The bound is tight at $R=1$ when $\p$ and $\q$ both put probability mass uniformly on
two non-intersecting same-sized subsets of elements.\footnote{An example of two vectors giving a tight bound in \eqref{eq:bound-appendix} is
$\p = [\frac{1}{2}, \frac{1}{2}, 0, 0, 0]$ and
$\q = [0, 0, \frac{1}{2}, \frac{1}{2}, 0]$.}

\subsection{Additional figures}
\label{sec:additional-figs}

%%%%%%%%%%%%%%% Game length distribution %%%%%%%%%%%%%%%

\begin{figure}[t!]
\centering
\begin{subfigure}[t]{\columnwidth}
\centering\captionsetup{width=.95\columnwidth}
\includegraphics[width=\columnwidth]{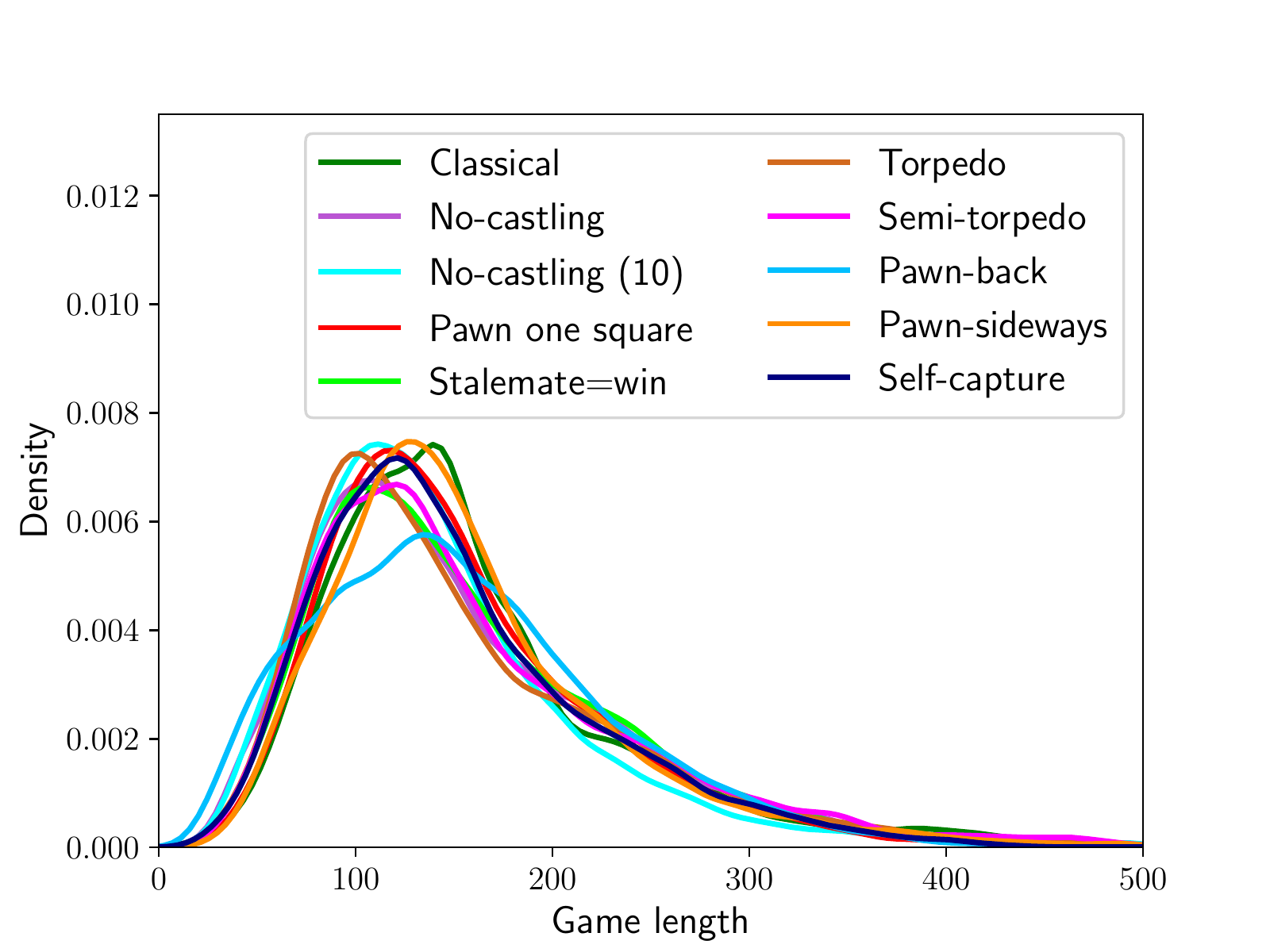}
\caption{The game length distributions of the total number of plies
for all self-play games for each variant.}
\label{fig:game-lengths-all}
\end{subfigure}%
\\
\begin{subfigure}[t]{\columnwidth}
\centering\captionsetup{width=.95\columnwidth}
\includegraphics[width=\columnwidth]{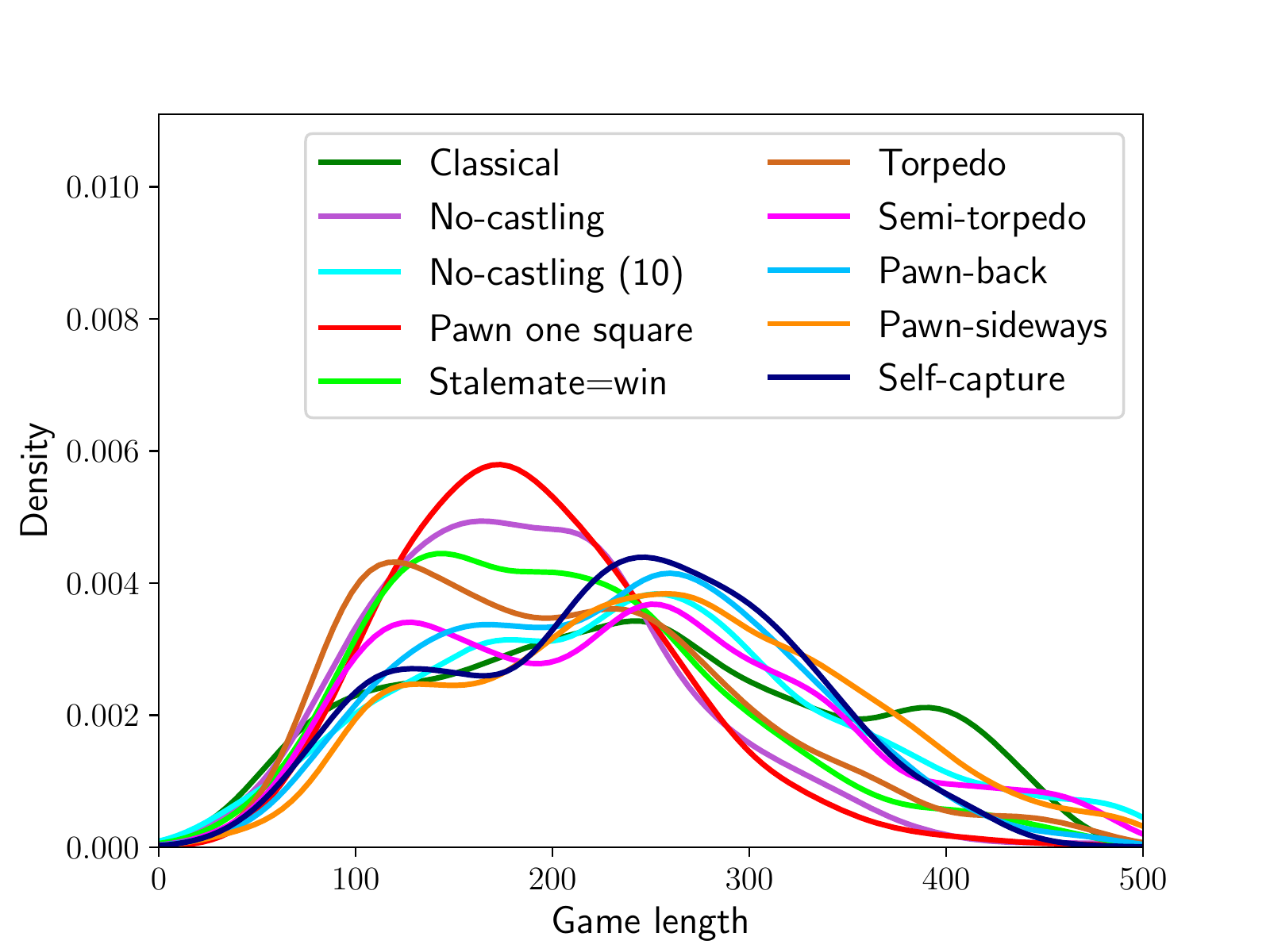}
\caption{The game length distributions of the total number of plies
for the subset of \emph{decisive} (not drawn) self-games for each variant.}
\label{fig:game-lengths-decisive}
\end{subfigure}
\caption{The game length distributions of the total number of plies of AlphaZero games in each chess variant, based on a sample of 10,000 games played at 1 second per move.
The experimental setup is described in Section \ref{sec:self-play-games}.
}
\label{fig:game-lengths}
\end{figure}

%%%%%%%%%%%%%%% Entropy breakdown %%%%%%%%%%%%%%%

\begin{figure*}[t]
\centering
\begin{subfigure}[t]{\columnwidth}
\centering\captionsetup{width=.95\columnwidth}
\includegraphics[width=\columnwidth]{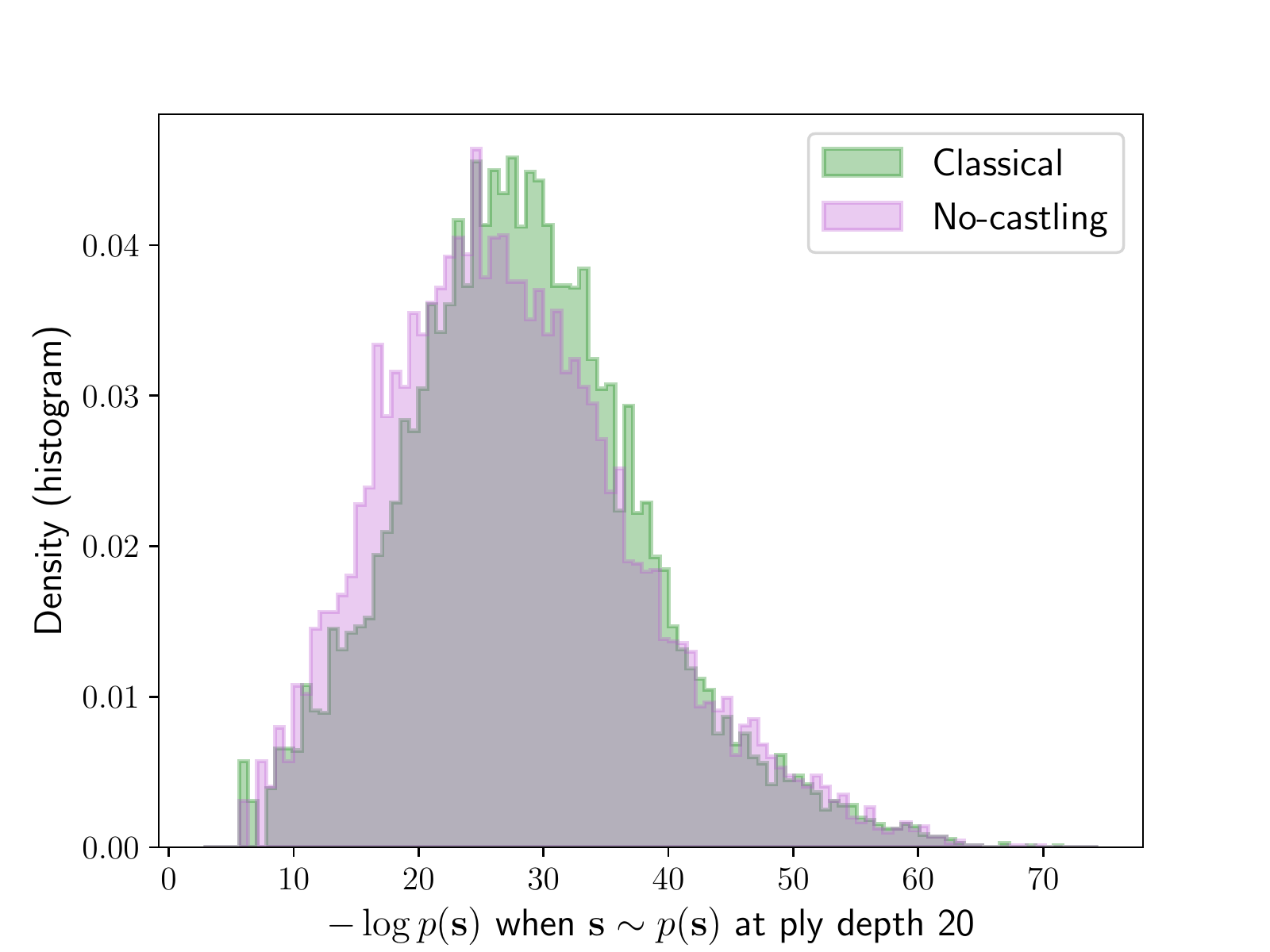}
\caption{No-castling and Classical chess}
\label{fig:entropy-breakdown-noc}
\end{subfigure}%
~ 
\begin{subfigure}[t]{\columnwidth}
\centering\captionsetup{width=.95\columnwidth}
\includegraphics[width=\columnwidth]{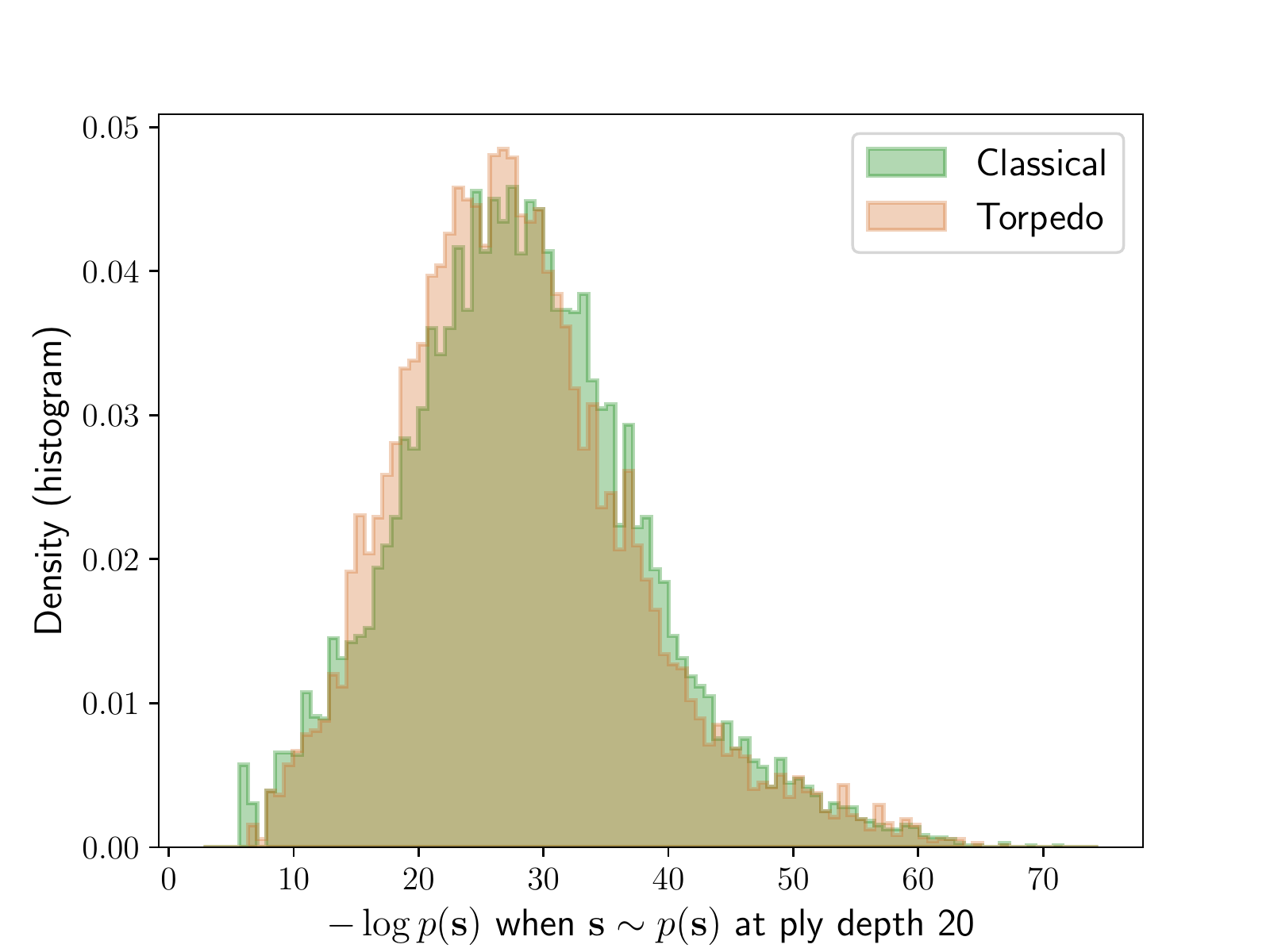}
\caption{Torpedo and Classical chess}
\label{fig:entropy-breakdown-t}
\end{subfigure}
\\
\begin{subfigure}[t]{\columnwidth}
\centering\captionsetup{width=.95\columnwidth}
\includegraphics[width=\columnwidth]{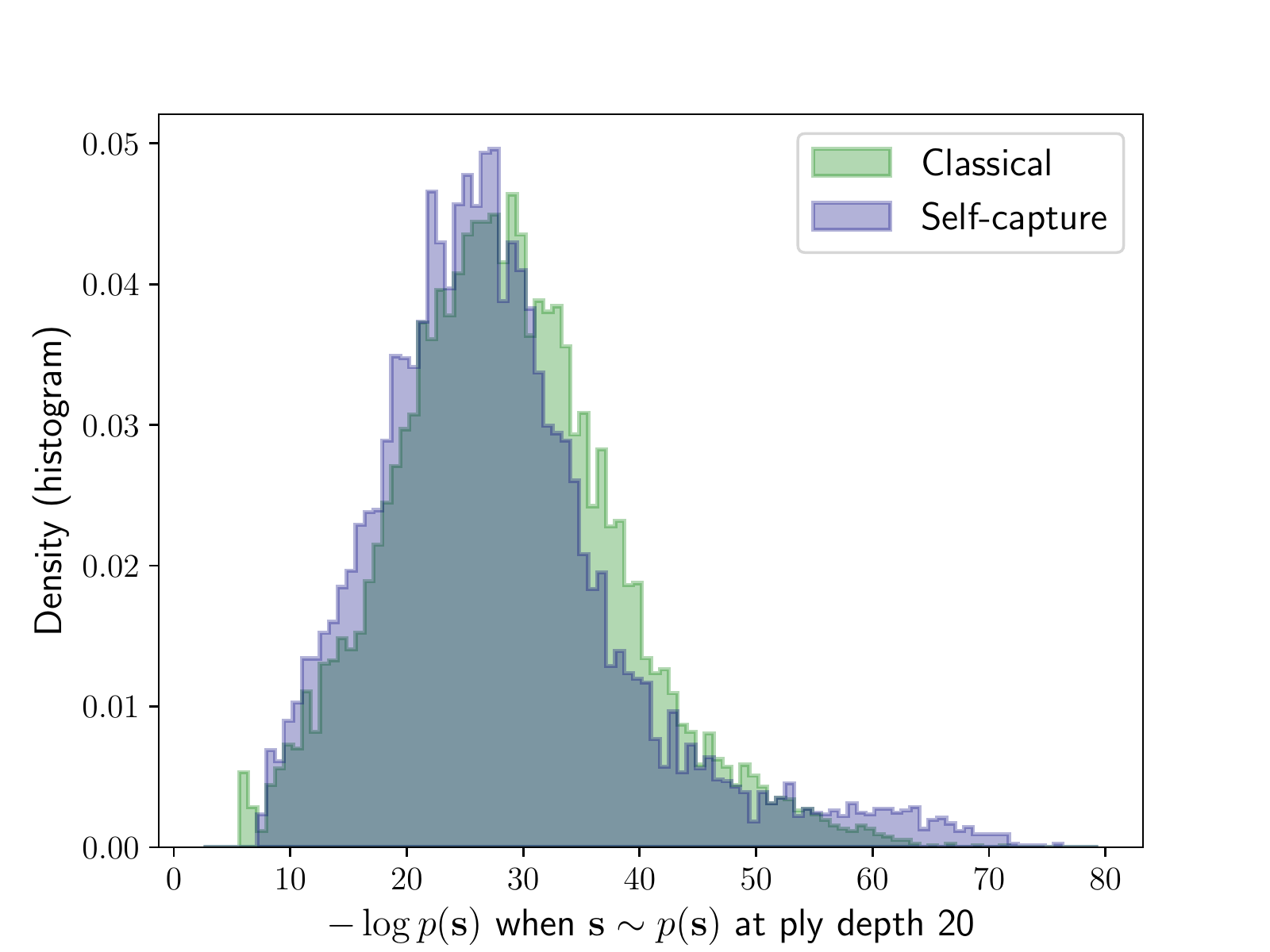}
\caption{Self-capture and Classical chess}
\label{fig:entropy-breakdown-sc}
\end{subfigure}%
~ 
\begin{subfigure}[t]{\columnwidth}
\centering\captionsetup{width=.95\columnwidth}
\includegraphics[width=\columnwidth]{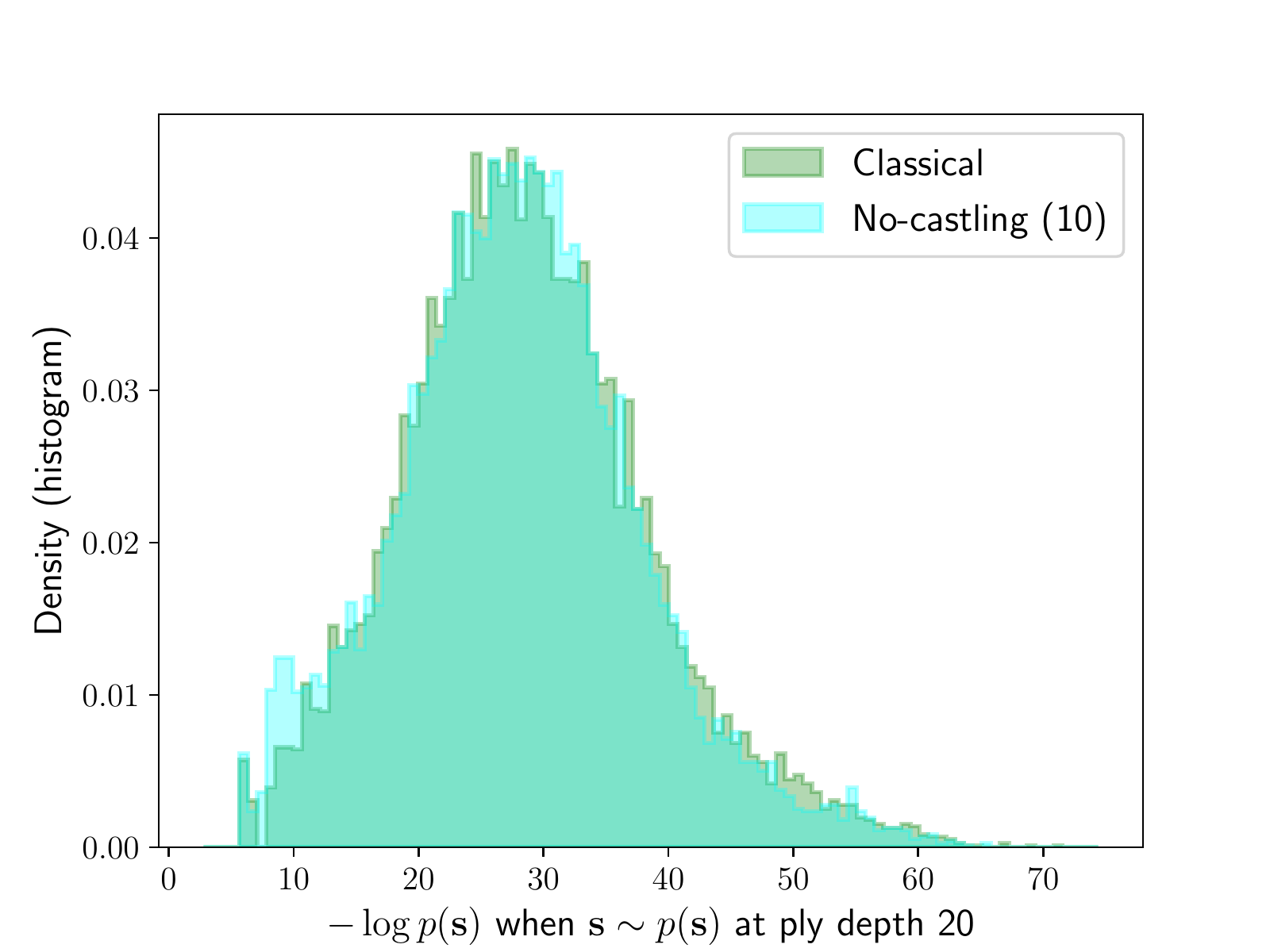}
\caption{No-castling (10) and Classical chess}
\label{fig:entropy-breakdown-noc10}
\end{subfigure}
\\
\begin{subfigure}[t]{\columnwidth}
\centering\captionsetup{width=.95\columnwidth}
\includegraphics[width=\columnwidth]{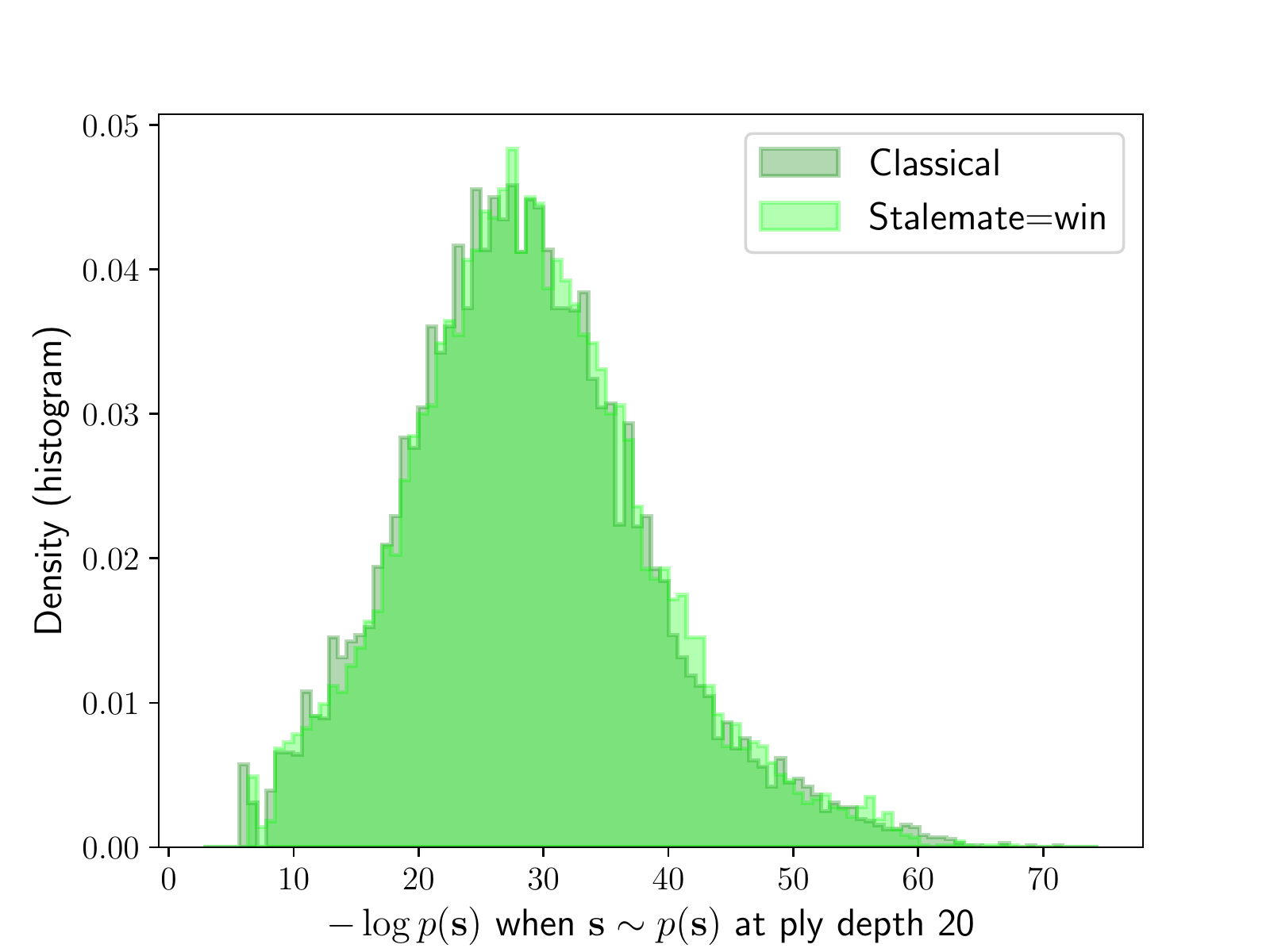}
\caption{Stalemate=win and Classical chess}
\label{fig:entropy-breakdown-sm}
\end{subfigure}%
~ 
\begin{subfigure}[t]{\columnwidth}
\centering\captionsetup{width=.95\columnwidth}
\includegraphics[width=\columnwidth]{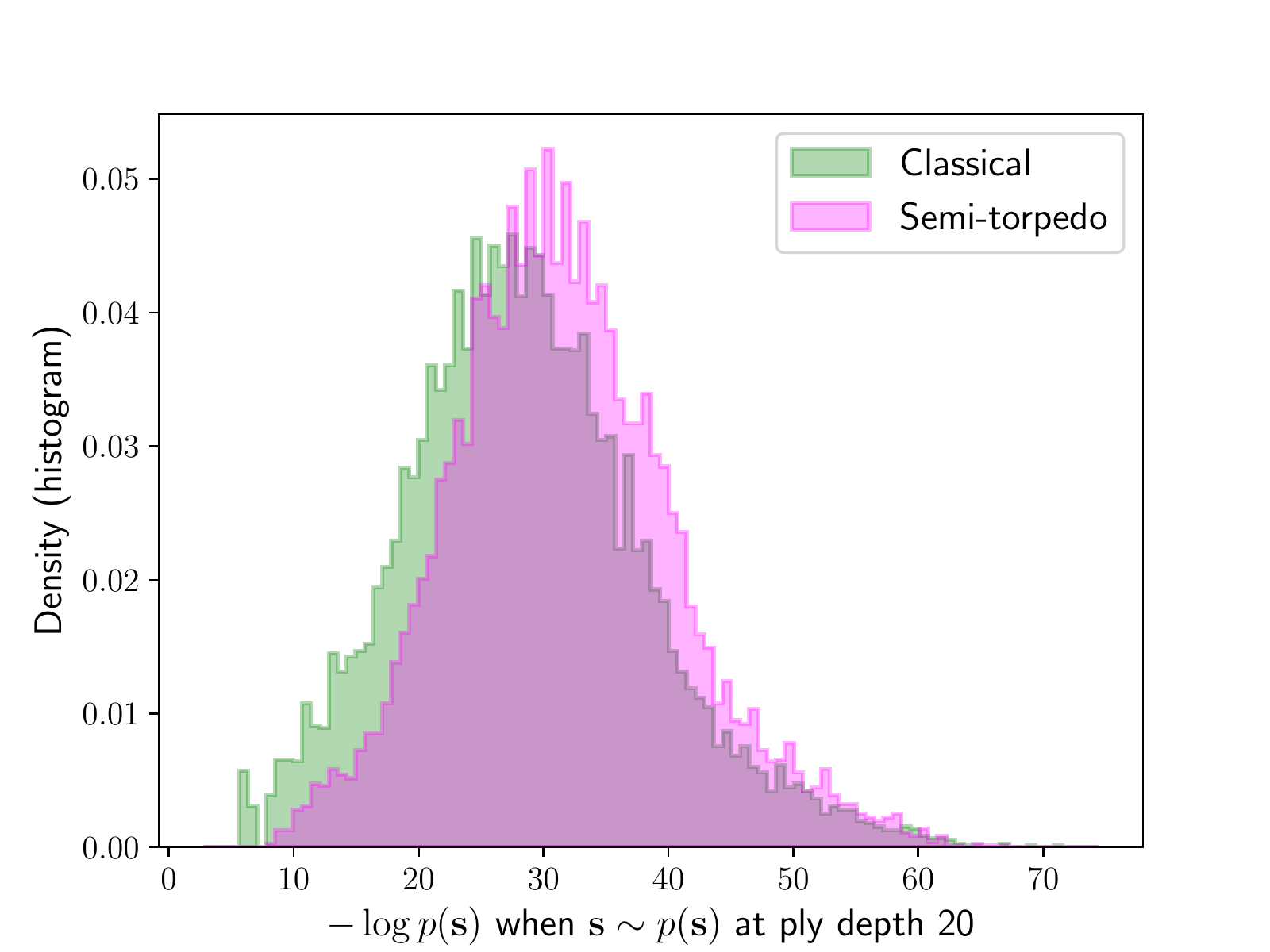}
\caption{Semi-torpedo and Classical chess}
\label{fig:entropy-breakdown-st}
\end{subfigure}
\caption{The density of (negative) log likelihoods for the prior opening lines for Classical chess and each of the variants.
The mean of each histogram gives the entropy or average information content for each variant's prior $p(\s)$, as given in \eqref{eq:entropy}.
The subfigures are ordered by entropy, following Table \ref{tab:entropy}.
\figref{fig:entropy-breakdown-bp} continues on the next page.
}
\end{figure*}

\addtocounter{figure}{-1}    %works

\begin{figure*}[t]
\vspace{-10pt}
\centering
\begin{subfigure}[t]{\columnwidth}
\addtocounter{subfigure}{6} % works here
\centering\captionsetup{width=.95\columnwidth}
\includegraphics[width=\columnwidth]{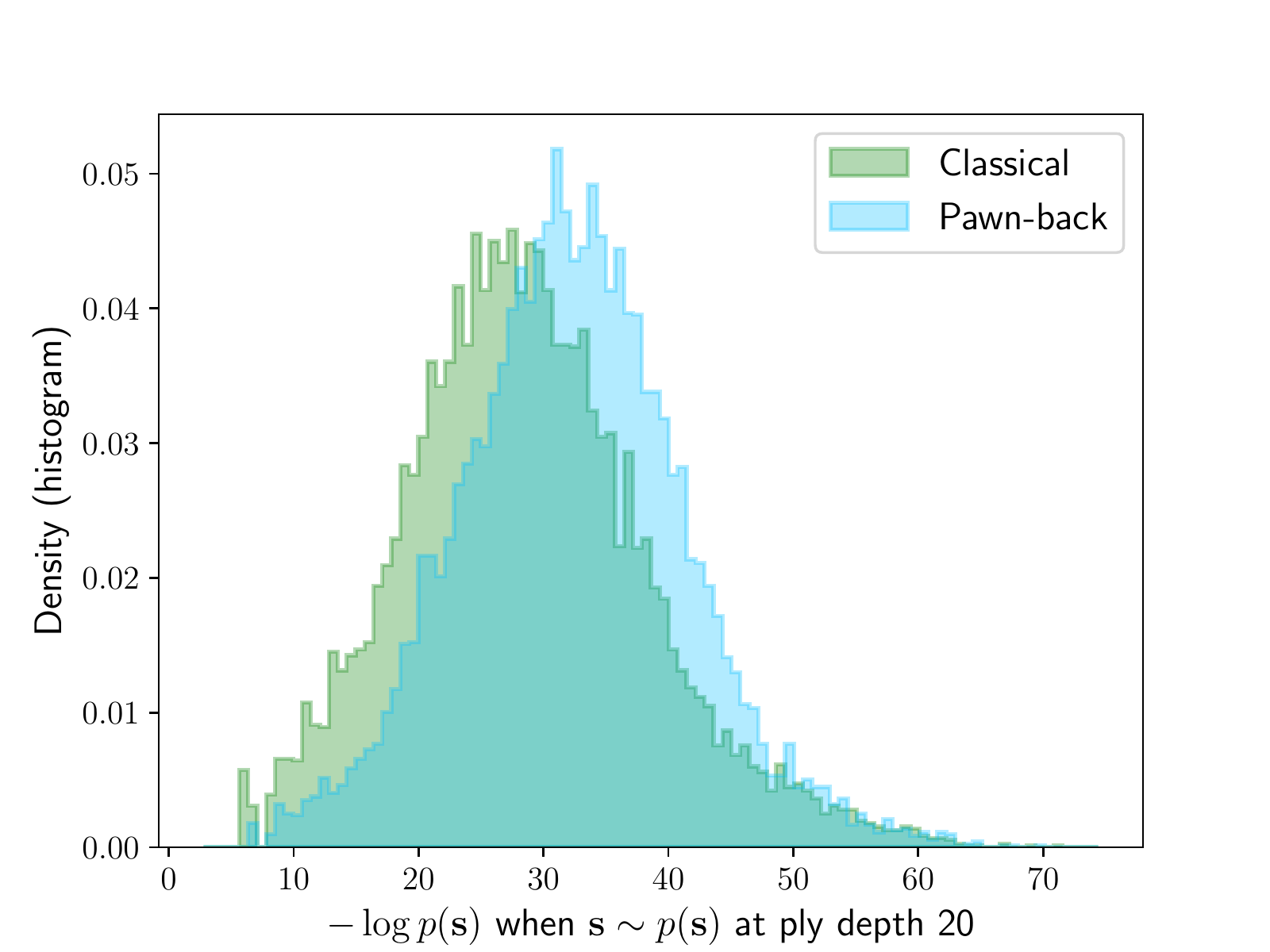}
\caption{Pawn-back and Classical chess}
\label{fig:entropy-breakdown-bp}
\end{subfigure}%
~ 
\begin{subfigure}[t]{\columnwidth}
\centering\captionsetup{width=.95\columnwidth}
\includegraphics[width=\columnwidth]{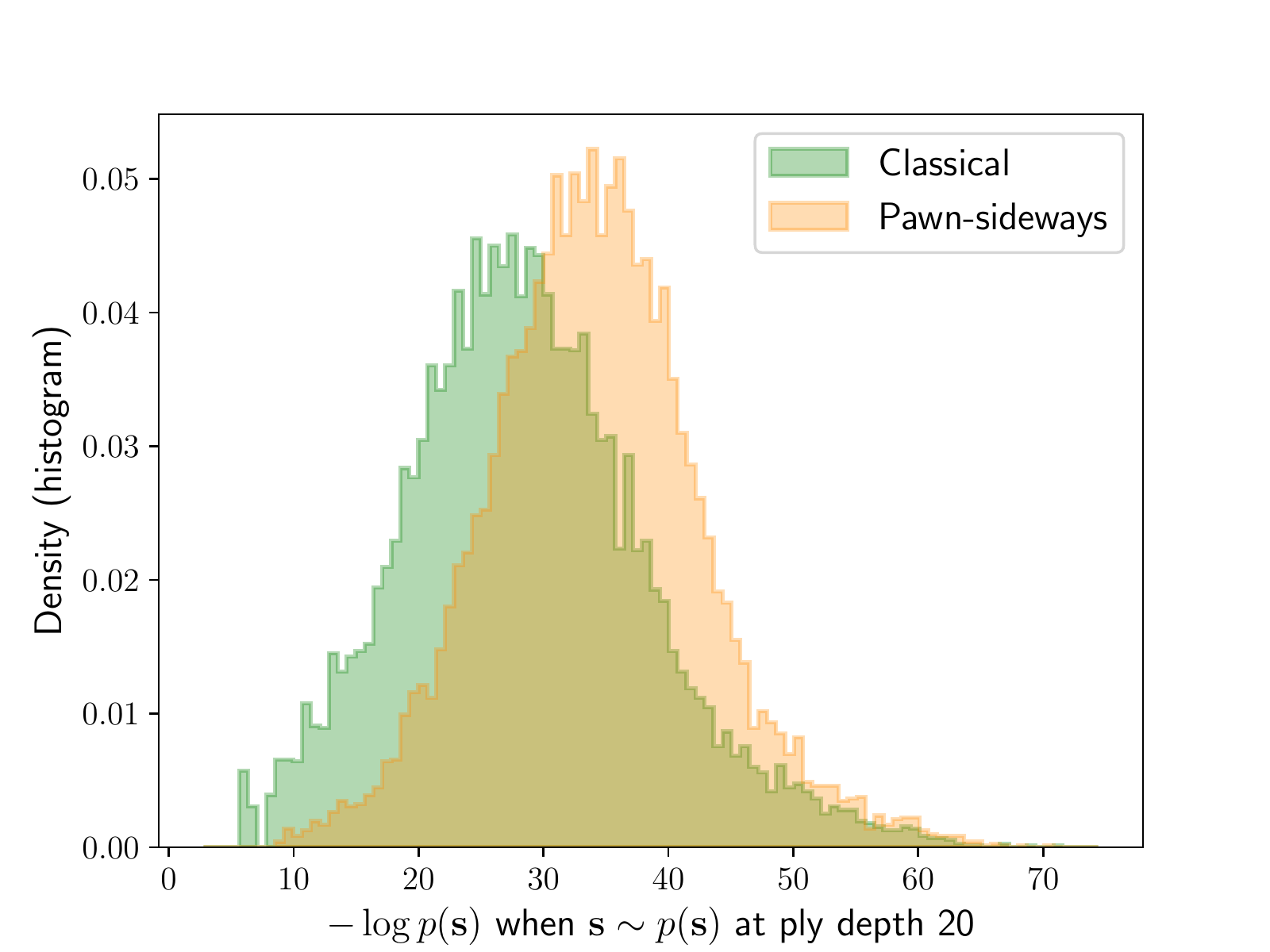}
\caption{Pawn-sideways and Classical chess}
\label{fig:entropy-breakdown-ps}
\end{subfigure}
\\
\begin{subfigure}[t]{\columnwidth}
\centering\captionsetup{width=.95\columnwidth}
\includegraphics[width=\columnwidth]{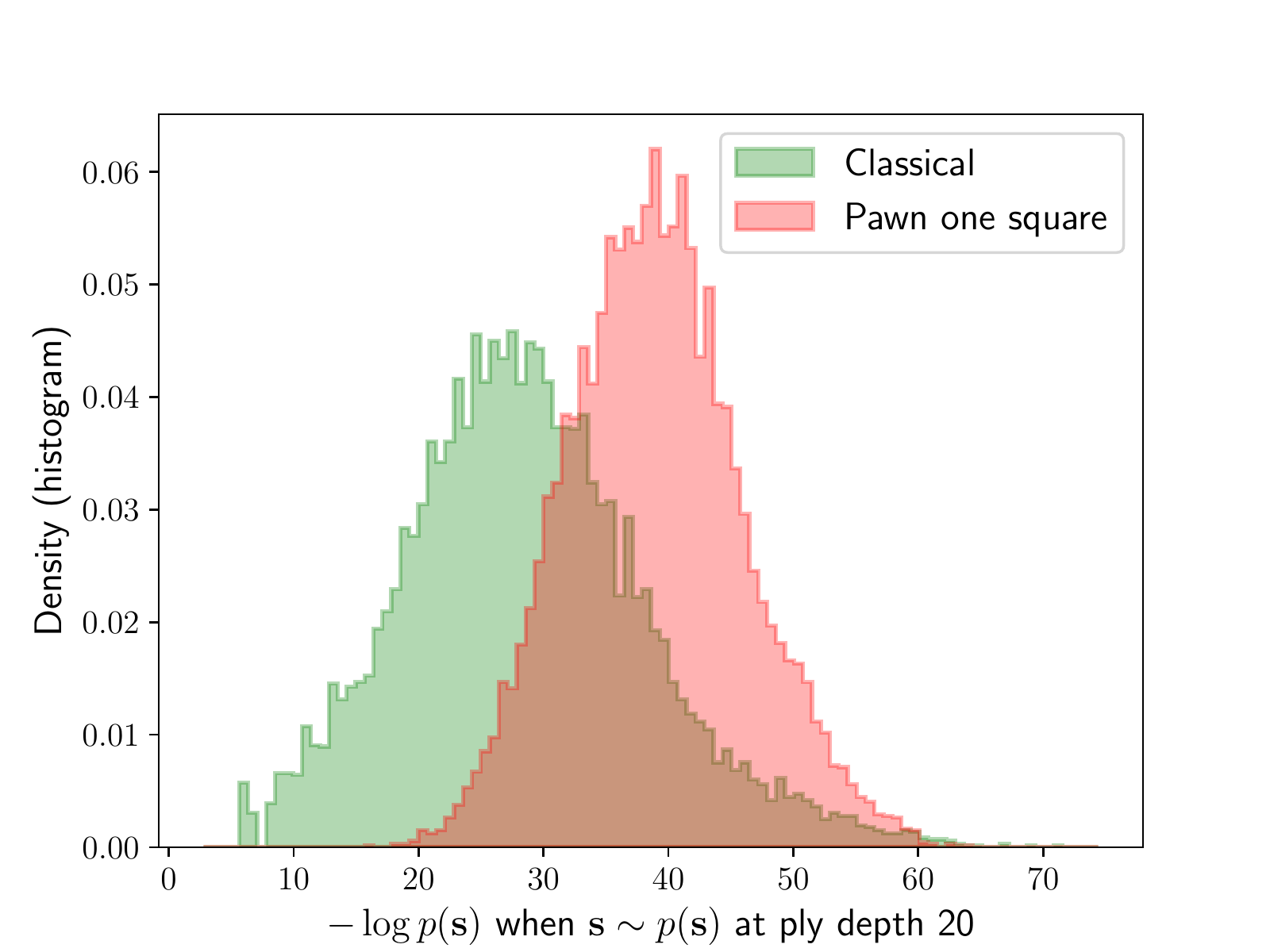}
\caption{Pawn one square and Classical chess}
\label{fig:entropy-breakdown-p1}
\end{subfigure}%
~ 
\begin{subfigure}[t]{\columnwidth}
\centering\captionsetup{width=.95\columnwidth}
\includegraphics[width=\columnwidth]{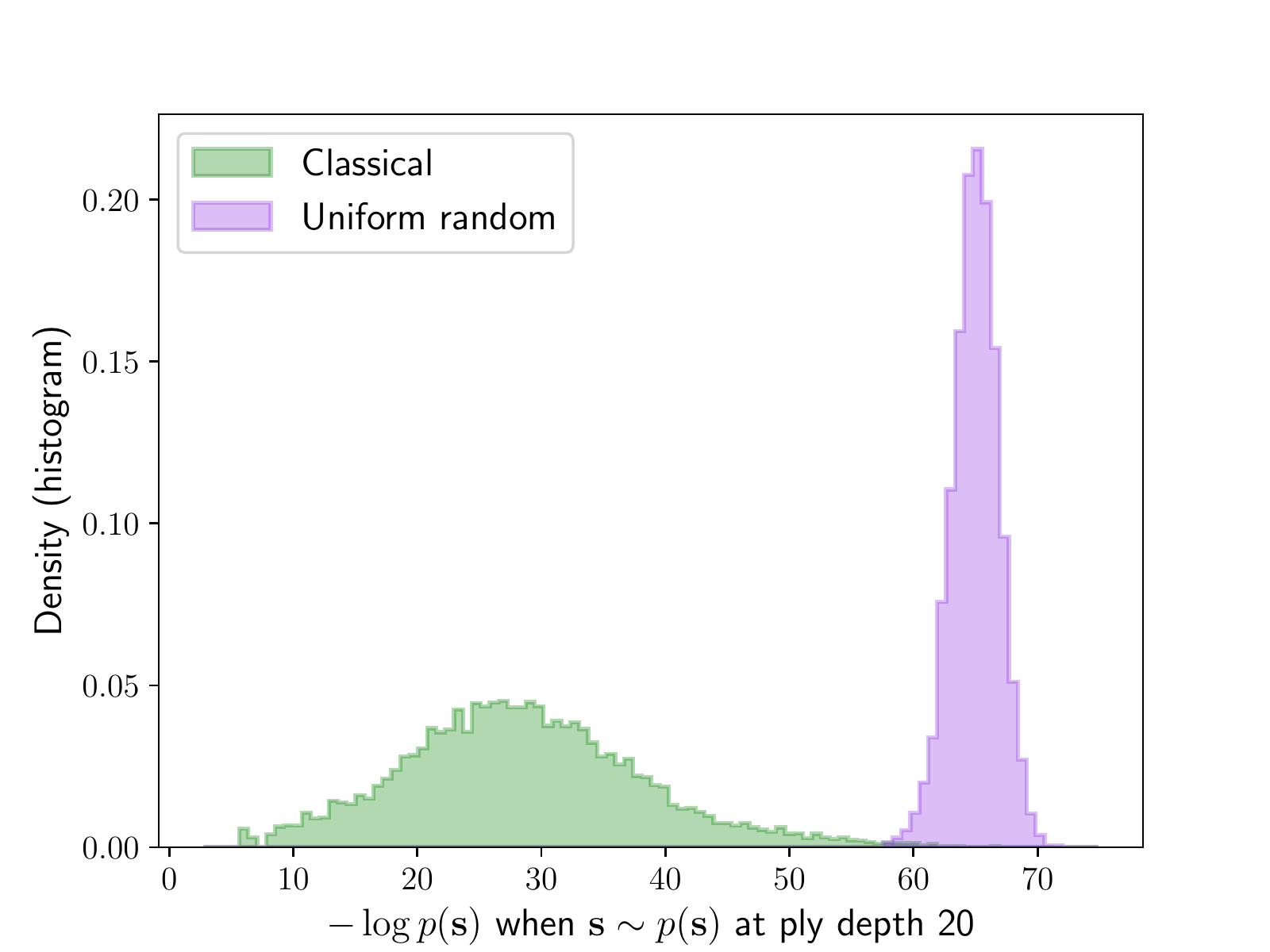}
\caption{Uniform random classical moves and Classical chess}
\label{fig:entropy-breakdown-u}
\end{subfigure}
\caption{\emph{(Continued from previous page.)}
The density of (negative) log likelihoods for the prior opening lines for Classical chess and each of the variants.
The mean of each histogram gives the entropy or average information content for each variant's prior $p(\s)$, as given in \eqref{eq:entropy}.
The subfigures are ordered by entropy, following Table \ref{tab:entropy}.
}
\label{fig:entropy-breakdown}
\end{figure*}

%%%%%%%%%%%%%%%%%%%%%%%%%%%%%%%%%%%%%%%%%%%%%%%%%%%%%%
%%%%%%%%% KL Divergence %%%%%%%%%%

\begin{figure*}[h!]
\vspace{-10pt}
\centering
\begin{subfigure}[t]{\columnwidth}
\centering\captionsetup{width=.95\columnwidth}
\includegraphics[width=\columnwidth]{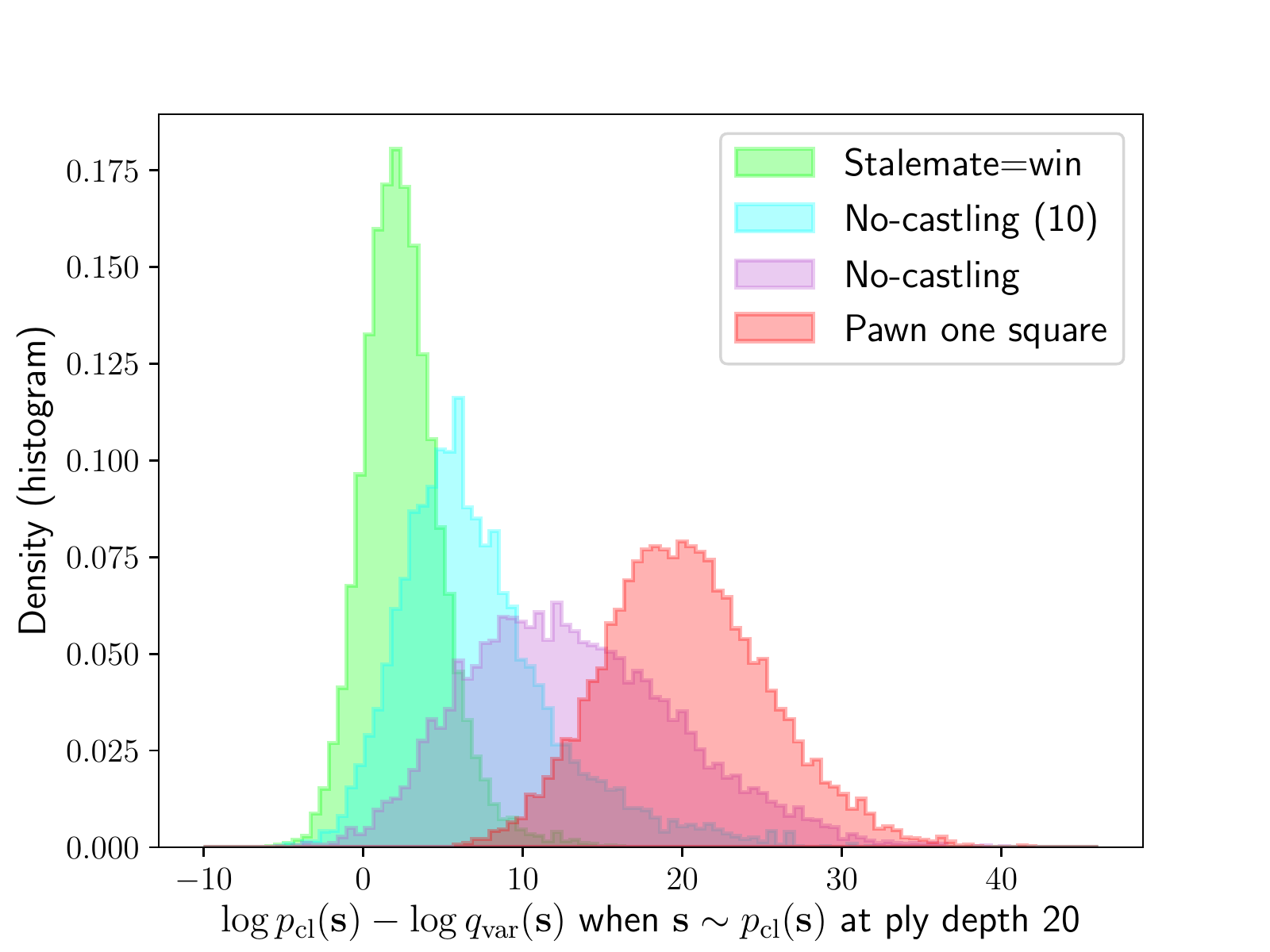}
\caption{
A decomposition of the entropy of subset variants of Classical chess relative to Classical chess.}
\label{fig:kl-subsets}
\end{subfigure}%
~ 
\begin{subfigure}[t]{\columnwidth}
\centering\captionsetup{width=.95\columnwidth}
\includegraphics[width=\columnwidth]{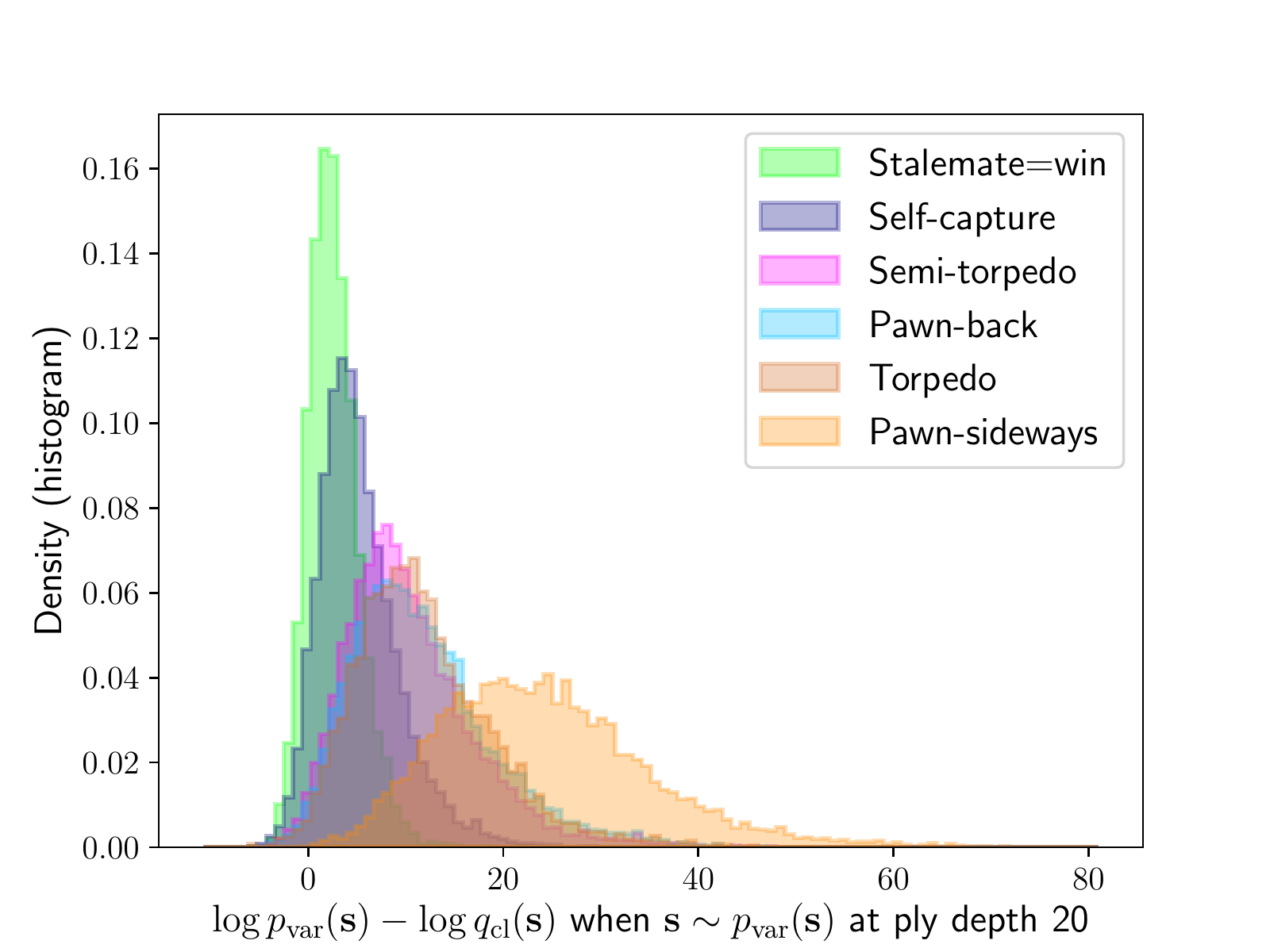}
\caption{A decomposition of the entropy of Classical chess relative to its superset variants.}
\label{fig:kl-supersets}
\end{subfigure}
\caption{Histograms of the density of terms $\log p(\s) - \log q(\s)$
whose mean under $p(\s)$ is the Kullback-Leibler divergence
in \eqref{eq:kl}.}
\label{fig:kl}
\end{figure*}

%%%%%%%%%%%%%%%%%%%%%%%%%%%%%%%%%%%%%%%%%%%%%%%%%%%%%%
%%%%%%%%% Average number of candidate moves %%%%%%%%%%

\begin{figure*}[t]
\centering
\begin{subfigure}[t]{\columnwidth}
\centering\captionsetup{width=.95\columnwidth}
\includegraphics[width=\columnwidth]{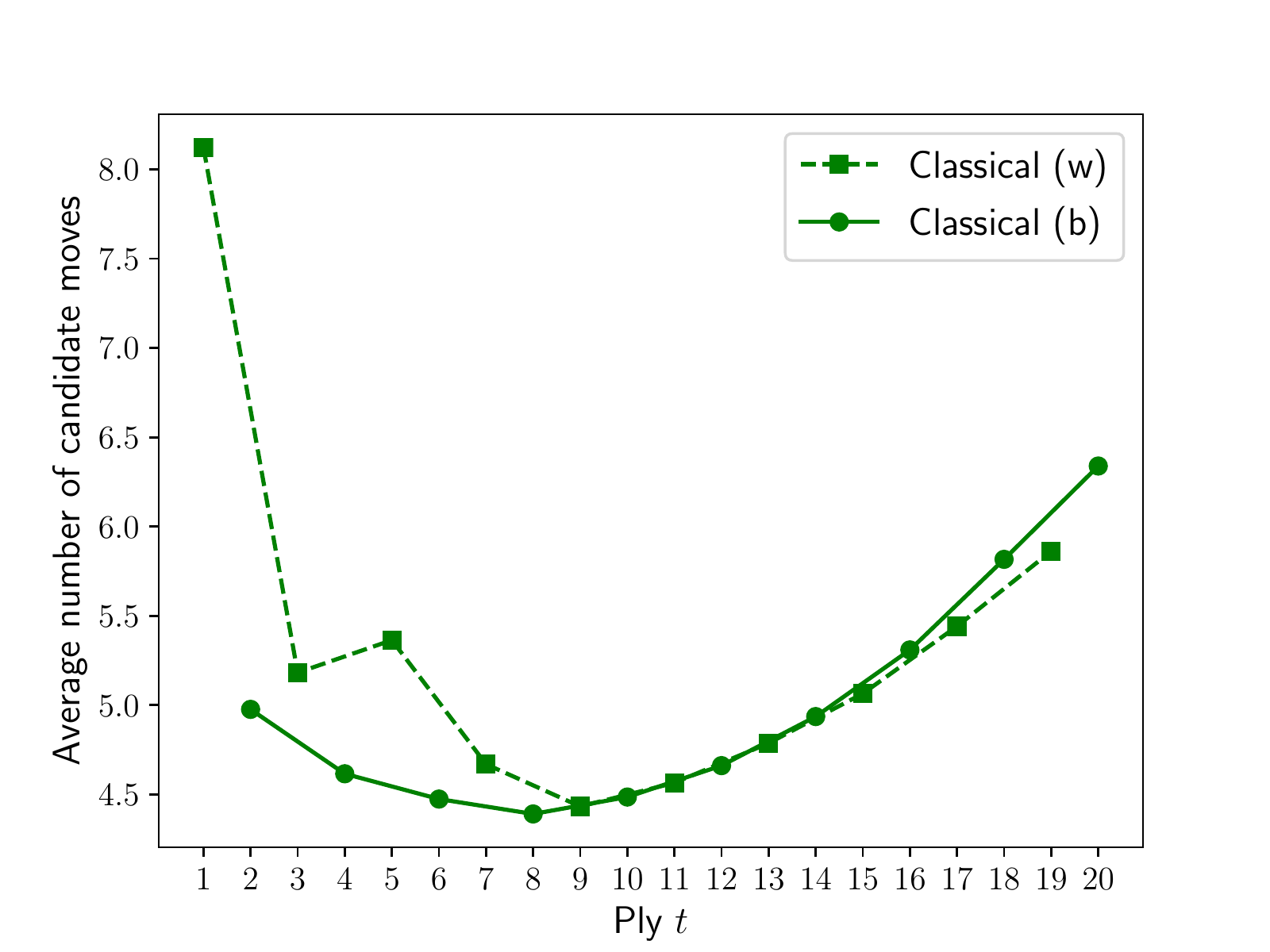}
\caption{Classical chess}
\label{fig:average-candidates-cl}
\end{subfigure}%
~ 
\begin{subfigure}[t]{\columnwidth}
\centering\captionsetup{width=.95\columnwidth}
\includegraphics[width=\columnwidth]{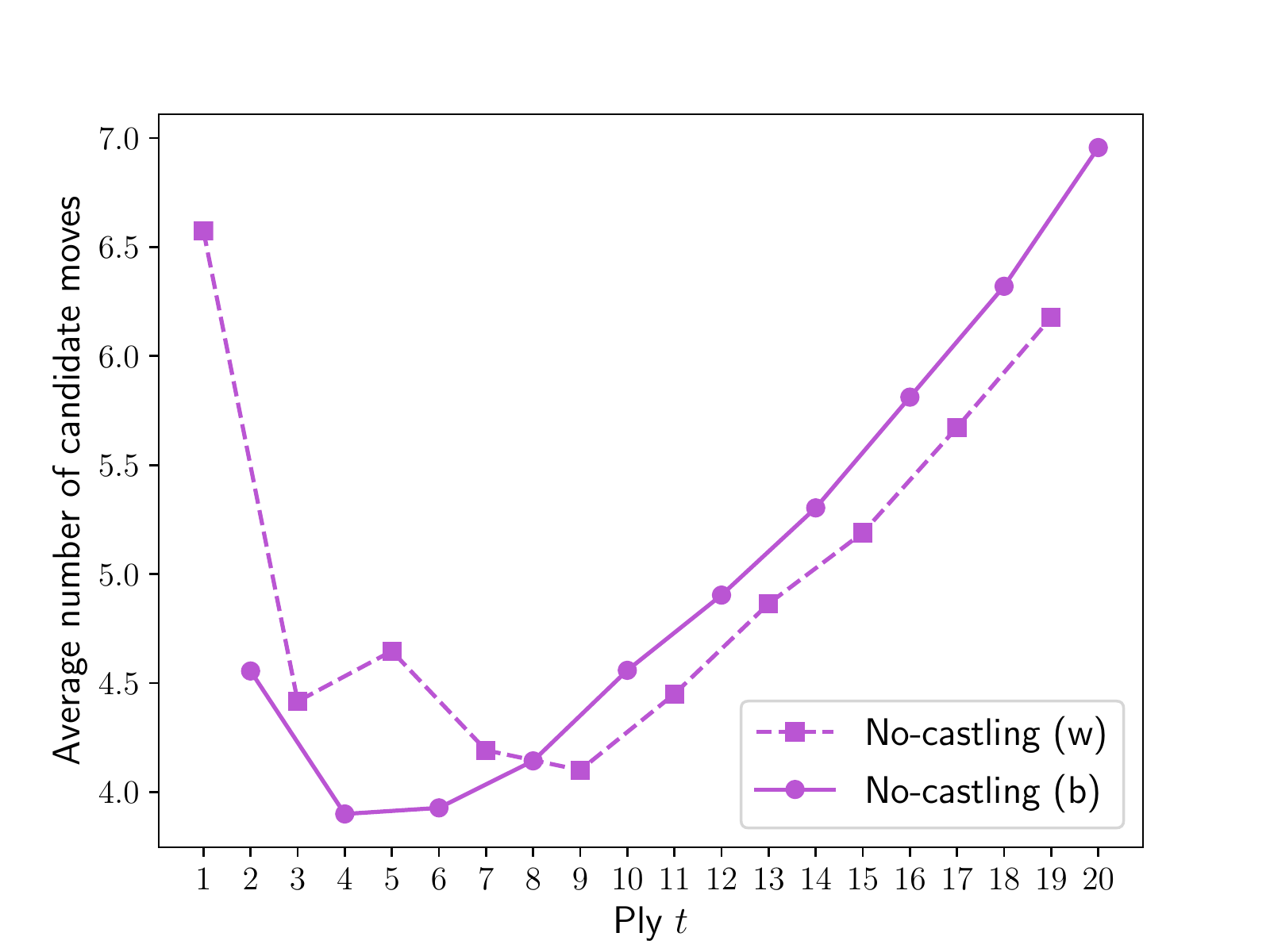}
\caption{No-castling chess}
\label{fig:average-candidates-noc}
\end{subfigure}
\\
\begin{subfigure}[t]{\columnwidth}
\centering\captionsetup{width=.95\columnwidth}
\includegraphics[width=\columnwidth]{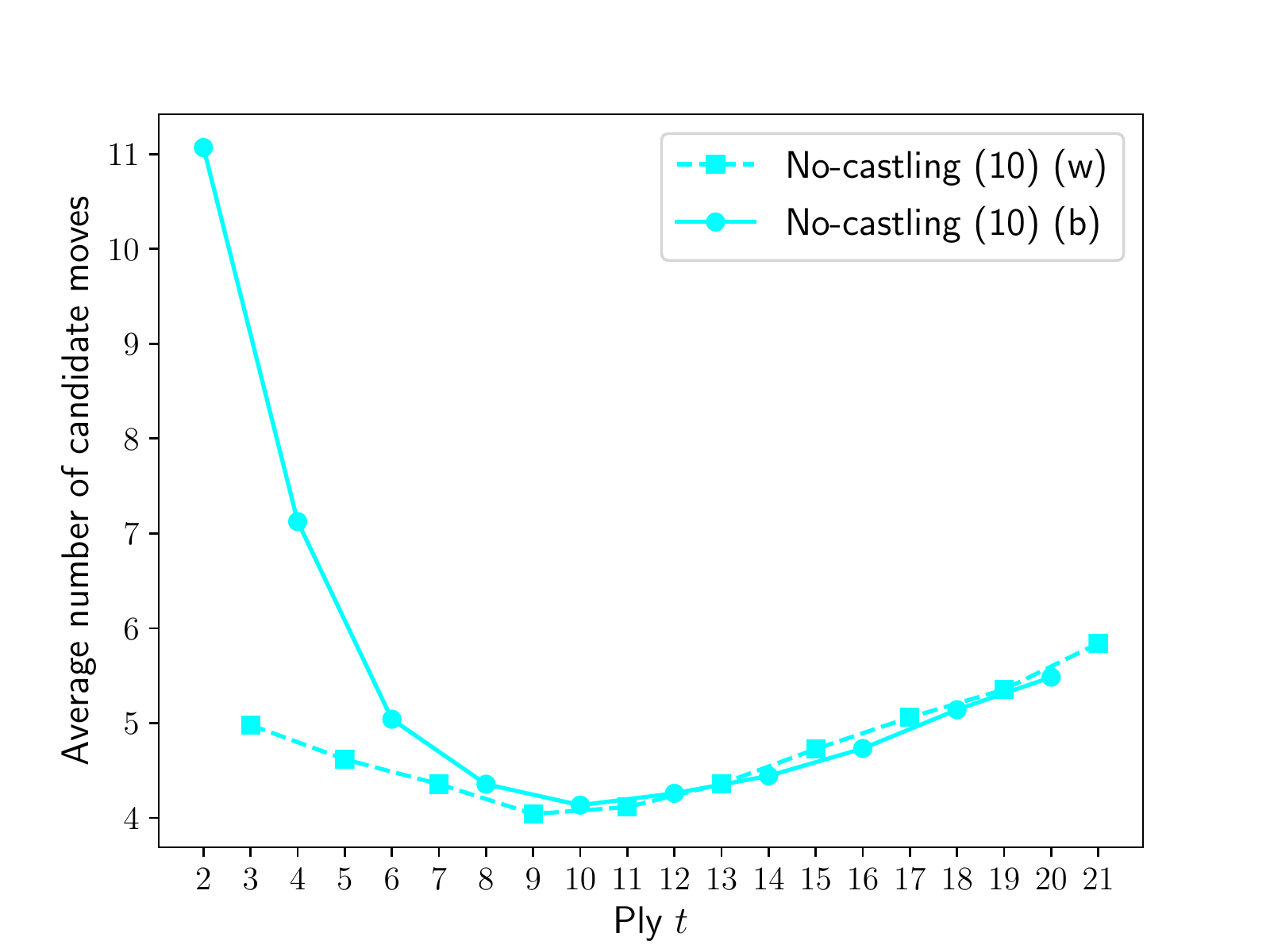}
\caption{No-castling (10) chess}
\label{fig:average-candidates-noc10}
\end{subfigure}%
~ 
\begin{subfigure}[t]{\columnwidth}
\centering\captionsetup{width=.95\columnwidth}
\includegraphics[width=\columnwidth]{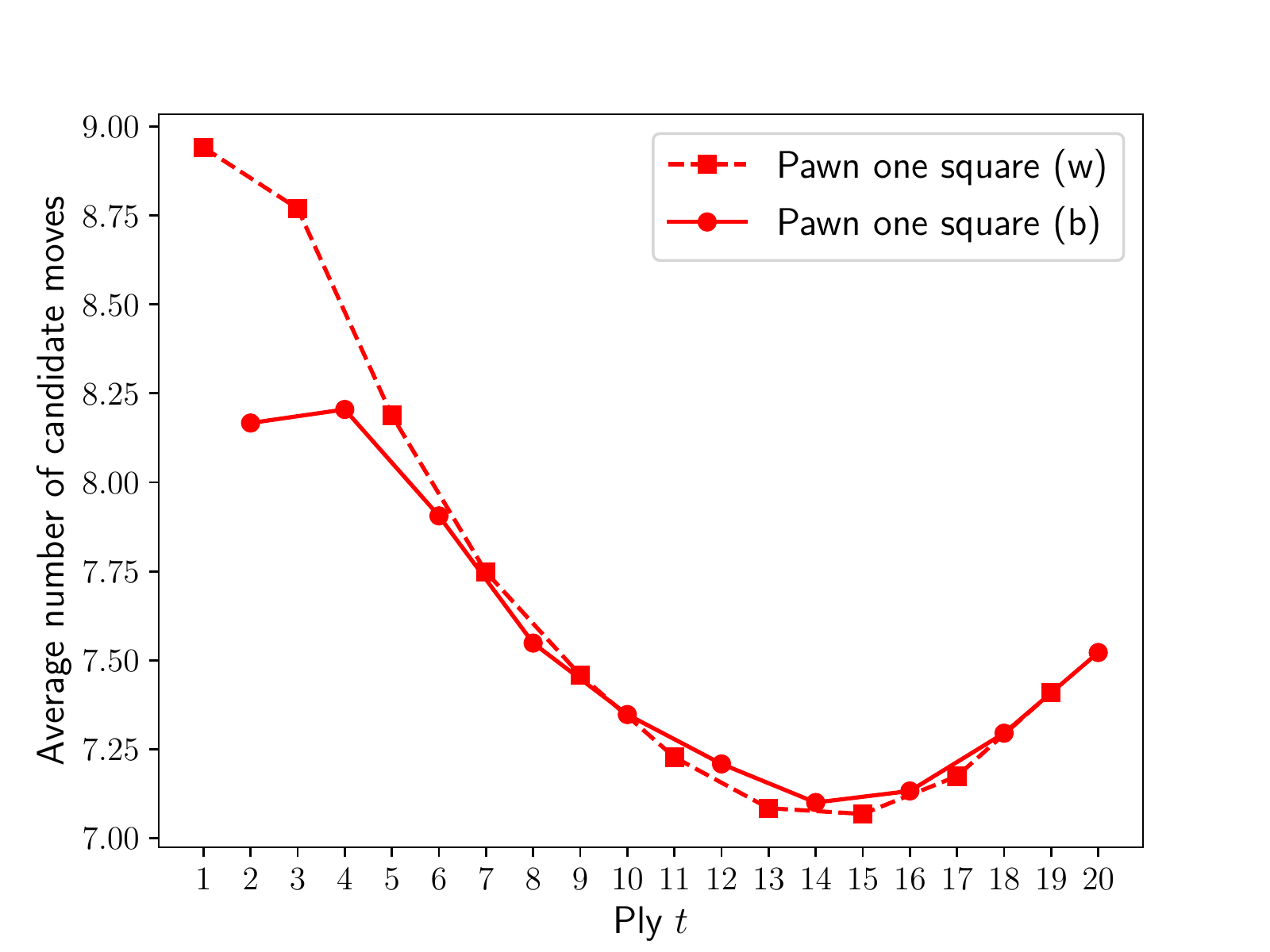}
\caption{Pawn one square chess}
\label{fig:average-candidates-p1}
\end{subfigure}
\\
\begin{subfigure}[t]{\columnwidth}
\centering\captionsetup{width=.95\columnwidth}
\includegraphics[width=\columnwidth]{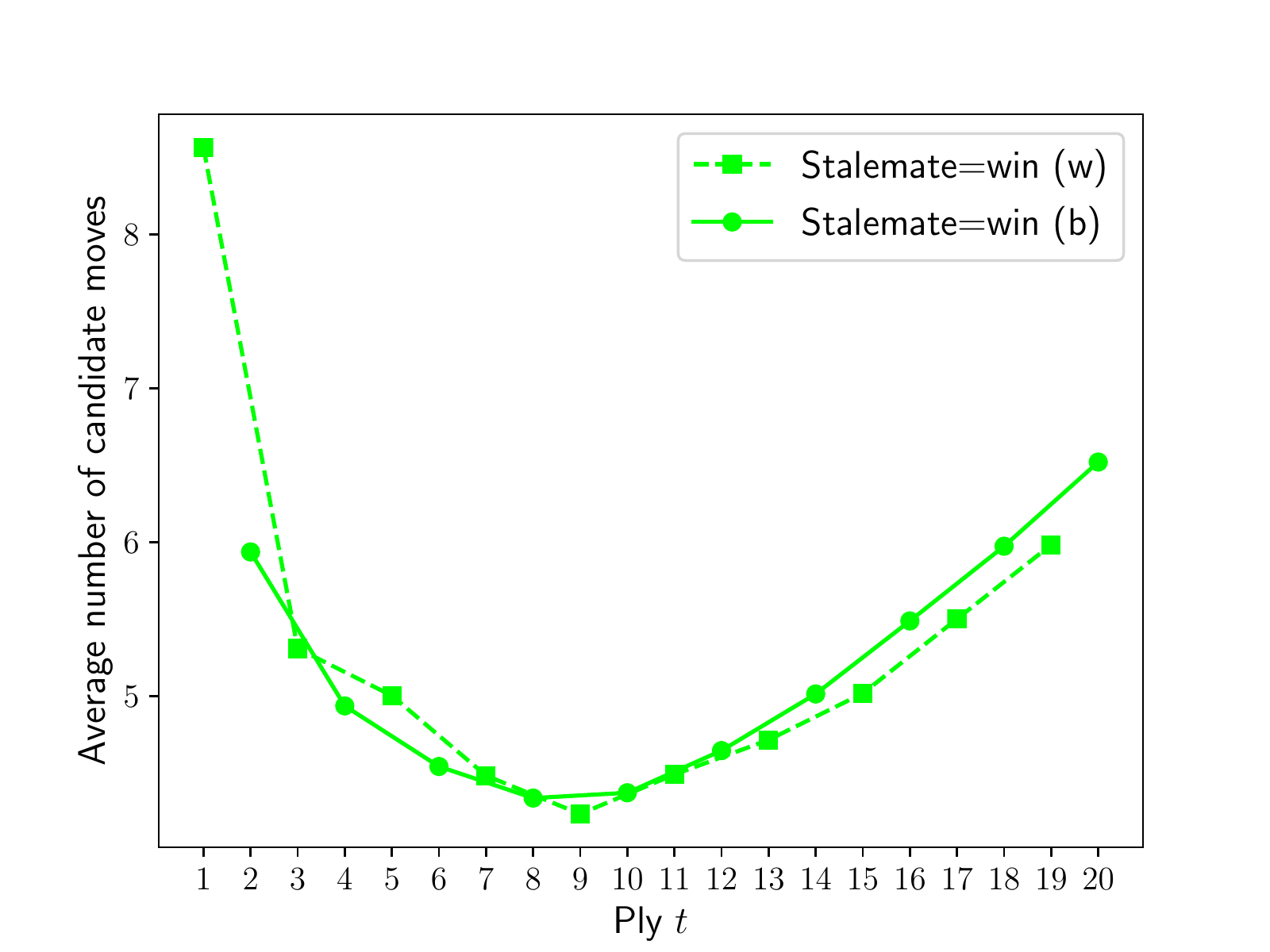}
\caption{Stalemate=win chess}
\label{fig:average-candidates-sm}
\end{subfigure}%
~ 
\begin{subfigure}[t]{\columnwidth}
\centering\captionsetup{width=.95\columnwidth}
\includegraphics[width=\columnwidth]{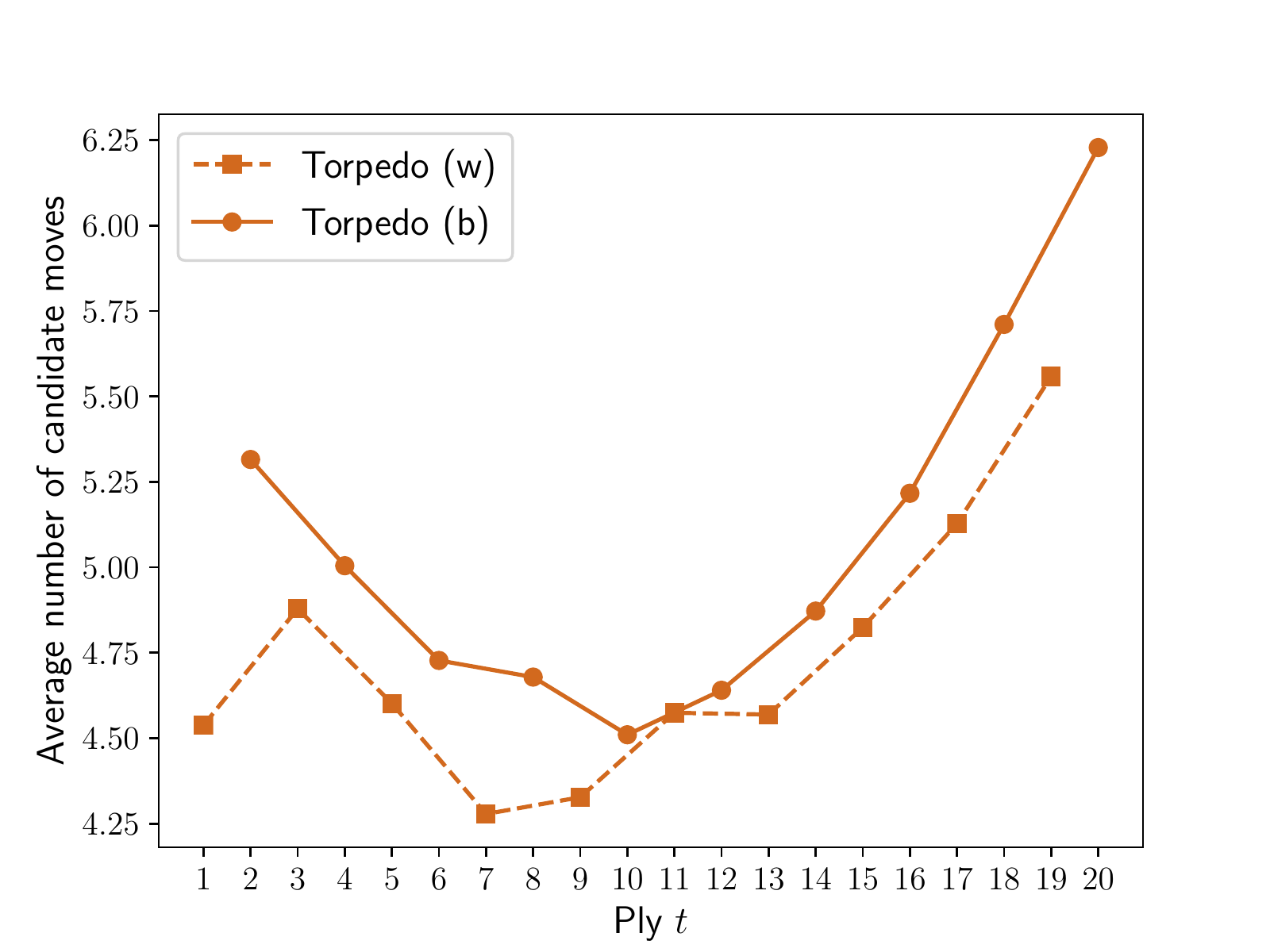}
\caption{Torpedo chess}
\label{fig:average-candidates-t}
\end{subfigure}
\caption{The average number of candidate moves $\Mcal(t)$ from
\eqref{eq:average-candidates} for each of the variants, as computed from their prior distributions $p(\s)$.
\figref{fig:average-candidates-st} continues on the next page.}
\end{figure*}

\addtocounter{figure}{-1}    %works

\begin{figure*}[t]
\centering
\begin{subfigure}[t]{\columnwidth}
\addtocounter{subfigure}{6} % works here
\centering\captionsetup{width=.95\columnwidth}
\includegraphics[width=\columnwidth]{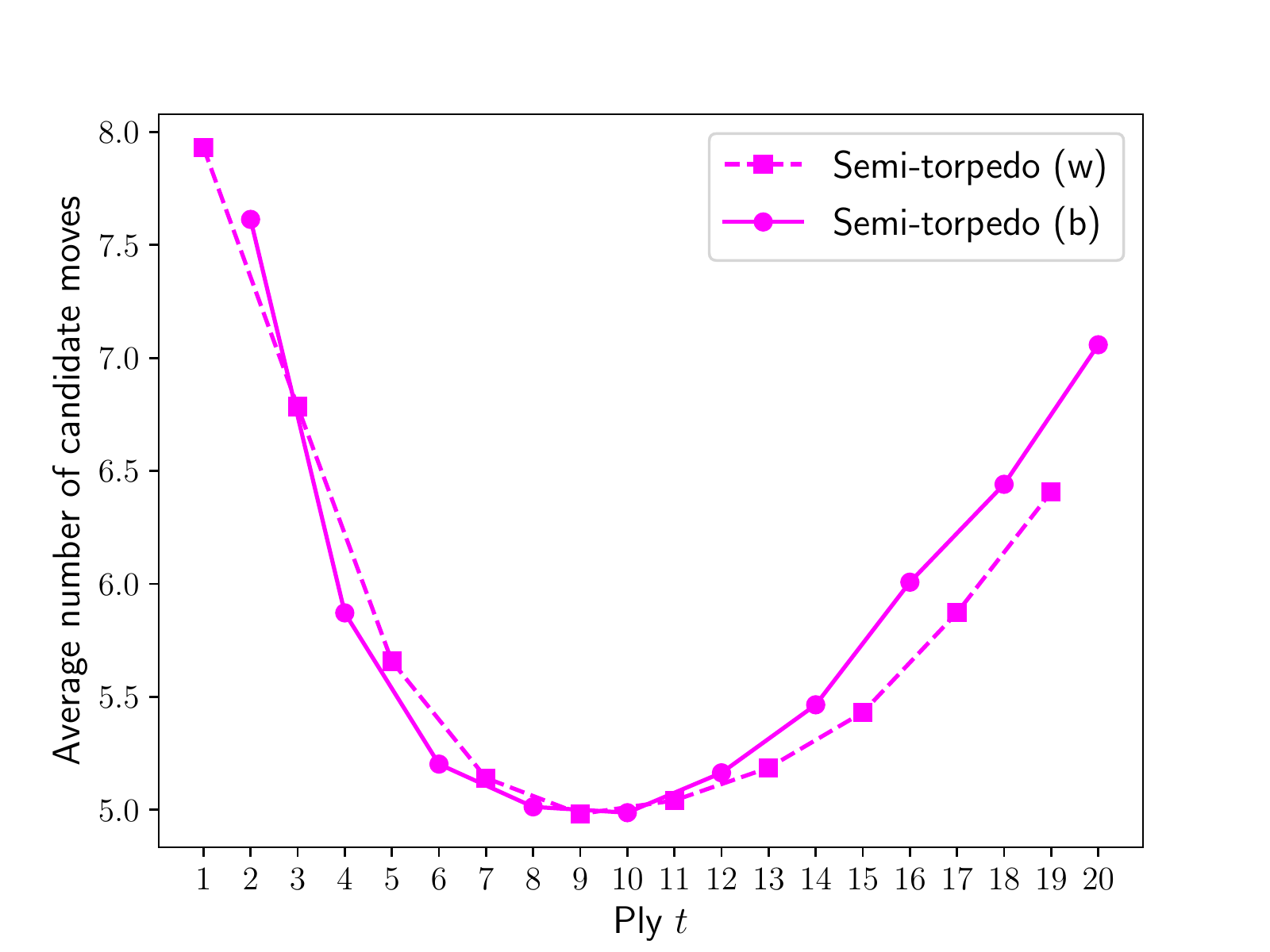}
\caption{Semi-torpedo chess}
\label{fig:average-candidates-st}
\end{subfigure}%
~ 
\begin{subfigure}[t]{\columnwidth}
\centering\captionsetup{width=.95\columnwidth}
\includegraphics[width=\columnwidth]{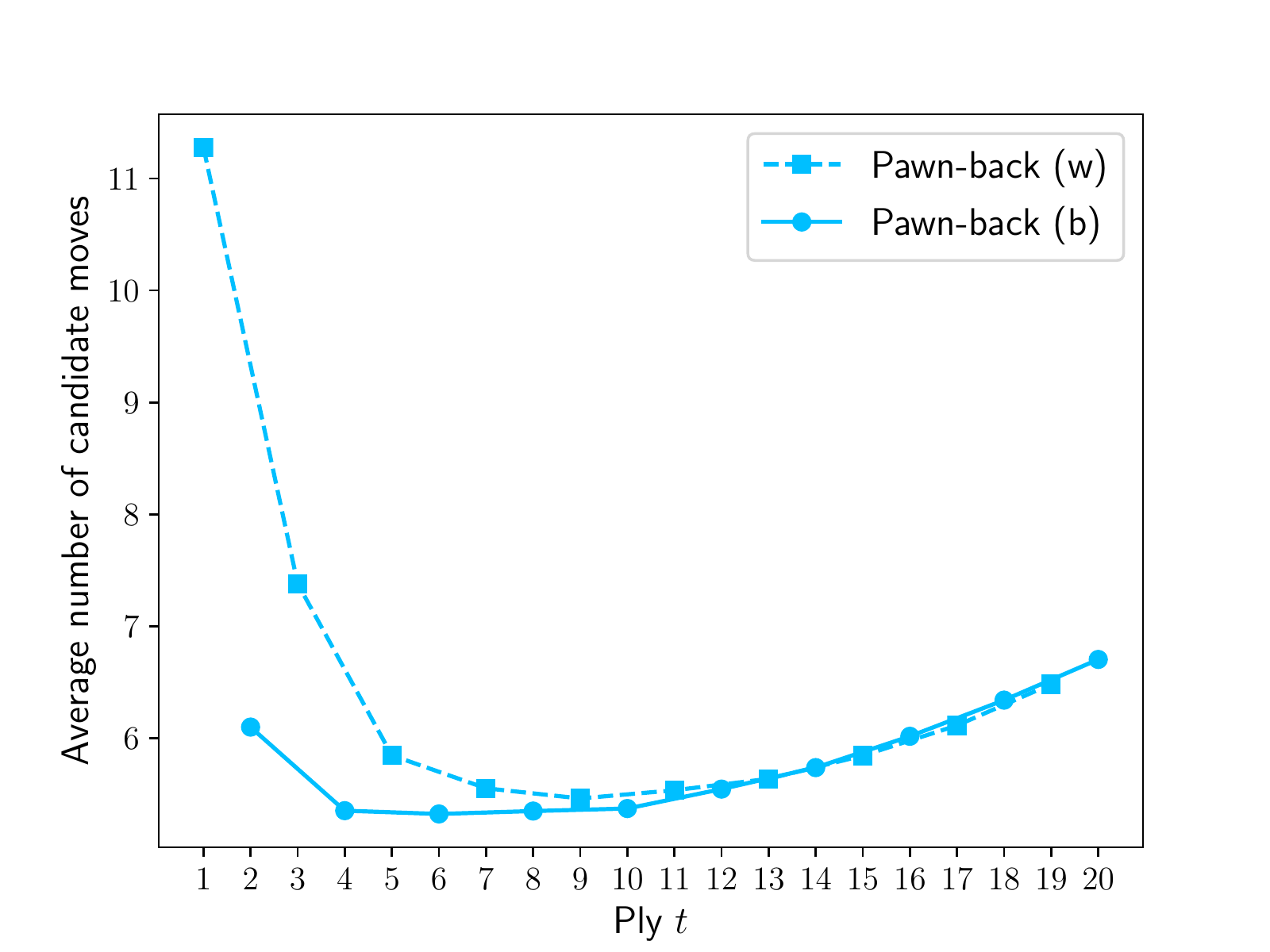}
\caption{Pawn-back chess}
\label{fig:average-candidates-pb}
\end{subfigure}
\\
\begin{subfigure}[t]{\columnwidth}
\centering\captionsetup{width=.95\columnwidth}
\includegraphics[width=\columnwidth]{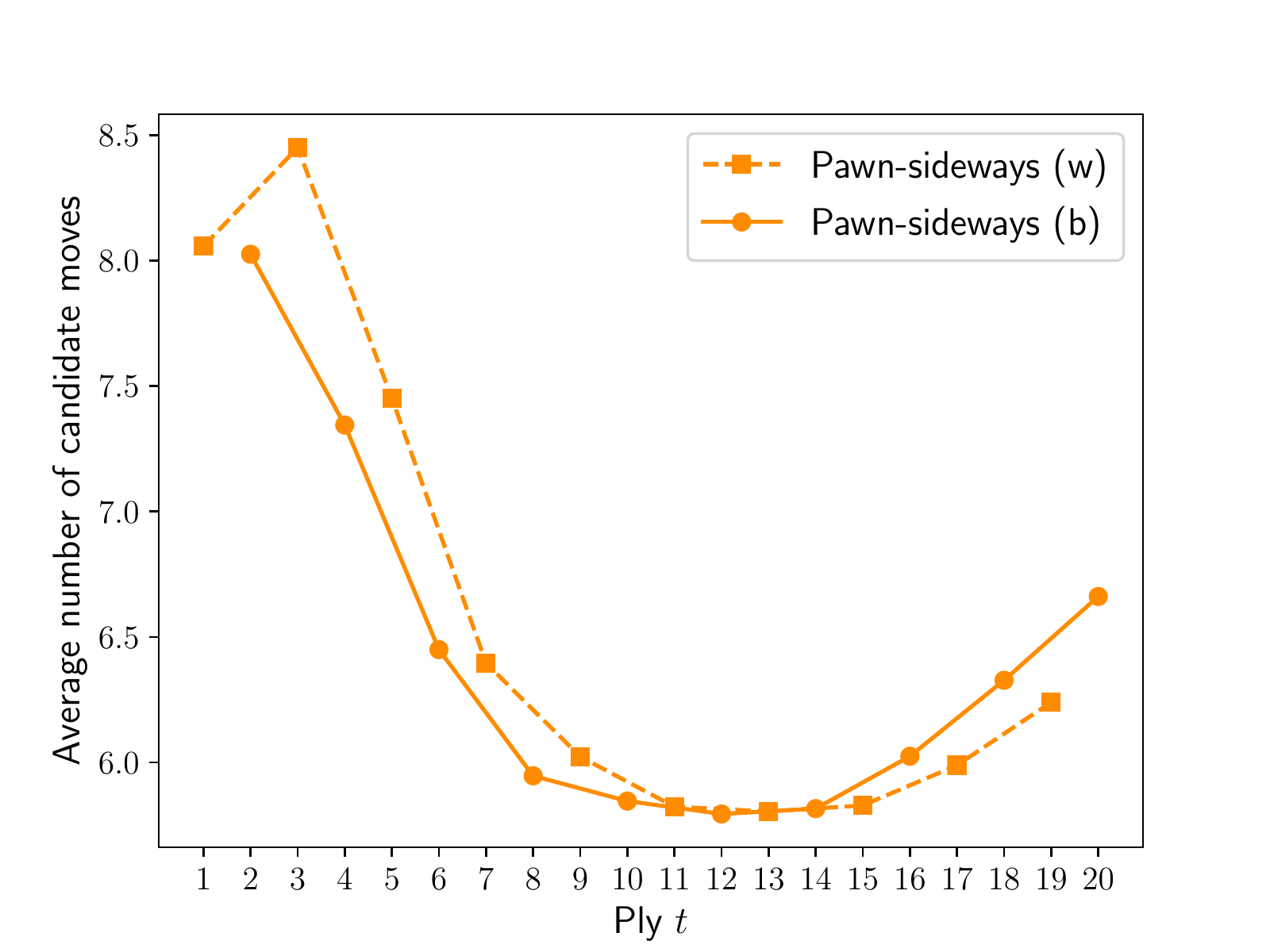}
\caption{Pawn-sideways chess}
\label{fig:average-candidates-ps}
\end{subfigure}%
~ 
\begin{subfigure}[t]{\columnwidth}
\centering\captionsetup{width=.95\columnwidth}
\includegraphics[width=\columnwidth]{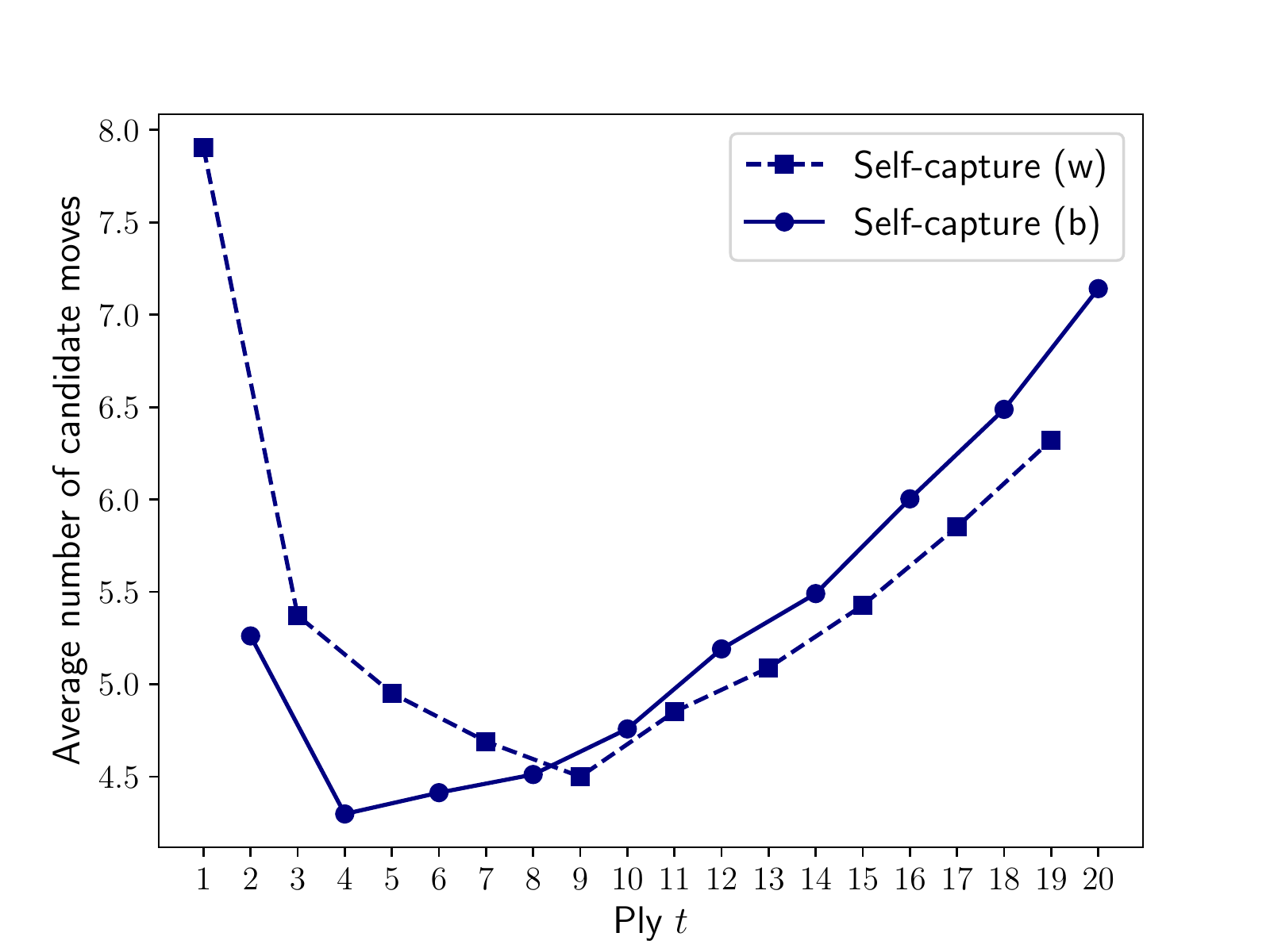}
\caption{Self-capture chess}
\label{fig:average-candidates-sc}
\end{subfigure}
\\
\begin{subfigure}[t]{\columnwidth}
\centering\captionsetup{width=.95\columnwidth}
\includegraphics[width=\columnwidth]{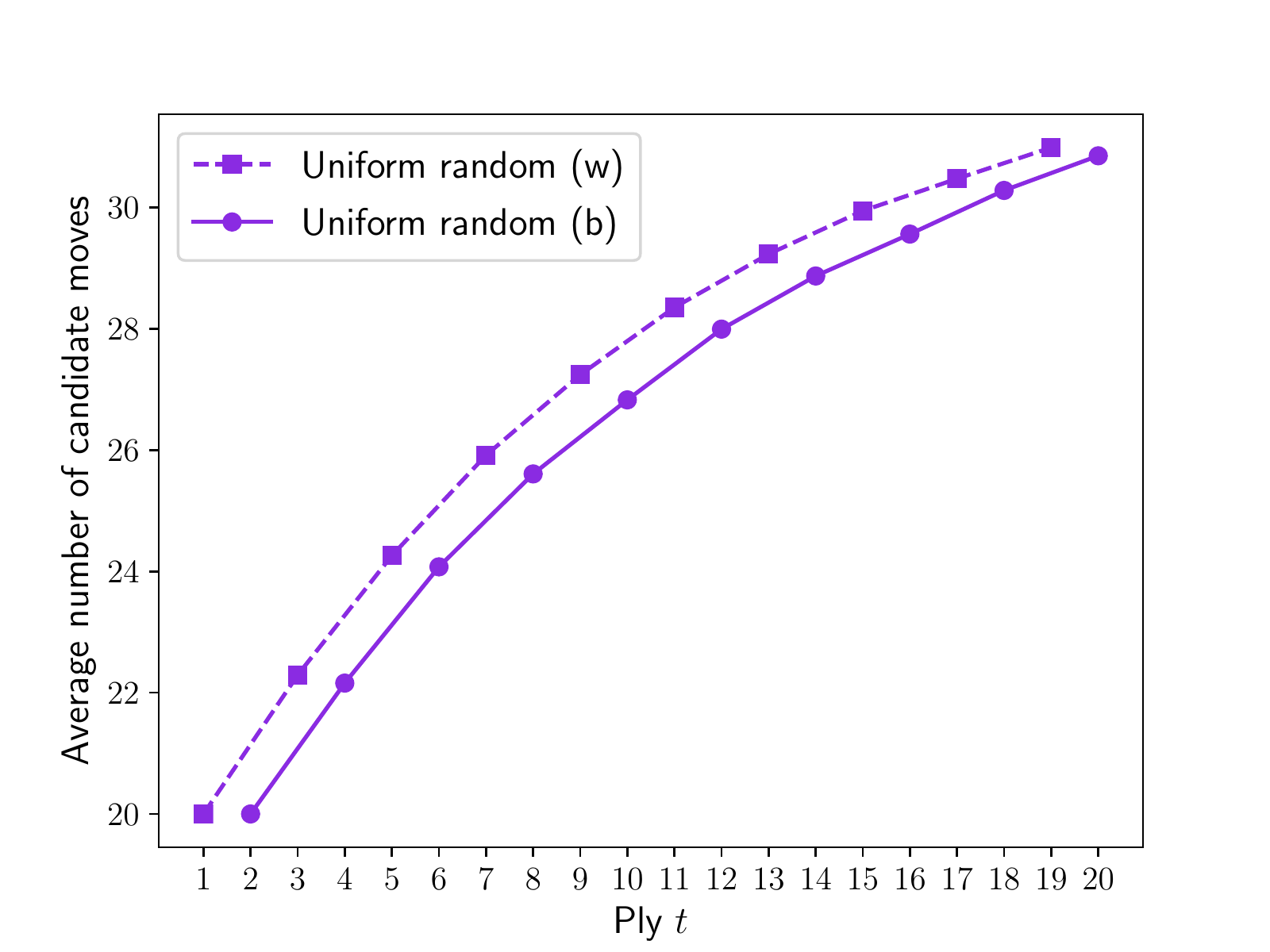}
\caption{Uniform random moves under classical chess rules}
\label{fig:average-candidates-u}
\end{subfigure}%
\caption{\emph{(Continued from previous page.)}
The average number of candidate moves $\Mcal(t)$ from
\eqref{eq:average-candidates} for each of the variants, as computed from their prior distributions $p(\s)$.}
\label{fig:average-candidates}
\end{figure*}

%% file: appendix-SAN.tex
\section{Appendix}
\label{sec:chess_appendix}

Here we present a selection of instructive games for each of the chess variations considered in the study, along with a detailed assessment of the variations by Vladimir Kramnik.

Given that different rule changes that we examined had led to a different degree of departure from existing chess theory and patterns, we do not present an equal amount of instructive positions and games for each chess variation, and rather focus on those that have either been assessed to be of greater immediate interest or simply employ patterns that are unfamiliar and novel and require more time to introduce and understand.

The Appendix is organised into sections corresponding to each of the chess variations and rule alterations examined in this study, in the following order:
No-castling chess (Page \pageref{sec-nocas}),
No-castling (10) chess (Page \pageref{sec-nocas10}),
Pawn one square chess (Page \pageref{sec-ponesq}), Stalemate=win chess (Page \pageref{sec-stalemate}),
Torpedo (Section~\pageref{sec-torpedo}),
Semi-torpedo (Page \pageref{sec-semitorpedo}), Pawn-back chess (Page ~\pageref{sec-pawnback}),
Pawn-sideways chess (Page \pageref{sec-pawnside}) and Self-capture chess (Page \pageref{sec-selfcapture}).

Each of the variants-specific sections first introduces the rule change, sets out the motivation for why it seemed of interest to be tried out, gives a qualitative assessment and a high-level conceptual overview of the dynamics of arising play by Vladimir Kramnik and then concludes with several instructive games and positions, selected to illustrate the typical motifs that arise in AlphaZero play in these variations.

\input{variants/SAN/no-castling.tex}
\input{variants/SAN/no-castling-10.tex}
\input{variants/SAN/pawn-one-square.tex}
\input{variants/SAN/stalemate-win.tex}

\input{variants/SAN/torpedo.tex}
\input{variants/SAN/semi-torpedo.tex}
\input{variants/SAN/pawn-back.tex}
\input{variants/SAN/pawn-sideways.tex}
\input{variants/SAN/self-capture.tex}

%% file: variants/SAN/no-castling.tex
\subsection{No-castling}
\label{sec-nocas}

In No-castling chess, the adjustment to the original rules involved a full removal of castling as an option.

\subsubsection{Motivation}

The motivation for the No-castling chess variant, as provided by Vladimir Kramnik:

\begin{shadequote}[]{}
\textit{
Adjustments to castling rules were chronologically the first type of changes implemented and assessed in this study. Firstly, excluding a single existing rule makes it comparatively easy for human players to adjust, as there is no need to learn an additional rule.
Secondly, the right to castle is relatively new in the long history of the game of chess. Arguably, it stands out amongst the rules of chess, by providing the only legal opportunity for a player to move two of their own pieces at the same time.}
\end{shadequote}

\subsubsection{Assessment}

The assessment of the no-castling chess variant, as provided by Vladimir Kramnik:

\begin{shadequote}[]{}
\textit{
I was expecting that abandoning the castling rule would make the game somewhat more favorable for White, increasing the existing opening advantage. Statistics of AlphaZero games confirmed this intuition, though the observed difference was not substantial to the point of unbalancing the game. Nevertheless, when considering human practice,
and considering that players would find themselves in unknown territory at the very early stage of the game, I would expect White to have a higher expected score in practice than under regular circumstances.}

\textit{
One of the main advantages of no-castling chess is that it eliminates the nowadays overwhelming importance of the opening preparation in professional chess, for years to come, and makes players think creatively from the very beginning of each game.
This would inevitably lead to a considerably higher amount of decisive games in chess tournaments until the new theory develops, and more creativity would be required in order to win. These factors could also increase the following of professional chess tournaments among chess enthusiasts.}

\textit{
With late middlegame and endgame patterns staying the same as in regular chess, there is a major difference in the opening phase of a no-castling chess game. The main conceptual rules of piece development and king safety are still valid, but most concrete opening variations of regular chess no longer apply, as castling is usually an essential part of existing chess opening variations.}

\textit{
For example, possibly opening a game with 1.~f4, which is not a great idea in classical chess, might be one of the better options already, since it might make it easier to evacuate the king after Nf3, g3, Bg2, Kf2, Rf1, Kg1. Some completely new patterns of playing the openings start to make sense, like pushing the side pawns in order to develop the rooks via the
``h'' file or ``a'' file, as well as ``artificial castling'' by means of Ke2, Re1, Kf1 and others. Many new conceptual questions arise in this chess variation.}

\textit{
For instance, one has to think about what ought to be preferable: evacuating the king out of the center of the board as soon as possible or aiming to first develop all the pieces and claim space and central squares.
Years of practice are
likely required to give a clear answer on the guiding principles of early play and best opening strategies. Even with the help of chess engines, it would likely take decades to develop the opening theory to the same level and to the same depth as we have in regular chess today. The engines can be helpful with providing initial recommendations of plausible opening lines of play, but the right understanding and timing of the implementation of new patterns is crucial in practical play.}

\textit{
Studying the numerous no-castling games played by AlphaZero, I have noticed one major conceptual change. Since both kings have a harder time finding a safe place, the dynamic positional factors (e.g.~initiative, piece activity, attack), seem to have more importance than in regular chess. In other words, a game becomes sharper, with both sides attacking the opponent king at the same time.}

\textit{
I am convinced that because of the aforementioned reasons we would see many interesting games, and many more decisive games at the top level chess tournaments in case the organisers decide to give it a try.
Due to the simplicity of the adjustment compared to regular chess, it is also easy to implement this variation at any other level, including the online chess playing platforms, as it merely requires an agreement between the two players not to play castling in their game.}
\end{shadequote}

\subsubsection{Main Lines}

Here we discuss ``main lines'' of AlphaZero under No-castling chess, when playing with roughly one minute per move from a particular fixed first move. Note that these are not purely deterministic, and each of the given lines is merely one of several highly promising and likely options. Here we give the first 20 moves in each of the main lines, regardless of the position.

\paragraph{Main line after e4}
The main line of AlphaZero after \move{1}e4 in No-castling chess is:

\move{1}e4 \bookmove c5 \move{2}Nf3  Nc6
\move{3}Nc3  Nf6  \move{4}d4  cxd4 \move{5}Nxd4  e6  \move{6}Ndb5  d6  \move{7}Bf4  e5  \move{8}Bg5  a6  \move{9}Na3  b5  \move{10}Nd5  Be7
\move{11}Bxf6  Bxf6  \move{12}c4  Ne7  \move{13}Nxf6+  gxf6  \move{14}cxb5  h5  \move{15}Qd2  Kf8  \move{16}Bc4  Kg7  \move{17}Rd1  d5  \move{18}exd5  Qb6  \move{19}bxa6  Rd8  \move{20}Nc2  Bxa6

\begin{center}
\newgame
\fenboard{r2r4/4npk1/bq3p2/3Pp2p/2B5/8/PPNQ1PPP/3RK2R w K - 0 21}
\showboard
\end{center}

\paragraph{Main line after d4}
The main line of AlphaZero after \move{1}d4 in No-castling chess is:

\move{1}d4 \bookmove d5 \move{2}c4 e6  \move{3}Nc3  c5  \move{4}cxd5  exd5  \move{5}Nf3  Nf6  \move{6}g3  Nc6  \move{7}Bg2  h6  \move{8}Kf1  Be6  \move{9}Bf4  Rc8  \move{10}h4  a6  \move{11}Rc1  Rg8  \move{12}a3  c4  \move{13}Ne5  Bd6  \move{14}e4  dxe4  \move{15}Nxe4  Bxe5  \move{16}dxe5  Nxe4  \move{17}Bxe4  Qa5  \move{18}Kg2  Rd8  \move{19}Qe2  Nd4  \move{20}Qe3  Nf5 

\begin{center}
\newgame
\fenboard{3rk1r1/1p3pp1/p3b2p/q3Pn2/2p1BB1P/P3Q1P1/1P3PK1/2R4R w - - 7 21}
\showboard
\end{center}

\paragraph{Main line after c4}
The main line of AlphaZero after \move{1}c4 in No-castling chess is:

\move{1}c4 \bookmove e5 \move{2}Nc3  Nf6  \move{3}Nf3  Nc6  \move{4}d4  exd4  \move{5}Nxd4  Bb4  \move{6}Bf4  Bxc3  \move{7}bxc3  d6  \move{8}g3  Ne5  \move{9}Bg2  Kf8  \move{10}c5  Ng6  \move{11}Be3  dxc5  \move{12}Nb3  Qe8  \move{13}h4  h5
\move{14}Bxc5+  Kg8  \move{15}Qc2  a5  \move{16}Bd4  Ne7  \move{17}Bxf6  gxf6  \move{18}a4  Kg7  \move{19}Nd4  Rb8  \move{20}Kf1  Bd7 

\begin{center}
\newgame
\fenboard{1r2q2r/1ppbnpk1/5p2/p6p/P2N3P/2P3P1/2Q1PPB1/R4K1R w - - 5 21}
\showboard
\end{center}

\subsubsection{Instructive games}

\paragraph{Game AZ-1: AlphaZero No-castling vs AlphaZero No-castling}
The first ten moves for White and Black have been sampled randomly from AlphaZero's opening ``book'', with the probability proportional to the time spent calculating each move. The remaining moves follow best play, at roughly one minute per move.
% game 25, on-policy

\move{1}d4  d5  \move{2}c4  e6  \move{3}Nc3  c5  \move{4}cxd5  exd5  \move{5}Nf3  Nf6  \move{6}g3  Nc6  \move{7}Bg2  h6  \move{8}h4  Be6  \move{9}Kf1  Rc8  \move{10}Be3  Ng4  \move{11}Qd2  b5

\begin{center}
\newgame
\fenboard{2rqkb1r/p4pp1/2n1b2p/1ppp4/3P2nP/2N1BNP1/PP1QPPB1/R4K1R w k - 0 12}
\showboard
\end{center}

\move{12}Nxb5  Qb6  \move{13}a4  a6  \move{14}dxc5  Nxe3+  \move{15}Qxe3  Bxc5  \move{16}Nd6+

\begin{center}
\newgame
\fenboard{2r1k2r/5pp1/pqnNb2p/2bp4/P6P/4QNP1/1P2PPB1/R4K1R b k - 1 16}
\showboard
\end{center}

\blackmove{16}Ke7  \move{17}Nxc8+  Rxc8  \move{18}a5  Qa7  \move{19}Qb3  Bxf2  \move{20}Bh3  Rb8

\begin{center}
\newgame
\fenboard{1r6/q3kpp1/p1n1b2p/P2p4/7P/1Q3NPB/1P2Pb2/R4K1R w - - 2 21}
\showboard
\end{center}

\move{21}Qa3+  Bc5  \move{22}Qd3  Nb4  \move{23}Qh7  Qd7  \move{24}Bxe6  Qxe6  \move{25}Rc1  Be3  \move{26}Rc3  d4  \move{27}Rc5  Kd6  \move{28}Re5  Qg4 \move{29}Qf5  Qxg3  \move{30}Rh2

\begin{center}
\newgame
\fenboard{1r6/5pp1/p2k3p/P3RQ2/1n1p3P/4bNq1/1P2P2R/5K2 b - - 1 30}
\showboard
\end{center}

\blackmove{30}Qg6  \move{31}Rg2  Qxf5  \move{32}Rxf5  Ke6  \move{33}Rc5  Kd6  \move{34}Rf5  Ke6 
\move{35}Re5+  Kf6  \move{36}h5  Rc8  \move{37}Rg4  Rc1+
\move{38}Kg2  Nc6  \move{39}Re8  Rc2  \move{40}Kh3  Rc5 
\move{41}Kh4  Bf2+  \move{42}Kh3  Be3  \move{43}Rh4  Rxa5  \move{44}Kg3  Ra1  \move{45}Rhe4  Kf5  \move{46}Nh4+  Kf6  \move{47}Rc8  Ne7  \move{48}Re8  Nc6  \move{49}Nf3  Kf5  \move{50}b3  Rb1  \move{51}Nh4+  Kf6  \move{52}Ra8  Ra1  \move{53}Kh3  Ne5  \move{54}Re8  Rh1+  \move{55}Kg3  Nc6  \move{56}Ra8  Ra1
\move{57}b4  Nxb4  \move{58}Rd8  Rg1+  \move{59}Kh3  Rh1+  \move{60}Kg2  Rg1+  \move{61}Kh3  Rh1+  \move{62}Kg3  Rg1+  \move{63}Ng2  Nc2  \move{64}Kh2  Rf1  \move{65}Rc8  Kf5  \move{66}Nxe3+  Nxe3  \move{67}Rxd4  Kg5  \move{68}Rc5+  f5  \move{69}Kg3  Kxh5  \move{70}Re4  Ng4  \move{71}Kg2  Rf2+  \move{72}Kg1  Nf6  \move{73}Re7  Rf4  \move{74}Rxg7  Ng4  \move{75}Rc3  Kh4  \move{76}Re7  Kg5  \move{77}Ra3  h5  \move{78}Rxa6  Rb4  \move{79}Ra5  h4  \move{80}Ra3  Nf6  \move{81}Rg7+  Kh5  \move{82}Rf7  Kg5  \move{83}Rg7+  Kh5  \move{84}Kh1  Rb2  \move{85}Ra5  Kh6  \move{86}Rg2  Rb1+  \move{87}Rg1  Rxg1+  \move{88}Kxg1  Kg5  \move{89}Ra8  Ne4  \move{90}Kg2  Kg4  \move{91}Ra4  Kg5  \move{92}Rb4  Kg4  \move{93}Rd4  Kh5  \move{94}Kh3  Ng5+  \move{95}Kh2  Ne4  \move{96}Kg2  Kg4  \move{97}Rb4  Kg5  \move{98}Kf3  Nd2+  \move{99}Ke3  Ne4  \move{100}Rb7  Kg4  \move{101}Rg7+  Ng5  \move{102}Rg8  h3  \move{103}Kf2  f4
\gamedrawn

\paragraph{Game AZ-2: AlphaZero No-castling vs AlphaZero No-castling}
The first ten moves for White and Black have been sampled randomly from AlphaZero's opening ``book'', with the probability proportional to the time spent calculating each move. The remaining moves follow best play, at roughly one minute per move.
% Game 23, on-policy

\move{1}Nf3  e6  \move{2}d4  d5  \move{3}e3  c5  \move{4}b3  h5  \move{5}dxc5  Bxc5
\move{6}Bb2  Kf8  \move{7}c4  Nf6  \move{8}h4  Bd7  \move{9}Nc3  Nc6  \move{10}Be2  Rc8  \move{11}Rc1  Qa5  \move{12}cxd5  Nxd5  \move{13}Kf1  Bxe3

\begin{center}
\newgame
\fenboard{2r2k1r/pp1b1pp1/2n1p3/q2n3p/7P/1PN1bN2/PB2BPP1/2RQ1K1R w - - 0 14}
\showboard
\end{center}

\move{14}Rc2  Bh6  \move{15}Ng5  Ncb4  \move{16}Rc1  Ke7  \move{17}Rh3  Rhd8  \move{18}a3  Nxc3  \move{19}Bxc3  Rxc3

\begin{center}
\newgame
\fenboard{3r4/pp1bkpp1/4p2b/q5Np/1n5P/PPr4R/4BPP1/2RQ1K2 w - - 0 20}
\showboard
\end{center}

\move{20}Rhxc3  Bc6  \move{21}Qe1  Qxa3  \move{22}Kg1  g6  \move{23}Bf1  Bg7  \move{24}Re3  Rd6  \move{25}Rc4  Nd5  \move{26}Rf3  Nf6

\begin{center}
\newgame
\fenboard{8/pp2kpb1/2brpnp1/6Np/2R4P/qP3R2/5PP1/4QBK1 w - - 8 27}
\showboard
\end{center}

\move{27}Rff4  Qxb3  \move{28}Be2  Rd7  \move{29}Rc1  Qb2  \move{30}Bf3  Bxf3  \move{31}Rxf3  Qd2  \move{32}Qf1  Qd5  \move{33}Qe1  Qd2  \move{34}Qf1  Qd5  \move{35}Qe2  Bh6  \move{36}Qb2  Bxg5  \move{37}hxg5  Ng4  \move{38}Re1  Qd2  \move{39}Qa3+  Rd6  \move{40}Rb1  Kf8  \move{41}g3  Ne5  \move{42}Rf4  Qd3  \move{43}Qxd3  Nxd3  \move{44}Ra4  Rd5  \move{45}Rxb7  Rxg5  \move{46}Ra3  Rd5  \move{47}Rbxa7  Ne5  \move{48}R7a5  Rd1+  \move{49}Kg2  Ng4  \move{50}Ra1  Rd4  \move{51}R5a4  Rd3  \move{52}R4a3  Rd4  \move{53}Ra4  Rd3  \move{54}R4a3  Rd2  \move{55}R3a2  Rd7  \move{56}Ra7  Rd6  \move{57}R7a6  Rd7  \move{58}Ra7  Rd6  \move{59}R7a6  Rd5  \move{60}R6a5  Rd2  \move{61}R5a2  Rd5  \move{62}Ra5  Rd2  \move{63}R5a2
\gamedrawn

\paragraph{Game AZ-3: AlphaZero No-castling vs AlphaZero No-castling}
The first ten moves for White and Black have been sampled randomly from AlphaZero's opening ``book'', with the probability proportional to the time spent calculating each move. The remaining moves follow best play, at roughly one minute per move.
% game 19, on-policy

\move{1}c4  e5  \move{2}Nc3  Nf6  \move{3}Nf3  Nc6  \move{4}d4  exd4  \move{5}Nxd4  Bb4  \move{6}g3  Ne4  \move{7}Qd3  Nc5  \move{8}Qe3+  Kf8  \move{9}Bg2  Qf6  \move{10}Ndb5  Ne6  \move{11}Kd1  Bc5  \move{12}Qe4  d6  \move{13}Nd5

\begin{center}
\newgame
\fenboard{r1b2k1r/ppp2ppp/2npnq2/1NbN4/2P1Q3/6P1/PP2PPBP/R1BK3R b - - 1 13}
\showboard
\end{center}

\blackmove{13}Qd8  \move{14}f4  Ned4  \move{15}Qd3  Bf5  \move{16}e4  Bg4+  \move{17}Ke1  Be2

\begin{center}
\newgame
\fenboard{r2q1k1r/ppp2ppp/2np4/1NbN4/2PnPP2/3Q2P1/PP2b1BP/R1B1K2R w - - 3 18}
\showboard
\end{center}

\move{18}Qc3  Nxb5  \move{19}cxb5  Bxb5  \move{20}Be3  Bxe3  \move{21}Nxe3  Ne7  \move{22}Kf2  h5  \move{23}h4  Rh6  \move{24}Rac1  c6  \move{25}Rhd1  Qb6  \move{26}Rd4  a5  \move{27}Rcd1  d5  \move{28}exd5  cxd5  \move{29}Qa3

\begin{center}
\newgame
\fenboard{r4k2/1p2npp1/1q5r/pb1p3p/3R1P1P/Q3N1P1/PP3KB1/3R4 b - - 1 29}
\showboard
\end{center}

The game soon ended in a draw.
\gamedrawn

\subsubsection{Human games}

Here we take a brief look at a couple of recently played blitz games between professional chess players from the tournament that took place in Chennai in January 2020~\cite{nocastlingtournament}.
We focus on new motifs in the opening stage of the game, and show how these might be counter-intuitive compared to similar patterns in classical chess.

\paragraph{Game H-1: Arjun, Kalyan (2477) vs D.~Gukesh (2522) (blitz)}

\move{1}d4 d5
\move{2}c4 c6
\move{3}Nc3 Nf6
\move{4}Nf3 

\begin{center}
\newgame
\fenboard{rnbqkb1r/pp2pppp/2p2n2/3p4/2PP4/2N2N2/PP2PPPP/R1BQKB1R b KQkq - 0 4}
\showboard
\end{center}

Interestingly, even at an early stage we can see an example of a difference in patterns that originate in Classical chess and those that arise in No-castling chess.
The positioning of the knight on f3 is very natural, but is in fact an imprecision. AlphaZero prefers keeping the option open of playing the pawn to f3 instead, in order to tuck the king away to safety. It gives the following line as its favored continuation: \move{4}e3 Bf5 \move{5}Bd3 g6 \move{6}h3 e6 \move{7}Nge2 Be7 \move{8}f3 Bxd3 \move{9}Qxd3 Kf8 \move{10}Kf2 Bg7 \move{11}Rd1.

\begin{center}
\newgame
\fenboard{rn1q3r/pp2bpkp/2p1pnp1/3p4/2PP4/2NQPP1P/PP2NKP1/R1BR4 b - - 4 11}
\showboard \\
\textsf{analysis diagram}
\end{center}

Yet, \move{4}Nf3 was played in the game, which continued:

\blackmove{4}e6 \move{5}e3 Nbd7 \move{6}Qc2 Bd6 \move{7}b3 b6

\begin{center}
\newgame
\fenboard{r1bqk2r/p2n1ppp/1ppbpn2/3p4/2PP4/1PN1PN2/P1Q2PPP/R1B1KB1R w KQkq - 0 8}
\showboard
\end{center}

Here AlphaZero suggests
that it was instead time to move the king to safety. Deciding on when exactly to initiate the evacuation of the king from the centre and choosing the best way of achieving it is one of the key motifs of No-castling chess.
This decision is less clear than the decision to castle in Classical chess, due to a larger number of options and the fact that the sequence takes more moves that all need to be staged accordingly. Instead of moving the pawn to b6, AlphaZero suggests the following instead:
\blackmove{7}h5 \move{8}Bb2 Kf8 \move{9}Rd1 Kg8.

\begin{center}
\newgame
\fenboard{r1bq2kr/pp1n1pp1/2pbpn2/3p3p/2PP4/1PN1PN2/PBQ2PPP/3RKB1R w K - 4 10}
\showboard \\
\textsf{analysis diagram}
\end{center}

Going back to the game continuation, after
\blackmove{7}b6 White has the upper hand. The game continued: \move{8}Bb2 Bb7 \move{9}Bd3 Qe7 \move{10}e4 

\begin{center}
\newgame
\fenboard{r3k2r/pb1nqppp/1ppbpn2/3p4/2PPP3/1PNB1N2/PBQ2PPP/R3K2R b KQkq - 0 10}
\showboard
\end{center}

This is another example of mistiming the evacuation of the king. Instead of playing \move{10}e4, it was the right time to move the king to safety instead, retaining a large plus for White after: \move{10}Kf1 Kf8 \move{11}h4 h5 \move{12}a4 Ng4 \move{13}Rh3 Rh6

\begin{center}
\newgame
\fenboard{r4k2/pb1nqpp1/1ppbp2r/3p3p/P1PP2nP/1PNBPN1R/1BQ2PP1/R4K2 w - - 3 14}
\showboard \\
\textsf{analysis diagram}
\end{center}

Going back to the position after \move{10}e4,
the game continuation goes as follows:

\blackmove{10}dxe4 \move{11}Nxe4 (Giving away the advantage. Recapturing with the bishop was correct, even though it might seem as otherwise counter-intuitive.)
\blackmove{11}Nxe4 \move{12}Bxe4 f5.
(This is looking bad for Black; \blackmove{12}Nf6 is the preferred move.)
\move{13}Bd3 c5 (At this point, AlphaZero assesses the position as winning for White.)
\move{14}Kf1 (The advantage could have been kept with \move{14}d5.)
\blackmove{14}Bxf3 \move{15}gxf3 cxd4 (\blackmove{15}Rf8 may have been equalizing)
\move{16}Bxd4 (Gives the advantage to Black.
White ought to have captured on f5 instead. The right way to respond to the game move would have been \blackmove{16}Qh4.)
\blackmove{16}Be5 \move{17}Bxe5 Nxe5 \move{18}Bxf5

\begin{center}
\newgame
\fenboard{r3k2r/p3q1pp/1p2p3/4nB2/2P5/1P3P2/P1Q2P1P/R4K1R b kq - 0 18}
\showboard
\end{center}

A brilliant piece sacrifice.

\blackmove{18}exf5 \move{19}Re1 Kd8 \move{20}Qxf5
(\move{20}Qd2+ may have been stronger) \blackmove{20}Re8 \move{21}f4 Qb7 \move{22}Rg1 Ng6 (The final mistake, it appears that \move{22}Nf7 might hold) \move{23}Rd1+ Ke7 \move{24}Rg3 Qh1+ \move{25}Ke2 Qe4+ \move{26}Re3 Qxe3+ \move{27}fxe3 Rad8 \move{28}Rxd8 Rxd8 \move{29}Qe4+ Kf8 \move{30}Qb7
\whitewins

\paragraph{Game H-2: Gelfand, Boris vs Kramnik, Vladimir (blitz)}

\move{1}f4 h5
Already Kramnik demonstrates a motif that is quite strong in no-castling chess, pushing one of the side pawns early.

\begin{center}
\newgame
\fenboard{rnbqkbnr/ppppppp1/8/7p/5P2/8/PPPPP1PP/RNBQKBNR w KQkq - 0 2}
\showboard
\end{center}

\move{2}Nf3 e6
\move{3}e3 Nf6
\move{4}b3
(Interestingly, AlphaZero doesn't like this very normal-looking move, giving Black a slight plus after 
\blackmove{4}c5 \move{5}Bb2 Be7 \move{6}Be2 d5 \move{7}Rf1 Kf8 \move{8}Kf2 Nc6 \move{9}Kg1 Kg8 \move{10}a4 Bd7.)
\blackmove{4}b6
\move{5}Bb2 Bb7
\move{6}Bd3
(\move{5}Be2 might have been better.)
\blackmove{6}h4
(Not the most precise, according to AlphaZero, suggesting that \blackmove{6}c5
\move{7}Rf1 Be7 \move{8}Kf2 h4 \move{9}Ng5 Kf8 \move{10}Kg1 Rh6 \move{11}Be2 Nc6 was still slightly better for Black.)
\move{7}h3
(This turns out to be the wrong reaction, giving the advantage back to Black again.)
\blackmove{7}Nh5
\move{8}Kf2 Be7
(Here, there was an opportunity to play \blackmove{8}Bc5 instead: 
\begin{center}
\newgame
\fenboard{rn1qk2r/pbpp1pp1/1p2p3/2b4n/5P1p/1P1BPN1P/PBPP1KP1/RN1Q3R w kq - 3 9}
\showboard \\
\textsf{analysis diagram}
\end{center}
which would have kept a big plus for Black.)

 \move{9}Re1 Bf6
 \move{10}Bxf6 (\move{10}Nc3) \blackmove{10}Qxf6
 (\blackmove{10}gxf6 was the better recapture)
 \move{11}Nc3 Ng3
 \move{12}Kg1 d6 (\blackmove{12}Ke7 was the correct plan)
 \move{13}Ng5 Nd7
 \move{14}Nce4 Nxe4
 \move{15}Bxe4 Bxe4
 \move{16}Nxe4 Qg6
 \move{17}Ng5 Ke7
 \move{18}e4 (\move{18}Qe2) \blackmove{18}e5
 \move{19}d4 exf4
 \move{20}Nf3 Kf8
 \move{21}Qd2 Qg3
 \move{22}Re2 Rh6 

\begin{center}
\newgame
\fenboard{r4k2/p1pn1pp1/1p1p3r/8/3PPp1p/1P3NqP/P1PQR1P1/R5K1 w - - 6 23}
\showboard
\end{center}

\move{23}Rf1 (Black gains the upper hand.) \blackmove{23}Re6
\move{24}Nh2 (A mistake, \move{24}e5 was required.) \blackmove{24}Rae8
\move{25}Rxf4 Nf6 \move{26}e5 dxe5 \move{27}Rf3 (Another mistake, \move{27}Rxe5 was correct.) \blackmove{27}Qg6 \move{28}d5 (Taking on e5 was still a better continuation.)
\blackmove{28}R6e7 \move{29}c4 e4 \move{30}Rc3 Nh5 \move{31}Nf1 Kg8 \move{32}Qe1 Nf4 \move{33}Rd2 e3 \move{34}Rxe3 Rxe3 \move{35}Nxe3 Qe4
\blackwins

%% file: variants/SAN/no-castling-10.tex
\subsection{No-castling (10)}
\label{sec-nocas10}

In the No-castling (10) variant of chess, castling is only allowed from move 11 onwards, both for the first and the second player.

\subsubsection{Motivation}

When it comes to limit the impact of castling on the game, it is possible to consider different types of partial limitations, the easiest of which is disallowing it for a fixed number of opening moves.
In this variation, we have explored the impact of disallowing castling for the first 10 moves, but any other number could have been used instead.
Each choice leads to a slightly different body of opening theory, as particular lines either become viable or stop being viable under different circumstances.

\subsubsection{Assessment}

The assessment of the No-castling (10) chess variant, as provided by Vladimir Kramnik:

\begin{shadequote}[]{}
\textit{The main purpose of the partial restriction to castling, as a hypothetical adjustment to the rules of chess, would be to sidestep opening theory.
As such, it is aimed at professional chess as an option to potentially consider. The game itself does not change in other meaningful ways, and AlphaZero usually aims at playing slower lines where castling does indeed take place after the first 10 moves.
This makes sense, given that castling is a fast an powerful move, so aiming to take advantage of it if available makes for a good approach. Yet, the slowing down of the game could as a side-effect lead to an increased number of draws. Another disadvantage is the need to count and keep track of the move number when considering variations.}
\end{shadequote}

\subsubsection{Main Lines}

Here we discuss ``main lines'' of AlphaZero under No-castling (10) chess,
when playing with roughly one minute per move from a particular fixed first move. Note that these are not purely deterministic, and each of the given lines is merely one of several highly promising and likely options. Here we give the first 20 moves in each of the main lines, regardless of the position.

\paragraph{Main line after e4}
The main line of AlphaZero after \move{1}e4 in No-castling (10) chess is:

\move{1}e4 \bookmove e5  \move{2}Bc4  Nc6  \move{3}Nf3  Nf6  \move{4}Qe2  Bc5  \move{5}c3  Qe7  \move{6}b4  Bb6  \move{7}a4  a6  \move{8}a5  Ba7  \move{9}d3  d6  \move{10}Na3  Be6  \move{11}Nc2  O-O  \move{12}O-O  h6  \move{13}Be3  Qd7  \move{14}Bxa7  Nxa7  \move{15}Rfe1  Nc6 
\move{16}h3  Rfe8  \move{17}Bxe6  Qxe6  \move{18}Ne3  d5  \move{19}Qc2  Rad8  \move{20}Rab1  Qd7

\begin{center}
\newgame
\fenboard{3rr1k1/1ppq1pp1/p1n2n1p/P2pp3/1P2P3/2PPNN1P/2Q2PP1/1R2R1K1 w - - 4 21}
\showboard
\end{center}

\paragraph{Main line after d4}
The main line of AlphaZero after \move{1}d4 in No-castling (10) chess is:

\move{1}d4 \bookmove d5  \move{2}Nf3  Nf6  \move{3}c4  dxc4  \move{4}Nc3  e6
\move{5}Qa4+  c6  \move{6}Qxc4  b5  \move{7}Qd3  Bb7  \move{8}e4  b4  \move{9}Na4  Qa5  \move{10}b3  c5  \move{11}Ne5  cxd4  \move{12}Qb5+  Qxb5  \move{13}Bxb5+  Nfd7  \move{14}Bb2  f6  \move{15}Nxd7  Nxd7  \move{16}Bxd4  Bxe4  \move{17}O-O  Bd6  \move{18}Rfe1  Bd5  \move{19}Nc5  Bxc5  \move{20}Bxc5  Rb8 

\begin{center}
\newgame
\fenboard{1r2k2r/p2n2pp/4pp2/1BBb4/1p6/1P6/P4PPP/R3R1K1 w k - 1 21}
\showboard
\end{center}

\paragraph{Main line after c4}
The main line of AlphaZero after \move{1}c4 in No-castling (10) chess is:

\move{1}c4 \bookmove e5  \move{2}Nc3  Nf6  \move{3}Nf3  Nc6  \move{4}e4  Bb4  \move{5}d3  d6  \move{6}a3  Bc5  \move{7}b4  Bb6  \move{8}Be3  Bg4  \move{9}Be2  Bxf3  \move{10}Bxf3  Nd4  \move{11}Na4  Nxf3+  \move{12}Qxf3  Bxe3  \move{13}fxe3  Nd7  \move{14}O-O  O-O  \move{15}Nc3  c6  \move{16}h3  Qb6
\move{17}Rab1  Rae8  \move{18}a4  Re6  \move{19}a5  Qd8  \move{20}Qg3  Rf6 

\begin{center}
\newgame
\fenboard{3q1rk1/pp1n1ppp/2pp1r2/P3p3/1PP1P3/2NPP1QP/6P1/1R3RK1 w - - 3 21}
\showboard
\end{center}

\subsubsection{Instructive games}

\paragraph{Game AZ-4: AlphaZero No-castling (10) vs AlphaZero No-castling (10)}
The first ten moves for White and Black have been sampled randomly from AlphaZero's opening ``book'', with the probability proportional to the time spent calculating each move. The remaining moves follow best play, at roughly one minute per move.
% Game 6, on policy

\move{1}c4  e5  \move{2}d4  exd4  \move{3}Qxd4  Nc6  \move{4}Qe3+  Nge7  \move{5}Nf3  d5  \move{6}cxd5  Qxd5  \move{7}Nc3  Qa5  \move{8}Qg5  Bf5  \move{9}Bd2  f6  \move{10}Qh5+  g6  \move{11}Qh4  Nb4  \move{12}Rc1  O-O-O  \move{13}Qxf6  Bh6  

\begin{center}
\newgame
\fenboard{2kr3r/ppp1n2p/5Qpb/q4b2/1n6/2N2N2/PP1BPPPP/2R1KB1R w K - 1 14}
\showboard
\end{center}

A stunning move, offering up a piece on h6. Accepting would be disastrous for White, as Black pieces mobilise quickly via Ned5. The h8 rook can also potentially come to e8, and this justifies the material investment.

\move{14}e3  Rhe8  \move{15}Qh4  Bg7  \move{16}Nb5  Rxd2 

\begin{center}
\newgame
\fenboard{2k1r3/ppp1n1bp/6p1/qN3b2/1n5Q/4PN2/PP1r1PPP/2R1KB1R w K - 0 17}
\showboard
\end{center}

The fireworks continue\ldots

\move{17}Rxc7+  Qxc7  \move{18}Nxc7  Rxb2  \move{19}Nxe8  Rb1+

\begin{center}
\newgame
\fenboard{2k1N3/pp2n1bp/6p1/5b2/1n5Q/4PN2/P4PPP/1r2KB1R w K - 1 20}
\showboard
\end{center}

Leading to a draw by perpetual check.
\gamedrawn

The next game is less tactically rich, but rather interesting from the perspective of showcasing differences in opening play and the overall approach, when castling is not possible in the first ten moves.

\paragraph{Game AZ-5: AlphaZero No-castling (10) vs AlphaZero No-castling (10)}
The first ten moves for White and Black have been sampled randomly from AlphaZero's opening ``book'', with the probability proportional to the time spent calculating each move. The remaining moves follow best play, at roughly one minute per move.
% Game 8, on policy

\move{1}c4  e5  \move{2}Nc3  Nf6  \move{3}Nf3  Nc6  \move{4}Qa4

\begin{center}
\newgame
\fenboard{r1bqkb1r/pppp1ppp/2n2n2/4p3/Q1P5/2N2N2/PP1PPPPP/R1B1KB1R b KQkq - 5 4}
\showboard
\end{center}

This is a slightly unusual move, showcasing that the style of play in this variation of chess involves opting for moves that do not necessarily achieve as much immediately and are somewhat less direct,
potentially trying to wait for the right time to castle, when possible. In this game, however, castling does not end up being critical.

\blackmove{4}e4  \move{5}Ng5  Qe7  \move{6}c5  e3

\begin{center}
\newgame
\fenboard{r1b1kb1r/ppppqppp/2n2n2/2P3N1/Q7/2N1p3/PP1PPPPP/R1B1KB1R w KQkq - 0 7}
\showboard
\end{center}

\move{7}dxe3  Qxc5  \move{8}Nge4  Nxe4  \move{9}Qxe4+  Qe5  \move{10}Qxe5  Nxe5  \move{11}e4  Bb4  \move{12}f4  Nc4  \move{13}e3  Nd6  \move{14}Bd3  Bxc3  \move{15}bxc3  f6  \move{16}Ba3  Nf7  \move{17}Bc4  b6 

\begin{center}
\newgame
\fenboard{r1b1k2r/p1pp1npp/1p3p2/8/2B1PP2/B1P1P3/P5PP/R3K2R w KQkq - 0 18}
\showboard
\end{center}

\move{18}Bd5  c6  \move{19}Bb3  Rb8  \move{20}Bxf7+  Kxf7  \move{21}Bd6  Ra8  \move{22}e5  c5  \move{23}O-O-O  Ba6  \move{24}e4  h5  

\begin{center}
\newgame
\fenboard{r6r/p2p1kp1/bp1B1p2/2p1P2p/4PP2/2P5/P5PP/2KR3R w - - 0 25}
\showboard
\end{center}

And the game eventually ended in a draw.
\gamedrawn

%% file: variants/SAN/pawn-one-square.tex
\subsection{Pawn one square}
\label{sec-ponesq}

\subsubsection{Motivation}

Restricting the pawn movement to one square only is interesting to consider, as the double-move from the second (or seventh rank) seems like a ``special case''
and an exception from the rule that pawns otherwise only move by one square. In addition, slowing down the game could make it more strategic and less forcing.

\subsubsection{Assessment}

The assessment of the Pawn one square chess variant, as provided by Vladimir Kramnik:

\begin{shadequote}[]{}
\textit{The basic rules and patterns are still mostly the same as in classical chess, but the opening theory changes and becomes completely different. Intuitively it feels that it ought to be more difficult for White to gain a lasting opening advantage and convert it into a win, but since new opening theory would first need to be developed, this would not pertain to human play at first. In most AlphaZero games one can notice the rather typical middlegame positions arise after the opening phase.}

\textit{This variation of chess can be a good pedagogical tool when teaching and practicing slow, strategic play and learning about how to set up and commit to pawn structures. Since the pawns are unable to advance very fast, many attacking ideas that involve rapid pawn advances are no longer relevant, and the play is instead much slower and ultimately more positional. Additionally, this variation of chess could simply be of interest for those wishing for an easy way of side-stepping opening theory.}
\end{shadequote}

\subsubsection{Main Lines}

Here we discuss ``main lines'' of AlphaZero under Pawn one square
chess, when playing with roughly one minute per move from a particular fixed first move.
Note that these are not purely deterministic, and each of the given lines is merely one of several highly promising and likely options. Here we give the first 20 moves in each of the main lines, regardless of the position.

\paragraph{Main line after e3}
The main line of AlphaZero after \move{1}e3 in Pawn one square chess is:

\move{1}e3 \bookmove Nf6 \move{2}d3 d6 \move{3}Nf3  h6 \move{4}e4 b6 \move{5}c3 Bb7 \move{6}Qc2 e6 \move{7}c4 e5 \move{8}g3 g6 \move{9}Nc3 Bg7 \move{10}Bg2 Nc6 \move{11}Be3 Nd7 \move{12}Ne2 Nc5 \move{13}a3 Na5 

\begin{center}
\newgame
\fenboard{r2qk2r/pbp2pb1/1p1p2pp/n1n1p3/2P1P3/P2PBNP1/1PQ1NPBP/R3K2R w KQkq - 1 14}
\showboard
\end{center}

An instructive position, as it looks optically like Black is blundering material. In this variation of chess, however, b2-b4 is not a legal move, because pawns can only move one square. This justifies the move sequence.

\move{14}Nd2 Nc6 \move{15}b3 a6 \move{16}Nf3  Ne6 \move{17}h3 O-O \move{18}O-O Ncd4 \move{19}Nfxd4 exd4 \move{20}Bd2 c6

\begin{center}
\newgame
\fenboard{r2q1rk1/1b3pb1/ppppn1pp/8/2PpP3/PP1P2PP/2QBNPB1/R4RK1 w - - 0 21}
\showboard
\end{center}

\paragraph{Main line after d3}
The main line of AlphaZero after \move{1}d3 in Pawn one square chess is:

\move{1}d3 \bookmove d6  \move{2}e3  Nf6  \move{3}Nd2  e6  \move{4}Ngf3  Nbd7  \move{5}d4  g6  \move{6}Bd3  Bg7  \move{7}O-O  O-O  \move{8}h3  e5  \move{9}c3  Re8  \move{10}e4  b6  \move{11}Re1  Bb7  \move{12}a3  a6  \move{13}Qc2  h6  \move{14}Nb3  a5  \move{15}dxe5  Nxe5  \move{16}Nxe5  dxe5  \move{17}Be3  Nh5  \move{18}Bb5  Rf8  \move{19}Nd2  Qf6  \move{20}g3  Rfd8

\begin{center}
\newgame
\fenboard{r2r2k1/1bp2pb1/1p3qpp/pB2p2n/4P3/P1P1B1PP/1PQN1P2/R3R1K1 w - - 1 21}
\showboard
\end{center}

\paragraph{Main line after c3}
The main line of AlphaZero after \move{1}c3 in Pawn one square chess is:

\move{1}c3 \bookmove d6  \move{2}d3  Nf6  \move{3}Nf3  h6  \move{4}d4  Bf5  \move{5}c4  e6  \move{6}Nc3  c6  \move{7}e3  d5  \move{8}Bd3  dxc4  \move{9}Bxc4  Bd6  \move{10}O-O  Nbd7  \move{11}Re1  Ne4  \move{12}Bd3  Nxc3  \move{13}bxc3  Bxd3  \move{14}Qxd3  Qe7  \move{15}c4  e5  \move{16}Qf5  O-O  \move{17}Rb1  b6  \move{18}c5  Bc7  \move{19}Ba3  b5  \move{20}d5  cxd5

\begin{center}
\newgame
\fenboard{r4rk1/p1bnqpp1/7p/1pPppQ2/8/B3PN2/P4PPP/1R2R1K1 w - - 0 21}
\showboard
\end{center}

\subsubsection{Instructive games}

Here we present some examples of AlphaZero play in Pawn one square chess.

\paragraph{Game AZ-6: AlphaZero Pawn One Square vs AlphaZero Pawn One Square}
The first ten moves for White and Black have been sampled randomly from AlphaZero's opening ``book'', with the probability proportional to the time spent calculating each move. The remaining moves follow best play, at roughly one minute per move.
% game 5 from on-policy selfplay, 50sec

\move{1}d3  Nf6  \move{2}Nd2  d6  \move{3}e3  e6
\move{4}Ngf3  g6  \move{5}h3  Bg7  \move{6}c3  O-O  \move{7}c4  Nbd7  \move{8}Rb1  e5  \move{9}b3  c6  \move{10}Bb2  Qe7  \move{11}Be2  b6  \move{12}b4  Bb7  \move{13}a3  h6  \move{14}O-O  h5  \move{15}Qc2  Rfd8  \move{16}Rfd1  c5 

\begin{center}
\newgame
\fenboard{r2r2k1/pb1nqpb1/1p1p1np1/2p1p2p/1PP5/P2PPN1P/1BQNBPP1/1R1R2K1 w - - 0 17}
\showboard
\end{center}

Here we have a rather normal middlegame position. The game continued:

\move{17}Ne4  Rac8  \move{18}b5  d5  \move{19}cxd5  Nxd5  \move{20}Nfd2  f6  \move{21}Nc4  Nf8  \move{22}a4  Kh7  \move{23}a5  Ne6  \move{24}Ra1  Nb4  \move{25}Qb1  Rb8  \move{26}axb6  axb6  \move{27}Nc3  Qe8  \move{28}Ra7  Nc7  \move{29}Na3  Rd7  \move{30}Ba1  Nbd5  \move{31}Na4  Ne6  \move{32}e4  Ndf4  \move{33}Bf1  Bc8  \move{34}Rxd7  Qxd7  \move{35}Nc4  Qa7  \move{36}Naxb6  Rxb6  \move{37}Nxb6  Qxb6  \move{38}g3

\begin{center}
\newgame
\fenboard{2b5/6bk/1q2npp1/1Pp1p2p/4Pn2/3P2PP/5P2/BQ1R1BK1 b - - 0 38}
\showboard
\end{center}

\blackmove{38}c4  \move{39}gxf4  Nxf4  \move{40}Bc3  Bxh3  \move{41}Bd2  Qe6  \move{42}Bxf4  exf4  \move{43}f3

\begin{center}
\newgame
\fenboard{8/6bk/4qpp1/1P5p/2p1Pp2/3P1P1b/8/1Q1R1BK1 b - - 0 43}
\showboard
\end{center}

\blackmove{43}Bg4  \move{44}Bg2  Bxf3  \move{45}Bxf3  Qh3  \move{46}dxc4  Qxf3  \move{47}Qd3  Qg4+  \move{48}Kf2  Qh4+  \move{49}Ke2  Qh2+  \move{50}Kf1  Qh1+  \move{51}Kf2  Qh4+  \move{52}Ke2  Qh2+  \move{53}Ke1  Bf8  \move{54}Qf3  Bc5  \move{55}Kf1  Qg1+  \move{56}Ke2  Qh2+  \move{57}Kf1  Qg1+  \move{58}Ke2  Qh2+  \move{59}Kf1
\gamedrawn

\paragraph{Game AZ-7: AlphaZero Pawn One Square vs AlphaZero Pawn One Square}
The first ten moves for White and Black have been sampled randomly from AlphaZero's opening ``book'', with the probability proportional to the time spent calculating each move. The remaining moves follow best play, at roughly one minute per move.

% Game 30 from on-policy games

\move{1}d3  c6  \move{2}e3  d6  \move{3}c3  g6  \move{4}d4  Nf6  \move{5}Nf3  Bf5  \move{6}Be2  e6  \move{7}O-O  Nbd7  \move{8}c4  Bg7  \move{9}b3  O-O  \move{10}Ba3  Ne4
\move{11}Nfd2  c5  \move{12}Nxe4  Bxe4  \move{13}Nd2  Bc6  \move{14}Rc1  Qa5  \move{15}Bb2  cxd4  \move{16}exd4  d5 

\begin{center}
\newgame
\fenboard{r4rk1/pp1n1pbp/2b1p1p1/q2p4/2PP4/1P6/PB1NBPPP/2RQ1RK1 w - - 0 17}
\showboard
\end{center}

This is a very normal-looking position, and one would be hard-pressed to guess that it originated from a different variation of chess, as it looks pretty ``classical''.

\move{17}Re1  Rfe8  \move{18}h3  Bh6  \move{19}Bc3  Qc7  \move{20}Bf1  b6  \move{21}Bb2  Qb7  \move{22}a3  Rac8  \move{23}Rc2  Bg7  \move{24}Qc1  Bh6
\move{25}Qd1  Bg7  \move{26}Qc1  h6  \move{27}c5  bxc5  \move{28}dxc5  e5  \move{29}Qb1  h5  \move{30}Qa2  a6  \move{31}b4  Ba4  \move{32}Rcc1  Bh6  \move{33}Ba1  e4  \move{34}Rb1  Ne5  \move{35}Nb3  Kh7  \move{36}Nd4  Nd3  \move{37}Re3 

\begin{center}
\newgame
\fenboard{2r1r3/1q3p1k/p5pb/2Pp3p/bP1Np3/P2nR2P/Q4PP1/BR3BK1 b - - 7 37}
\showboard
\end{center}

A very instructive position, reminiscent of a famous classical game between Petrosian and Reshevsky from Zurich in 1953, where Petrosian was playing Black.
The positional exchange sacrifice allows White easy play on the dark squares.

\blackmove{37}Bxe3  \move{38}fxe3  f6  \move{39}Be2  Rc7  \move{40}Rf1  Rf7  \move{41}Qd2  Ne5  \move{42}Qe1  Bb5  \move{43}Nxb5  axb5  \move{44}a4  Nd3  \move{45}Qh4  bxa4  \move{46}Bxh5  Re5  \move{47}Be2+  Kg7  \move{48}Qg3  Qc7  \move{49}Bxe5  Qxe5  \move{50}Qxe5  fxe5  \move{51}c6  Rxf1+
\move{52}Bxf1  a3  \move{53}c7  a2  \move{54}c8=Q  a1=Q  \move{55}Qb7+  Kh6  \move{56}Qxd5  Qe1  \move{57}Qf7  Qxb4  \move{58}Qa2  Qc5  \move{59}Qd2  Nb4  \move{60}Kf2  Nd5  \move{61}g3  Qf8+  \move{62}Kg1  Qc5  \move{63}Kf2  Qf8+  \move{64}Ke1  Nb4  \move{65}Bc4  Kh7  \move{66}Qd7+  Kh6  \move{67}Qd2  Kg7  \move{68}Qf2  Qe7  \move{69}Kf1  Nd3  \move{70}Qe2  Qf6+  \move{71}Kg2  Qc6  \move{72}Bb3  Qc5  \move{73}h4  Qc1  \move{74}Kh2  Ne1  \move{75}Bd1  Nf3+  \move{76}Kg2  Kh6  \move{77}Kf1  g5  \move{78}hxg5  Kxg5  \move{79}Kf2  Qd2  \move{80}Bc2  Qxe2+  \move{81}Kxe2  Kf5  \move{82}Kf2  Ng5  \move{83}Kg2  Nh7  \move{84}Kh3  Nf6  \move{85}Bb3  Kg5  \move{86}Be6  Kh5  \move{87}Bb3  Kg5  \move{88}Be6  Kh5  \move{89}g4+  Kg5  \move{90}Kg3  Nh7  \move{91}Kh3  Nf6  \move{92}Kg3  Nh7  \move{93}Kh3  Nf6
\gamedrawn

%% file: variants/SAN/stalemate-win.tex
\subsection{Stalemate=win}
\label{sec-stalemate}

In this variation of chess, achieving a stalemate position is considered a win for the attacking side, rather than a draw.

\subsubsection{Motivation}

The stalemate rule in classical chess allows for additional drawing resources for the defending side, and has been a subject of debate, especially when considering ways of making the game potentially more decisive. Yet, due to its potential effect on endgames, it was unclear whether such a rule would also discourage some attacking ideas that involve material sacrifices, if being down material in endgames ends up being more dangerous and less likely to lead to a draw than in classical chess.

\subsubsection{Assessment}

The assessment of the Stalemate=win chess variant, as provided by Vladimir Kramnik:

\begin{shadequote}[]{}
\textit{I was at first somewhat surprised that the decisive game percentage in this variation was roughly equal to that of classical chess, with similar levels of performance for White and Black. I was personally expecting the change to lead to more decisive games and a higher winning percentage for White.}

\textit{It seems that the openings and the middlegame remain very similar to regular chess, with very few exceptions, but that there is a significant difference in endgame play since some basic endgame like K+P vs K are already winning instead of being drawn depending on the position.}
% \end{shadequote}

\begin{center}
\newgame
\fenboard{4k3/4P3/5K2/8/8/8/8/8 w - - 0 1}
\showboard
\end{center}

% \begin{shadequote}[]{}
\textit{In the position above, with White to move, in classical chess the position would be a draw due to stalemate after Ke6. Yet, the same move wins in this variation of chess, so the defending side needs to steer away from these types of endgames.}

\textit{Similarly, the stalemates that arise in K+N+N vs K are now wins rather than draws, for example:}
% \end{shadequote}

\begin{center}
\newgame
\fenboard{6N1/8/8/5K1k/8/5N2/8/8 b - - 0 1}
\showboard
\end{center}

% \begin{shadequote}[]{}
\textit{Looking at the games of AlphaZero, it seems that there are enough defensive resources in most middlegame positions
that certain types of inferior endgame positions, now possible under this rule chance, could be avoided and defended.
A strong player can in principle learn to navigate to these positions to take advantage of them, or find ways to escape them.}

\textit{In terms of the anticipated effect on human play, I would still expect this rule change to lead to a higher percentage of wins in endgames where one side has a clear advantage, but probably not as much as one would otherwise have been expecting. This may be a nice variation of chess for chess enthusiasts with an interest in endgame patterns.}
\end{shadequote}

\subsubsection{Main Lines}

Here we discuss ``main lines'' of AlphaZero under Stalemate=win
chess, when playing with roughly one minute per move from a particular fixed first move. Note that these are not purely deterministic, and each of the given lines is merely one of several highly promising and likely options. Here we give the first 20 moves in each of the main lines, regardless of the position.

\paragraph{Main line after e4}
The main line of AlphaZero after \move{1}e4 in Stalemate=win chess is:

\move{1}e4 \bookmove e5 \move{2}Nf3  Nc6  \move{3}Bb5  Nf6  \move{4}O-O  Nxe4  \move{5}Re1  Nd6  \move{6}Nxe5  Be7  \move{7}Bf1  Nxe5  \move{8}Rxe5  O-O  \move{9}d4  Bf6  \move{10}Re1  Re8  \move{11}c3  Rxe1  \move{12}Qxe1  Ne8  \move{13}Bf4  d5  \move{14}Nd2  Bf5  \move{15}Qe2  Nd6  \move{16}Re1  Qd7  \move{17}Qd1  c6  \move{18}Nb3  b6  \move{19}Nd2  Ne4  \move{20}Nf3  Bg4

\begin{center}
\newgame
\fenboard{r5k1/p2q1ppp/1pp2b2/3p4/3PnBb1/2P2N2/PP3PPP/3QRBK1 w - - 4 21}
\showboard
\end{center}

\paragraph{Main line after d4}
The main line of AlphaZero after \move{1}d4 in Stalemate=win chess is:

\move{1}d4 \bookmove Nf6 \move{2}c4 e6 \move{3}g3  Bb4+
\move{4}Bd2  Be7  \move{5}Qc2  c6  \move{6}Bg2  d5  \move{7}Nf3  b6  \move{8}O-O  O-O  \move{9}Bf4  Bb7  \move{10}Rd1  Nbd7  \move{11}Ne5  Nh5  \move{12}Bd2  Nhf6  \move{13}cxd5  cxd5  \move{14}Nc6  Qe8  \move{15}Nxe7+  Qxe7  \move{16}Qc7  Ba6  \move{17}Nc3  Rfc8  \move{18}Qf4  Nf8  \move{19}Be1  h6  \move{20}Qd2  Ng6

\begin{center}
\newgame
\fenboard{r1r3k1/p3qpp1/bp2pnnp/3p4/3P4/2N3P1/PP1QPPBP/R2RB1K1 w - - 2 21}
\showboard
\end{center}

\paragraph{Main line after c4}
The main line of AlphaZero after \move{1}c4 in Stalemate=win chess is:

\move{1}c4 \bookmove e5  \move{2}g3  Nf6  \move{3}Bg2  Bc5  \move{4}d3  d5  \move{5}cxd5  Nxd5
\move{6}Nf3  Nc6  \move{7}O-O  O-O  \move{8}Nc3  Nxc3  \move{9}bxc3  Rb8  \move{10}Bb2  Re8  \move{11}d4  Bd6  \move{12}e4  Bg4  \move{13}h3  Bh5  \move{14}Qc2  f6  \move{15}d5  Na5  \move{16}c4  b6  \move{17}Nh4  Nb7  \move{18}Rae1  Rf8  \move{19}f4  Nc5  \move{20}Re3  Qd7

\begin{center}
\newgame
\fenboard{1r3rk1/p1pq2pp/1p1b1p2/2nPp2b/2P1PP1N/4R1PP/PBQ3B1/5RK1 w - - 3 21}
\showboard
\end{center}

\subsubsection{Instructive games}

The games in Stalemate=win chess are at the first glance almost indistinguishable from those of classical chess, as the lines are merely a subset of the lines otherwise playable and plausible under classical rules.

\paragraph{Game AZ-8: AlphaZero Stalemate=win vs AlphaZero Stalemate=win\footnote{Game AZ-8 was  labelled wrongly in arXiv:2009.04374v1,
and this game replaces it.}}
The first ten moves for White and Black have been sampled randomly from AlphaZero's opening ``book'', with the probability proportional to the time spent calculating each move. The remaining moves follow best play, at roughly one minute per move.

\move{1}e4 c6 \move{2}d4 d5 \move{3}e5 Bf5
\move{4}Be2 h6
\move{5}Nf3 e6 \move{6}O-O a6 \move{7}c4 dxc4
\move{8}Bxc4 Nd7
\move{9}Bd3 Ne7 \move{10}Nc3 Nb6 \move{11}Re1 Nbd5 \move{12}Ne4 Ng6
\move{13}Bd2 Qb6

\begin{center}
\newgame
\fenboard{r3kb1r/1p3pp1/pqp1p1np/3nPb2/3PN3/3B1N2/PP1B1PPP/R2QR1K1 w kq - 11 14}
\showboard
\end{center}

\move{14}h3 Be7 \move{15}Bc4 Ngf4 \move{16}Bf1 g5
\move{17}Ng3 Rg8 \move{18}Bc3 Qc7 \move{19}Nh2 Bg6 \move{20}Ne4 O-O-O

\begin{center}
\newgame
\fenboard{2kr2r1/1pq1bp2/p1p1p1bp/3nP1p1/3PNn2/2B4P/PP3PPN/R2QRBK1 w - - 8 21}
\showboard
\end{center}

\move{21}g3 Nxc3 \move{22}bxc3 Nd5 \move{23}Rc1 Qa5 \move{24}Qb3 Qa3
\move{25}Qc2 Kb8 \move{26}Nf3 Bb4

\begin{center}
\newgame
\fenboard{1k1r2r1/1p3p2/p1p1p1bp/3nP1p1/1b1PN3/q1P2NPP/P1Q2P2/2R1RBK1 w - - 9 27}
\showboard
\end{center}

\move{27}cxb4 Qxf3 \move{28}Rb1 Bxe4
\move{29}Rxe4 Qf5 \move{30}Bg2 Nc7 \move{31}Rb2 Rd7 \move{32}a4 Rgd8
\move{33}Qc3 h5 \move{34}Kh2 Qh7 \move{35}Qd2 Qg8

\begin{center}
\newgame
\fenboard{1k1r2q1/1pnr1p2/p1p1p3/4P1pp/PP1PR3/6PP/1R1Q1PBK/8 w - - 4 36}
\showboard
\end{center}

\move{36}h4 gxh4
\move{37}Rxh4 f5 \move{38}exf6 e5 \move{39}b5 cxb5 \move{40}Qe3 exd4
\move{41}Qf4 Ka7
\move{42}Qf3 Rb8 \move{43}Rb4 Qd8 \move{44}Bh3 Rd5
\move{45}f7 bxa4 \move{46}Rxa4 d3

\begin{center}
\newgame
\fenboard{1r1q4/kpn2P2/p7/3r3p/R6R/3p1QPB/5P1K/8 w - - 0 47}
\showboard
\end{center}

\move{47}Qe3+ Ka8 \move{48}Rad4 Qf8
\move{49}Be6 Rxd4 \move{50}Rxd4 h4 \move{51}Bc4 hxg3+ \move{52}fxg3 Nb5
\move{53}Bxb5 axb5 \move{54}Rxd3 b4 \move{55}Rd5 Rc8

\begin{center}
\newgame
\fenboard{k1r2q2/1p3P2/8/3R4/1p6/4Q1P1/7K/8 w - - 2 56}
\showboard
\end{center}

White is clearly winning here,
and Ra5+ is good and tempting.
AlphaZero is only optimised for achieving an end result. Even though a slower approach achieves the same outcome, a win is a win!
This game ultimately finishes with checkmate.

\move{56}Rf5 Kb8
\move{57}g4 Ka8 \move{58} Rf2 b3 \move{59} Qxb3 Qe7 \move{60}Ra2+ Kb8
\move{61}Qg3+ Rc7 \move{62}Rf2 Ka7 \move{63}f8=Q Qh7+ \move{64}Qh3 Qb1
\move{65}Qf5 Qa1 \move{66}Qf1 Qh8+ \move{67}Qh5 Qg7 \move{68}Qh4 Rc5
\move{69}Kh1 Ra5 \move{70}Qg3 Ra4 \move{71}Rf3 Ra2 \move{72}Qff2+ Rxf2
\move{73}Qxf2+ Ka6 \move{74}Qg3 b5 \move{75}Qf4 Qg8 \move{76}Qf6+ Ka5
\move{77}Qf5 Ka4 \move{78}Qf8 Qh7+ \move{79}Kg2 b4 \move{80}Qe8+ Ka5
\move{81}Qh5+ Qxh5 \move{82}gxh5 Ka4 \move{83}Rf1 Ka3 \move{84}Kf2 Kb2
\move{85}Ke1 b3 \move{86}Kd1 Kc3 \move{87}Kc1 Kd4 \move{88}h6 b2+
\move{89}Kxb2 Ke5 \move{90}Re1+ Kf6 \move{91}Rd1 Kg6 \move{92}Rc1 Kxh6
\move{93}Rc3 Kg5 \move{94}Rc5+ Kh4 \move{95}Rc7 Kg3 \move{96}Rb7 Kh4
\move{97}Ra7 Kh3 \move{98}Kc3 Kg3 \move{99}Ra1 Kh4 \move{100}Ra3 Kh3
\move{101}Kd4+ Kg4 \move{102}Ra5 Kf4 \move{103}Ra7 Kg5 \move{104}Ra1 Kf5
\move{105}Ra3 Kg4 \move{106}Ke5 Kh5 \move{107}Ra5 Kg5 \move{108}Rd5 Kh6
\move{109}Rd7 Kg5 \move{110}Rc7 Kh6 \move{111}Kf5 Kh5 \move{112}Rh7\#
\whitewins

%% file: variants/SAN/torpedo.tex
\subsection{Torpedo}
\label{sec-torpedo}

In the variation of chess that we've named Torpedo chess, the pawns can move by either one or two squares forward from anywhere on the board rather than just from the initial squares, which is the case in Classical chess. We will refer to the pawn moves that involve advancing them by two squares as ``torpedo'' moves.

We have also looked at a Semi-torpedo variant in our experiments, where we only add a partial extension to the original rule and have the pawns be able to move by two squares from the 2nd/3rd and 6th/7th rank for White and Black respectively.
In this section we will focus on the universal motifs of full Torpedo chess, and cover the sub-motifs and sub-patterns that correspond to Semi-torpedo chess in its own dedicated section in Appendix \ref{sec-semitorpedo}.

\subsubsection{Motivation}

In a sense, having the pawns always be able to move by one or two squares makes the pawn movement more consistent, as it removes a ``special case'' of them only being able to do the ``double move'' from their initial position. Increasing pawn mobility has the potential of speeding up all stages of the game. It adds additional attacking motifs to the openings and changes opening theory, it makes middlegames more complicated, and changes endgame theory in cases where pawns are involved.

\subsubsection{Assessment}

The assessment of the Torpedo chess variant, as provided by Vladimir Kramnik:

\begin{shadequote}[]{}
\textit{The pawns become quite powerful in Torpedo chess. Passed pawns are in particular a very strong asset and the value of pawns changes based on the circumstances and closer to the endgame. All of the attacking opportunities increase and this strongly favours the side with the initiative, which makes taking initiative a crucial part of the game. Pawns are very fast, so less of a strategical asset and much more tactical instead. The game becomes more tactical and calculative compared to standard chess.}

\textit{There is a lot of prophylactic play, which is why some games don't feature many ``torpedo'' moves -- ``torpedo'' moves are simply quite powerful and the play often proceeds in a way where each player positions their pawn structure so as to disincentivise ``torpedo'' moves, either by the virtue of directly blocking their advance, or by placing their own pawns on squares that would be able to capture ``en passant'' if ``torpedo'' moves were to occur.}

\textit{This seems to favour the ``classical'' style of play in classical chess, which advocates for strong central control rather than conceding space to later attack the center once established. It seems like it is more difficult to play openings like the Grunfeld or the King's Indian defence.}

\textit{In summary, this is an interesting chess variant, leading to lots of decisive games and a potentially high entertainment value, involving lots of tactical play.}
\end{shadequote}

\subsubsection{Main Lines}

Here we discuss ``main lines'' of AlphaZero under Torpedo chess, when playing with roughly one minute per move from a particular fixed first move. Note that these are not purely deterministic, and each of the given lines is merely one of several highly promising and likely options. Here we give the first 20 moves in each of the main lines, regardless of the position.

\paragraph{Main line after e4}
The main line of AlphaZero after \move{1}e4 in Torpedo chess is:

\move{1}e4 \bookmove c5  \move{2}Nf3  d6  \move{3}d4  Nf6  \move{4}Nc3  cxd4  \move{5}Nxd4  a6  \move{6}g3  h6  \move{7}Bg2  e5  \move{8}Nde2  Be7  \move{9}Be3  Be6  \move{10}Nd5  Nbd7  \move{11}c4  Rc8  \move{12}b3  Ng4  \move{13}O-O  Nxe3  \move{14}Nxe3  \specialmove{h4}  \move{15}Nf5  Kf8  \move{16}Qd2  Nf6  \move{17}Nc3  g6  \move{18}Nxe7  Qxe7  \move{19}Rad1  Rc6  \move{20}Rc1  Kg7

\begin{center}
\newgame
\fenboard{7r/1p2qpk1/p1rpbnp1/4p3/2P1P2p/1PN3P1/P2Q1PBP/2R2RK1 w - - 0 1}
\showboard
\end{center}

\paragraph{Main line after d4}
The main line of AlphaZero after \move{1}d4 in Torpedo chess is:

\move{1}d4 \bookmove d5  \move{2}c4  e6  \move{3}Nf3  Nf6  \move{4}Nc3  a6  \move{5}e3  b6  \move{6}Bd3  Bb7  \move{7}O-O  Bd6  \move{8}cxd5  exd5  \move{9}Ne5  O-O  \move{10}a3  Nbd7  \move{11}f4  Ne4  \move{12}Bd2  c5  \move{13}Be1  cxd4  \move{14}exd4  b5  \move{15}h3  Rc8  \move{16}Qe2  Ndf6  \move{17}a4  b4  \move{18}Nxe4  dxe4  \move{19}Bxa6  Bxa6  \move{20}Qxa6  Bxe5

\begin{center}
\newgame
\fenboard{2rq1rk1/5ppp/Q4n2/4b3/Pp1PpP2/7P/1P4P1/R3BRK1 w Q - 0 1}
\showboard
\end{center}

\paragraph{Main line after c4}
The main line of AlphaZero after \move{1}c4 in Torpedo chess is:

\move{1}c4 \bookmove c5  \move{2}e3  e6  \move{3}d4  d5  \move{4}Nc3  Nc6  \move{5}Nf3  Nf6  \move{6}a3  h6  \move{7}dxc5  Bxc5  \move{8}cxd5  exd5  \move{9}b4  Bd6  \move{10}Bb2  O-O  \move{11}Be2  a5  \move{12}b5  Ne7  \move{13}O-O  Re8  \move{14}Rc1  Be6  \move{15}Bd3  Ng6  \move{16}Ne2  a4  \move{17}Rc2  Qe7  \move{18}Qa1  Nf8  \move{19}Nfd4  N8d7  \move{20}Ng3  Ng4

\begin{center}
\newgame
\fenboard{r3r1k1/1p1nqpp1/3bb2p/1P1p4/p2N2n1/P2BP1N1/1BR2PPP/Q4RK1 w q - 0 1}
\showboard
\end{center}

\subsubsection{Instructive games}

Here we showcase several instructive games that illustrate the type of play that frequently arises in Torpedo chess, along with some selected extracted game positions in cases where particular (endgame) move sequences are of interest.

\paragraph{Game AZ-9: AlphaZero Torpedo vs AlphaZero Torpedo}
The first ten moves for White and Black have been sampled randomly from AlphaZero's opening ``book'', with the probability proportional to the time spent calculating each move. The remaining moves follow best play, at roughly one minute per move.
% game 10 from on-policy games

\move{1}d4  d5  \move{2}Nf3  Nf6  \move{3}c4  e6  \move{4}Nc3  c6  \move{5}e3  Nbd7  \move{6}g3  Ne4  \move{7}Nxe4  dxe4  \move{8}Nd2  f5  \move{9}c5  Be7  \move{10}h4  O-O

\begin{center}
\newgame
\fenboard{r1bq1rk1/pp1nb1pp/2p1p3/2P2p2/3Pp2P/4P1P1/PP1N1P2/R1BQKB1R w KQq - 0 1}
\showboard
\end{center}

\move{11}\specialmove{g5}  b6  \move{12}b4  a5  \move{13}Bc4  axb4  \move{14}Bxe6+  Kh8  \move{15}Bb2  Ne5

\begin{center}
\newgame
\fenboard{r1bq1r1k/4b1pp/1pp1B3/2P1npP1/1p1Pp2P/4P3/PB1N1P2/R2QK2R w KQq - 0 1}
\showboard
\end{center}

\move{16}Bc4  Ng4  \move{17}\specialmove{d6} \specialmove{cxd5}  \move{18}\specialmove{h6}  Rg8

\begin{center}
\newgame
\fenboard{r1bq2rk/4b1pp/1p5P/2Pp1pP1/1pB1p1n1/4P3/PB1N1P2/R2QK2R w KQq - 0 1}
\showboard
\end{center}

\move{19}hxg7+  Rxg7  \move{20}\specialmove{c7}  Qd7  \move{21}Bxd5  Qxd5  \move{22}Nc4

\begin{center}
\newgame
\fenboard{r1b4k/2P1b1rp/1p6/3q1pP1/1pN1p1n1/4P3/PB3P2/R2QK2R w KQq - 0 1}
\showboard
\end{center}

\blackmove{22}Qg8  \move{23}Ne5  Nxe5  \move{24}Bxe5  Bxg5  \move{25}Qh5

\begin{center}
\newgame
\fenboard{r1b3qk/2P3rp/1p6/4BpbQ/1p2p3/4P3/P4P2/R3K2R w KQq - 0 1}
\showboard
\end{center}

\blackmove{25}\specialmove{b2}  \move{26}\specialmove{axb3}  Rxa1+  \move{27}Bxa1  Be7  \move{28}f4  exf3  \move{29}Rg1  Bf8  \move{30}Qg5 

\begin{center}
\newgame
\fenboard{2b2bqk/2P3rp/1p6/5pQ1/8/1P2Pp2/8/B3K1R1 w - - 0 1}
\showboard
\end{center}

\blackmove{30}h6  \move{31}Qxh6+  Qh7  \move{32}Bxg7+  Bxg7  \move{33}Qxh7+  Kxh7  \move{34}Kf2  Be5  \move{35}Rd1  Bb7  \move{36}Rc1  Bc8  \move{37}Kxf3  Kg6  \move{38}e4 \specialmove{b4}  \move{39}Rc4  Kf6  \move{40}Rc6+  Kg5  \move{41}Ke3  f4+
\move{42}Kf2  Bd4+
\move{43}Kg2  Be5  \move{44}Rc5  Kf6  \move{45}Kf3  Ke6  \move{46}Rb5  Bd7  \move{47}Rxb4  Bxc7  \move{48}Rd4  Ke7 
\move{49}Rd2  Be8  \move{50}Rh2  Bd6  \move{51}Rh7+  Ke6  \move{52}Rh6+  Ke7  \move{53}Rh2  Kf6  \move{54}Rh8  Ke7  \move{55}Rh2  Kf6  \move{56}Rh6+  Ke7  \move{57}Rh1  Kf6  \move{58}Rh8  Ke7  \move{59}Rh6  Be5  
\move{60}b4  Kd7  \move{61}Kf2  Bc3
\move{62}\specialmove{b6}  Kc8  \move{63}Rd6  Kb7  \move{64}Kf3  Be5  \move{65}Rd5  Bb8  \move{66}Rd8  Bc6  \move{67}Rf8  Ba4  \move{68}Ke2  Be5  \move{69}Rf5  Bb8  \move{70}Rf6  Be5  \move{71}Rf5  Bb8  \move{72}e5  Kxb6  \move{73}Rxf4  Bb5+  \move{74}Kd1  Bxe5  \move{75}Rf5  Bh2  \move{76}Rxb5+  Kxb5
\gamedrawn

\paragraph{Game AZ-10: AlphaZero Torpedo vs AlphaZero Torpedo}
The first ten moves for White and Black have been sampled randomly from AlphaZero's opening ``book'', with the probability proportional to the time spent calculating each move. The remaining moves follow best play, at roughly one minute per move.
% game 13 from on-policy games

\move{1}d4  d5  \move{2}c4  e6  \move{3}Nf3  Nf6  \move{4}Nc3  a6  \move{5}e3  b6  \move{6}Bd3  Bb7  \move{7}O-O  Bd6  \move{8}cxd5  exd5  \move{9}a3  O-O  \move{10}Ne5  c5  \move{11}f4  Nbd7  \move{12}Bd2  cxd4  \move{13}exd4  Ne4  \move{14}Be1  b5  \move{15}h3  Rc8  \move{16}Qe2  Ndf6

\begin{center}
\newgame
\fenboard{2rq1rk1/1b3ppp/p2b1n2/1p1pN3/3PnP2/P1NB3P/1P2Q1P1/R3BRK1 w Q - 0 1}
\showboard
\end{center}

A normal-looking position arises in the middlegame (this is one of AlphaZero's main lines in this variation of chess), but the board soon explodes in tactics.

\move{17}a4  b4  \move{18}Nxe4  dxe4  \move{19}Bxa6  Bxa6  \move{20}Qxa6  Bxe5  \move{21}dxe5  Qd4+  \move{22}Kh1  Qxb2  \move{23}Bh4  \specialmove{e2}

\begin{center}
\newgame
\fenboard{2r2rk1/5ppp/Q4n2/4P3/Pp3P1B/7P/1q2p1P1/R4R1K w Q - 0 1}
\showboard
\end{center}

\move{24}Rfe1  Nd5  \move{25}\specialmove{e7}  Rfe8  \move{26}Qd6  Qd2  \move{27}\specialmove{a6}  Ra8  \move{28}Qc6  \specialmove{b2}

\begin{center}
\newgame
\fenboard{r3r1k1/4Pppp/P1Q5/3n4/5P1B/7P/1p1qp1P1/R3R2K w Qq - 0 1}
\showboard
\end{center}

A series of consecutive torpedo moves had given rise to this incredibly sharp position, with multiple passed pawns for White and Black, and the threats are culminating, as demonstrated by the following tactical sequence.

\move{29}Qxe8+  Rxe8  \move{30}\specialmove{a8=Q}  Nc7

\begin{center}
\newgame
\fenboard{Q3r1k1/2n1Pppp/8/8/5P1B/7P/1p1qp1P1/R3R2K w Q - 0 1}
\showboard
\end{center}

\move{31}Qa5  bxa1=Q  \move{32}Qxd2  Qa4  \move{33}Rxe2  f6  \move{34}Re3  Kf7  \move{35}Bf2  Qb5  \move{36}Qd6  Qf1+  \move{37}Bg1  Qc4  \move{38}f5  g6  \move{39}Rg3  gxf5  \move{40}Qd1

\begin{center}
\newgame
\fenboard{4r3/2n1Pk1p/5p2/5p2/2q5/6RP/6P1/3Q2BK w - - 0 1}
\showboard
\end{center}

Here Black utilizes a torpedo move to give back the pawn and protect h5 via d5.

\blackmove{40}\specialmove{f3}  \move{41}Qxf3  Qd5  \move{42}Qg4  Ne6  \move{43}Be3  Rb8  \move{44}Qa4  Kxe7  \move{45}Bc1  Kf7  \move{46}Qc2  f5  \move{47}Kh2  Rb5  \move{48}Qa4  h5  \move{49}Qa7+  Qb7  \move{50}Qa4  Qd5  \move{51}Ba3  f4  \move{52}Rf3  Ra5  \move{53}Qb4  Rc5

\begin{center}
\newgame
\fenboard{8/5k2/4n3/2rq3p/1Q3p2/B4R1P/6PK/8 w - - 0 1}
\showboard
\end{center}

\move{54}Rf2  Qf5  \move{55}Qb2  Rd5  \move{56}Qb7+  Kg6  \move{57}Qc6  Kh7  \move{58}Bb2  Rd8  \move{59}Qb7+  Kg8  \move{60}Rf3  Qg6  \move{61}Be5  Qf5  \move{62}Ba1  Rd3  \move{63}Qb1  Rd5  \move{64}Qb8+  Rd8  \move{65}Qb2  Nd4  \move{66}Rf2  Ne6  \move{67}Qb3  Kh7  \move{68}Qb7+  Kg8  \move{69}Qa6  Kh7  \move{70}Qa7+  Kg6  \move{71}Qb7  Rd1  \move{72}Qa8  Rd8  \move{73}Qc6  Kh7  \move{74}Qb7+  Kg6  \move{75}Bb2  Rd1  \move{76}Qb8  Re1  \move{77}Bc3  Re3  \move{78}Qg8+  Kh6  \move{79}Qh8+  Kg6  \move{80}Qe8+  Kh7  \move{81}Qd7+  Kg6  \move{82}Qc6  Kh7  \move{83}Qb7+  Kg8  \move{84}Qa8+  Kh7  \move{85}Qb7+  Kg8  \move{86}Qc8+  Kh7  \move{87}Qh8+  Kg6  \move{88}Qg8+  Kh6  \move{89}Qc8  Kh7  \move{90}Qd7+  Kg6  \move{91}Qe8+  Kh7  \move{92}Bd2  Rd3  \move{93}Qb8  Rd8  \move{94}Qb7+  Kg8  \move{95}Qc6  Qg6  \move{96}Qc3  Rd4  \move{97}Qa5  Kh7  \move{98}Kg1  Re4  \move{99}h4  Nd4  \move{100}Qc7+  Qg7  \move{101}Qa5  Qf7  \move{102}Qa6  f3
\move{103}Qh6+  Kg8  \move{104}Qg5+  Kh7  \move{105}Be3  Ne2+

\begin{center}
\newgame
\fenboard{8/5q1k/8/6Qp/4r2P/4Bp2/4nRP1/6K1 w - - 0 1}
\showboard
\end{center}

And the game soon ends in a draw.

\move{106}Kh2  Qc7+  \move{107}Kh1  Ng3+  \move{108}Kg1  Ne2+  \move{109}Kf1  Ng3+  \move{110}Kg1  Ne2+  \move{111}Rxe2  fxe2  \move{112}Qxh5+  Kg8  \move{113}Qxe2  Rxh4  \move{114}Qa6  Qe7  \move{115}Qc8+  Kh7  
\move{116}Qf5+  Kg7  \move{117}Qg5+  Qxg5  \move{118}Bxg5  Re4  \move{119}Kf2  Kg6  \move{120}Kf3  Re8  \move{121}Bf4  Re7  \move{122}g4  Rf7  \move{123}Kg3  Rg7  \move{124}Bb8  Kg5  \move{125}Bd6  Rb7  \move{126}Bf4+  Kg6  \move{127}Kf3  Ra7  \move{128}Bb8  Rb7  \move{129}Bf4  Ra7  \move{130}Bb8  Rb7  \move{131}Bf4
\gamedrawn

\paragraph{Game AZ-11: AlphaZero Torpedo vs AlphaZero Torpedo}
The first ten moves for White and Black have been sampled randomly from AlphaZero's opening ``book'', with the probability proportional to the time spent calculating each move. The remaining moves follow best play, at roughly one minute per move.
% game 7 from on-policy games

\move{1}d4  d5  \move{2}Nf3  Nf6  \move{3}c4  e6  \move{4}Nc3  a6  \move{5}e3  b6  \move{6}Bd3  Bb7  \move{7}a3  g6  \move{8}cxd5  exd5  \move{9}h4  Nbd7  \move{10}O-O  Bd6  \move{11}Nxd5

\begin{center}
\newgame
\fenboard{r2qk2r/1bpn1p1p/pp1b1np1/3N4/3P3P/P2BPN2/1P3PP1/R1BQ1RK1 w Qkq - 0 1}
\showboard
\end{center}

An interesting tactical motif, made possible by torpedo moves. One has to wonder, after \blackmove{11}Nxd5 \move{12}e4,
what happens on \blackmove{12}Nf4? The game would have followed \move{13}e5 Nxd3 \move{14}exd6 Nxc1 \move{15}dxc7 Qxc7

\begin{center}
\newgame
\fenboard{r3k2r/1bqn1p1p/pp4p1/8/3P3P/P4N2/1P3PP1/R1nQ1RK1 w Qkq - 0 1}
\showboard \\
\textsf{analysis diagram}
\end{center}

and here, White would have played \move{16}\specialmove{d6}, a torpedo move -- gaining an important tempo while weakening the Black king.
\blackmove{16}Qc4 \move{17}Rxc1, followed by Re1+ once the queen has moved. AlphaZero evaluates this position as being strongly in White's favour, despite the material deficit. 

Going back to the game continuation,

\blackmove{11}Nxd5  \move{12}e4  O-O  \move{13}exd5  Bxd5  \move{14}Bg5  Qb8  \move{15}Re1  Re8  \move{16}Nd2  Rxe1+  \move{17}Qxe1  Bf8  \move{18}Qe3  c6

\begin{center}
\newgame
\fenboard{rq3bk1/3n1p1p/ppp3p1/3b2B1/3P3P/P2BQ3/1P1N1PP1/R5K1 w Qq - 0 1}
\showboard
\end{center}

Now we see several torpedo moves taking place. First White takes the opportunity to plant a pawn on h6, weakening the Black king, then Black responds by a4 and b4, getting the queenside pawns in motion and creating counterplay on the other side of the board.

\move{19}\specialmove{h6}  \specialmove{a4}  \move{20}Re1  Qa7  \move{21}Bf5 \specialmove{b4}

\begin{center}
\newgame
\fenboard{r4bk1/q2n1p1p/2p3pP/3b1BB1/pp1P4/P3Q3/1P1N1PP1/4R1K1 w q - 0 1}
\showboard
\end{center}

\move{22}Bg4  Qb7  \move{23}Bh3  Rc8  \move{24}Qd3  Ra8  \move{25}Qe3  Nb6  \move{26}Rc1  Nd7  \move{27}g3  b3  \move{28}Bg4  Qc7  \move{29}Re1  Nb6  \move{30}Ne4  Bxe4  \move{31}Qxe4  Qd6  \move{32}Qd3  Qd5  \move{33}Re5  Qc4  \move{34}Qe4  c5  \move{35}Re8  Rxe8  \move{36}Qxe8  cxd4

\begin{center}
\newgame
\fenboard{4Qbk1/5p1p/1n4pP/6B1/p1qp2B1/Pp4P1/1P3P2/6K1 w - - 0 1}
\showboard
\end{center}

The position is getting sharp again, with Black having gained a passed pawn, and White making threats around the Black king.

\move{37}Be7  Qc1+  \move{38}Kg2  Qxh6  \move{39}Bc5

\begin{center}
\newgame
\fenboard{4Qbk1/5p1p/1n4pq/2B5/p2p2B1/Pp4P1/1P3PK1/8 w - - 0 1}
\showboard
\end{center}

A critical moment, and a decision which shows just how valuable the advanced pawns are in this chess variation. Normally it would make sense to save the knight, but AlphaZero decides to keep the pawn instead, and rely on promotion threats coupled with checks on d5.

\blackmove{39}d3  \move{40}Bxb6  Qg5  \move{41}Bd1  Qd5+  \move{42}Kh2  Qe6

\begin{center}
\newgame
\fenboard{4Qbk1/5p1p/1B2q1p1/8/p7/Pp1p2P1/1P3P1K/3B4 w - - 0 1}
\showboard
\end{center}

Being a piece down, Black offers an exchange of queens, an unusual sight, but tactically justified -- Black is also threatening to capture on a3, and that threat is hard to meet.
White can't passively ignore the capture and defend the b2 pawn with the bishop, because Black could capture on b2, offering the piece for the second time -- and then follow up by an immediate a3, knowing that bxa3 would allow for
\specialmove{b1=Q}. In addition, Black could retreat the bishop instead of capturing on b2, to make room for \specialmove{a2} bxa3 and again \specialmove{b1=Q}. So, it's again a torpedo move that makes a difference and justifies the tactical sequence. 

\move{43}Qb5  Qf6  \move{44}Kg2  h5  \move{45}Be3  Qxb2  \move{46}Qxa4  Qxa3  \move{47}Qxb3  Qxb3  \move{48}Bxb3

\begin{center}
\newgame
\fenboard{5bk1/5p2/6p1/7p/8/1B1pB1P1/5PK1/8 w - - 0 1}
\showboard
\end{center}

White is a piece up for two pawns, and has the bishop pair. Yet, Black is just in time to use a torpedo move to shut the White king out and exchange a pair of pawns on the h-file (by another torpedo move).

\blackmove{48}\specialmove{g4}  \move{49}Bd2  Kg7  \move{50}Kf1  f5  \move{51}Ke1  Be7  \move{52}Bc4  \specialmove{h3}

\begin{center}
\newgame
\fenboard{8/4b1k1/8/5p2/2B3p1/3p2Pp/3B1P2/4K3 w - - 0 1}
\showboard
\end{center}

\move{53}\specialmove{gxh4}  Bxh4  \move{54}Kf1  Bg5  \move{55}Bc3+  Bf6  \move{56}Bd2  Bg5  \move{57}Bc3+  Bf6  \move{58}Bxf6  Kxf6  \move{59}Bxd3  f4  \move{60}Be4  g3  \move{61}Bg2  gxf2  \move{62}Ba8  Kf5  \move{63}Kxf2  Kg4  \move{64}Bb7  Kh5  \move{65}Ba6  f3  \move{66}Kxf3
\gamedrawn

\paragraph{Game AZ-12: AlphaZero Torpedo vs AlphaZero Torpedo}
Playing from a predefined Nimzo-Indian opening position (the first 3 moves for each side). The remaining moves follow best play, at roughly one minute per move.
% game 4 from div games

\move{1}d4 \bookmove Nf6 \bookmove
\move{2}c4 \bookmove e6 \bookmove
\move{3}Nc3 \bookmove Bb4 \bookmove
\move{4}e3  Bxc3  \move{5}bxc3  d6  \move{6}Nf3  O-O  \move{7}Ba3  Re8  \move{8}\specialmove{e5}

\begin{center}
\newgame
\fenboard{rnbqr1k1/ppp2ppp/3ppn2/4P3/2PP4/B1P2N2/P4PPP/R2QKB1R w KQq - 0 1}
\showboard
\end{center}

Already we see the first torpedo move, keeping the initiative.

\blackmove{8}dxe5  \move{9}Nxe5  Nbd7  \move{10}Bd3  c5  \move{11}O-O  Qa5  \move{12}Bb2  Nxe5  \move{13}dxe5  Nd7  \move{14}Re1  f5  \move{15}Qh5  Re7

\begin{center}
\newgame
\fenboard{r1b3k1/pp1nr1pp/4p3/q1p1Pp1Q/2P5/2PB4/PB3PPP/R3R1K1 w Qq - 0 1}
\showboard
\end{center}

\move{16}Qh4  Re8  \move{17}Qg3  Nf8  \move{18}Bc1  Kh8  \move{19}Be2  Bd7  \move{20}Rd1  Bc6  \move{21}Bh5  h6  \move{22}Bxe8  Rxe8  \move{23}Rd6  \specialmove{f3}

\begin{center}
\newgame
\fenboard{4rn1k/pp4p1/2bRp2p/q1p1P3/2P5/2P2pQ1/P4PPP/R1B3K1 w Q - 0 1}
\showboard
\end{center}

Here we see an effect of another torpedo move, after the exchange sacrifice earlier, taking over the initiative and creating a dangerous pawn.

\move{24}Bd2  fxg2  \move{25}f4  Qc7  \move{26}Be3  Qf7  \move{27}Bxc5  Qf5  \move{28}Rxc6  bxc6  \move{29}Bxa7  Ng6  \move{30}Be3  Ra8  \move{31}a4  Qd3

\begin{center}
\newgame
\fenboard{r6k/6p1/2p1p1np/4P3/P1P2P2/2PqB1Q1/6pP/R5K1 w Qq - 0 1}
\showboard
\end{center}

\move{32}Re1  Kh7  \move{33}a5  Rxa5  \move{34}Bb6  Qxg3  \move{35}hxg3  Ra2  \move{36}f5  exf5

\begin{center}
\newgame
\fenboard{8/6pk/1Bp3np/4Pp2/2P5/2P3P1/r5p1/4R1K1 w - - 0 1}
\showboard
\end{center}

The following move shows the power of advanced pawns -- \move{37}e6!, in order to create a threat of \move{38}\specialmove{e8=Q}, so Black has to block with the knight. If instead \move{37}e7, Black responds by first giving the knight for the pawn --
\blackmove{37}Nxe7, and then after \move{38}Rxe7 follows it up with \blackmove{38}\specialmove{h4}!, similar to the game continuation.

\move{37}e6  Ne7  \move{38}Bc5  \specialmove{h4}  \move{39}Bxe7  hxg3  \move{40}Re3  f4

\begin{center}
\newgame
\fenboard{8/4B1pk/2p1P3/8/2P2p2/2P1R1p1/r5p1/6K1 w - - 0 1}
\showboard
\end{center}

and Black manages to force a draw, as the pawns are just too threatening.

\move{41}Rf3  Ra1+  \move{42}Kxg2  Ra2+  \move{43}Kg1  Ra1+  \move{44}Kg2  Ra2+  \move{45}Kh1  Ra1+  \move{46}Kg2
\gamedrawn

\paragraph{Game AZ-13: AlphaZero Torpedo vs AlphaZero Torpedo}
The game starts from a predefined Ruy Lopez opening position (the first 5 plies). The remaining moves follow best play, at roughly one minute per move.
% game 13 from div games

\move{1}e4 \bookmove e5 \bookmove
\move{2}Nf3 \bookmove Nc6 \bookmove
\move{3}Bb5 \bookmove a6  \move{4}Bxc6  dxc6  \move{5}O-O  f6  \move{6}d4  exd4  \move{7}Nxd4  Bd6  \move{8}Be3  c5  \move{9}Ne2  Ne7  \move{10}Nbc3  b6  \move{11}Qd3  Be6  \move{12}Rad1  Be5  \move{13}Nd5  O-O  \move{14}Bf4  c6  \move{15}Nxe7+  Qxe7  \move{16}Bxe5  fxe5  \move{17}a3  

\begin{center}
\newgame
\fenboard{r4rk1/4q1pp/ppp1b3/2p1p3/4P3/P2Q4/1PP1NPPP/3R1RK1 w q - 0 1}
\showboard
\end{center}

Here comes the first torpedo move (b6-b4), gaining space on the queenside.

\blackmove{17}\specialmove{b4}  \move{18}a4  h6  \move{19}Qe3  Rad8  \move{20}f3  a5  \move{21}\specialmove{f5} 

\begin{center}
\newgame
\fenboard{3r1rk1/4q1p1/2p1b2p/p1p1pP2/Pp2P3/4Q3/1PP1N1PP/3R1RK1 w - - 0 1}
\showboard
\end{center}

Here we see an effect of another torpedo move, f3-f5, advancing towards the Black king.

\blackmove{21}Bc4  \move{22}Rde1  Qd6  \move{23}Rd1  Qxd1  \move{24}Rxd1  Rxd1+  \move{25}Kf2  Rd4  \move{26}g3  Rfd8

\begin{center}
\newgame
\fenboard{3r2k1/6p1/2p4p/p1p1pP2/PpbrP3/4Q1P1/1PP1NK1P/8 w - - 0 1}
\showboard
\end{center}

\move{27}Nxd4  cxd4  \move{28}Qd2  Rd6  \move{29}\specialmove{g5}  

\begin{center}
\newgame
\fenboard{6k1/6p1/2pr3p/p3pPP1/PpbpP3/8/1PPQ1K1P/8 w - - 0 1}
\showboard
\end{center}

White uses a torpedo move to generate play on the kingside.

\blackmove{29}hxg5  \move{30}b3  Bd5

\begin{center}
\newgame
\fenboard{6k1/6p1/2pr4/p2bpPp1/Pp1pP3/1P6/2PQ1K1P/8 w - - 0 1}
\showboard
\end{center}

The Black bishop can't be taken, due to a torpedo threat \specialmove{e3}+!

\move{31}Qxg5  Bxe4  \move{32}\specialmove{f7}+

\begin{center}
\newgame
\fenboard{6k1/5Pp1/2pr4/p3p1Q1/Pp1pb3/1P6/2P2K1P/8 w - - 0 1}
\showboard
\end{center}

And yet another torpedo strike, in order to capture on e5.

\blackmove{32}Kxf7  \move{33}Qxe5  Rf6+  \move{34}Ke1  Bxc2  \move{35}Qxd4  Bxb3  \move{36}Qd7+  Kg8  \move{37}Qd8+  Kh7  \move{38}Qd3+  g6  \move{39}Qxb3  c5

\begin{center}
\newgame
\fenboard{8/7k/5rp1/p1p5/Pp6/1Q6/7P/4K3 w - - 0 1}
\showboard
\end{center}

White ends up with the queen against the rook and two pawns, but this ends up being a draw, as the pawns are simply too fast and need to remain blocked.
Normally the queen on b3 would prevent the c5 pawn from moving, but a c5-c3 torpedo move shows that this is no longer the case!

\move{40}Kd1  \specialmove{c3}  \move{41}Qc4  Rf5  \move{42}Kc2  Rf2+  \move{43}Kb1  Rb2+  \move{44}Kc1  Rd2  \move{45}Kb1  Kh6  \move{46}h3  Rd1+  \move{47}Kc2  Rd2+  \move{48}Kb1  Rd1+  \move{49}Kc2  Rh1  \move{50}Qf4+  Kh7  \move{51}Qc7+  Kh6  \move{52}Qb8  Rf1  \move{53}Qh8+  Kg5  \move{54}Qd8+  Kh6  \move{55}h4  Rf2+  \move{56}Kb1  Rf1+  \move{57}Kc2  Rf2+  \move{58}Kb1  Rf1+  \move{59}Kc2
\gamedrawn

\paragraph{Game AZ-14: AlphaZero Torpedo vs AlphaZero Torpedo}
The position below, with Black to move, is taken from a game that was played with roughly one minute per move:
% game 42 from on policy games

\begin{center}
\newgame
\fenboard{8/6pk/Q6p/4Pp2/p3p1q1/4P1P1/P4P1P/B1b3K1 b - - 0 1}
\showboard
\end{center}

A dynamic position from an endgame reached in one of the AlphaZero games. White has an advanced passed pawn, which is quite threatening -- and Black tries to respond by creating threats around the White king. To achieve that, Black starts with a torpedo move:

\blackmove{31}\specialmove{h4}  \move{32}e6  hxg3  \move{33}hxg3  Bxe3

\begin{center}
\newgame
\fenboard{8/6pk/Q3P3/5p2/p3p1q1/4b1P1/P4P2/B5K1 b - - 0 1}
\showboard
\end{center}

White is one torpedo move away from queening, but has to first try to safeguard the king.

\move{34}Be5  Qd1+  \move{35}Qf1  Bxf2+

\begin{center}
\newgame
\fenboard{8/6pk/4P3/4Bp2/p3p3/6P1/P4b2/3q1QK1 b - - 0 1}
\showboard
\end{center}

Black is in time, due to the torpedo threats involving the e-pawn.

\move{36}Kxf2  e3+  \move{37}Kxe3  Qxf1

\begin{center}
\newgame
\fenboard{8/6pk/4P3/4Bp2/p7/4K1P1/P7/5q2 b - - 0 1}
\showboard
\end{center}

Black captures White's queen, but White creates a new one, with a torpedo move.

\move{38}\specialmove{e8=Q}  Qe1+  \move{39}Kd3  Qb1+  \move{40}Kc3  Qa1+  \move{41}Kb4  Qxa2

\begin{center}
\newgame
\fenboard{4Q3/6pk/8/4Bp2/pK6/6P1/q7/8 b - - 0 1}
\showboard
\end{center}

An interesting endgame arises, where White is up a piece, given that Black had to give away its bishop in the tactics earlier, and Black will soon only have a single pawn in return.
Yet, after a long struggle, AlphaZero manages to defend as Black and achieve a draw.

\move{42}Qe7  Qb3+  \move{43}Ka5  Qg8  \move{44}Kxa4  Kg6  \move{45}Bf4  Qc4+  \move{46}Ka5  Qd5+  \move{47}Kb4  Qd4+  \move{48}Kb3  Qd3+  \move{49}Kb2  Qd4+  \move{50}Kc2  Qd5  \move{51}Qe3  Qc4+  \move{52}Kd2  Qb4+  \move{53}Kd3  Qb3+  \move{54}Kd4  Qb4+  \move{55}Kd5  Qb5+  \move{56}Kd6  Qa6+  \move{57}Kd7  Qb5+
\move{58}Ke7  Qb7+  \move{59}Kd8  Qd5+  \move{60}Kc7  Qc4+  \move{61}Kb6  Qb4+  \move{62}Kc6  Qa4+  \move{63}Kb7  Qd7+  \move{64}Kb6  Kh5  \move{65}Qf3+  Kg6  \move{66}Kc5  Qa7+  \move{67}Kb4  Qb6+  \move{68}Kc3  Qa5+  \move{69}Kc2  Qa4+  \move{70}Kd2  Qa5+  \move{71}Ke2  Qb5+  \move{72}Kf2  Qb2+  \move{73}Kg1  Qb1+  \move{74}Kg2  Qb2+  \move{75}Kh3  Qc2  \move{76}Bd6  Qc1  \move{77}Bf4  Qc2  \move{78}Qe3  Qc6  \move{79}Qe1  Kf6  \move{80}Kh4  Kg6  \move{81}Kh3  Kf6  \move{82}Kh2  Qc2+  \move{83}Bd2  Qd3  \move{84}Bf4  Qd5  \move{85}Qe2  Qd4  \move{86}Bd2  Kf7  \move{87}Bg5  Qd5  \move{88}Bc1  Qc6  \move{89}Bf4  Kf6  \move{90}Qd3  Qe6  \move{91}Bd2  Kf7  \move{92}Kh3  Qf6  \move{93}Qd5+  Ke7  \move{94}\specialmove{g5}  \specialmove{fxg4}+  \move{95}Kxg4  Qe6+  \move{96}Qxe6  Kxe6  
\gamedrawn

\paragraph{Game AZ-15: AlphaZero Torpedo vs AlphaZero Torpedo}
The position below, with Black to move, is taken from a game that was played with roughly one minute per move:
% game 41 from on policy games

\begin{center}
\newgame
\fenboard{1r4k1/2pq2pp/3r4/1p1P1p2/8/3QP1P1/5P1P/2RR2K1 w - - 0 1}
\showboard
\end{center}

A position from one of the AlphaZero games, illustrating the utilization of pawns in a heavy piece endgame. The b-pawn is fast, and it gets pushed down the board via a torpedo move.

\blackmove{26}h5  \move{27}h4  \specialmove{b3}  \move{28}Qc4  Rb7  \move{29}Rb1  Qb5  \move{30}Qd4  c6  \move{31}dxc6  
\begin{center}
\newgame
\fenboard{6k1/1r4p1/2Pr4/1q3p1p/3Q3P/1p2P1P1/5P2/1R1R2K1 w - - 0 1}
\showboard
\end{center}

Unlike in Classical chess, this capture is possible, even though it seemingly hangs the queen. If Black were to capture it with the rook, the c-pawn would queen with check in a single move! The threat of \specialmove{c8=Q} forces Black to recapture the pawn instead. 

\blackmove{31}Rxc6  \move{32}\specialmove{e5}  \specialmove{fxe4}  \move{33}Qxe4  Rb8  \move{34}Rb2  Qc4  \move{35}Qe5  and the game soon ended in a draw.
\gamedrawn

\paragraph{Game AZ-16: AlphaZero Torpedo vs AlphaZero Torpedo}
The first ten moves for White and Black were sampled randomly from AlphaZero's opening ``book'', with the probability proportional to the time spent calculating each move. The remaining moves follow best play, at roughly one minute per move.

% manymoves 2 game, on-policy

\move{1}d4  Nf6  \move{2}c4  e6  \move{3}Nc3  d5  \move{4}Nf3  a6  \move{5}e3  b6  \move{6}g3  dxc4  \move{7}\specialmove{e5}  Nd5  \move{8}Bxc4  Be7  \move{9}O-O  Bb7  \move{10}Re1  h6  \move{11}a3  b5  \move{12}Bb3  Nxc3  \move{13}bxc3  \specialmove{a4}

\begin{center}
\newgame
\fenboard{rn1qk2r/1bp1bpp1/4p2p/1p2P3/p2P4/PBP2NP1/5P1P/R1BQR1K1 w Qkq - 0 1}
\showboard
\end{center}

In the early stage of the game, we see White using a torpedo e3-e5 move to expand in the center and Black responding by an a6-a4 torpedo move to gain space on the queenside.

\move{14}Bc2  Bd5  \move{15}Qe2  c6  \move{16}Nd2  Qa5  \move{17}Ne4  Nd7  \move{18}Bd2  Qc7  \move{19}Qg4  g6  \move{20}h3  O-O-O  \move{21}Qf4  Nb6  \move{22}\specialmove{c5}

\begin{center}
\newgame
\fenboard{2kr3r/2q1bp2/1np1p1pp/1pPbP3/p2PNQ2/P5PP/2BB1P2/R3R1K1 b Qk - 0 1}
\showboard
\end{center}

White moves forward with a c3-c5 torpedo move.

\blackmove{22}Nc4  \move{23}Bc3  Rdf8  \move{24}Nd6+  Bxd6  \move{25}exd6  g5

\begin{center}
\newgame
\fenboard{2k2r1r/2q2p2/2pPp2p/1pPb2p1/p1nP1Q2/P1B3PP/2B2P2/R3R1K1 w Qk - 0 1}
\showboard
\end{center}

\move{26}Qf6  Qd8  \move{27}Qxd8+  Rxd8  \move{28}Bd3  Rdg8  \move{29}f3  h5  \move{30}Be4  Re8  \move{31}Bd3  Rh6  \move{32}Rf1  f5  \move{33}Bxc4  Bxc4  \move{34}Rf2  Bd5  \move{35}Kh2  Rg6  \move{36}Rg1  Reg8  \move{37}Bd2  R6g7  \move{38}Bb4  Kd7  \move{39}f4  gxf4  \move{40}Rxf4  Kc8  \move{41}Be1  \specialmove{b3}

\begin{center}
\newgame
\fenboard{2k3r1/6r1/2pPp3/2Pb1p1p/p2P1R2/Pp4PP/7K/4B1R1 w - - 0 1}
\showboard
\end{center}

Black uses two consecutive torpedo moves (b5-b3, a4-a2) on the queenside to create a dangerous passed pawn on a2.

\move{42}\specialmove{axb4}  \specialmove{a2}  \move{43}Bc3  Ra7  \move{44}R4f1  Kd8  \move{45}Ra1  Kd7  \move{46}Bb2  Ra4  \move{47}Bc3  Ra3  \move{48}Rac1  Be4  \move{49}h4  Rg4  \move{50}Bd2  \specialmove{f3}

\begin{center}
\newgame
\fenboard{8/3k4/2pPp3/2P4p/1P1Pb1rP/r4pP1/p2B3K/2R3R1 w - - 0 1}
\showboard
\end{center}

Black uses another torpedo move (f5-f3) to advance further on the kingside and create another passed pawn.

\move{51}Rf1  Rg8  \move{52}Ra1  Rga8  \move{53}Rf2  Rb3  \move{54}Kh3  Rb1  \move{55}Bc3  Bd5  \move{56}g4  Rb3  \move{57}Be1  hxg4+
\move{58}Kg3  Rb1  \move{59}Bc3  Rb3  \move{60}Bd2  Rb1  \move{61}Bc3  Rb3  \move{62}Bd2  Rb2  \move{63}\specialmove{h6}

\begin{center}
\newgame
\fenboard{r7/3k4/2pPp2P/2Pb4/1P1P2p1/5pK1/pr1B1R2/R7 b - - 0 1}
\showboard
\end{center}

White advances the h-pawn with an h4-h6 torpedo move, seeking counterplay.

\blackmove{63}Rg8  \move{64}Raf1  Bc4  \move{65}h7  Rf8  \move{66}Rh1  Rh8  \move{67}Bc3  Rxf2  \move{68}Kxf2  \specialmove{g2}

\begin{center}
\newgame
\fenboard{7r/3k3P/2pPp3/2P5/1PbP4/2B2p2/p4Kp1/7R w - - 0 1}
\showboard
\end{center}

The torpedo move g4-g2 forces the White rook away from the h-file.

\move{69}Re1  Rxh7  \move{70}\specialmove{b6}  

\begin{center}
\newgame
\fenboard{8/3k3r/1PpPp3/2P5/2bP4/2B2p2/p4Kp1/4R3 b - - 0 1}
\showboard
\end{center}

White needs to generate immediate counterplay, and does so via b4-b6, another torpedo move.
White then uses a b6-b8=Q torpedo move to promote to a queen in the next move, demonstrating how fast the pawns are in this variation of chess.

\blackmove{70}Rh1  \move{71}\specialmove{b8=Q}  Rf1+

\begin{center}
\newgame
\fenboard{1Q6/3k4/2pPp3/2P5/2bP4/2B2p2/p4Kp1/4Rr2 w - - 0 1}
\showboard
\end{center}

\move{72}Rxf1  gxf1=Q+  \move{73}Kg3  and the game eventually ended in a draw due to mutual threats and ensuing checks.
\gamedrawn

\paragraph{Game AZ-17: AlphaZero Torpedo No-castling vs AlphaZero Torpedo No-castling}
This game was an experiment combining the No-castling chess with Torpedo chess, resulting in a highly tactical position. The first ten moves for White and Black were sampled randomly from AlphaZero's opening ``book'', with the probability proportional to the time spent calculating each move. The remaining moves follow best play, at roughly one minute per move.

\begin{center}
\newgame
\fenboard{r4k2/p3n2b/1q2Pp1P/3p2p1/rPpP1B2/5P2/3Q1KN1/6RR w - - 0 1}
\showboard
\end{center}

Here White executes a stunning 'double attack':

27. Qc2!! Kg8

Black can't afford to capture the Queen, due to the powerful attack following 27... Bxc2 28. h8=Q+. White also had to assess the consequences of 27... gxf4

28. Qxa4 Qxd4+ 29. Kxg3 gxf4+ 30. Kh2 Qf2 31. Rf1 Qg3+ 32. Kg1

\begin{center}
\newgame
\fenboard{r5k1/p3n2b/4Pp1P/3p4/QPp2p2/5Pq1/6N1/5RKR w - - 0 1}
\showboard
\end{center}

32... Qg6 33. Rh4 Kh8 34. Rg4 Qe8 35. Qa1 Qf8 36. Qc3 Bf5 37. Rxf4 a6 38. Re1 d3 39. Rxc4 Bxe6 40. Rd4 Bf5 41. Qc7 Ng6 42. Kf2 Qxh6

\begin{center}
\newgame
\fenboard{r6k/2Q5/p4pnq/5b2/1P1R4/3p1P2/5KN1/4R3 w - - 0 1}
\showboard
\end{center}

43. Qc1 Qxc1 44. Rxc1 Ne7 45. Ne3 Bg6 46. Ra1 Nc6 47. Rh4+ Kg7 48. b5 Nb8 49. Rc4 Bf7 50. Rc7 f4 51. Nd1 a4 52. Nc2 a2 53. Nxd3 Kf6 54. Rc8 Ra3 55. Nxf4 Nd7 56. Ne2 Ne5 57. b7 Rxf3+ 58. Kg2 Rb3 59. b8=Q Rxb8 60. Rxb8 and White went on to win the game easily. 1-0

%% file: variants/SAN/semi-torpedo.tex
\subsection{Semi-torpedo}
\label{sec-semitorpedo}

In Semi-torpedo chess, we consider a partial extension to the rules of pawn movement, where the pawns are allowed to move by two squares from the 2nd/3rd and 6th/7th rank for White and Black respectively. This is a restricted version of another variant we have considered (Torpedo chess) where the option is extended to cover the entire board. Yet, even this partial extension adds lots of dynamic options and here we independently evaluate its impact on the arising play.

\subsubsection{Motivation}

As with Torpedo chess, the motivation in extending the possibilities for rapid pawn movement lies in adding dynamic, attacking options to the middlegame. Yet, given that it is only a partial extension, adding an extra rank for each side from which the pawns can move by two squares, its impact on endgame patterns is much more limited.

\subsubsection{Assessment}

The assessment of the Semi-torpedo chess variant, as provided by Vladimir Kramnik:

\begin{shadequote}[]{}
\textit{Compared to Classical chess, the pawns that have been played to the 3rd/6th rank become much more useful, which manifests in several ways. First, prophylactic pawn moves to h3/h6 and a3/a6 now allow for a subsequent torpedo push. Having played h3 for example, it is now possible to play the pawn to h5 in a single move.
This also means, if the goal was to push the pawn to h5 in two moves, that there are two ways of achieving it -- either via h4 and h5 or via h3 and h5 -- and doing the latter does not expose a weakness on the g4 square and can thus be advantageous.
Secondly, fianchetto setups now allow for additional dynamic options. The g3 pawn can now be pushed to g5 in a single move, to attack a knight on f6 -- and vice versa. Thirdly, openings where one of the central pawns is on the 3rd/6th rank change -- consider the Meran for example -- the e3 pawn can now go to e5 in a single move.}

\textit{Theory might change in other openings as well, like for instance the Ruy Lopez with a7-a6, given that there would be some lines where the torpedo option of playing a6-a4 might force White to adopt a slightly different setup. AlphaZero also likes playing g6 early for Black, with a threat of g4 in some lines, aimed against a knight on f3 if White starts expanding in the center. As another example, consider a pretty standard opening sequence in the Sicilian defence:
\move{1}e4 c5 \move{2}Nf3 Nc6
\move{3}d4 cxd4 \move{4}Nxd4 Nf6
\move{5}Nc3 e5 \move{6}Ndb5 d6 -- it turns out that here \move{7}Bg5 no longer keeps the advantage, because of \blackmove{7}a6 \move{8}Na3 followed up by a torpedo move \blackmove{8}\specialmove{d4}:}

\begin{center}
\newgame
\fenboard{r1bqkb1r/1p3ppp/p1n2n2/4p1B1/3pP3/N1N5/PPP2PPP/R2QKB1R w KQkq - 0 1}
\showboard
\end{center}

\textit{Here, the game could continue \move{9}\specialmove{exd5} Bxa3 \move{10}bxa3 Nd4 \move{11}Bd3 Qa5, and the position is assessed as equal by AlphaZero. This variation illustrates nicely how the torpedo moves provide not only an additional attacking option for White, but also additional equalizing options for Black, depending on the position.}

\textit{Semi-torpedo chess seems to be more decisive than Classical chess, and less decisive than Torpedo chess. It is an interesting variation, to be potentially considered by those who like the general middlegame flavor of Torpedo chess, but are unwilling to abandon existing endgame theory.}
\end{shadequote}

\subsubsection{Main Lines}

Here we discuss ``main lines'' of AlphaZero under Semi-torpedo chess, when playing with roughly one minute per move from a particular fixed first move.
Note that these are not purely deterministic, and each of the given lines is merely one of several highly promising and likely options. Here we give the first 20 moves in each of the main lines.

\paragraph{Main line after e4}
The main line of AlphaZero after \move{1}e4 in Semi-torpedo chess is:

\move{1}e4 \bookmove c5  \move{2}c3  Nf6  \move{3}e5  Nd5  \move{4}Bc4  e6  \move{5}Nf3  Be7  \move{6}d4  d6  \move{7}O-O  O-O  \move{8}Re1  Nc6  \move{9}exd6  Qxd6  \move{10}dxc5  Qxc5  \move{11}Nbd2  b6  \move{12}b4  Qd6  \move{13}Qc2  Bb7  \move{14}a3  Nf6  \move{15}Ne4  Qc7  \move{16}Bd3  h6
\move{17}\specialmove{c5}  bxc5  \move{18}Nxc5  Bxc5  \move{19}bxc5  Na5  \move{20}Ne5  Rac8 

\begin{center}
\newgame
\fenboard{2r2rk1/pbq2pp1/4pn1p/n1P1N3/8/P2B4/2Q2PPP/R1B1R1K1 w Q - 0 1}
\showboard
\end{center}

and after \move{21}Bb2 White would have compensation for the pawn. There are also tactical resources in this position, for instance White could consider a more forcing line of play -- \move{21}Bxh6!? gxh6 \move{22}Qd2 Kg7 \move{23}Re3 Rh8 \move{24}Rg3+ Kf8 \move{25}Rae1 h4 \move{26}Rg7! Kxg7 \move{27}Qg5+ Kf8 \move{28}Qxf6 Rg8 \move{29}Ng6+ Rxg6 \move{30}Bxg6 -- potentially leading to a draw by perpetual check.

\paragraph{Main line after d4}
The main line of AlphaZero after \move{1}d4 in Semi-torpedo chess is:

\move{1}d4 \bookmove Nf6  \move{2}c4  e6  \move{3}e3  d5  \move{4}cxd5  exd5  \move{5}Nc3  Bd6  \move{6}Bd3  O-O  \move{7}Nge2  a6  \move{8}O-O  Re8  \move{9}b3  Nc6  \move{10}Ng3  Bg4  \move{11}f3  Bc8  \move{12}a3  Ne7  \move{13}Bb2  h6  \move{14}Qd2  c6
\move{15}\specialmove{e5}  \specialmove{dxe4}  \move{16}Ncxe4  Ned5  \move{17}Nxd6  Qxd6  \move{18}Rae1  Qd8  \move{19}Rxe8+  Nxe8  \move{20}Re1  Bd7

\begin{center}
\newgame
\fenboard{r2qn1k1/1p1b1pp1/p1p4p/3n4/3P4/PP1B1PN1/1B1Q2PP/4R1K1 w q - 0 1}
\showboard
\end{center}

\paragraph{Main line after c4}
The main line of AlphaZero after \move{1}c4 in Semi-torpedo chess is:

\move{1}c4 \bookmove c5  \move{2}g3  g6  \move{3}Bg2  Bg7  \move{4}e3  e6  \move{5}d4  cxd4  \move{6}exd4  Ne7  \move{7}Nc3  O-O  \move{8}Nge2  d5  \move{9}cxd5  Nxd5  \move{10}h4  Bd7  \move{11}Nxd5  exd5  \move{12}Be3  Re8  \move{13}Nc3  Nc6  \move{14}O-O  Be6  \move{15}h5  h6  \move{16}hxg6  fxg6  \move{17}Qd2  Kh7  \move{18}Ne2  Qf6  \move{19}Nf4  Bf7  \move{20}Nxd5  Bxd5

\begin{center}
\newgame
\fenboard{r3r3/pp4bk/2n2qpp/3b4/3P4/4B1P1/PP1Q1PB1/R4RK1 w - - 0 21}
\showboard
\end{center}

\subsubsection{Instructive games}

\paragraph{Game AZ-18: AlphaZero Semi-torpedo vs AlphaZero Semi-torpedo}
The first ten moves for White and Black have been sampled randomly from AlphaZero's opening ``book'', with the probability proportional to the time spent calculating each move. The remaining moves follow best play, at roughly one minute per move.
% Game 10 from on policy games

\move{1}d4  Nf6  \move{2}c4  e6  \move{3}e3  d5  \move{4}cxd5  exd5  \move{5}Nc3  Bd6  \move{6}Nb5  c6  \move{7}Nxd6+  Qxd6
\move{8}Bd3  Ne4  \move{9}f3  Qb4+  \move{10}Bd2  Nxd2  \move{11}Qxd2  Qd6  \move{12}Ne2  O-O  \move{13}O-O  Nd7  \move{14}g4  Nf6  \move{15}Kg2  Bd7  \move{16}Ng3  Kh8  \move{17}Rae1  Rae8  \move{18}Bb1  Ng8  \move{19}h3  Ne7  \move{20}\specialmove{f5}  

\begin{center}
\newgame
\fenboard{4rr1k/pp1bnppp/2pq4/3p1P2/3P2P1/4P1NP/PP1Q2K1/1B2RR2 w - - 0 1}
\showboard
\end{center}

Here we see the first torpedo move of the game, f3-f5, claiming space before Black has the chance to play f5.

\blackmove{20}f6  \move{21}a3  b6  \move{22}Nh5  Rb8  \move{23}Qf2  \specialmove{b4}  

\begin{center}
\newgame
\fenboard{1r3r1k/p2bn1pp/2pq1p2/3p1P1N/1p1P2P1/P3P2P/1P3QK1/1B2RR2 w - - 0 1}
\showboard
\end{center}

Black utilizes a torpedo move of its own, b6-b4, to initiate counterplay on the queenside.

\move{24}Qf4  Nc8  \move{25}a4  b3  \move{26}Qf3  Nb6  \move{27}Nf4  Rbe8  \move{28}Re2  \specialmove{c4}

And c6-c4 comes as another torpedo move, speeding up the queenside expansion. White chooses not to take \emph{en passant}, but to play a5 instead in reply.

\move{29}a5  Nc8

\begin{center}
\newgame
\fenboard{2n1rr1k/p2b2pp/3q1p2/P2p1P2/2pP1NP1/1p2PQ1P/1P2R1K1/1B3R2 w - - 0 1}
\showboard
\end{center}

\move{30}Rfe1  Ne7  \move{31}Rd1  Rc8  \move{32}\specialmove{e5}

\begin{center}
\newgame
\fenboard{2r2r1k/p2bn1pp/3q1p2/P2pPP2/2pP1NP1/1p3Q1P/1P2R1K1/1B1R4 w - - 0 1}
\showboard
\end{center}

White expands in the center with another torpedo move, e3-e5.

\blackmove{32}\specialmove{dxe4}  \move{33}Bxe4  Rfe8  \move{34}Ne6  Nc6  \move{35}Bxc6  Bxc6  \move{36}d5  Ba8  \move{37}Re3  Bb7  \move{38}\specialmove{h5}

\begin{center}
\newgame
\fenboard{2r1r2k/pb4pp/3qNp2/P2P1P1P/2p3P1/1p2RQ2/1P4K1/3R4 w - - 0 1}
\showboard
\end{center}

Here comes another torpedo advance, h3-h5, creating threats on the kingside.

\blackmove{38}h6  \move{39}Kh3  Qd7  \move{40}Rc3  Re7  \move{41}Qf1  Qb5  \move{42}d6  Rd7  \move{43}Qf4  Ba6

\begin{center}
\newgame
\fenboard{2r4k/p2r2p1/b2PNp1p/Pq3P1P/2p2QP1/1pR4K/1P6/3R4 w - - 0 1}
\showboard
\end{center}

\move{44}Qd2  Qe5  \move{45}Rg3  Rc6  \move{46}Nf8

\begin{center}
\newgame
\fenboard{5N1k/p2r2p1/b1rP1p1p/P3qP1P/2p3P1/1p4RK/1P1Q4/3R4 w - - 0 1}
\showboard
\end{center}

\blackmove{46}Rcxd6  \move{47}Qb4  Qb5  \move{48}Qxd6  Rxd6  \move{49}Rxd6  Qb8  \move{50}Ng6+  Kh7  \move{51}Rxa6  Qb7

\begin{center}
\newgame
\fenboard{8/pq4pk/R4pNp/P4P1P/2p3P1/1p4RK/1P6/8 w - - 0 1}
\showboard
\end{center}

\move{52}Re3  Qh1+  \move{53}Kg3  Qg1+  \move{54}Kf3  Qf1+  \move{55}Kg3  Qg1+  \move{56}Kf3  Qf1+  \move{57}Ke4  Qg2+  \move{58}Kf4  Qf2+  \move{59}Ke4  Qg2+  \move{60}Kf4  Qf2+  \move{61}Ke4  Qg2+
\gamedrawn

\paragraph{Game AZ-19: AlphaZero Semi-torpedo vs AlphaZero Semi-torpedo}
The position below, with Black to move, is taken from a game that was played with roughly one minute per move:

% Game 6 from on policy games

\begin{center}
\newgame
\fenboard{2rqr1k1/1b1n2b1/1p1p2pp/p2PpP1n/P7/2NBB1PP/1PQN1P2/3RK2R b K - 0 1}
\showboard
\end{center}

\blackmove{18}Bxd5  \move{19}Nde4  Bxe4  \move{20}Qb3+  Kh8  \move{21}Nxe4  \specialmove{d4}

\begin{center}
\newgame
\fenboard{2rqr2k/3n2b1/1p4pp/p3pP1n/P2pN3/1Q1BB1PP/1P3P2/3RK2R w K - 0 1}
\showboard
\end{center}

Here, a torpedo move (d6-d4) unleashes a tactical sequence.

\move{22}Nd6  Rf8  \move{23}Be2  Rc7  \move{24}fxg6  Nc5

\begin{center}
\newgame
\fenboard{3q1r1k/2r3b1/1p1N2Pp/p1n1p2n/P2p4/1Q2B1PP/1P2BP2/3RK2R w K - 0 1}
\showboard
\end{center}

\move{25}Qxb6  Nxa4  \move{26}Nf7+

\begin{center}
\newgame
\fenboard{3q1r1k/2r2Nb1/1Q4Pp/p3p2n/n2p4/4B1PP/1P2BP2/3RK2R b K - 0 1}
\showboard
\end{center}

\blackmove{26}R8xf7  \move{27}Qxa5  Rfd7  \move{28}Qxa4  dxe3

\begin{center}
\newgame
\fenboard{3q3k/2rr2b1/6Pp/4p2n/Q7/4p1PP/1P2BP2/3RK2R w K - 0 1}
\showboard
\end{center}

\move{29}Bxh5  Rxd1+  \move{30}Qxd1  exf2+  \move{31}Kxf2  Rd7  \move{32}Qc1  Rd2+  \move{33}Ke1  Rd3  \move{34}Kf2  Rd2+  \move{35}Ke1  Rd3  \move{36}Rg1  e4

\begin{center}
\newgame
\fenboard{3q3k/6b1/6Pp/7B/4p3/3r2PP/1P6/2Q1K1R1 w - - 0 1}
\showboard
\end{center}

\move{37}Rg2  Qa5+  \move{38}Kf1  Qf5+  \move{39}Qf4  Qxf4+  \move{40}gxf4  Rxh3  \move{41}Bd1  Rh4  \move{42}Kf2  Rxf4+  \move{43}Ke3  Rf1  \move{44}Bg4  Rf6  with a draw soon to follow.
\gamedrawn

\paragraph{Game AZ-20: AlphaZero Semi-torpedo vs AlphaZero Semi-torpedo}
The first ten moves for White and Black have been sampled randomly from AlphaZero's opening ``book'', with the probability proportional to the time spent calculating each move. The remaining moves follow best play, at roughly one minute per move.
% Game 9 from on policy games

\move{1}d4  Nf6  \move{2}Nf3  d5  \move{3}c4  e6  \move{4}a3  dxc4  \move{5}e3  c6  \move{6}Bxc4  b5  \move{7}Bd3  Bb7  \move{8}Nc3  a6  \move{9}\specialmove{e5}

\begin{center}
\newgame
\fenboard{rn1qkb1r/1b3ppp/p1p1pn2/1p2P3/3P4/P1NB1N2/1P3PPP/R1BQK2R b KQkq - 0 1}
\showboard
\end{center}

Here we see another typical central torpedo move (e3-e5), claiming space.

\blackmove{9}Nd5  \move{10}Be4  Be7  \move{11}h3  Nxc3  \move{12}bxc3  Nd7  \move{13}O-O  Rb8  \move{14}Qe2  \specialmove{c4}

\begin{center}
\newgame
\fenboard{1r1qk2r/1b1nbppp/p3p3/1p2P3/2pPB3/P1P2N1P/4QPP1/R1B2RK1 w Qk - 0 1}
\showboard
\end{center}

Black uses a torpedo move as a counter (c6-c4), expanding on the queenside.

\move{15}Bxb7  Rxb7  \move{16}Qe4  Rc7  \move{17}Qg4  g6  \move{18}\specialmove{a5}

\begin{center}
\newgame
\fenboard{3qk2r/2rnbp1p/p3p1p1/Pp2P3/2pP2Q1/2P2N1P/5PP1/R1B2RK1 w Qk - 0 1}
\showboard
\end{center}

Another torpedo move follows (a3-a5), giving rise to a thematic pawn structure.

\blackmove{18}h5  \move{19}Qg3  Nb8  \move{20}d5  Qxd5  \move{21}Bg5  Qd8  \move{22}Rad1  Rd7  \move{23}Bxe7  Qxe7  \move{24}Ng5  O-O  \move{25}Ne4  Rxd1  \move{26}Rxd1  Rd8  \move{27}Rd6  Rxd6  \move{28}exd6  Qd8  \move{29}Qe5  Nd7  \move{30}Qd4  Qh4 and the game eventually ended in a draw.
\gamedrawn

\paragraph{Game AZ-21: AlphaZero Semi-torpedo vs AlphaZero Semi-torpedo}
The position below, with White to move, is taken from a game that was played with roughly one minute per move:
% Game 18 from on policy games

\begin{center}
\newgame
\fenboard{r2q1rk1/1p1b1p2/p2bp2p/3pNnp1/3Pn3/2PBB2N/PP3PPP/R2QR1K1 w Qq - 0 1}
\showboard
\end{center}

\move{16}f3  Bxe5  \move{17}dxe5  Nxe3  \move{18}Rxe3  f5

\begin{center}
\newgame
\fenboard{r2q1rk1/1p1b4/p3p2p/3pPpp1/4n3/2PBRP1N/PP4PP/R2Q2K1 w Qq - 0 1}
\showboard
\end{center}

\move{19}exf6  Qb6  \move{20}Qc1  Nxf6  \move{21}Nf2  \specialmove{e4}

\begin{center}
\newgame
\fenboard{r4rk1/1p1b4/pq3n1p/3p2p1/4p3/2PBRP2/PP3NPP/R1Q3K1 w Qq - 0 1}
\showboard
\end{center}

Here we see a torpedo move e6-e4 being used in a tactical sequence in center of the board.

\move{22}fxe4  Rae8  \move{23}e5  Ng4  \move{24}Nxg4  Bxg4  \move{25}Kh1  Rf2  \move{26}b4  Ref8

\begin{center}
\newgame
\fenboard{5rk1/1p6/pq5p/3pP1p1/1P4b1/2PBR3/P4rPP/R1Q4K w Q - 0 1}
\showboard
\end{center}

\move{27}Qe1  Be6  \move{28}Re2  R2f4  \move{29}a3  Kg7  \move{30}h3  Qd8  \move{31}Re3  h5  \move{32}Rd1  g4  \move{33}Rd2  h4  \move{34}hxg4  Qg5

\begin{center}
\newgame
\fenboard{5r2/1p4k1/p3b3/3pP1q1/1P3rPp/P1PBR3/3R2P1/4Q2K w - - 0 1}
\showboard
\end{center}

\move{35}Rh3  Rxg4  \move{36}Qe3  d4  \move{37}cxd4  R8f4  \move{38}Rf3  Bd5

\begin{center}
\newgame
\fenboard{8/1p4k1/p7/3bP1q1/1P1P1rrp/P2BQR2/3R2P1/7K w - - 0 1}
\showboard
\end{center}

\move{39}Rxf4  Qxf4  \move{40}Qxf4  Rxf4  \move{41}Kg1  Rxd4  and the game soon ended in a draw.
\gamedrawn

%% file: variants/SAN/pawn-back.tex
\subsection{Pawn-back}
\label{sec-pawnback}

In the Pawn-back variation of chess, the pawns are allowed to move one square backwards, up to the 2nd/7th rank for White and Black respectively.
In addition, if the pawn moves back to its starting rank, it is allowed to move by two squares again on its next move. In this particular implementation, the two-square pawn move is always allowed from the 2nd or the 7th rank, regardless of whether the pawn has moved before. A different implementation of this variation of chess might consider disallowing this, though it is unlikely to make a big difference.
Because the pawns are allowed to move backwards and
pawn moves are now reversible in this implementation of chess, the 50 move rule is modified so that 50 moves without captures lead to a draw, regardless of whether any pawn moves were made in the meantime.

\subsubsection{Motivation}

In Classical chess, pawns that move forwards leave weaknesses behind.
Some of these remain long-term weaknesses, resulting in squares that can be easily occupied by the opponent's pieces. If the pawns could move backwards, they could come back to help fight for those squares and therefore reduce the number of weaknesses in a position. Allowing the pawns to move backwards would therefore make it easier to push them forward, as the effect would not be irreversible.
This might make advancing in a position easier, but equally, it could provide defensive options for the weaker side, such as retreating from a less favourable situation and covering a weaknesses in front of the king.

\subsubsection{Assessment}

The assessment of the Pawn-back chess variant, as provided by Vladimir Kramnik:

\begin{shadequote}[]{}
\textit{There are quite a few educational motifs in this variation of chess. The backward pawn moves can be used to open the diagonals for the bishops, or make squares available for the knights. The bishops can therefore become more powerful, as they are easier to activate. The pawns can be pushed in the center more aggressively than in classical chess, as they can always be pulled back. Exposing the king is not as big of an issue, as the pawns can always move back to protect. Weak squares are much less important for positional assessment in this variation, given that they can almost always be protected via moving the pawns back.}

\textit{It was interesting to see AlphaZero's strong preference for playing the French defence under these rules, the point being that the light-squared bishop is no longer bad, as it can be developed via c8-b7 followed by a timely d5-d6 back-move.}

\textit{Other openings change as well. After the standard \move{1}e4 e5 \move{2}Nf3 Nc6, there comes a surprise: \move{3}c4!}

\begin{center}
\newgame
\fenboard{r1bqkbnr/pppp1ppp/2n5/4p3/2P1P3/5N2/PP1P1PPP/RNBQKB1R b KQkq - 0 1}
\showboard
\end{center}

\textit{It is followed by \blackmove{3}Bc5 \move{4}\specialmove{e3} (a back-move!) Bb6 \move{5}d4 d6}

\begin{center}
\newgame
\fenboard{r1bqk1nr/ppp2ppp/1bnp4/4p3/2PP4/4PN2/PP3PPP/RNBQKB1R b KQkq - 0 1}
\showboard
\end{center}

\textit{Who would have guessed that we are on move 5, after the game having started with e4 e5?}

\textit{The Pawn-back version of chess allows for more fluid and flexible pawn structures and could potentially be interesting for players who like such strategic manoeuvring.
Given that Pawn-back chess offers additional defensive resources, winning with White seems to be slightly harder, so the variant might also appeal to players who enjoy defending and attackers looking for a challenge.}
\end{shadequote}

\subsubsection{Main Lines}

Here we discuss ``main lines'' of AlphaZero under Pawn-back chess, when playing with roughly one minute per move from a particular fixed first move. Note that these are not purely deterministic, and each of the given lines is merely one of several highly promising and likely options. Here we give the first 20 moves in each of the main lines, regardless of the position.

\paragraph{Main line after e4}
The main line of AlphaZero after \move{1}e4 in Pawn-back chess is:

\move{1}e4 \bookmove e6  \move{2}Nc3  d5  \move{3}d4  Nf6  \move{4}e5  Nfd7  \move{5}f4  c5  \move{6}Nf3  a6  \move{7}Be3  b5  \move{8}f5  Nc6  \move{9}fxe6  fxe6  \move{10}\specialmove{e4}  cxd4  \move{11}Nxd4  Nxd4  \move{12}Qxd4  b4  \move{13}Ne2  Nf6  \move{14}exd5  Qxd5  \move{15}Nf4  Qxd4  \move{16}Bxd4  Bd6  \move{17}Nd3  a5  \move{18}Be5  Ke7  \move{19}Bxd6+  Kxd6  \move{20}O-O-O  Ke7

\begin{center}
\newgame
\fenboard{r1b4r/4k1pp/4pn2/p7/1p6/3N4/PPP3PP/2KR1B1R w K - 0 1}
\showboard
\end{center}

\paragraph{Main line after d4}
The main line of AlphaZero after \move{1}d4 in Pawn-back chess is:

\move{1}d4 \bookmove d5  \move{2}e3  Nf6  \move{3}Nf3  e6  \move{4}c4  Be7  \move{5}b3  O-O  \move{6}cxd5  exd5  \move{7}Bd3  Re8  \move{8}Bb2  a5  \move{9}O-O  Bf8  \move{10}Nc3  c6  \move{11}Qc2  b6  \move{12}Ne2  Ra7  \move{13}Rac1  Rc7  \move{14}Rfe1  Bb4  \move{15}Nc3  Ba6  \move{16}Bxa6  Nxa6  \move{17}h4  \specialmove{b7}  \move{18}a3  Bf8  \move{19}Ne2  Rc8  \move{20}Nf4  Nc7

\begin{center}
\newgame
\fenboard{2rqrbk1/1pn2ppp/2p2n2/p2p4/3P1N1P/PP2PN2/1BQ2PP1/2R1R1K1 w - - 0 1}
\showboard
\end{center}

\paragraph{Main line after c4}
The main line of AlphaZero after \move{1}c4 in Pawn-back chess is:

\move{1}c4 \bookmove e5  \move{2}e3  c5  \move{3}Nc3  Nc6  \move{4}Nf3  f5  \move{5}d4  e4  \move{6}Nd2  Nf6  \move{7}d5  Ne5  \move{8}Be2  g6  
\move{9}\specialmove{d4}  Nf7  \move{10}dxc5  Bxc5  \move{11}a3  Bf8  \move{12}b4  Bg7  \move{13}Bb2  O-O  \move{14}O-O  d6  \move{15}a4  Be6  \move{16}Qb3  a5  \move{17}Rfd1  b6  \move{18}bxa5  bxa5  \move{19}Qa3  \specialmove{e5}  \move{20}c5  Qb8 

\begin{center}
\newgame
\fenboard{rq3rk1/5nbp/3pbnp1/p1P1pp2/P7/Q1N1P3/1B1NBPPP/R2R2K1 w Qq - 0 1}
\showboard
\end{center}

\subsubsection{Instructive games}

\paragraph{Game AZ-22: AlphaZero Pawn-back vs AlphaZero Pawn-back}
The first ten moves for White and Black have been sampled randomly from AlphaZero's opening ``book'', with the probability proportional to the time spent calculating each move. The remaining moves follow best play, at roughly one minute per move.

% Game 43 from on policy games

\move{1}e4  c6  \move{2}d4  d5  \move{3}e5  Bf5  \move{4}h4  h5  \move{5}c4  \specialmove{d6}

\begin{center}
\newgame
\fenboard{rn1qkbnr/pp2ppp1/2pp4/4Pb1p/2PP3P/8/PP3PP1/RNBQKBNR w KQkq - 0 1}
\showboard
\end{center}

Here we see d5-d6 as the first back-move of the game, challenging White's (over)extended center -- an option that would not have been available in classical chess.

\move{6}exd6  exd6  \move{7}d5  Be7  \move{8}Nc3  Bxh4  \move{9}Be3  Qe7  \move{10}g3  Bf6  \move{11}Rxh5  Rxh5  \move{12}Qxh5  Bg6  \move{13}Qe2  Bxc3+  \move{14}bxc3  Nd7  \move{15}f3  O-O-O  \move{16}Kf2  Ngf6

\begin{center}
\newgame
\fenboard{2kr4/pp1nqpp1/2pp1nb1/3P4/2P5/2P1BPP1/P3QK2/R4BN1 w - - 0 1}
\showboard
\end{center}

Black is putting pressure on d5, so White uses the back-move d5-d4 option to reconfigure the central pawn structure, rather than release the tension.

\move{17}\specialmove{d4}  d5  \move{18}c5

\begin{center}
\newgame
\fenboard{2kr4/pp1nqpp1/2p2nb1/2Pp4/3P4/2P1BPP1/P3QK2/R4BN1 w - - 0 1}
\showboard
\end{center}

Black and White repeat back-moves a couple of times.
Each time that Black challenges the c5 pawn via a d5-d6 back-move, White responds by c5-c4, refusing to exchange on that square.

\blackmove{18}\specialmove{d6}  \move{19}\specialmove{c4}  d5  \move{20}Rc1  Rh8  \move{21}c5  \specialmove{d6}  \move{22}\specialmove{c4}  d5  \move{23}Bf4  Qxe2+  \move{24}Bxe2  dxc4  \move{25}Bxc4  b5  \move{26}Bf1  Nd5  \move{27}Bd2  N7b6

\begin{center}
\newgame
\fenboard{2k4r/p4pp1/1np3b1/1p1n4/3P4/2P2PP1/P2B1K2/2R2BN1 w k - 0 1}
\showboard
\end{center}

Here we see an example of how back-moves can help cover weak squares. Black is threatening to invade on the light squares on the queenside at an opportune moment, but White utilizes a back-move d4-d3 and protects c4. This, however, enables Black to go forward and Black takes the opportunity to play c6-c5.

\move{28}\specialmove{d3}  c5  \move{29}f4  c4

\begin{center}
\newgame
\fenboard{2k4r/p4pp1/1n4b1/1p1n4/2p2P2/2PP2P1/P2B1K2/2R2BN1 w k - 0 1}
\showboard
\end{center}

White decides to keep retreating here and not give up the light squares with a back-move c3-c2.

\move{30}\specialmove{c2}  Rh2+

\begin{center}
\newgame
\fenboard{2k5/p4pp1/1n4b1/1p1n4/2p2P2/3P2P1/P1PB1K1r/2R2BN1 w - - 0 1}
\showboard
\end{center}

At this point it should come as no surprise how White should respond to the rook invasion using a back-move g3-g2!

\move{31}\specialmove{g2}  c3  \move{32}Be1  Bf5  \move{33}Nf3  Rh6  \move{34}g3  Rd6  \move{35}Bg2  a6  \move{36}a3  Na4  \move{37}Ng5  f6  \move{38}Ne4  Rd7  \move{39}Kf3  Kc7  \move{40}Bf2

\begin{center}
\newgame
\fenboard{8/2kr2p1/p4p2/1p1n1b2/n3NP2/P1pP1KP1/2P2BB1/2R5 b - - 0 1}
\showboard
\end{center}

\blackmove{40}Bxe4  \move{41}Kxe4  Kd6  \move{42}Re1  Rc7  \move{43}Kf5  Ne7+  \move{44}Kg4  \specialmove{c4}  \move{45}\specialmove{d2}

\begin{center}
\newgame
\fenboard{8/2r1n1p1/p2k1p2/1p6/n1p2PK1/P5P1/2PP1BB1/4R3 b - - 0 1}
\showboard
\end{center}

Here we see both Black and White having retreated from the interaction on the queenside, Black via a back-move c3-c4 and White by playing the d-pawn back to d2. The game soon ended in a draw.

\blackmove{45}f5+  \move{46}Kf3  c3  \move{47}d3  c4  \move{48}d2  c3  \move{49}d3  \specialmove{c4}  \move{50}g4  cxd3  \move{51}cxd3  Rc3  \move{52}gxf5  Rxd3+  \move{53}Kg4  Rd2  \move{54}Re6+  Kd7  \move{55}Kf3  Rd3+  \move{56}Kg4  Rd2  \move{57}Kf3  Rd3+  \move{58}Kg4  Rd2
\gamedrawn

\paragraph{Game AZ-23: AlphaZero Pawn-back vs AlphaZero Pawn-back}
The position below, with Black to move, is taken from a game that was played with roughly one minute per move:

% Game 8, on policy games

\begin{center}
\newgame
\fenboard{r1bqkb1r/1p3pp1/2n2n1p/pN1p1P2/P3pB2/1N4P1/1PP1P1BP/R2Q1RK1 b Qkq - 0 1}
\showboard
\end{center}

White is targeting c7 with the bishop and the knight, but here Black plays a back-move, e4-e5.
It initiates a long forced tactical sequence, showcasing that things can indeed get quite tactical in this variation of chess, depending on the line of play.

\blackmove{13}\specialmove{e5}  \move{14}e4

\begin{center}
\newgame
\fenboard{r1bqkb1r/1p3pp1/2n2n1p/pN1ppP2/P3PB2/1N4P1/1PP3BP/R2Q1RK1 b Qkq - 0 1}
\showboard
\end{center}

AlphaZero decides to sacrifice a piece for the initiative!

\blackmove{14}exf4  \move{15}exd5  Qb6+  \move{16}Kh1  Na7  \move{17}Qe1+

\begin{center}
\newgame
\fenboard{r1b1kb1r/np3pp1/1q3n1p/pN1P1P2/P4p2/1N4P1/1PP3BP/R3QR1K b Qkq - 0 1}
\showboard
\end{center}

\blackmove{17}Kd8  \move{18}N5d4  Bd7  \move{19}Nxa5  fxg3  \move{20}Rd1  Bb4  \move{21}Nxb7+

\begin{center}
\newgame
\fenboard{r2k3r/nN1b1pp1/1q3n1p/3P1P2/Pb1N4/6p1/1PP3BP/3RQR1K b kq - 0 1}
\showboard
\end{center}

Sacrificing another piece!

\blackmove{21}Qxb7  \move{22}Qxg3  Rg8  \move{23}Ne6+

\begin{center}
\newgame
\fenboard{r2k2r1/nq1b1pp1/4Nn1p/3P1P2/Pb6/6Q1/1PP3BP/3R1R1K b q - 0 1}
\showboard
\end{center}

Third consecutive piece sacrifice by White!

\blackmove{23}fxe6  \move{24}dxe6  Qc7  \move{25}Bxa8  Qxg3  \move{26}hxg3  Kc7  \move{27}Bg2 Bc6

\begin{center}
\newgame
\fenboard{6r1/n1k3p1/2b1Pn1p/5P2/Pb6/6P1/1PP3B1/3R1R1K w - - 0 1}
\showboard
\end{center}

It's time to take stock -- White has a rook and 4 pawns for 3 pieces, a very unusual material imbalance.

\move{28}Rf4  Rb8  \move{29}Rc4  Bd6  \move{30}Rb1  Kb6  \move{31}Re1  Nh5  \move{32}g4  Nf6  \move{33}Re3  Rc8  \move{34}Rec3  Be5  \move{35}a5+  Ka6  \move{36}Rxc6+  Rxc6  \move{37}Rxc6+  Nxc6  \move{38}Bxc6  Bxb2  \move{39}e7  Kxa5  

\begin{center}
\newgame
\fenboard{8/4P1p1/2B2n1p/k4P2/6P1/8/1bP5/7K w - - 0 1}
\showboard
\end{center}

And the game soon ended in a draw.
\gamedrawn

\paragraph{Game AZ-24: AlphaZero Pawn-back vs AlphaZero Pawn-back}
The first ten moves for White and Black have been sampled randomly from AlphaZero's opening ``book'', with the probability proportional to the time spent calculating each move. The remaining moves follow best play, at roughly one minute per move.

% Game 47 from on policy games

\move{1}e4  e6  \move{2}Nc3  d5  \move{3}d4  Nf6  \move{4}e5  Nfd7  \move{5}f4  c5  \move{6}Nf3  a6  \move{7}a3  Nc6  \move{8}Be3  b5  \move{9}Ne2  Bb7  \move{10}c3

\begin{center}
\newgame
\fenboard{r2qkb1r/1b1n1ppp/p1n1p3/1pppP3/3P1P2/P1P1BN2/1P2N1PP/R2QKB1R b KQkq - 0 1}
\showboard
\end{center}

This looks like a pretty normal French position, but here comes Black's main equalizing resource, a back move d5-d6! Maybe that's all that was needed to make the French an undeniably good opening for Black?

\blackmove{10}\specialmove{d6}

\begin{center}
\newgame
\fenboard{r2qkb1r/1b1n1ppp/p1npp3/1pp1P3/3P1P2/P1P1BN2/1P2N1PP/R2QKB1R w KQkq - 0 1}
\showboard
\end{center}

This completely changes the nature of the position, as the center is suddenly not static and Black's light-squared bishop can find good use on the a8-h1 diagonal.

\move{11}Ng3  dxe5  \move{12}fxe5  Qb6  \move{13}Bf2  Rd8  \move{14}Qb1  cxd4  \move{15}cxd4  b4  \move{16}Be2  bxa3  \move{17}bxa3  Qa5+  \move{18}Kf1  Rb8  \move{19}h4  Be7  \move{20}Kg1  O-O  \move{21}Qd3  Rfc8  \move{22}Kh2  Qd8

\begin{center}
\newgame
\fenboard{1rrq2k1/1b1nbppp/p1n1p3/4P3/3P3P/P2Q1NN1/4BBPK/R6R w - - 0 1}
\showboard
\end{center}

Here AlphaZero prefers a solid back-move h4-h3 to a further expansion with h5.

\move{23}\specialmove{h3}  Na5  \move{24}Rhc1  Nf8  \move{25}Qe3

\begin{center}
\newgame
\fenboard{1rrq1nk1/1b2bppp/p3p3/n3P3/3P4/P3QNNP/4BBPK/R1R5 w - - 0 1}
\showboard
\end{center}

The a6 pawn is under pressure from the e2 bishop, and simply moves back to a7. The game soon fizzles out to a draw.

\blackmove{25}\specialmove{a7}  \move{26}a4  Ng6  \move{27}Rab1  Rxc1  \move{28}Qxc1  Rc8  \move{29}Qd1  Ba8  \move{30}Ba6  Rb8  \move{31}Bf1  Rxb1  \move{32}Qxb1  Bc6  \move{33}Bb5  Qb8  \move{34}Qd3  Qb7  \move{35}Ne2  h6  \move{36}Bg3  Be4  \move{37}Qe3  Bb4  \move{38}Bf2  Bd5  \move{39}Qd3  Be4  \move{40}Qe3  Bc6  \move{41}Qd3  Be7  \move{42}Bg3  Be4  \move{43}Qe3  Bb4  \move{44}Bf2  \gamedrawn

\paragraph{Game AZ-25: AlphaZero Pawn-back vs AlphaZero Pawn-back}
The first ten moves for White and Black have been sampled randomly from AlphaZero's opening ``book'', with the probability proportional to the time spent calculating each move. The remaining moves follow best play, at roughly one minute per move.

% Game 20 from on policy games

\move{1}e4  e6  \move{2}Nc3  d5  \move{3}d4  Nf6  \move{4}e5  Nfd7  \move{5}Be3  c5  \move{6}f4  a6  \move{7}Nf3  b5  \move{8}f5  Nc6  \move{9}fxe6  fxe6  \move{10}Bd3  g6  \move{11}O-O  cxd4  \move{12}Nxd4  Ndxe5  \move{13}Kh1  Ne7  \move{14}Rf6  Bg7

\begin{center}
\newgame
\fenboard{r1bqk2r/4n1bp/p3pRp1/1p1pn3/3N4/2NBB3/PPP3PP/R2Q3K w Qkq - 0 1}
\showboard
\end{center}

\move{15}Nxe6  Bxe6  \move{16}Rxe6  O-O  \move{17}Bg5  Ra7  \move{18}Be2  Nf7  \move{19}Bh4  g5

\begin{center}
\newgame
\fenboard{3q1rk1/r3nnbp/p3R3/1p1p2p1/7B/2N5/PPP1B1PP/R2Q3K w Q - 0 1}
\showboard
\end{center}

Here we see that moves like g5, that would potentially otherwise be quite weakening, are perfectly playable, given that the g-pawn can (and soon will) move back to g6, and in the meantime the threatening bishop is forced to move back and unpin the Black knight on e7.

\move{20}Bf2  d4  \move{21}Ne4  Nf5  \move{22}Qd3  \specialmove{g6}

\begin{center}
\newgame
\fenboard{3q1rk1/r4nbp/p3R1p1/1p3n2/3pN3/3Q4/PPP1BBPP/R6K w Q - 0 1}
\showboard
\end{center}

After moving the pawn back to g6 with a back-move, Black safeguards the kingside, justifying the previous g5 pawn push, which was helpful in achieving development.

\move{23}g4  Ne3  \move{24}Bxe3  dxe3  \move{25}Qxe3  Re7  \move{26}Rxe7  Qxe7  \move{27}a4  Nd6  \move{28}Bd3  Bxb2  \move{29}Rb1  Qe5  \move{30}axb5  axb5  \move{31}Qe2  Ba3

\begin{center}
\newgame
\fenboard{5rk1/7p/3n2p1/1p2q3/4N1P1/b2B4/2P1Q2P/1R5K w - - 0 1}
\showboard
\end{center}

As a mirror-motif to Black's g5-g6, here White plays g4-g3 to improve the safety of its king.

\move{32}\specialmove{g3}  Nxe4  \move{33}Qxe4  Qxe4  \move{34}Bxe4  b4  and the game soon ended in a draw.
\gamedrawn

\paragraph{Game AZ-26: AlphaZero Pawn-back vs AlphaZero Pawn-back}
The first ten moves for White and Black have been sampled randomly from AlphaZero's opening ``book'', with the probability proportional to the time spent calculating each move. The remaining moves follow best play, at roughly one minute per move.

% Game 11 from on policy games

\move{1}d4  Nf6  \move{2}c4  e6  \move{3}Nf3  a6  \move{4}Nc3  d5  \move{5}cxd5  exd5  \move{6}b3  Bb4  \move{7}Bd2  Be7  \move{8}e3  O-O  \move{9}Bc1  Bf5  \move{10}Bd3  Bxd3  \move{11}Qxd3  c6  \move{12}Qc2  Re8  \move{13}O-O  a5  \move{14}h4  Na6  \move{15}Ne2  Nb4  \move{16}Qd1  Bd6  \move{17}Bb2  h6

\begin{center}
\newgame
\fenboard{r2qr1k1/1p3pp1/2pb1n1p/p2p4/1n1P3P/1P2PN2/PB2NPP1/R2Q1RK1 w Qq - 0 1}
\showboard
\end{center}

Here we see the first back-move of the game, opening the diagonal for the White bishop -- d4-d3!

\move{18}\specialmove{d3}  Nd7  \move{19}a4  c5

\begin{center}
\newgame
\fenboard{r2qr1k1/1p1n1pp1/3b3p/p1pp4/Pn5P/1P1PPN2/1B2NPP1/R2Q1RK1 w Qq - 0 1}
\showboard
\end{center}

Just having played a4 on the previous move, White plays a back-move a4-a3 to challenge the b4 knight, given that the circumstances have changed due to Black having played c5.

\move{20}\specialmove{a3}  Na6  \move{21}d4

\begin{center}
\newgame
\fenboard{r2qr1k1/1p1n1pp1/n2b3p/p1pp4/3P3P/PP2PN2/1B2NPP1/R2Q1RK1 w Qq - 0 1}
\showboard
\end{center}

White goes back to the previous plan and plays the pawn to d4 again, despite having moved it back before, showcasing the fluidity of pawn structures
Black responds by moving the c-pawn back, to avoid having an isolated pawn.

\blackmove{21}\specialmove{c6}  \move{22}g3  Nc7  \move{23}a4  Ne6  \move{24}Kg2  Nf6  \move{25}Rc1  Bf8  \move{26}\specialmove{d3} 

\begin{center}
\newgame
\fenboard{r2qrbk1/1p3pp1/2p1nn1p/p2p4/P6P/1P1PPNP1/1B2NPK1/2RQ1R2 b q - 0 1}
\showboard
\end{center}

White opens the Bishop's diagonal with a back-move, again.

\blackmove{26}Rc8  \move{27}Nfd4  Nxd4  \move{28}Bxd4  c5  \move{29}Ba1  Nh5  \move{30}Ng1  g6  \move{31}Nf3  Ng7  \move{32}Bxg7  Bxg7  \move{33}d4  \specialmove{c6}  \move{34}Qd2  Bf8  \move{35}h5  g5

\begin{center}
\newgame
\fenboard{2rqrbk1/1p3p2/2p4p/p2p2pP/P2P4/1P2PNP1/3Q1PK1/2R2R2 w - - 0 1}
\showboard
\end{center}

Having just played h5, White plays h5-h4 now, to attack Black's g-pawn again. They repeat once before continuing with other plans.

\move{36}\specialmove{h4}  \specialmove{g6}  \move{37}h5  g5  \move{38}Qd3  Rc7  \move{39}Qf5  Qc8  \move{40}Qxc8  Rcxc8  \move{41}\specialmove{h4}  f6  \move{42}g4  Re4

\begin{center}
\newgame
\fenboard{2r2bk1/1p6/2p2p1p/p2p2p1/P2Pr1PP/1P2PN2/5PK1/2R2R2 w - - 0 1}
\showboard
\end{center}

Black is attacking White's pawn on g4, so it just moves back to g3.

\move{43}\specialmove{g3}  Kf7  \move{44}Rh1  g4  \move{45}Ne1  Bd6  \move{46}\specialmove{h3}  \specialmove{g5}

\begin{center}
\newgame
\fenboard{2r5/1p3k2/2pb1p1p/p2p2p1/P2Pr3/1P2P1PP/5PK1/2R1N2R w - - 0 1}
\showboard
\end{center}

After having been challenged by a h4-h3 back-move, Black retreats with g4-g5 as well.

\move{47}Nd3  Ke8  \move{48}h4  g4  \move{49}\specialmove{h3}  \specialmove{g5}  \move{50}h4  \specialmove{g6}  \move{51}Kf3  Kd7  \move{52}Nf4  Rg8  \move{53}Ne2  h5  \move{54}Nf4  Bxf4  \move{55}gxf4  b6  \move{56}\specialmove{a3}  \specialmove{f7}  \move{57}Rc2  Ra8  \move{58}Rb1  Re6  \move{59}Ke2  Rf6  \move{60}2f3  Rf5  \move{61}Kf2  \specialmove{d6}  \move{62}a4  Re8  \move{63}Rbc1  \specialmove{b7}  

\begin{center}
\newgame
\fenboard{4r3/1p1k1p2/2pp2p1/p4r1p/P2P1P1P/1P2PP2/2R2K2/2R5 w - - 0 1}
\showboard
\end{center}

White takes aim at the c6 pawn, but Black simply plays b6-b7, guarding it. With no clear way forward in this position, and after many more pawn structure reconfigurations, the game unsurprisingly ended in a draw. 
\gamedrawn

%% file: variants/SAN/pawn-sideways.tex
\subsection{Pawn-sideways}
\label{sec-pawnside}

In the Pawn-sideways version of chess, pawns are allowed an additional option of moving sideways by one square, when available.

\subsubsection{Motivation}

Allowing the pawns to move laterally introduces lots of new tactics into chess, while keeping the pawn structures very flexible and fluid.
It makes pawns much more powerful than before and drastically increases the complexity of the game, as there are many more moves to consider at each juncture -- and no static weaknesses to exploit.

\subsubsection{Assessment}

The assessment of the Pawn-sideways chess variant, as provided by Vladimir Kramnik:

\begin{shadequote}[]{}
\textit{This is the most perplexing and ``alien'' of all variants of chess that we have considered. Even after having looked at how AlphaZero plays Pawn-side chess, the principles of play remain somewhat mysterious -- it is not entirely clear what each side should aim for. The patterns are very different and this makes many moves visually appear very strange, as they would be mistakes in Classical chess.}

\textit{Lateral pawn moves change all stages of the game. Endgame theory changes entirely, given that the pawns can now ``run away'' laterally to the edge of the board, and it is hard to block them and pin them down. Consider, for instance, the following position, with White to move:}

\begin{center}
\newgame
\fenboard{8/1P6/8/8/3k1K2/1r6/8/8 w - - 0 1}
\showboard
\end{center}

\textit{In classical chess, White would be completely lost. Here, White can play b7-a7 or b7-c7, changing files. The rook can follow, but the pawn can always step aside. In this particular position, after b7-c7, Rc3, c7-d7 -- Black has no way of stopping the pawn from queening, and instead of losing -- White actually wins!}

\textit{It almost appears as if being a pawn up might give better chances of winning than being up a piece for a pawn. In fact, AlphaZero often chooses to play with two pawns against a piece, or a minor piece and a pawn against a rook, suggesting that pawns are indeed more valuable here than in classical chess.}

\textit{This variant of chess is quite different and at times hard to understand, but could be interesting for players who are open to experimenting with few attachments to the original game!}
\end{shadequote}

\subsubsection{Main Lines}

Here we discuss ``main lines'' of AlphaZero under Pawn-sideways chess, when playing with roughly one minute per move from a particular fixed first move. Note that these are not purely deterministic, and each of the given lines is merely one of several highly promising and likely options. Here we give the first 20 moves in each of the main lines, regardless of the position.

\paragraph{Main line after e4}
The main line of AlphaZero after \move{1}e4 in Pawn-sideways chess is:

\move{1}e4 \bookmove c5  \move{2}c3  b6  \move{3}dd4  Bb7  \move{4}Nd2  g6  \move{5}Bd3  Bg7  \move{6}Ngf3  a5

\begin{center}
\newgame
\fenboard{rn1qk1nr/1b1pppbp/1p4p1/p1p5/3PP3/2PB1N2/PP1N1PPP/R1BQK2R w KQkq - 0 1}
\showboard
\end{center}

The previous move (a5) seems very unusual to a Classical chess player's eye. Black chooses to disregard the centre, while creating a glaring weakness on b5.
Yet, there is method to this ``madness''.
It seems that rushing to grab space early is not good in this setup, so White's most promising plan according to AlphaZero is to prepare b4. Apart from fighting against that advance, a5 prepares for playing a5-b5!~later in this line, as we will see. Yet, this whole line of play is hard to grasp as it violates the Classical chess principles.

\move{7}O-O  d6  \move{8}Rb1  Nf6  \move{9}a3  O-O  \move{10}b4

\begin{center}
\newgame
\fenboard{rn1q1rk1/1b2ppbp/1p1p1np1/p1p5/1P1PP3/P1PB1N2/3N1PPP/1RBQ1RK1 b q - 0 1}
\showboard
\end{center}

White has achieved the desired advance, to which Black responds with a lateral move -- c5-d5!

\blackmove{10}\specialmove{cd5}

\begin{center}
\newgame
\fenboard{rn1q1rk1/1b2ppbp/1p1p1np1/p2p4/1P1PP3/P1PB1N2/3N1PPP/1RBQ1RK1 w q - 0 1}
\showboard
\end{center}

\move{11}Qc2  Nxe4  \move{12}Nxe4  dxe4  \move{13}Bxe4  Bxe4  \move{14}Qxe4  Nd7  \move{15}Be3  \specialmove{ab5}

\begin{center}
\newgame
\fenboard{r2q1rk1/3nppbp/1p1p2p1/1p6/1P1PQ3/P1P1BN2/5PPP/1R3RK1 w q - 0 1}
\showboard
\end{center}

As mentioned earlier, the a5 pawn finds a new purpose -- on b5! The b6 pawn will soon move to c6, in the process of reconfiguring the pawn structure. 

\move{16}\specialmove{ab3}  Nf6  \move{17}Qd3  \specialmove{bc6}  \move{18}\specialmove{cc4}  Qb8  \move{19}\specialmove{a4}  b4  \move{20}\specialmove{c3} Rxa4

\begin{center}
\newgame
\fenboard{1q3rk1/4ppbp/2pp1np1/8/rpPP4/2PQBN2/5PPP/1R3RK1 w - - 0 1}
\showboard
\end{center}

\paragraph{Main line after d4}
The main line of AlphaZero after \move{1}d4 in Pawn-sideways chess is:

\move{1}d4 \bookmove d5  \move{2}e3  e6  \move{3}cc4  dxc4  \move{4}Bxc4  a6  \move{5}a4  c5  \move{6}Nf3  Nc6  \move{7}Be2  cxd4  \move{8}exd4  g6  \move{9}b3  Nge7  \move{10}Bb2  Bg7  \move{11}Na3

\begin{center}
\newgame
\fenboard{r1bqk2r/1p2npbp/p1n1p1p1/8/P2P4/NP3N2/1B2BPPP/R2QK2R b KQkq - 0 1}
\showboard
\end{center}

Here Black has a way of opening the light-squared bishop while safeguarding the e5 square, by playing:

\blackmove{11}\specialmove{d6}  \move{12}O-O  O-O  \move{13}\specialmove{c3}  d5  \move{14}Re1  Qc7  \move{15}Bf1  Be6  \move{16}h3 
\begin{center}
\newgame
\fenboard{r4rk1/1pq1npbp/p1n1b1p1/3p4/P2P4/N1P2N1P/1B3PP1/R2QRBK1 b Qq - 0 1}
\showboard
\end{center}

In this position, Black utilizes a rather unique defensive resource:

\blackmove{16}\specialmove{gf6}

\begin{center}
\newgame
\fenboard{r4rk1/1pq1npbp/p1n1bp2/3p4/P2P4/N1P2N1P/1B3PP1/R2QRBK1 w Qq - 0 1}
\showboard
\end{center}

\move{17}Nc2  Rfd8  \move{18}Qb1  Rab8  \move{19}\specialmove{hg3}  b5  \move{20}\specialmove{b4}  Ra8

\begin{center}
\newgame
\fenboard{r2r2k1/2q1npbp/p1n1bp2/1p1p4/1P1P4/2P2NP1/1BN2PP1/RQ2RBK1 w Qq - 0 1}
\showboard
\end{center}

\paragraph{Main line after c4}
The main line of AlphaZero after \move{1}c4 in Pawn-sideways chess is:

\move{1}c4 \bookmove c5  \move{2}Nc3  Nc6  \move{3}g3  g6  \move{4}Bg2  e6  \move{5}e4  a6  \move{6}a3  Rb8  \move{7}Nge2  Bg7  \move{8}Rb1  dd6  \move{9}bb4

\begin{center}
\newgame
\fenboard{1rbqk1nr/1p3pbp/p1npp1p1/2p5/1PP1P3/P1N3P1/3PNPBP/1RBQK2R b Kk - 0 1}
\showboard
\end{center}

Here, Black responds with a typical lateral move.

\blackmove{9}\specialmove{c7}

\begin{center}
\newgame
\fenboard{1rbqk1nr/2p2pbp/p1npp1p1/2p5/1PP1P3/P1N3P1/3PNPBP/1RBQK2R w Kk - 0 1}
\showboard
\end{center}

\move{10}O-O  Nge7  \move{11}bxc5  Rxb1  \move{12}cxd6

\begin{center}
\newgame
\fenboard{2bqk2r/2p1npbp/p1nPp1p1/8/2P1P3/P1N3P1/3PNPBP/1rBQ1RK1 b k - 0 1}
\showboard
\end{center}

White fights for the advantage by going for this kind of a material imbalance, an exchange down.

\blackmove{12}Rb8  \move{13}dxe7  Qxe7  \move{14}dd4  O-O  \move{15}h4  Rd8  \move{16}d5  Qc5

\begin{center}
\newgame
\fenboard{1rbr2k1/2p2pbp/p1n1p1p1/2qP4/2P1P2P/P1N3P1/4NPB1/2BQ1RK1 w - - 0 1}
\showboard
\end{center}

Here another lateral move proves useful:

\move{17}\specialmove{b4}  Qc4  \move{18}Bf4  e5  \move{19}Bg5  \specialmove{gf6}  \move{20}Be3  \specialmove{e6}

\begin{center}
\newgame
\fenboard{1rbr2k1/2p2pbp/p1n1p3/3Pp3/1Pq1P2P/P1N1B1P1/4NPB1/3Q1RK1 w - - 0 1}
\showboard
\end{center}

Black moves the g6 pawn first to f6 and then to e6, reaching this position. The continuation shown here is not forced, and in some of its games, AlphaZero opts for slightly different lines with Black, as this seems to be a very rich opening.

\subsubsection{Instructive games}

\paragraph{Game AZ-27: AlphaZero Pawn-sideways vs AlphaZero Pawn-sideways}
The game is played from a fixed opening position that arises after: \move{1}e4 e5 \move{2}Nf3 Nc6 \move{3}Bc4. The remaining moves follow best play, at roughly one minute per move.

% Game 45 from the div directory.

\move{1}e4 \bookmove e5 \bookmove
\move{2}Nf3 \bookmove Nc6 \bookmove
\move{3}Bc4 \bookmove d6  \move{4}O-O  Be6  \move{5}Bb3  g5  \move{6}dd4

\begin{center}
\newgame
\fenboard{r2qkbnr/ppp2p1p/2npb3/4p1p1/3PP3/1B3N2/PPP2PPP/RNBQ1RK1 b Qkq - 0 1}
\showboard
\end{center}

\blackmove{6}Bxb3  \move{7}axb3  g4  \move{8}Nxe5  

\begin{center}
\newgame
\fenboard{r2qkbnr/ppp2p1p/2np4/4N3/3PP1p1/1P6/1PP2PPP/RNBQ1RK1 b Qkq - 0 1}
\showboard
\end{center}

Already, things are getting very tactical and very unorthodox.

\blackmove{8}dxe5  \move{9}d5

\begin{center}
\newgame
\fenboard{r2qkbnr/ppp2p1p/2n5/3Pp3/4P1p1/1P6/1PP2PPP/RNBQ1RK1 b Qkq - 0 1}
\showboard
\end{center}

Black leaves the knight on c6 and goes on with creating counter-threats.

\blackmove{9}\specialmove{hg7}  \move{10}Qxg4  Nf6  \move{11}Qf3  Ne7  \move{12}Re1  Ng6  \move{13}\specialmove{d4}

\begin{center}
\newgame
\fenboard{r2qkb1r/ppp2pp1/5nn1/3Pp3/3P4/1P3Q2/1PP2PPP/RNB1R1K1 b Qkq - 0 1}
\showboard
\end{center}

White uses a lateral move (e4-d4) to create threats on the e-file.

\blackmove{13}e4  \move{14}cc4  Bd6  \move{15}g3  Kf8  \move{16}Qg2  Ng4

\begin{center}
\newgame
\fenboard{r2q1k1r/ppp2pp1/3b2n1/3P4/2PPp1n1/1P4P1/1P3PQP/RNB1R1K1 w Qkq - 0 1}
\showboard
\end{center}

Black goes for the attack.

\move{17}Rxe4  Nxh2  \move{18}Nd2  f5  \move{19}Re1  Bf4

\begin{center}
\newgame
\fenboard{r2q1k1r/ppp3p1/6n1/3P1p2/2PP1b2/1P4P1/1P1N1PQn/R1B1R1K1 w Qkq - 0 1}
\showboard
\end{center}

Offering a piece on f4.

\move{20}gxf4  Nxf4  \move{21}Qg3  gg5

\begin{center}
\newgame
\fenboard{r2q1k1r/ppp5/8/3P1pp1/2PP1n2/1P4Q1/1P1N1P1n/R1B1R1K1 w Qkq - 0 1}
\showboard
\end{center}

White uses a lateral pawn move to safeguard the king.

\move{22}\specialmove{g2}  Qd6  \move{23}Nf1  Nxf1  \move{24}Qxg5  Nh3+

\begin{center}
\newgame
\fenboard{r4k1r/ppp5/3q4/3P1pQ1/2PP4/1P5n/1P4P1/R1B1RnK1 w Qkq - 0 1}
\showboard
\end{center}

\move{25}gxh3  Rg8  \move{26}Re5  Rxg5+  \move{27}Bxg5  Ng3  \move{28}Be7+  Qxe7  \move{29}Rxe7  Kxe7

\begin{center}
\newgame
\fenboard{r7/ppp1k3/8/3P1p2/2PP4/1P4nP/1P6/R5K1 w Q - 0 1}
\showboard
\end{center}

Finally the dust has settled: White having two pawns for the piece.

\move{30}\specialmove{c3}  a6  \move{31}Re1+  Kd7  \move{32}Kg2  Ne4
\move{33}\specialmove{e5}  Ke6  \move{34}\specialmove{d3}  Rg8+  \move{35}Kf3  Ng5+  \move{36}Kg3  c6  \move{37}Re3  Rd8  \move{38}\specialmove{ed5}+  Kf6  \move{39}h4

\begin{center}
\newgame
\fenboard{3r4/1p6/p1p2k2/3P1pn1/2PP3P/3PR1K1/1P6/8 b - - 0 1}
\showboard
\end{center}

Here Black decides to take on d5 rather than try to move the knight, and White recaptures on d5 as well rather than taking on g5!

\blackmove{39}cxd5  \move{40}cxd5  Rxd5  \move{41}\specialmove{e4}  fxe4  \move{42}dxe4  Nxe4  \move{43}Rxe4  a5

\begin{center}
\newgame
\fenboard{8/1p6/5k2/p2r4/4R2P/6K1/1P6/8 w - - 0 1}
\showboard
\end{center}

And now the game moves towards a draw.

\move{44}Ra4  Rd2  \move{45}\specialmove{a2}  \specialmove{ab5}  \move{46}Rf4+  Ke5  \move{47}Rb4  \specialmove{c5}  \move{48}Rxb7  Rxa2  \move{49}h5  Ra6  \move{50}Rb5  \specialmove{d5}  \move{51}Kg4  Ke4  \move{52}Kg5  d4  

with a draw to follow soon.
\gamedrawn

\paragraph{Game AZ-28: AlphaZero Pawn-sideways vs AlphaZero Pawn-sideways}
The game is played from a fixed opening position that arises after \move{1}c4 c5. The remaining moves follow best play, at roughly one minute per move.

% Game 21 from the div directory

\move{1}c4 \bookmove c5 \bookmove \move{2}Nc3  Nc6  \move{3}g3  g6  \move{4}Bg2  e6  \move{5}e4  a6  \move{6}a3  dd6  \move{7}Nge2  Bg7  \move{8}Rb1  Nge7  \move{9}O-O  Rb8  \move{10}bb4  \specialmove{c7}  \move{11}bxc5  Rxb1  \move{12}cxd6  Rb8  \move{13}dxe7  Qd7  \move{14}c5  \specialmove{b6}

\begin{center}
\newgame
\fenboard{1rb1k2r/2pqPpbp/1pn1p1p1/2P5/4P3/P1N3P1/3PNPBP/2BQ1RK1 w k - 0 1}
\showboard
\end{center}

To Black's a6-b6, White responds with c5-b5, another lateral move.

\move{15}\specialmove{b5}  Nd4  \move{16}Nxd4  Bxd4  \move{17}Ne2  Bg7  \move{18}a4  Qxe7  \move{19}dd4  O-O
\move{20}Qb3  \specialmove{a6}  \move{21}Be3  axb5  \move{22}axb5  Ba6

\begin{center}
\newgame
\fenboard{1r3rk1/2p1qpbp/b3p1p1/1P6/3PP3/1Q2B1P1/4NPBP/5RK1 w - - 0 1}
\showboard
\end{center}

White uses a lateral move to protect the pawn

\move{23}\specialmove{c4}  c6  \move{24}b6  c5  \move{25}\specialmove{ed4}  \specialmove{d5}  \move{26}cxd5

\begin{center}
\newgame
\fenboard{1r3rk1/4qpbp/bP2p1p1/3P4/3P4/1Q2B1P1/4NPBP/5RK1 b - - 0 1}
\showboard
\end{center}

Not minding to give up the piece, for getting strong passed pawns in return.

\blackmove{26}Bxe2  \move{27}Re1

\begin{center}
\newgame
\fenboard{1r3rk1/4qpbp/1P2p1p1/3P4/3P4/1Q2B1P1/4bPBP/4R1K1 b - - 0 1}
\showboard
\end{center}

And yet, Black agrees and decides to return the piece instead.

\blackmove{27}exd5  \move{28}Rxe2  Bxd4  \move{29}Bxd4  Qxe2  \move{30}Bxd5

White opts to have the bishop pair and a pawn for two exchanges, an unbalanced position.

\blackmove{30} \specialmove{gf6}  \move{31}h4  \specialmove{hg7}  \move{32}b7  Qa6  \move{33}Bg2  Rfe8  \move{34}Bc5  gg6  \move{35}Qf3  Kg7  \move{36}\specialmove{a7}  Rbd8  \move{37}Be3  Rh8

\begin{center}
\newgame
\fenboard{3r3r/P4pk1/q4pp1/8/7P/4BQP1/5PB1/6K1 w - - 0 1}
\showboard
\end{center}

\move{38}Qf4  Rd7  \move{39}Qb8  Rdd8  \move{40}Qc7  Qa1+  \move{41}Kh2  g5  \move{42}Qc4  Qe5  \move{43}Kh3  Rc8  \move{44}Qg4  Qe6  \move{45}Bb7  Qxg4+  \move{46}Kxg4  Rc4+  \move{47}Kf3  gxh4  \move{48}a8=R  Rxa8  \move{49}Bxa8  \specialmove{g4+}

\begin{center}
\newgame
\fenboard{B7/5pk1/5p2/8/2r3p1/4BKP1/5P2/8 w - - 0 1}
\showboard
\end{center}

\move{50}Ke2  \specialmove{e6}  \move{51}ff3  gxf3+  \move{52}Bxf3  f5  \move{53}Kd3  Ra4  \move{54}Bd1  Ra3+  \move{55}Ke2  ee5  \move{56}\specialmove{f3}  Kf6  \move{57}Bc1  Ra2+  \move{58}Bd2  \specialmove{g5}  \move{59}Bb3  Ra3  \move{60}Bd5  \specialmove{ef5}  \move{61}Be3  f4  \move{62}Bd4+  Kf5  \move{63}Be4+  Ke6  \move{64}\specialmove{e3}  \specialmove{fg4}  \move{65}Kf2  \specialmove{f5}  \move{66}Bb7  \specialmove{e5}  \move{67}Bc8+  Kd5  \move{68}Bxe5  Kxe5  \move{69}Bxg4

and the game soon ended in a draw.
\gamedrawn

\paragraph{Game AZ-29: AlphaZero Pawn-sideways vs AlphaZero Pawn-sideways}
Position from an AlphaZero game played at roughly one minute per move, from a predefined position.

% Game 3, div directory

\begin{center}
\newgame
\fenboard{r3kbnr/ppp1qp1p/2npb3/3Pp3/2B1P1p1/N4N2/PPP2PPP/R1BQR1K1 b Qkq - 0 1}
\showboard
\end{center}

\blackmove{8}gxf3 \move{9}Qxf3 Bd7  \move{10}Nb5

Instead of capturing the knight, White has something else in mind\ldots 

\begin{center}
\newgame
\fenboard{r3kbnr/pppbqp1p/2np4/1N1Pp3/2B1P3/5Q2/PPP2PPP/R1B1R1K1 b Qkq - 0 1}
\showboard
\end{center}

\blackmove{10}Nd4  \move{11}Nxd4 exd4 \move{12}Bg5

\begin{center}
\newgame
\fenboard{r3kbnr/pppbqp1p/3p4/3P2B1/2BpP3/5Q2/PPP2PPP/R3R1K1 b Qkq - 0 1}
\showboard
\end{center}

with a motif of a lateral (e4-f4) discovery! In the game, Black didn't take the bishop.
So, how would have the game proceeded if Black took the bishop?
Here is one possible continuation from AlphaZero:
\blackmove{12}Qxg5 \move{13}\specialmove{f4+} Qe7 \move{14}Rxe7+ Nxe7 \move{15}\specialmove{c5} dxc5 \move{16}Qxb7 Rc8 \move{17}Re1 Kd8 \move{18}Qxa7 Nc6 \move{19}Qa4 \specialmove{hg7} \move{20}c3 Rh6 \move{21}Bb5 Rb8 \move{22}g3 Rd6 \move{23}\specialmove{d3} f6 \move{24}h4 \specialmove{e6} \move{25}h5 \specialmove{f7} \move{26}Rb1 Rb6 \move{27}Kg2.
The continuation is assessed as better for White.

\blackmove{12}f6 \move{13}\specialmove{f4} \specialmove{de6}

\begin{center}
\newgame
\fenboard{r3kbnr/pppbq2p/4pp2/3P2B1/2Bp1P2/5Q2/PPP2PPP/R3R1K1 w Qkq - 0 1}
\showboard
\end{center}

Black uses lateral moves to cover the file as well.

\move{14}dxe6 Bc6 \move{15}Bd5 O-O-O \move{16}Bxc6 bxc6 \move{17}Qxc6

\begin{center}
\newgame
\fenboard{2kr1bnr/p1p1q2p/2Q1Pp2/6B1/3p1P2/8/PPP2PPP/R3R1K1 b Qk - 0 1}
\showboard
\end{center}

White has gained several pawns for the piece, has a dangerous attack and a substantial advantage, according to AlphaZero. Yet, Black uses a lateral pawn move here to prevent immediate disaster:

\blackmove{17}\specialmove{ab7} \move{18}Qa4 Kb8
\move{19}Bh4 Qb4 \move{20}Qb3 \specialmove{g7}
\move{21}Bg3 Bd6 \move{22}c3 Qxb3
\move{23}axb3 dxc3 \move{24}bxc3 Nh6
\move{25}h3 Nf5 \move{26}Bh2 Rhe8 \move{27}Re2 Ne7 \move{28}gg3 g5

\begin{center}
\newgame
\fenboard{1k1rr3/1pp1n3/3bPp2/6p1/5P2/1PP3PP/4RP1B/R5K1 w Q - 0 1}
\showboard
\end{center}

\move{29}fxg5 fxg5 \move{30}h4 gxh4 \move{31}gxh4 Rh8 \move{32}Bxd6 cxd6 \move{33}Ra4 Rc8 \move{34}Rg4 Nf5 \move{35}e7 Kc7 \move{36}Rf4 Nh6 \move{37}\specialmove{g4} Kd7 \move{38}f3 Rhe8 \move{39}Rh2 Rxc3 \move{40}Rxh6 Rxe7 \move{41}Kh2 d5 \move{42}b4 \specialmove{e5} \move{43}Rf8 Rb3 \move{44}g5 e4 \move{45}fxe4 Rxb4 \move{46}\specialmove{f4} Rb3 \move{47}Kg1 Re2 \move{48}Rh7+ Kd6 \move{49}Rd8+ Kc5 \move{50}Rd1 Rg3+ \move{51}Kh1 Re4 \move{52}Rf1 Rg4 \move{53}Rxb7 Rexf4 \move{54}Rxf4 Rxf4 \move{55}Kh2 Rg4 \move{56}Rg7 Kd6 \move{57}Kh3 Rg1 \move{58}Kh4

\begin{center}
\newgame
\fenboard{8/6R1/3k4/6P1/7K/8/8/6r1 b - - 0 1}
\showboard
\end{center}

and White soon won the game.
\whitewins

\paragraph{Game AZ-30: AlphaZero Pawn-sideways vs AlphaZero Pawn-sideways}
The first ten moves for White and Black have been sampled randomly from AlphaZero's opening ``book'', with the probability proportional to the time spent calculating each move. The remaining moves follow best play, at roughly one minute per move.

\move{1}c4  c5  \move{2}Nc3  Nc6  \move{3}g3  g6  \move{4}Bg2  e6  \move{5}e4  dd6  \move{6}Rb1  a6  \move{7}a3  Bg7  \move{8}Nge2  Rb8  \move{9}bb4  \specialmove{c7}  \move{10}O-O  Nge7  \move{11}bxc5  Rxb1  \move{12}cxd6  Rb8  \move{13}dxe7  Qd7 

% Game 76 from on policy games

\begin{center}
\newgame
\fenboard{1rb1k2r/2pqPpbp/p1n1p1p1/8/2P1P3/P1N3P1/3PNPBP/2BQ1RK1 w k - 0 1}
\showboard
\end{center}

In this game (unlike in the main lines section before), Black decides to recapture on e7 with the knight instead.

\move{14}Re1  Bb7  \move{15}\specialmove{b3}  Nxe7  \move{16}dd4  O-O  \move{17}Be3

\begin{center}
\newgame
\fenboard{1r3rk1/1bpqnpbp/p3p1p1/8/2PPP3/1PN1B1P1/4NPBP/3QR1K1 b - - 0 1}
\showboard
\end{center}

Here Black plays a lateral move (a6-b6) to improve its pawn structure:

\blackmove{17}\specialmove{b6}  \move{18}h4  Ba8  \move{19}Qc2

But the pawn marches on, although not forward, opening the line for the rook with:

\blackmove{19}\specialmove{bc6}

\begin{center}
\newgame
\fenboard{br3rk1/2pqnpbp/2p1p1p1/8/2PPP2P/1PN1B1P1/2Q1NPB1/4R1K1 w - - 0 1}
\showboard
\end{center}

\move{20}Bh3  Qc8  \move{21}Rd1  Qa6  \move{22}Bg2  Rfd8  \move{23}Rb1  cd6  \move{24}d5  exd5  \move{25}Nxd5  Nxd5  \move{26}exd5

\begin{center}
\newgame
\fenboard{br1r2k1/2p2pbp/q2p2p1/3P4/2P4P/1P2B1P1/2Q1NPB1/1R4K1 b - - 0 1}
\showboard
\end{center}

The d5 pawn is locking out the a8 bishop, so Black challenges the center with a lateral move, only to decide to push forward on the next move.
This perhaps reveals a fluidity of plans as well as structures.

\blackmove{26}\specialmove{e6}  \move{27}Nf4  e5  \move{28}Ne2  Bf8  \move{29}Nc3  c6

The center is challenged again, this time from the other side, but White has a lateral response to keep things locked:

\move{30}\specialmove{dc5}

\begin{center}
\newgame
\fenboard{br1r1bk1/5p1p/q1p3p1/2P1p3/2P4P/1PN1B1P1/2Q2PB1/1R4K1 b - - 0 1}
\showboard
\end{center}

And Black responds with a lateral move as well, bringing the h-pawn towards the center.

\blackmove{30}\specialmove{hg7}  \move{31}Na4  Qa5  \move{32}\specialmove{hg4}  \specialmove{gf6}  \move{33}g5  \specialmove{fg6}  \move{34}\specialmove{f5}  \specialmove{gf6}

\begin{center}
\newgame
\fenboard{br1r1bk1/5pp1/2p2p2/q1P1pP2/N1P5/1P2B1P1/2Q2PB1/1R4K1 w - - 0 1}
\showboard
\end{center}

After a sequence of lateral moves, the situation has settled on the kingside.

\move{35}Rb2  \specialmove{d5}  \move{36}Bf4  \specialmove{e5}  \move{37}Bd2  Qc7  \move{38}Be3  Qa5  \move{39}Ra2  Qb4  \move{40}Rb2  Qa5  \move{41}Kh2  \specialmove{d5}  \move{42}Bf4  \specialmove{e5}  \move{43}Bd2  Qc7  \move{44}Be3  \specialmove{d6}  \move{45}\specialmove{d5} 
\begin{center}
\newgame
\fenboard{br1r1bk1/2q2pp1/3p1p2/3PpP2/N1P5/1P2B1P1/1RQ2PBK/8 w - - 0 1}
\showboard
\end{center}

Black and White keep reconfiguring the central pawns.

\blackmove{45}\specialmove{c6}  \move{46}\specialmove{dc5}  \specialmove{d6}  \move{47}cxd6  Rxd6
\move{48}c5  Rdd8  \move{49}b4  Bxg2  \move{50}Kxg2  Qc6+  \move{51}f3  \specialmove{d5}  \move{52}Bd4  \specialmove{e5}  \move{53}Bf2  \specialmove{d5}  \move{54}Bd4  \specialmove{e5}  \move{55}Bf2  \specialmove{d5}  \move{56}Qb3  Qb5  \move{57}Nc3  Qc4  \move{58}bb5  Rdc8  \move{59}Nxd5  Qxb3  \move{60}Rxb3  Bxc5  \move{61}b6 Bd6

\begin{center}
\newgame
\fenboard{1rr3k1/5pp1/1P1b1p2/3N1P2/8/1R3PP1/5BK1/8 w - - 0 1}
\showboard
\end{center}

An interesting endgame arises.

\move{62}Nc3  Bc5  \move{63}Nd5  Bd6  \move{64}Rb2  \specialmove{e6}  \move{65}\specialmove{e5}

\begin{center}
\newgame
\fenboard{1rr3k1/5pp1/1P1bp3/3NP3/8/5PP1/1R3BK1/8 b - - 0 1}
\showboard
\end{center}

Both sides using lateral move to create threats.

\blackmove{65}Bf8  \move{66}Nf4  Bc5  \move{67}b7  Rc7  \move{68}Rc2  Bb6  \move{69}Rxc7  Bxc7

\begin{center}
\newgame
\fenboard{1r4k1/1Pb2pp1/4p3/4P3/5N2/5PP1/5BK1/8 w - - 0 1}
\showboard
\end{center}

But the pawn can switch files!

\move{70}\specialmove{a7}

\begin{center}
\newgame
\fenboard{1r4k1/P1b2pp1/4p3/4P3/5N2/5PP1/5BK1/8 b - - 0 1}
\showboard
\end{center}

\blackmove{70}Ra8  \move{71}\specialmove{d5}  g5  \move{72}\specialmove{b7}

\begin{center}
\newgame
\fenboard{r5k1/1Pb2p2/4p3/3P2p1/5N2/5PP1/5BK1/8 b q - 0 1}
\showboard
\end{center}

\blackmove{72}Rb8  \move{73}\specialmove{a7}  Ra8  \move{74}Nd3  exd5  \move{75}Nb4  \specialmove{e5}  \move{76}\specialmove{b7}  Rb8  \move{77}\specialmove{a7}  Ra8  \move{78}Na6  Bd6  \move{79}Bc5

\begin{center}
\newgame
\fenboard{r5k1/P4p2/N2b4/2B1p1p1/8/5PP1/6K1/8 b q - 0 1}
\showboard
\end{center}

\blackmove{79}Bxc5  \move{80}\specialmove{b7}

\begin{center}
\newgame
\fenboard{r5k1/1P3p2/N7/2b1p1p1/8/5PP1/6K1/8 b q - 0 1}
\showboard
\end{center}

\blackmove{80}Rd8  \move{81}Nxc5  f6  \move{82}Ne6  Rb8  \move{83}\specialmove{c7}  Ra8  \move{84}Nd8  Rc8  \move{85}Ne6  Kf7  \move{86}\specialmove{d7}  

\begin{center}
\newgame
\fenboard{2r5/3P1k2/4Np2/4p1p1/8/5PP1/6K1/8 b - - 0 1}
\showboard
\end{center}

\blackmove{86}Rb8  \move{87}d8=Q  Rxd8  \move{88}Nxd8+  Ke7  \move{89}Nc6+  Kd6  \move{90}Nd8  Ke7  \move{91}Nb7  ff5  \move{92}\specialmove{e3}  g4  \move{93}Kf2  e4  \move{94}Ke2  Ke6  \move{95}Nd8+  Ke7  \move{96}Nc6+  Kf6  \move{97}Nd4

\begin{center}
\newgame
\fenboard{8/8/5k2/5p2/3Np1p1/4P1P1/4K3/8 b - - 0 1}
\showboard
\end{center}

However, this position is a draw!

\blackmove{97}\specialmove{g5}  \move{98}Kf1  Ke5  \move{99}Ne2  \specialmove{f5}  \move{100}Kg1  Kd5  \move{101}Kf2  Ke5  \move{102}Kf1  Kd5  \move{103}Kf2  Ke5  \move{104}Kf1  Kd5  \move{105}Kg2  Kc4  \move{106}Nd4  \specialmove{e5}  \move{107}Nc6  Kd5  \move{108}Ne7  Ke6  \move{109}Nc8  \specialmove{f5}  \move{110}Na7  Ke5  \move{111}Nc6+  Kd5  \move{112}Nd4  \specialmove{e5}  \move{113}Nf5  Ke6  \move{114}Ng7+  Kf7  \move{115}Nf5  Ke6  \move{116}Nh6  \specialmove{f5}  \move{117}Kf1  Kf6  \move{118}Ke1  Kg6  \move{119}Ng8  Kf7
\move{120}Nh6+  Kg6  \move{121}Nxg4  fxg4  \move{122}Kd2  Kf5  \move{123}Kc3  \specialmove{ef4}  \move{124}exf4  \specialmove{h4}  \move{125}\specialmove{e4+}  Kxe4  \move{126}gxh4  Kf4  \move{127}Kc2  Kg4  \move{128}Kc1  Kxh4
\gamedrawn

\paragraph{Game AZ-31: AlphaZero Pawn-sideways vs AlphaZero Pawn-sideways}
The first ten moves for White and Black have been sampled randomly from AlphaZero's opening ``book'', with the probability proportional to the time spent calculating each move. The remaining moves follow best play, at roughly one minute per move.

% Game 0 from on policy games 

\move{1}c4  c5  \move{2}e3  e6  \move{3}dd4  cxd4  \move{4}exd4  g6  \move{5}Nc3  Bg7  \move{6}Nb5  \specialmove{bc7}  \move{7}Bf4  Na6  \move{8}Nf3  Nf6  \move{9}h3  d5  \move{10}Bd3  O-O  \move{11}O-O  \specialmove{ab7}  \move{12}Re1  c6  \move{13}Nd6  cc5  \move{14}Be5  cxd4  \move{15}Nxd4  Nc5  \move{16}Bf1  Nce4  \move{17}N4b5

\begin{center}
\newgame
\fenboard{r1bq1rk1/1p3pbp/3Npnp1/1N1pB3/2P1n3/7P/PP3PP1/R2QRBK1 w Qq - 0 1}
\showboard
\end{center}

Here we see a new kind of tactic, made possible by a lateral pawn move!

\blackmove{17}Nxf2  \move{18}Kxf2  \specialmove{e7} 

\begin{center}
\newgame
\fenboard{r1bq1rk1/1p2p1bp/3Npnp1/1N1pB3/2P5/7P/PP3KP1/R2QRB2 w q - 0 1}
\showboard
\end{center}

\move{19}Kg1  exd6  \move{20}Nxd6  Nh5  \move{21}Bxg7  Nxg7  \move{22}Nxc8  Qxc8  \move{23}cxd5  Qc5+  \move{24}Kh1  exd5  \move{25}Qb3  b6

\begin{center}
\newgame
\fenboard{r4rk1/6np/1p4p1/2qp4/8/1Q5P/PP4P1/R3RB1K w Qq - 0 1}
\showboard
\end{center}

The dust has settled, and the game soon ended in a draw.

\move{26}g4  Qd6  \move{27}Bg2  Rad8  \move{28}Rac1  Ne6  \move{29}Qxd5  Qxd5  \move{30}Bxd5  Rxd5  \move{31}Rxe6  Rf2  \move{32}Rxb6  Rdd2  \move{33}g5  \specialmove{hg7}  \move{34}a4  Rh2+  \move{35}Kg1  Rdg2+  \move{36}Kf1  Rf2+  \move{37}Kg1  Rfg2+  \move{38}Kf1  Rf2+  \move{39}Ke1  Rfg2  \move{40}Rb8+  Kh7  \move{41}Kf1  Rf2+  \move{42}Kg1  Rfg2+  \move{43}Kf1  Rf2+  \move{44}Kg1  Rfg2+  \move{45}Kf1
\gamedrawn

\paragraph{Game AZ-32: AlphaZero Pawn-sideways vs AlphaZero Pawn-sideways}
The first ten moves for White and Black have been sampled randomly from AlphaZero's opening ``book'', with the probability proportional to the time spent calculating each move. The remaining moves follow best play, at roughly one minute per move.

% Game 37 from on policy games.

\move{1}c4  c5  \move{2}Nc3  g6  \move{3}e3  e6  \move{4}dd4  \specialmove{bc7}  \move{5}dxc5  Bxc5  \move{6}g4  

\begin{center}
\newgame
\fenboard{rnbqk1nr/p1pp1p1p/4p1p1/2b5/2P3P1/2N1P3/PP3P1P/R1BQKBNR b KQkq - 0 1}
\showboard
\end{center}

Now that is an unusual sight, the early advance of the g-pawn.

\blackmove{6}\specialmove{hg7}  \move{7}Bg2  c6  \move{8}Nf3  d5  \move{9}O-O  Qc7  \move{10}\specialmove{d4}

\begin{center}
\newgame
\fenboard{rnb1k1nr/p1q2pp1/2p1p1p1/2bp4/3P2P1/2N1PN2/PP3PBP/R1BQ1RK1 b Qkq - 0 1}
\showboard
\end{center}

White plays c4-d4, a lateral move, to reinforce the center.

\blackmove{10}Bd6  \move{11}h3  f5  \move{12}\specialmove{f4}

\begin{center}
\newgame
\fenboard{rnb1k1nr/p1q3p1/2pbp1p1/3p1p2/3P1P2/2N1PN1P/PP3PB1/R1BQ1RK1 b Qkq - 0 1}
\showboard
\end{center}

The g-pawn, advanced earlier in what seemed to be weakening, now finds its place on f4, where it shuts out the activity on the b8-h2 diagonal.

\blackmove{12}Nf6  \move{13}a3  \specialmove{b7}  \move{14}Rb1  \specialmove{f7}  \move{15}b4  O-O  \move{16}Bb2  Rd8  \move{17}Rc1  Bf8  \move{18}Qb3  Bd7  \move{19}\specialmove{dc4}  dxc4  \move{20}Qxc4  Be8  \move{21}\specialmove{g3}  Bg7  \move{22}Qb3  Qb6  \move{23}Nd4  Nbd7  
\move{24}aa4  Bf8  \move{25}Ba3  ee5  \move{26}fxe5  Nxe5  \move{27}Rfd1  Neg4  \move{28}\specialmove{gf3}  f4

\begin{center}
\newgame
\fenboard{r2rbbk1/1p3p2/1qp2np1/8/PP1N1pn1/BQN1PP2/5PB1/2RR2K1 w q - 0 1}
\showboard
\end{center}

The game gets quite tactical here.

\move{29}a5  Qc7  \move{30}exf4  Qxf4  \move{31}fxg4  Rxd4  \move{32}Rxd4  Qxd4  \move{33}g5  Ng4  \move{34}Ne4  Qe5  \move{35}Bb2  Qh2+  \move{36}Kf1  Bd7  \move{37}f3  Qf4  \move{38}Re1  Re8  \move{39}Qc4  Nh2+

\begin{center}
\newgame
\fenboard{4rbk1/1p1b1p2/2p3p1/P5P1/1PQ1Nq2/5P2/1B4Bn/4RK2 w - - 0 1}
\showboard
\end{center}

\move{40}Kg1  Nxf3+  \move{41}Bxf3  Qxf3  \move{42}Nf6+

\begin{center}
\newgame
\fenboard{4rbk1/1p1b1p2/2p2Np1/P5P1/1PQ5/5q2/1B6/4R1K1 b - - 0 1}
\showboard
\end{center}

Black needs to give away its queen to stop the attack.

\blackmove{42}Qxf6  \move{43}Bxf6  Rxe1+  \move{44}Kf2  Rd1  \move{45}Qf4  Bf5  \move{46}Qb8  Rd7

\begin{center}
\newgame
\fenboard{1Q3bk1/1p1r1p2/2p2Bp1/P4bP1/1P6/8/5K2/8 w - - 0 1}
\showboard
\end{center}

Is this a fortress? As we will see, the question is slightly more complicated by the fact that the pawn structure isn't fixed,  and things will eventually open up.

\move{47}Be5  Rd2+  \move{48}Ke1  Rd7  \move{49}Bc3  Bd3  \move{50}\specialmove{a4}  Bc2  \move{51}Qc8  Re7  \move{52}Kd2  Bf5  \move{53}Qb8  Rd7+  \move{54}Kc1  Rd3  \move{55}Bf6  Rd7  \move{56}Bb2  \specialmove{e7}

\begin{center}
\newgame
\fenboard{1Q3bk1/1p1rp3/2p3p1/P4bP1/P7/8/1B6/2K5 w - - 0 1}
\showboard
\end{center}

This resource is what Black was keeping in reserve, as a potential way of responding to the threats on the a3-f8 diagonal while the f8 bishop was pinned.

\move{57}Qh2  Bg7  \move{58}\specialmove{b4}  Bxb2+  \move{59}Kxb2  \specialmove{f7}

\begin{center}
\newgame
\fenboard{6k1/1p1r1p2/2p3p1/P4bP1/1P6/8/1K5Q/8 w - - 0 1}
\showboard
\end{center}

The pawn has served its purpose on e7 and moves back.

\move{60}\specialmove{c4}  Be6  \move{61}Qb8+  Kh7  \move{62}Kb3  Kg7  \move{63}Kc3  f6  \move{64}gxf6+ Kxf6  \move{65}Qf8+  Bf7  \move{66}Kb4  g5 \move{67}Qh6+  Bg6  \move{68}Qh8+  Kf7 \move{69}Qh3  Re7  \move{70}Kc5  \specialmove{f5}  \move{71}Kd6  Re6+  \move{72}Kd7  Re7+  \move{73}Kd8  Re8+  \move{74}Kc7  Re7+  \move{75}Kb6  \specialmove{e5}

\begin{center}
\newgame
\fenboard{8/1p2rk2/1Kp3b1/P3p3/2P5/7Q/8/8 w - - 0 1}
\showboard
\end{center}

\move{76}Qf3+  Kg7  \move{77}a6  bxa6  \move{78}Kxc6  e4  \move{79}Qe3  Rf7  \move{80}c5  \specialmove{f4}  \move{81}Qf3  Bf5  \move{82}Kd6  Rf6+  \move{83}Ke5  Bd7  \move{84}Qg2+  Kf7  \move{85}Qh1  Kg7  \move{86}Qb7  Rf7  \move{87}Qg2+  Kh7  \move{88}Qf3  Kg7  \move{89}Kd6  Kh6  \move{90}Qb3  Kg7  \move{91}Qc3+  Kh7  \move{92}Qd3+  Kg7  \move{93}Qd4+  Kh7  \move{94}Qd5  Kg7  \move{95}Qg2+  Kh7  \move{96}Qh2+  Kg8  \move{97}Qg1+  Kh7  \move{98}Qb1+  Kg7  \move{99}Qb2+  Kg8  \move{100}Qb8+  Kg7  \move{101}Qb2+  Kg8  \move{102}Qg2+  Kh7  \move{103}Qc2+  Kg8  \move{104}Qg6+  Kf8  \move{105}Qh5  Kg7  \move{106}Qe5+  Kg8  \move{107}Qg5+  Kh7  \move{108}Qd5  Kg7  \move{109}Qe5+  Kg8  \move{110}Qg5+  Kh7
\move{111}Qh5+  Kg7  \move{112}Qf3  Kh6  \move{113}c6

\begin{center}
\newgame
\fenboard{8/3b1r2/p1PK3k/8/5p2/5Q2/8/8 b - - 0 1}
\showboard
\end{center}

\blackmove{113}Bxc6  \move{114}Kxc6  Kg5  \move{115}Kd6  Rf5  \move{116}Ke6  Rf6+  \move{117}Ke5  Rf5+  \move{118}Ke4  Rf7  \move{119}Kd4  Rd7+  \move{120}Kc4  \specialmove{b6}  \move{121}Qg2+  \specialmove{g4}  \move{122}Qf1  Rd6  \move{123}Qc1+  \specialmove{f4}  \move{124}Qg1+  \specialmove{g4}  \move{125}Qe3+  Kf5  \move{126}Qf2+  Kg5  \move{127}Qe3+  Kf5  \move{128}Qg3  Rf6  \move{129}Qh4  \specialmove{c6}  \move{130}Kd4  \specialmove{d6}  \move{131}Kd5  \specialmove{c6+}  \move{132}Kc5  Rg6  \move{133}Qg3  Rf6  \move{134}Qh4  Rg6  \move{135}Qg3  Rf6  \move{136}Kb6  Kg5  \move{137}Kc7  Rf3  \move{138}Qe5+  Rf5  \move{139}Qe1  c5  \move{140}Kd6

\begin{center}
\newgame
\fenboard{8/8/3K4/2p2rk1/6p1/8/8/4Q3 w - - 0 1}
\showboard
\end{center}

\blackmove{140}c4  \move{141}Qe7+  Kf4  \move{142}Qe2  g3  \move{143}Ke6  Kg5
\move{144}Qxc4  Rf6+  \move{145}Ke5  Rf5+  \move{146}Kd6  Rf6+  \move{147}Ke5  Rf5+  \move{148}Ke6  Rf6+  \move{149}Kd7  Rf4  \move{150}Qe2  Kh4  \move{151}Kd6  Kh3  \move{152}Ke5  Rf2  \move{153}Qh5+  Kg2  \move{154}Ke4  Kg1  \move{155}Ke3  g2  \move{156}Qh4  Rf8  \move{157}Ke2  Rf1  \move{158}Qg3  Kh1  \move{159}Qh3+  Kg1  \move{160}Qh4  \specialmove{h2}  \move{161}Qd4+  Kh1  \move{162}Qh4  Kg1  \move{163}Qg5+  \specialmove{g2}  \move{164}Qh6  Rf2+  \move{165}Ke3  Rf1  \move{166}Ke2  Rf2+  \move{167}Ke3  Rf1  \move{168}Qh3

\begin{center}
\newgame
\fenboard{8/8/8/8/8/4K2Q/6p1/5rk1 w - - 0 1}
\showboard
\end{center}

And the game ended in a draw in a couple of moves.

\gamedrawn

%% file: variants/SAN/self-capture.tex
\subsection{Self-capture}
\label{sec-selfcapture}

In Self-capture chess, we have considered extending the rules of chess to allow players to capture their own pieces.

\subsubsection{Motivation}

The ability to capture one's own pieces could help break ``deadlocks'' and offer additional ways of infiltrating the opponent's position, as well as quickly open files for the attack. Self-captures provide additional defensive resources as well, given that the King that is under attack can consider escaping by self-capturing its own adjacent pieces.

\subsubsection{Assessment}

The assessment of the Self-capture chess variant, as provided by Vladimir Kramnik:

\begin{shadequote}[]{}
\textit{I like this variation a lot, I would even go as far as to say that to me this is simply an improved version of regular chess.}

\textit{Self-captures make a minor influence on the opening stage of a chess game, though we have seen examples of lines that become possible under this rule change that were not possible before. For example, consider the following line \move{1}e4 e5 \move{2}Nf3 Nc6 \move{3}Bb5 a6 \move{4}Ba4 Nf6 \move{5}0-0 Nxe4 \move{6}d4 exd4 \move{7}Re1 f5 \move{8}Nxd4 Qh4 \move{9}g3 in the Ruy Lopez.}

\begin{center}
\newgame
\fenboard{r1b1kb1r/1ppp2pp/p1n5/5p2/B2Nn2q/6P1/PPP2P1P/RNBQR1K1 b kq - 0 9}
\showboard
\end{center}

\textit{While not the main line, it is possible to play in Self-capture chess and AlphaZero assesses it as equal. In classical chess, however, this position is much better for White.
The key difference is that in self-capture chess Black can respond to g3 by taking its own pawn on h7 with the queen,
gaining a tempo on the open file. In fact, White can gain the usual opening advantage earlier in the variation, by playing \move{8}Ng5 d5 \move{9}f3 Bd6 \move{10}fxe4 dxe4, which AlphaZero assesses as giving the 60\% expected score for White after about a minute's thought, which is usually possible to defend with precise play.
In fact, there are multiple improvements for both sides in the original line, but discussing these is beyond the scope of this example. It is worth noting that AlphaZero prefers to utilise the setup of the Berlin Defence, similar to its style of play in classical chess.}

\textit{Regardless of its relatively minor effect on the openings, self-captures add aesthetically beautiful motifs in the middlegames and provide additional options and winning motifs in the endgames.}

\textit{Taking one's own piece represents another way of sacrificing in chess, and material sacrifices make chess games more spectacular and enjoyable both for public and for the players. Most of the times this is used as an attacking idea, to gain initiative and compromise the opponent's king.}

\textit{For example, consider the Dragon Sicilian, as an example of a sharp opening. After \move{1}e4 c5 \move{2}Nf3 d6 \move{3}d4 cxd4 \move{4}Nxd4 Nf6 \move{5}Nc3 g6 \move{6}Be3 Bg7 \move{7}f3 0-0 \move{8}Qd2 Nc6 \move{9}0-0-0 d5 something like \move{10}g4 e5 \move{11}Nxc6 bxc6 is possible, at which point there is already \specialmove{Qxh2}, a self-capture, opening the file against the enemy king. Of course, Black can (and probably should) play differently.}

\begin{center}
\newgame
\fenboard{r1bq1rk1/p4pbp/2p2np1/3pp3/4P1P1/2N1BP2/PPP4Q/2KR1B1R b Kq - 0 1}
\showboard
\end{center}

\textit{The possibilities for self-captures in this example don't end, as after \blackmove{12}d4, White could even consider a self-capture \move{13}\specialmove{Nxe4}, sacrificing another pawn. This is not the best continuation though, and AlphaZero evaluates that as being equal.
It is just an illustration of the ideas which become available, and which need to be taken into account in tactical calculations.}

\textit{In terms of endgames, self-captures affect a wide spectrum of otherwise drawish endgame positions winning for the stronger side. Consider the following examples:}

\begin{center}
\newgame
\fenboard{1b6/1P6/8/5B2/3k4/8/6K1/8 w - - 0 1}
\showboard
\end{center}

\textit{In this position, under Classical rules, the game would be an easy draw for Black.
In Self-capture chess, however, this is a trivial win for White, who can play Bc8 and then capture the bishop with the b7 pawn, promoting to a queen!}

\begin{center}
\newgame
\fenboard{8/4bk2/3pRp2/p1pP1Pp1/PpP3Pp/1P3K1P/8/8 w - - 0 1}
\showboard
\end{center}

\textit{This endgame, which represents a fortress in classical chess, becomes a trivial win in self-capture chess, due to the possibilities for the White king to infiltrate the Black position either via e4 and a self-capture on d5 or via e2, d3 and a self-capture on c4.}

\textit{To conclude, I would highly recommend this variation for chess lovers who value beauty in the game on top of everything else.}
\end{shadequote}

\subsubsection{Main Lines}

Here we discuss ``main lines'' of AlphaZero under
Self-capture chess, when playing with roughly one minute per move from a particular fixed first move. Note that these are not purely deterministic, and each of the given lines is merely one of several highly promising and likely options. Here we give the first 20 moves in each of the main lines, regardless of the position.

\paragraph{Main line after e4}
The main line of AlphaZero after \move{1}e4 in Self-capture chess is:

\move{1}e4 \bookmove e5 \move{2}Nf3 Nc6 \move{3}Bb5 Nf6 \move{4}O-O Nxe4 \move{5}Re1 Nd6 \move{6}Nxe5 Be7 \move{7}Bf1 Nxe5 \move{8}Rxe5 O-O \move{9}Nc3 Ne8 \move{10}Nd5 Bd6 \move{11}Re1 c6 \move{12}Ne3 Be7 \move{13}c4 Nc7 \move{14}d4 d5 \move{15}cxd5 Bb4 \move{16}Bd2  Bxd2 \move{17}Qxd2 Nxd5 \move{18}Nxd5 Qxd5 \move{19}Re5 Qd6 \move{20}Bc4 Bd7

\begin{center}
\newgame
\fenboard{r4rk1/pp1b1ppp/2pq4/4R3/2BP4/8/PP1Q1PPP/R5K1 w Qq - 0 1}
\showboard
\end{center}

\paragraph{Main line after d4}
The main line of AlphaZero after \move{1}d4 in Self-capture chess is:

\move{1}d4 \bookmove d5 \move{2}c4 e6 \move{3}Nc3 Nf6  \move{4}cxd5 exd5 \move{5}Bg5 c6 \move{6}Qc2 Nbd7 \move{7}e3 Be7 \move{8}Nf3 Nh5 \move{9}Bxe7 Qxe7 \move{10}Be2 
% O-O
\mbox{O-O} % to not line break halfway through O-O
\move{11}O-O Ndf6 \move{12}Ne5 g6 \move{13}Qa4 Be6 \move{14}b4 a6 \move{15}Qb3 Ng7 \move{16}Na4 Ne4 \move{17}Qb2 Qg5 \move{18}Nf3 Qe7 \move{19}Ne5 Qg5 \move{20}Nf3 Qe7

\begin{center}
\newgame
\fenboard{r4rk1/1p2qpnp/p1p1b1p1/3p4/NP1Pn3/4PN2/PQ2BPPP/R4RK1 w Qq - 0 1}
\showboard
\end{center}

\paragraph{Main line after c4}
The main line of AlphaZero after \move{1}c4 in Self-capture chess is:

\move{1}c4 \bookmove e5 \move{2}g3 d5 \move{3}cxd5 Nf6 \move{4}Bg2 Nxd5 \move{5}Nc3 Nb6 \move{6}b3 Nc6 \move{7}Bb2 f6 \move{8}Rc1 Bf5 \move{9}Bxc6+ bxc6 \move{10}Nf3 Qd7 \move{11}O-O Be7 \move{12}d3 a5 \move{13}Ne4 O-O \move{14}Qc2 a4 \move{15}Qxc6 Qxc6 \move{16}Rxc6 Nd5 \move{17}Nc3 Nxc3 \move{18}Rxc3 axb3 \move{19}axb3 Rfb8 \move{20}Rxc7 Bd8

\begin{center}
\newgame
\fenboard{rr1b2k1/2R3pp/5p2/4pb2/8/1P1P1NP1/1B2PP1P/5RK1 w q - 0 1}
\showboard
\end{center}

\subsubsection{Instructive games}

\paragraph{Game AZ-33: AlphaZero Self-capture vs AlphaZero Self-capture}
The first ten moves for White and Black have been sampled randomly from AlphaZero's opening ``book'', with the probability proportional to the time spent calculating each move. The remaining moves follow best play, at roughly one minute per move.

% Game 38 from on policy games

\move{1}d4  Nf6  \move{2}c4  e6  \move{3}Nc3  d5  \move{4}cxd5  exd5  \move{5}Bg5  c6  \move{6}Qc2  Nbd7  \move{7}Nf3  
h6  \move{8}Bh4  Be7  \move{9}e3  O-O  \move{10}Bd3  Re8  \move{11}O-O  Ne4  \move{12}Bxe4  Bxh4  \move{13}Bh7+  Kh8  \move{14}Nxh4  Qxh4  \move{15}Bd3  Qe7  \move{16}a3  Nf6  \move{17}b4  Bd7  \move{18}h3  Kg8  \move{19}Rfb1  Rec8  \move{20}Qd1  Be6  \move{21}Ne2  a6  \move{22}Nf4  Ne8  \move{23}a4  Nd6  \move{24}Qb3  Qd7  \move{25}Be2  Bf5  \move{26}Rc1  Qd8  \move{27}Qb2  Be4  \move{28}Rc5  Bf5  \move{29}Rc3  Ra7  \move{30}Rcc1  Raa8  \move{31}Rc5  Qh4  \move{32}Bf1  Re8  \move{33}Rcc1  g5  \move{34}Nd3  Bxd3  \move{35}Bxd3  g4  \move{36}hxg4  Re6  \move{37}Qe2 

\begin{center}
\newgame
\fenboard{r5k1/1p3p2/p1pnr2p/3p4/PP1P2Pq/3BP3/4QPP1/R1R3K1 b - - 0 1}
\showboard
\end{center}

And here we see the first self-capture of the game, creating threats down the h-file:

\blackmove{37}\specialmove{Rxh6} \move{38}Qf3  Qh1+

\begin{center}
\newgame
\fenboard{r5k1/1p3p2/p1pn3r/3p4/PP1P2P1/3BPQ2/5PP1/R1R3Kq w - - 0 1}
\showboard
\end{center}

The end? Not really. In self-capture chess the king can escape by capturing its way through its own army, and hence here it just takes on f2 and gets out of check.

\move{39}\specialmove{Kxf2}  Qh4+  \move{40}Ke2  Re8  \move{41}Rh1  Qxh1  \move{42}Rxh1  Rxh1  \move{43}Qf4  Ne4  \move{44}Bxe4  Rxe4  \move{45}Qb8+  Kg7  \move{46}Qxb7  Rh6  \move{47}Qxa6  Rxg4  \move{48}Kf1  Rh1+  \move{49}Kf2  Rg6  \move{50}b5  Rh2
\move{51}b6  Rgxg2+  \move{52}Kf3  Rf2+

\begin{center}
\newgame
\fenboard{8/5pk1/QPp5/3p4/P2P4/4PK2/5r1r/8 w - - 0 1}
\showboard
\end{center}

Unlike in classical chess, White can still play on here, and AlphaZero does, by advancing the king forward with a self-capture!

\move{53}\specialmove{Kxe3}  Rb2  \move{54}a5  Rb3+

\begin{center}
\newgame
\fenboard{8/5pk1/QPp5/P2p4/3P4/1r2K3/7r/8 w - - 0 1}
\showboard
\end{center}

And, as if one pawn was not enough, White self-captures another one by taking on d4.

\move{55}\specialmove{Kxd4}  Ra2  \move{56}Ke5  Rb5  \move{57}b7  Raxa5  \move{58}Qxa5  Rxa5  \move{59}b8=Q  Ra2

\begin{center}
\newgame
\fenboard{1Q6/5pk1/2p5/3pK3/8/8/r7/8 w - - 0 1}
\showboard
\end{center}

White manages to get a queen, but in the end, Black's defensive resources prove sufficient and the game eventually ends in a draw.

\move{60}Kd6  Re2  \move{61}Kxc6  Re6+  \move{62}Kd7  Rg6  \move{63}Qa8  Re6  \move{64}Qxd5  Kg8  \move{65}Qa8+  Kg7 

With draw soon to follow.

\gamedrawn

\paragraph{Game AZ-34: AlphaZero Self-capture vs AlphaZero Self-capture}
The first ten moves for White and Black have been sampled randomly from AlphaZero's opening ``book'', with the probability proportional to the time spent calculating each move. The remaining moves follow best play, at roughly one minute per move.

% Game 13 from on-policy

\move{1}d4  d5  \move{2}c4  e6  \move{3}Nc3  Nf6  \move{4}Nf3  c6  \move{5}Bg5  h6  \move{6}Bh4  dxc4  \move{7}e4  g5  \move{8}Bg3  b5  \move{9}Be2  Bb7  \move{10}Ne5  Nbd7  \move{11}Qc2  Bg7  \move{12}Rd1  Qe7  \move{13}h4  Nxe5  \move{14}Bxe5  a6  \move{15}a4  Rg8  \move{16}hxg5  hxg5

\begin{center}
\newgame
\fenboard{r3k1r1/1b2qpb1/p1p1pn2/1p2B1p1/P1pPP3/2N5/1PQ1BPP1/3RK2R w Kq - 0 1}
\showboard
\end{center}

\move{17}Qc1  O-O-O  \move{18}Qxg5  Nd5  \move{19}Qxe7  Nxe7  \move{20}g3  Bxe5  \move{21}dxe5  Rxd1+  \move{22}Kxd1  Rd8+  \move{23}Kc1  b4

\begin{center}
\newgame
\fenboard{2kr4/1b2np2/p1p1p3/4P3/Ppp1P3/2N3P1/1P2BP2/2K4R w K - 0 1}
\showboard
\end{center}

Here we come to the first self-capture of the game, White decides to give up the a4 pawn in order to get the knight to an active square.

\move{24}\specialmove{Nxa4}

\begin{center}
\newgame
\fenboard{2kr4/1b2np2/p1p1p3/4P3/Npp1P3/6P1/1P2BP2/2K4R b K - 0 1}
\showboard
\end{center}

And Black responds in turn with a self-capture of its own, on c6!

\blackmove{24}\specialmove{Nxc6}

\begin{center}
\newgame
\fenboard{2kr4/1b3p2/p1n1p3/4P3/Npp1P3/6P1/1P2BP2/2K4R w K - 0 1}
\showboard
\end{center}

\move{25}Nb6+  Kc7  \move{26}Nxc4  Nd4  \move{27}Bd3  Nf3  \move{28}Bc2  Rd4  \move{29}Nd6  Nxe5  \move{30}Nxb7  Kxb7  \move{31}f4  Nd3+  \move{32}Bxd3  Rxd3  \move{33}Rh7  Rxg3  \move{34}Rxf7+  Kc6  \move{35}Rf6  Kd7  \move{36}Rf7+  Kc6  \move{37}Rf6  Kd7  \move{38}f5  exf5  \move{39}exf5  Rf3 

\begin{center}
\newgame
\fenboard{8/3k4/p4R2/5P2/1p6/5r2/1P6/2K5 w - - 0 1}
\showboard
\end{center}

And the game eventually ended in a draw.
\gamedrawn

\textbf{Game AZ-35: AlphaZero Self-capture vs AlphaZero Self-capture}
The first ten moves for White and Black have been sampled randomly from AlphaZero's opening ``book'', with the probability proportional to the time spent calculating each move. The remaining moves follow best play, at roughly one minute per move.

% Game 9 from on-policy

\move{1}d4  e6  \move{2}Nf3  Nf6  \move{3}c4  d5  \move{4}Bg5  dxc4  \move{5}Nc3  a6  \move{6}e4  b5  \move{7}e5  h6  \move{8}Bh4  g5  \move{9}Nxg5  hxg5  \move{10}Bxg5  Nbd7

\begin{center}
\newgame
\fenboard{r1bqkb1r/2pn1p2/p3pn2/1p2P1B1/2pP4/2N5/PP3PPP/R2QKB1R w KQkq - 1 11}
\showboard
\end{center}

In this highly tactical position, self-captures provide additional resources, as AlphaZero quickly demonstrates, by a self-capture on g2, developing the bishop on the long diagonal at the price of a pawn.

\move{11}\specialmove{Bxg2}

\begin{center}
\newgame
\fenboard{r1bqkb1r/2pn1p2/p3pn2/1p2P1B1/2pP4/2N5/PP3PBP/R2QK2R b KQkq - 0 1}
\showboard
\end{center}

Yet, Black responds in turn by a self-capture on a6:

\blackmove{11}\specialmove{Rxa6}

\begin{center}
\newgame
\fenboard{2bqkb1r/2pn1p2/r3pn2/1p2P1B1/2pP4/2N5/PP3PBP/R2QK2R w KQk - 0 1}
\showboard
\end{center}

\move{12}exf6  Rg8  \move{13}h4  Nxf6  \move{14}Nxb5  Be7  \move{15}Qc2  Nd5  \move{16}Qh7

\begin{center}
\newgame
\fenboard{2bqk1r1/2p1bp1Q/r3p3/1N1n2B1/2pP3P/8/PP3PB1/R3K2R b KQ - 0 1}
\showboard
\end{center}

\blackmove{16}Rf8  \move{17}Bh6  Nf6  \move{18}Qc2  Rg8  \move{19}Bf3  c6  \move{20}Nc3  Qxd4  \move{21}Be3  Qe5  \move{22}O-O-O  Nd5

\begin{center}
\newgame
\fenboard{2b1k1r1/4bp2/r1p1p3/3nq3/2p4P/2N1BB2/PPQ2P2/2KR3R w K - 0 1}
\showboard
\end{center}

\move{23}Kb1  Nxc3+  \move{24}bxc3  c5  \move{25}Rhg1  Rh8  \move{26}Rg4  Qf5  \move{27}Qxf5  exf5  \move{28}Rxc4  Be6  \move{29}Bd5  Rd6  \move{30}Rxc5  Rb6+  \move{31}Kc2  Bxc5  \move{32}Bxc5  Ra6  \move{33}a3  Bxd5  \move{34}Rxd5  Rxh4  \move{35}Rxf5 

\begin{center}
\newgame
\fenboard{4k3/5p2/r7/2B2R2/7r/P1P5/2K2P2/8 b - - 0 1}
\showboard
\end{center}

and the game eventually ended in a draw.
\gamedrawn

\paragraph{Game AZ-36: AlphaZero Self-capture vs AlphaZero Self-capture}
The first ten moves for White and Black have been sampled randomly from AlphaZero's opening ``book'', with the probability proportional to the time spent calculating each move. The remaining moves follow best play, at roughly one minute per move.

% Game 53 from on-policy

In this game, self-captures happen towards the end, but the game itself is pretty tactical and entertaining.
We therefore included the full game.

\move{1}Nf3  d5  \move{2}d4  Nf6  \move{3}c4  e6  \move{4}Nc3  c6  \move{5}Bg5  h6  \move{6}Bh4  dxc4  \move{7}e4  g5  \move{8}Bg3  b5  \move{9}Be2  Bb7  \move{10}O-O  Nbd7  \move{11}Ne5  h5  \move{12}Nxd7  Qxd7

\begin{center}
\newgame
\fenboard{r3kb1r/pb1q1p2/2p1pn2/1p4pp/2pPP3/2N3B1/PP2BPPP/R2Q1RK1 w Qkq - 0 1}
\showboard
\end{center}

In the game, White played the pawn to a3, but it's interesting to note that potential self-captures factor in the lines that AlphaZero is calculating at this point. AlphaZero is initially considering the following line: \move{13}Qd2 Be7 \move{14}Qxg5 b4 \move{15}Na4 \specialmove{Qxc6}

\begin{center}
\newgame
\fenboard{r3k2r/pb2bp2/2q1pn2/6Qp/NppPP3/6B1/PP2BPPP/R4RK1 w Qkq - 0 1}
\showboard \\
\textsf{analysis diagram}
\end{center}

where Black has just self-captured its c6 pawn! \move{16}Nc5 Nxe4 \move{17}Qe5 with exchanges to follow. Going back to the game:

\move{13}a3  Rh6  \move{14}Qc1  h4

\begin{center}
\newgame
\fenboard{r3kb2/pb1q1p2/2p1pn1r/1p4p1/2pPP2p/P1N3B1/1P2BPPP/R1Q2RK1 w Qq - 0 1}
\showboard
\end{center}

\move{15}Be5  h3  \move{16}Qxg5  hxg2  \move{17}Rd1  Rg6  \move{18}Qf4  Qe7  \move{19}Qf3  Bg7  \move{20}h4  O-O-O 

\begin{center}
\newgame
\fenboard{2kr4/pb2qpb1/2p1pnr1/1p2B3/2pPP2P/P1N2Q2/1P2BPp1/R2R2K1 w Q - 0 1}
\showboard
\end{center}

\move{21}Bg3  Bh6  \move{22}h5  Rgg8  \move{23}b3  Rxg3

\begin{center}
\newgame
\fenboard{2kr4/pb2qp2/2p1pn1b/1p5P/2pPP3/PPN2Qr1/4BPp1/R2R2K1 w Q - 0 1}
\showboard
\end{center}

\move{24}Qxg3  Rg8  \move{25}Qh3  Bf4  \move{26}Qh4  Bg5  \move{27}Qh2  cxb3  \move{28}Rd3  a6  \move{29}Rb1  c5  \move{30}dxc5  Nd7

\begin{center}
\newgame
\fenboard{2k3r1/1b1nqp2/p3p3/1pP3bP/4P3/PpNR4/4BPpQ/1R4K1 w - - 0 1}
\showboard
\end{center}

\move{31}Rg3  Rg7  \move{32}Qxg2  f5  \move{33}Rxb3  Nxc5  \move{34}Rb4  Qf6  \move{35}Bf1  fxe4

\begin{center}
\newgame
\fenboard{2k5/1b4r1/p3pq2/1pn3bP/1R2p3/P1N3R1/5PQ1/5BK1 w - - 0 1}
\showboard
\end{center}

\move{36}Nxe4  Nxe4  \move{37}Rxe4  Bxe4  \move{38}Qxe4  Bf4  \move{39}Rg6  Rxg6+  \move{40}hxg6  Qg5+  \move{41}Bg2  Be5  \move{42}f4  Bxf4  \move{43}Qb7+  Kd8  \move{44}g7  Qc5+

\begin{center}
\newgame
\fenboard{3k4/1Q4P1/p3p3/1pq5/5b2/P7/6B1/6K1 w - - 0 1}
\showboard
\end{center}

What happens next is a rather remarkable self-capture, demonstrating that it's not only the pawns that can justifiably be self-captured, as the least valuable pieces.
Indeed, White self-captures the bishop on g2, in its attempt at avoiding perpetuals!

\move{45}\specialmove{Kxg2}  Qg5+  \move{46}Kf1

\begin{center}
\newgame
\fenboard{3k4/1Q4P1/p3p3/1p4q1/5b2/P7/8/5K2 b - - 0 1}
\showboard
\end{center}

Yet, Black responds in turn, by capturing its own bishop!
The game ultimately ends in a draw.

\blackmove{46}\specialmove{Qxf4+}  \move{47}Kg2  Qg4+  \move{48}Kf2  Qf4+  \move{49}Ke2  Qe5+  \move{50}Kf3  Qf5+  \move{51}Ke3  Qe5+  \move{52}Kd3  Qd6+  \move{53}Ke4  \specialmove{Qxe6+}  \move{54}Kf4  Qf6+  \move{55}Ke3  Qe5+  \move{56}Kd3  Qd6+  \move{57}Ke4  Qe6+  \move{58}Kd4  Qf6+  \move{59}Kd5  Qf3+  \move{60}Kd6  Qf6+  \move{61}Kc5  Qf2+  \move{62}Kb4  Qd2+  \move{63}Kb3  Qd3+  \move{64}Kb2  Qd2+  \move{65}Kb1  Qd3+  \move{66}Kb2  Qd2+  \move{67}Kb3  Qd3+  \move{68}Kb4  Qd2+
\move{69}\specialmove{Kxa3}  Qc3+  \move{70}Ka2  Qc2+  \move{71}Ka1  Qc1+  \move{72}Ka2  Qc2+  \move{73}Ka1  Qc1+  \move{74}Ka2  Qc2+
\gamedrawn

\paragraph{Game AZ-37: AlphaZero Self-capture vs AlphaZero Self-capture}
The first ten moves for White and Black have been sampled randomly from AlphaZero's opening ``book'',
with the probability proportional to the time spent calculating each move. The remaining moves follow best play, at roughly one minute per move.

% multicapture4, on policy

\move{1}d4  Nf6  \move{2}c4  e6  \move{3}Qc2  c5  \move{4}dxc5  h6  \move{5}Nf3  Bxc5  \move{6}a3  O-O  \move{7}Bf4  Qa5+  \move{8}Nbd2  Nc6  \move{9}e3  Re8  \move{10}Bg3  e5  \move{11}Bh4  g5

\begin{center}
\newgame
\fenboard{r1b1r1k1/pp1p1p2/2n2n1p/q1b1p1p1/2P4B/P3PN2/1PQN1PPP/R3KB1R w KQq - 0 1}
\showboard
\end{center}

\move{12}Nxg5  hxg5  \move{13}Bxg5  Re6  \move{14}O-O-O  Bf8  \move{15}h4  d5

\begin{center}
\newgame
\fenboard{r1b2bk1/pp3p2/2n1rn2/q2pp1B1/2P4P/P3P3/1PQN1PP1/2KR1B1R w Kq - 0 1}
\showboard
\end{center}

Here we see the first self-capture move of the game, creating threats along the h-file:

\move{16}\specialmove{Rxh4}

\begin{center}
\newgame
\fenboard{r1b2bk1/pp3p2/2n1rn2/q2pp1B1/2P4R/P3P3/1PQN1PP1/2KR1B2 b q - 0 1}
\showboard
\end{center}

It's interesting to note that White could have also tried opening the h-file a move earlier, by playing \move{15}\specialmove{Rxh2} instead of \move{15}h4, but AlphaZero prefers provoking \blackmove{15}d5 first and having its rook on the 4th rank, where it stands more active and controls additional squares.

\blackmove{16}Bg7  \move{17}Nb3  Qb6  \move{18}cxd5  Nxd5  \move{19}Rxd5  Rg6

\begin{center}
\newgame
\fenboard{r1b3k1/pp3pb1/1qn3r1/3Rp1B1/7R/PN2P3/1PQ2PP1/2K2B2 w q - 0 1}
\showboard
\end{center}

Here comes another self-capture:

\move{20}\specialmove{Bxe3}

\begin{center}
\newgame
\fenboard{r1b3k1/pp3pb1/1qn3r1/3Rp3/7R/PN2B3/1PQ2PP1/2K2B2 b q - 0 1}
\showboard
\end{center}

\blackmove{20}Qc7  \move{21}Rc5  b6  \move{22}Rc3  Bb7  \move{23}Bd3  Qd8  \move{24}g3  Rd6  \move{25}Bc4  Kf8  \move{26}Qh7  Qf6  \move{27}Nd2  Ne7  \move{28}Rg4  Rad8  \move{29}Bg5  Qxf2

\begin{center}
\newgame
\fenboard{3r1k2/pb2npbQ/1p1r4/4p1B1/2B3R1/P1R3P1/1P1N1q2/2K5 w - - 0 1}
\showboard
\end{center}

\move{30}Bxe7+  Kxe7  \move{31}Rxg7  Qe1+  \move{32}Kc2  Be4+

\begin{center}
\newgame
\fenboard{3r4/p3kpRQ/1p1r4/4p3/2B1b3/P1R3P1/1PKN4/4q3 w - - 0 1}
\showboard
\end{center}

\move{33}Nxe4  Qd1+ is what is played and made possible by a self-capture, avoiding mate:

\move{34}\specialmove{Kxb2} 

\begin{center}
\newgame
\fenboard{3r4/p3kpRQ/1p1r4/4p3/2B1N3/P1R3P1/1K6/3q4 b - - 0 1}
\showboard
\end{center}

Here Black responds by a self-capture on b6:

\blackmove{34}\specialmove{Rxb6+}

\begin{center}
\newgame
\fenboard{3r4/p3kpRQ/1r6/4p3/2B1N3/P1R3P1/1K6/3q4 w - - 0 1}
\showboard
\end{center}

The game soon ends in a draw.

\move{35}Rb3  Rxb3+  \move{36}Bxb3  Qe2+  \move{37}Kb1  Qf1+  \move{38}Ka2  Qe2+  \move{39}Ka1  Qe1+  \move{40}Ka2  Qe2+  \move{41}Kb1  Qe1+  \move{42}Kc2  Qe2+  \move{43}Kc3  Qe3+  \move{44}Kb2  Qe2+  \move{45}\specialmove{Kxa3}  Qa6+  \move{46}Ba4  Qd3+  \move{47}Ka2  Qe2+  \move{48}Ka1  Qe1+  \move{49}Kb2  Qe2+  \move{50}Ka1  Qe1+  \move{51}Ka2  Qe2+  \move{52}Ka3  Qd3+  \move{53}Bb3  Qa6+  \move{54}Kb2  Qe2+  \move{55}Kb1  Qf1+  \move{56}Ka2  Qa6+  \move{57}Kb2  Qe2+  \move{58}Ka1  Qf1+  \move{59}Ka2  Qa6+  \move{60}Kb1  Qf1+  \move{61}Kb2  Qe2+
\gamedrawn

\paragraph{Game AZ-38: AlphaZero Self-capture vs AlphaZero Self-capture}
The following position, with Black to play, arose in an AlphaZero game, played at roughly one minute per move.

% Game 28, on policy games

\begin{center}
\newgame
\fenboard{r2q1rk1/p2nbpp1/5n2/2p4p/2N2B1P/5Q2/P3NPP1/3R1RK1 b - - 0 1}
\showboard
\end{center}

In this position, with Black to play, in classical chess Black would struggle to find a good plan and activity.
Yet, here in self-capture chess, Black plays the obvious idea -- sacrificing the a7 pawn to open the a-file for its rook and initiate active play!

\blackmove{19}\specialmove{Rxa7} \move{20}Nc3 Qa8 \move{21}Qg3  Rfd8

\begin{center}
\newgame
\fenboard{q2r2k1/r2nbpp1/5n2/2p4p/2N2B1P/2N3Q1/P4PP1/3R1RK1 w - - 0 1}
\showboard
\end{center}

Black soon managed to equalize and eventually draw the game.
\gamedrawn

\paragraph{Game AZ-39: AlphaZero Self-capture vs AlphaZero Self-capture}
The following position, with White to play, arose in an AlphaZero game, played at roughly one minute per move.

\begin{center}
\newgame
\fenboard{8/pBp3k1/1pP3p1/8/2P2bbp/8/P5P1/R6K w - - 0 34}
\showboard
\end{center}

In the previous moves, AlphaZero had manoeuvred its light-squared bishop to b7 via a6, with a clear intention of setting up threats to self-capture on b7 and promote the pawn on b8.
Yet, if attempted immediately, Black can respond in turn by playing c6, c5, or even self-capturing on c7 with the bishop. If the bishop moves away from the b8-h2 diagonal, White can proceed with the plan. This explains why White plays the following next:

\move{34}Rc1

\begin{center}
\newgame
\fenboard{8/pBp3k1/1pP3p1/8/2P2bbp/8/P5P1/2R4K b - - 0 34}
\showboard
\end{center}

The rook can now be taken on c1, but this would allow the promotion of the c-pawn via a self-capture.

\blackmove{34}Be6 \move{35}Rf1 Bd6 \move{36}Rd1 Bf4 \move{37}Rd4 Bg3 \move{38}Rxh4

\begin{center}
\newgame
\fenboard{8/pBp3k1/1pP1b1p1/8/2P4R/6b1/P5P1/7K b - - 0 1}
\showboard
\end{center}

\blackmove{38}Bxh4 \move{39}\specialmove{cxb7}

\begin{center}
\newgame
\fenboard{8/pPp3k1/1p2b1p1/8/2P4b/8/P5P1/7K b - - 0 1}
\showboard
\end{center}

And White went on to eventually win the game.
\whitewins

\paragraph{Game AZ-40: AlphaZero Self-capture vs AlphaZero Self-capture}
The following position, with White to play, arose in an AlphaZero game, played at roughly one minute per move.

\begin{center}
\newgame
\fenboard{R7/1N6/P4b2/6k1/r6p/8/4K3/8 w - - 0 1}
\showboard
\end{center}

In this position, White plays a self-capture, \move{50}\specialmove{axb7},
giving away the knight, for an immediate threat of promoting on b8. This is a common pattern in endgames in this variation, where pieces can be used to help promote the passed pawns.

\paragraph{Game AZ-41: AlphaZero Self-capture vs AlphaZero Self-capture}
The following position, with Black to play, arose in an AlphaZero game, played at roughly one minute per move.

\begin{center}
\newgame
\fenboard{8/1p3p2/2pk1r1p/r2p1p1P/P2PnK2/3BP1P1/2R2P2/1R6 b - - 0 1}
\showboard
\end{center}

In this position, AlphaZero as Black plays another self-capture motif: \blackmove{75}\specialmove{fxe4+}, self-capturing its own knight with check, while attacking White's bishop on d3.
This highlights novel tactical opportunities where self-captures can be utilised not only as dynamic material sacrifices for the initiative, but rather a key part of tactical sequences where material gets immediately recovered.

\paragraph{Game AZ-42: AlphaZero Self-capture vs AlphaZero Self-capture}
The following position, with White to play, arose in a
fast-play AlphaZero game, played at roughly one second per move.

\begin{center}
\newgame
\fenboard{2r3rk/p4qpp/1p3p2/2n1n3/4P1R1/4QP2/PB2B2R/7K w - - 0 1}
\showboard
\end{center}

At the moment, White is two pawns down for the attack and has very strong threats against the Black king.
In Classical chess, those might prove fatal, but here Black uses a self-capture as a defensive resource, as can be seen in the following forcing sequence:

\move{34}Rxh7+ Kxh7 \move{35}Rh4+ \specialmove{Kxg8} -- Black is forced to capture its own rook to avoid checkmate -- \move{36}f4 Ng6 \move{37}Rh2 Qxa2 \move{38}Qc1 Qa4 \move{39}Qc4+ Qxc4 \move{40}Bxc4+

\begin{center}
\newgame
\fenboard{2r3k1/p5p1/1p3pn1/2n5/2B1PP2/8/1B5R/7K b - - 0 1}
\showboard
\end{center}

And here Black uses the second self-capture in this sequence, \blackmove{40}\specialmove{Kxg7}, to secure the king.

\paragraph{Game AZ-43: AlphaZero Self-capture vs AlphaZero Self-capture}
The following position, with White to play, arose in a fast-play AlphaZero game, played at roughly one second per move.

\begin{center}
\newgame
\fenboard{5rk1/1Q6/5b1R/5p2/3PnP2/8/7P/5qBK w - - 0 1}
\showboard
\end{center}

With White to play, in Classical chess this would result in a mate in one move, on h7. Yet, in Self-capture chess Black can escape by self-capturing its rook on f8, at Which point White has to attend to its own king's safety.

\move{45}Qh7+ \specialmove{Kxf8} \move{46}Rxf6+ Nxf6 \move{47}Qg6 Qf3+ \move{48}Qg2 Qxf4, leading to a simplified position.

\paragraph{Game AZ-44: AlphaZero Self-capture vs AlphaZero Self-capture}
The following position, with White to play, arose in a
fast-play AlphaZero game, played at roughly one second per move.

\begin{center}
\newgame
\fenboard{4kb1r/pp1b1ppp/4p3/4P3/3pN3/P5Q1/1qB2PPP/5R1K w - - 0 1}
\showboard
\end{center}

In this position, with White to move, White self-captures a pawn to open up dynamic possibilities against the Black king on the f-file.

\move{20}\specialmove{Qxf2} d3 \move{21}Qxf7+ Kd8 \move{22}Bxd3 Qxe5 \move{23}Rd1 Be7 \move{24}Bc4

\begin{center}
\newgame
\fenboard{3k3r/pp1bbQpp/4p3/4q3/2B1N3/P7/6PP/3R3K b - - 0 1}
\showboard
\end{center}

\blackmove{24}Rf8 \move{25}Ng5

\begin{center}
\newgame
\fenboard{3k1r2/pp1bbQpp/4p3/4q1N1/2B5/P7/6PP/3R3K b - - 0 1}
\showboard
\end{center}

\blackmove{25}Qxg5 \move{26}Qxe6

\begin{center}
\newgame
\fenboard{3k1r2/pp1bb1pp/4Q3/6q1/2B5/P7/6PP/3R3K b - - 0 1}
\showboard
\end{center}

Here, Black utilizes a self-capture for defensive purposes, giving up the e7 bishop

\blackmove{26}\specialmove{Qxe7} \move{27}Qh3 Rf6 \move{28}Qg3 Kc8 \move{29}Re1 Qd6 \move{30}Qxg7 Bc6

\begin{center}
\newgame
\fenboard{2k5/pp4Qp/2bq1r2/8/2B5/P7/6PP/4R2K w - - 0 1}
\showboard
\end{center}

with a roughly equal position.